%% file: main_pami.tex
\documentclass[lettersize,journal]{IEEEtran}
\usepackage{amsmath,amsfonts}
\usepackage{array}
\usepackage{subcaption} 
\usepackage{textcomp}
\usepackage{stfloats}
\usepackage{url}
\usepackage{verbatim}
\usepackage{cite}
\usepackage[pagebackref,breaklinks,colorlinks]{hyperref}
\usepackage{url}            
\usepackage{booktabs}       
\usepackage{amsfonts}       
\usepackage{nicefrac}       
\usepackage{microtype}      
\usepackage{xcolor}         
\usepackage{tabularx}       
\usepackage{arydshln} 
\usepackage{colortbl}      

\usepackage{import}
\subimport{./}{packages.tex}

\subimport{./}{macros.tex}
\graphicspath{ {./figures} }

\usepackage{pdfcomment}

\begin{document}
\title{Gaussian Splashing: \\ Direct Volumetric Rendering Underwater}


\author{
\IEEEauthorblockN{
Nir Mualem\IEEEauthorrefmark{1},
Roy Amoyal\IEEEauthorrefmark{1},
Oren Freifeld\IEEEauthorrefmark{1}, and
Derya Akkaynak\IEEEauthorrefmark{2}
}\\[0.5em]
\IEEEauthorblockA{
\IEEEauthorrefmark{1}Ben-Gurion University of the Negev, Israel\\
Email: nirmu@post.bgu.ac.il, amoyalr@post.bgu.ac.il, orenfr@cs.bgu.ac.il
}\\[0.25em]
\IEEEauthorblockA{
\IEEEauthorrefmark{2}The Inter-University Institute for Marine Sciences and the University of Haifa, Israel\\
Email: dakkaynak@univ.haifa.ac.il
}
}



\twocolumn[{%
\renewcommand\twocolumn[1][]{#1}%
\maketitle
\begin{center}
\vspace{-.4cm}
    \centering
   \newcommand{\myW}{0.32\linewidth}
\centering
\includegraphics[width=1\linewidth]{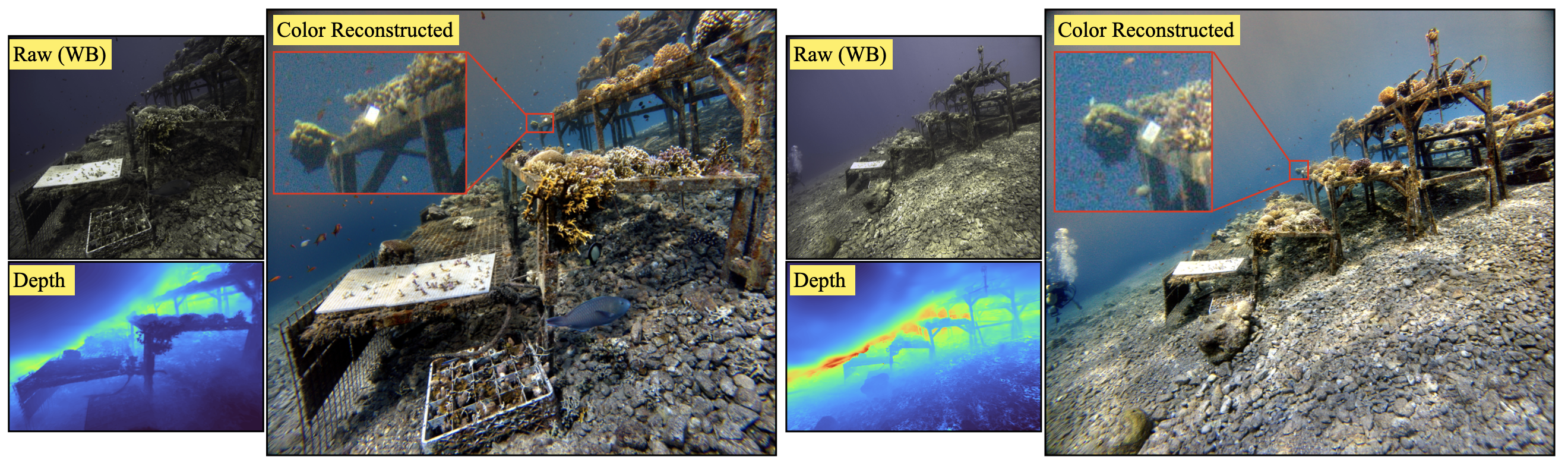}
    \captionsetup{type=figure}   
    \captionof{figure}{The proposed method, Gaussian Splashing, reconstructs accurate geometry for underwater scenes in minutes and achieves real-time rendering at 140 frames fer second. Here, we demonstrate an application in color reconstruction: we used the depth maps produced by our method from raw (but white balanced - WB) images as inputs to the original Sea-thru algorithm~\cite{akkaynak2019}, which requires an accurate depth map to estimate medium parameters. The results, shown here on images from two different distances from the  new \emph{TableDB} dataset we contribute, show excellent visual quality, even for distant scene details.
}
    \label{Fig:Intro}
\end{center}%
}]

\begin{abstract}
\input{content/abstract/abstract}

\end{abstract}

\begin{IEEEkeywords}
Underwater Imaging, 3D Reconstruction, Gaussian Splatting 
\end{IEEEkeywords}

\section{Introduction}
\input{content/intro/intro}

\section{Related Work}
\input{content/related_work/related}

\section{Preliminaries}
\input{content/preliminaries/prelims}

\input{figures/method/method_figure}

\section{The Proposed Method: Gaussian Splashing}\label{Sec:Method}
\input{content/method/method}

\input{content/experiments_results/comp_table}
\section{Experiments and Results}\label{Sec:Results}
\subsection{New Dataset}
\input{content/dataset/dataset}
\subsection{Experiments}

\input{content/experiments_results/experiments}

\subsection{Results}
\input{figures/experiments_results/fig_results}  
\input{figures/experiments_results/fig_results2} 
\input{figures/appendix/appendix_fig_examples3}
\input{figures/appendix/fig_appendix_color_reconstruction}
\input{content/experiments_results/results}

\subsection{Time Comparison}
\input{content/experiments_results/time_comparison}

\input{figures/experiments_results/table_compTime}

\section{Ablation Study}
\input{content/ablation/ablation_table}  
\input{content/ablation/ablation}

\section{Conclusion}\label{Sec:Conclusion}
\input{content/conclusion/conclusion}  
\subsection{Limitations}
\input{content/conclusion/limitations}

\subsection{Discussion}
\input{content/conclusion/discussion}

\section*{Acknowledgments}
\input{content/acknowledgments/acknowledgments}

\bibliographystyle{IEEEtran}
\input{refs.bbl}

\input{content/biography/biography}

\clearpage
\section{Appendix}\label{Sec:Appendix}

\input{content/appendix/appendix}

\vfill

\vfill

\end{document}

%% file: packages.tex
\usepackage{graphicx}

\usepackage{xspace} 

\usepackage{amsmath,amssymb,amsfonts}
\usepackage{bbm}
\usepackage{bm}

\usepackage{times}
\usepackage{subcaption}




\usepackage{enumitem}


\usepackage{color,soul}
\usepackage{xcolor}  

\usepackage{pifont}

\usepackage{tikz}
\usepackage{pgfplots}
\definecolor{lineBlue}{RGB}{57,106,177}
\definecolor{lineOrange}{RGB}{218,124,48}
\definecolor{lineGreen}{RGB}{62,150,81}
\definecolor{lineRed}{RGB}{204,37,41}
\definecolor{lineGray}{RGB}{83,81,84}
\definecolor{linePurple}{RGB}{107,76,154}
\definecolor{lineMaroon}{RGB}{146,36,40}
\definecolor{barBlue}{RGB}{114,147,203}
\definecolor{barOrange}{RGB}{225,151,76}
\definecolor{barGreen}{RGB}{132,186,91}
\definecolor{barRed}{RGB}{211,94,96}
\definecolor{barGray}{RGB}{128,133,133}
\definecolor{barPurple}{RGB}{144,103,167}
\definecolor{barMaroon}{RGB}{171,104,81}

\pgfplotsset{compat=1.18}
\usetikzlibrary{3d,decorations.text,shapes.arrows,positioning,fit,backgrounds}
\usetikzlibrary{intersections, pgfplots.fillbetween}

\usepackage{multirow}

\usetikzlibrary{arrows.meta}

\usepackage[linesnumbered,algoruled,noend,noline]{algorithm2e}
\usepackage[percent]{overpic}

%% file: content/abstract/abstract.tex
In underwater images, most useful features are occluded by water. 
The extent of the occlusion depends on imaging geometry and can vary even across a sequence of burst images. 
As a result, 3D reconstruction methods robust on in-air scenes, like Neural Radiance Field methods (NeRFs) or 3D Gaussian Splatting (3DGS), fail on underwater scenes. 
While a recent underwater adaptation of NeRFs achieved state-of-the-art results, it is impractically slow: reconstruction takes hours and its rendering rate, in frames per second (FPS), 
is less than 1. 
Here, we present a new method that takes only a few minutes for reconstruction and renders novel underwater scenes at 140 FPS. Named \emph{Gaussian Splashing}, our method unifies the strengths and speed of 3DGS with an image formation model for capturing scattering, introducing innovations in the rendering and depth estimation procedures
and in the 3DGS loss function. Despite the complexities of underwater adaptation, our method produces images at unparalleled speeds with superior details. Moreover, it reveals distant scene details with far greater clarity than other methods, dramatically improving reconstructed and rendered images. We demonstrate results on existing datasets and a new dataset we have collected.

%% file: content/intro/intro.tex
Understanding and interpreting underwater scenes pose unique challenges for computer vision. Image features that would be important for downstream tasks such as detection, segmentation, classification, and tracking, \etc, are commonly occluded by color distortions and haze, which arise due to the distance- and wavelength-dependent attenuation of light in the water. It may seem straightforward that, if we could consistently remove the degrading effects of water from underwater photographs, existing computer vision methods would be readily applicable. However, in practice, this is challenging. 

In the physical world, light attenuation in water is governed only by the constituents of the water body. In a photograph, the attenuation parameters have little relation to those of the real world scene; they are specific to the image, the imager, and the conditions under which the image was captured~\cite{akkaynak2017,akkaynak2018revised}. Consequently, in wideband terms, there are no universal medium parameters representing a water `type', or rules that can be generalized to all underwater images. Therefore, for successful color reconstruction—and, by extension, to perform downstream tasks like detection and segmentation—medium parameters must be estimated per image, alongside scene depth.

Recently, Akkaynak \& Treibitz demonstrated with the Sea-thru algorithm~\cite{akkaynak2019} that if the scene depth is known, medium parameters can easily be estimated (\autoref{Fig:Intro}). Depth for underwater scenes is most commonly obtained from multiple images in pre-processing, \eg, through Structure-From-Motion (SFM)~\cite{akkaynak2019,she2024refractive}, or more recently, neural radiance fields (NeRFs)~\cite{levy2023seathru}. 
Unfortunately, however, it is not yet possible to obtain scene depth underwater, in real-time, without using multiple cameras or specialized sensors. Underwater-specific monocular depth estimation methods (\eg,~\cite{gupta2019unsupervised,yu2023udepth,zhang2024atlantis,amitai2023self,varghese2023self}) are performing increasingly better, but do not yet produce accurate enough results when scenes are turbid or objects are heavily occluded by backscatter, and do not generalize to scenes captured under different optical conditions. Depth Anything~\cite{depthanything}, while not trained exclusively for underwater, also fails in generalizing to underwater scenes even though it boasts excellent zero-shot depth estimation ability for a diverse set of land scenes. Thus, robust estimation of scene geometry from multiple images\textemdash but using as few images as possible\textemdash remains the most promising direction to pursue.


\subimport{figures/intro}{fig_speed}

When it comes to recovering scene geometry from a sparse set of images, NeRFs have achieved immense success~\cite{mildenhall2020nerf,gao2022nerf}. While classic approaches for scene representation relied on explicit geometric models (\eg, meshes~\cite{verma20103drendering}), neural rendering employs functions \cite{mueller2022instant} or data structures (\cite{fridovichkeil2021plenoxels}, \cite{jablonski2016realtimevr}) to capture the scene's appearance and geometry. In a fresh perspective in neural rendering, NeRFs learn an implicit function to represent the scene's radiance at every point in space. Levy \etal~\cite{levy2023seathru} were the first to take NeRFs underwater, adapting the original rendering equation for image formation in a scattering medium. That method, Seathru-NeRF, achieved state-of-the-art (SOTA) results for novel-view synthesis and color reconstruction for underwater scenes. However, while successful as a prototype algorithm, the practicality of Seathru-NeRF is limited by its performance\textemdash requiring 15-20 images and approximately 3 hours for training~\cite{nerfstudio,subsea2023}. Even on more powerful hardware and with more optimized implementations, it is difficult to achieve faster training and near real-time rendering as the requirement to evaluate the implicit function along viewing rays during rendering makes NeRF-based methods inherently computationally expensive.

Fortunately, a recent (non-neural) radiance field method, 3D Gaussian Splatting (3DGS)~\cite{kerbl3Dgaussians}, emerges as an alternative 
that, we argue, is intrinsically more suitable for adaptation to underwater scenes. Briefly,
3DGS represents the scene via a finite set of anisotropic 3D Gaussians, augmented with opacity 
and color information. Importantly, 
the algorithm from
~\cite{kerbl3Dgaussians} achieves accelerated training ($\sim$ mins) and lends itself to real-time novel view synthesis. As with most computer-vision methods, however, 3DGS does not readily work on underwater scenes. To fill this gap, we proposed a new method, called \emph{Gaussian Splashing}, which is an adaptation of 3DGS  specific to underwater scenes,
and yet its speed is on par with 3DGS.
\emph{The key insight behind the proposed approach is that the entire underwater image formation formulation
can be  integrated seamlessly  
into the 3DGS data representation and the 3DGS rendering pipeline. Consequently, we were able to implement our
new algorithm, in its entirety, in CUDA,
without having to resort to external (\ie, 
non-CUDA) computations. In turn, this facilitates a smoother backpropagation process and lets the method achieve state-of-the-art performance with minimal rendering time, establishing it as the fastest in this domain. }
Moreover, unlike several recent methods, 
which try, and often fail, to estimate medium parameters directly (\eg, via an expensive neural net), we are able to recover those parameters by feeding the Sea-thru image formation model~\cite{akkaynak2018revised} with our estimated depth maps. 
Our main contributions are:

\textbf{1. New method}: The novelty of the method is fourfold. 1) At its core Gaussian Splashing uses a new rendering equation based on the Sea-thru image formation model~\cite{akkaynak2019} for scattering media. 
2) The new rendering equation alters the original loss function (from~\cite{kerbl3Dgaussians})
by making it also dependent on learnable backscatter coefficients and depth. 
3) We also add a new loss term related to those
coefficients. 4) We propose a new depth estimation procedure and combine it within the overall optimization.  
\textbf{2. SOTA results}: for geometry estimation (especially for far away pixels under high backscatter), scene color reconstruction, and reconstruction of high-frequency areas (\ie, sharp visual detail); see, \eg, \autoref{Fig:Intro}. 
\textbf{3. Speed}: Fast training ($\sim$mins) and real-time rendering  (see, \eg,~\autoref{fig:performance_comparison}) despite the fact that we work with the images in their original high resolution
(\ie, no down-sampling as is mandated by the NeRF-based methods), and real-time novel-view synthesis of original and color-reconstructed scenes. This speed is obtained by  virtue of both our algorithm design and the fact that the entire algorithm 
is implemented in pure CUDA (with a user-friendly Python wrapper). 
\textbf{4. New dataset:} We collected a dataset of images of an underwater 3D scene. Unlike existing such datasets, 
ours presents a large variability
in terms of camera-to-scene distances.


Our results and the speed of our method are a significant step forward towards overcoming the primary bottleneck\textemdash robust depth estimation\textemdash that has prevented underwater computer vision from achieving the progress and performance in-air computer vision has enjoyed in the last decade.


%% file: figures/intro/fig_speed.tex
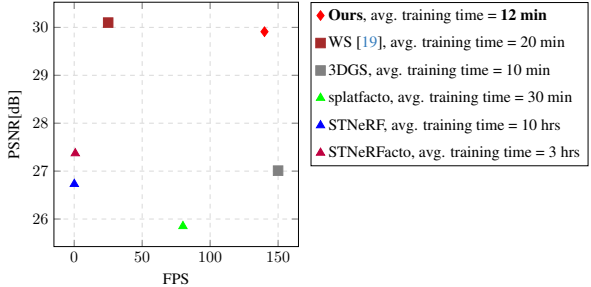
\begin{figure}[!t]
\centering
\begin{tikzpicture}[scale=0.6]
\begin{axis}[
    xlabel={FPS},
    ylabel={PSNR[dB]},
    title={},
    grid=both,
    grid style={dashed, gray!30},
    mark size=3pt,
    width=7cm, 
    height=7cm,
    enlargelimits=0.1,
    legend style={
        at={(1.05, 1)}, 
        anchor=north west,
        legend columns=1, 
        legend cell align={left}, 
        row sep=0.5em,
        font=\normalsize 
    }
]

\addplot [
    only marks,
    mark=diamond*,
    red
] coordinates {
    (140, 29.91)
};
\addlegendentry{\textbf{Gaussian Splashing}, avg. training time = \textbf{12 min}}

\addplot [
    only marks,
    mark=halfcircle*,
    color=brown   
] coordinates {
    (25, 30.1)
};
\addlegendentry{{WS~\cite{li2024watersplatting}}, avg. training time = 20 min}

\addplot [
    only marks,
    mark=halfcircle*,
    color=pink   
] coordinates {
    (110, 27.5)
};
\addlegendentry{{SeaSplat~\cite{yang2024seasplat}}, avg. training time = 1.5 hrs}

\addplot [
    only marks,
    mark=square*,
    color=gray
] coordinates { 
    (150, 27.01)
};
\addlegendentry{3DGS~\cite{kerbl3Dgaussians}, avg. training time = 10 min}

\addplot [
    only marks,
    mark=triangle*,
    green
] coordinates { 
    (80, 25.85)
};
\addlegendentry{Splatfacto ~\cite{liang2024analyticsplatting}, avg. training time = 30 min}

\addplot [
    only marks,
    mark=triangle*,
    blue
] coordinates { 
    (0.08, 26.73)
};
\addlegendentry{STNeRF ~\cite{levy2023seathru}, avg. training time = 10 hrs}

\addplot [
    only marks,
    mark=triangle*,
    purple
] coordinates { 
    (0.8, 27.37)
};
\addlegendentry{STNeRFacto, avg. training time = 3 hrs}

\end{axis}
\end{tikzpicture}
\caption{Comparison of methods by Frames Per Second (FPS) and Peak Signal-to-Noise Ratio (PSNR). Our approach (red diamonds) achieves an impressive PSNR average of $29.11$ while maintaining high inference speeds for real-time rendering. The PSNR values were averaged over the Red Sea, Curaçao, Panama and TableDB datasets (the FPS rates are fairly consistent across datasets).}
\label{fig:performance_comparison}
\end{figure}

%% file: content/related_work/related.tex

Given the different and more complex image formation process underwater, we focus here on works in the underwater realm. 
For the in-land case, 
recent reviews of NeRF-based methods and 3DGS methods can be found in~\cite{yao2024neural}
and~\cite{fei20243d}, respectively. 
%
Due to the logistical difficulties and costs of collecting high-quality, calibrated data at sea, there are markedly fewer studies on scene reconstruction and novel-view synthesis underwater than for scenes on land. 

\textbf{NeRF-based methods}. 
\texttt{WaterNeRF}~\cite{sethuraman2023waternerf} uses an atmospheric image formation model and histogram-equalized images to estimate medium parameters. 
\texttt{WaterHE-NeRF}~\cite{zhou2023waterhe} utilizes the Retinex theory for attenuation removal, which is meant to compensate for spatially-varying illumination. That method is limited to close-range scenes where backscatter is small. \texttt{Seathru-NeRF}~\cite{levy2023seathru} was the first NeRF using an image formation model for scattering. While it achieved SOTA results on novel-view synthesis and color reconstruction, both its training and inference were very slow. Moreover, that method has difficulties with accurately capturing scene details that are far from the camera (that said, in that aspect other existing methods do even worse as capturing such details underwater is hard). 
Lian \etal~\cite{lian2024uncertainty} proposed a framework to quantify the uncertainty in underwater scenes rendered via NeRFs, but did not propose a new way to render. Additionally, it focused only on scenes illuminated with white light (\ie, deep sea imagery), which makes it inapplicable to the outputs of all  other methods using physics-based image formation models, such as \texttt{Seathru-NeRF}. 

\textbf{3DGS}~\cite{kerbl3Dgaussians}, central to our paper, offers a different approach to scene representation by explicitly modeling scenes with  3D Gaussians,
augmented with color and opacity information. 
Unlike NeRFs, which rely on implicit representations, 3DGS offers both enhanced interpretability and intuitive manipulation of scene geometry. Its rendering method, utilizing splatting
(\ie, projecting 3D Gaussians to 2D), analytic derivatives,
and a tile-based processing, 
enables rapid estimation of the
parameters of the augmented Gaussians. Importantly, by replacing the complex processes of NeRF with a discrete representation and direct rendering, 3DGS achieves real-time rendering during inference, marking a significant improvement in rendering efficiency and performance. 

A work that applied 3D Gaussian Splatting (3DGS) to underwater scenes is \texttt{Z-Splat}~\cite{Qu2024ZSplatZG}, introduced by Qu \etal. It extends 3DGS by incorporating sonar data to address the ``missing cone problem''. However, that method requires both optical and acoustic data (and thus specialized instrumentation), in contrast to our purely vision-based multi-view approach.

Several other recent works have also addressed the problem of adapting 3DGS to underwater environments, including \texttt{WaterSplatting}~\cite{li2024watersplatting}, \texttt{UW-GS}~\cite{Wang2024UWGSD3}, and \texttt{SeaSplat}~\cite{yang2024seasplat}. While these methods tackle similar challenges of underwater rendering, they either combine 3DGS with neural networks (~\cite{li2024watersplatting}, ~\cite{Wang2024UWGSD3}) , which increases rendering time, or introduce additional learnable parameters and multiple loss terms (~\cite{yang2024seasplat}) to achieve model fitting.
Among those methods, \texttt{WaterSplatting}~\cite{li2024watersplatting} and \texttt{SeaSplat}~\cite{yang2024seasplat}, have released their code, allowing us to perform a direct comparison. 

Notably, our method, unlike these alternatives, (1) does not rely on deep neural networks and (2) is implemented entirely in CUDA, making it \emph{significantly faster}. In terms of reconstruction quality, our approach achieves results comparable to \texttt{WaterSplatting}---the best-performing method among these works---as shown in \autoref{fig:performance_comparison}.

%% file: content/preliminaries/prelims.tex
\subsection{3D Gaussian Splatting (3DGS)}
3DGS~\cite{kerbl3Dgaussians} represents a 3D scene by $N$ 3D Gaussians, each of which 
is augmented with two more attributes: color and opacity. 
Let $\theta_i=(\bmu_i,\bSigma_i,\bc_i,\sigma_i)$ denote the parameters of Gaussian $i$,
where $\bmu_i\in\RR^3$ is its 3D position (\ie, its mean), $\bSigma_i$, a 3-by-3 symmetric positive-definite matrix, is its covariance matrix,   $\bc_i\in\RR^3$ is the color (represented by Spherical Harmonics), 
and $\sigma_i\ge 0$ is the opacity. Thus, $\Theta\triangleq(\theta_i)_{i=1}^N$
represents the entire scene. 

If $\Theta$ and the camera pose are known, the corresponding 2D image can be generated by rendering the scene. The rendering process accumulates the contributions from each Gaussian (based on differentiable point-based rendering techniques~\cite{KPLD21}~\cite{Wiles2019SynSinEV}~\cite{Wang2019DifferentiableSS}) along viewing rays to generate the image as follows. 
Given the camera pose, the 2D location of a pixel defines a ray, in 3D,
from the camera origin to the pixel. Let $\bp_{i}\in\Rthree$ be the closest point, among all of the points on the ray, to $\bmu_i$. 
The pixel's color, denoted by $\bC$, is computed using
a technique called $\alpha$-compositing: 
\begin{align}
    \bC = \sum\nolimits_{i=1}^{M} \bc_i \alpha_{i}  
    T_i\,, \qquad 
    T_i \triangleq \prod\nolimits_{j=0}^{i-1}(1-\alpha_{j})\,,
    \label{eq:3dgs_orig}
\end{align}
where $M<N$, and, in a slight abuse of notation, possibly renaming indices,
the $M$ Gaussians participating in \autoref{eq:3dgs_orig} are sorted in an increasing order according to their depth (\ie, the distance from the camera),
and $\alpha_i>0$ is the opacity contribution of Gaussian $i$ to this pixel, defined by 
\begin{align}
    \alpha_i = \mathrm{sigmoid}(\sigma_i) e^{-\frac{1}{2} (\bp_i - \bmu_i)^T \bSigma_i^{-1} (\bp_i-\bmu_i)} \,.\label{eq:opacity_impact}
\end{align}
Typically, the pixel-dependent number $M\ll N$; \ie,  
only a small subset of the $N$ Gaussians affects the pixel's color.  
How the subset is chosen will be clear later when we discuss tiles. 
Also, in practice,  $(\bp_i - \bmu_i)^T \bSigma_i^{-1} (\bp_i-\bmu_i)$ is computed in 2D (not 3D) using an affine projection that approximates the camera's projection of $\bp_i$, $\bmu_i$, and $\bSigma_i$; see~\cite{Zwicker2002EWAS} for details. 
  Note that $\alpha_i\in[0,1]$ $\forall i$, and that the effect of $T_i$, which is called the \emph{transmittance} of Gaussian $i$ (for that particular pixel), 
  is that the first $i-1$ Gaussians (whose depths are smaller than that of Gaussian $i$)
  downweight the color contribution of Gaussian $i$ according to their opacity, covariance matrices, and distances from the ray.   

To achieve real-time rendering, 3DGS employs two strategies~\cite{chen2024survey}:
 \textbf{1) Tiles:} To avoid computing contribution of 
 every Gaussian for every pixel, the image is divided into non-overlapping patches, called \emph{tiles}, 
 of $16\times16$ pixels. 
 Among the $N$ 2D projected Gaussians, only those close enough (\eg, less than several standard deviations) to the tile are deemed relevant for that tile (the other Gaussians are ignored). Thus, during the rendering of a pixel, the $\alpha$-compositing (\autoref{eq:opacity_impact})
uses only the projected Gaussians associated with the tile containing the pixel. In other words,
all the pixels within the a tile share the value of $M$.  
 \textbf{2) Parallel Rendering:} 
 The tile-based approach, together with the direct rendering,
 facilitates massive parallelism that leads to huge speedups. 

 So far we discussed the forward problem; \ie, how a 2D image is rendered given  $\Theta$ and the camera pose. 
 The inverse problem, which uses the forward one as a sub-routine,
 is as follows: given a collection of 2D images and the associated estimated camera poses, 
 reconstruct the 3D scene by optimizing over $\Theta$.
 This reconstruction is also referred to as the \emph{learning} of the scene. 
 After the reconstruction, novel view synthesis (\ie, \emph{generalization}) is done by rendering the scene given a new  camera pose.
Finally, compositing techniques are often employed during post-processing of 3DGS to combine the contributions from various effects (\eg, lighting, materials, anti-aliasing~\cite{jiang2023gaussianshader},~\cite{liang2024analyticsplatting},~\cite{yang2023gs4d}).

\subsection{Underwater Image Formation}

By definition, clear air does not attenuate light so image formation models for clear air do not include the medium parameters. When the medium is attenuating (\eg, atmospheric haze, fog, smog, and all water bodies), an image formation model must include, at a minimum, two processes that affect light: out-scattering from the scene as light travels towards the camera (commonly referred to as \emph{attenuation}, which causes color distortions), and in-scattering from the water volume between the camera and the scene (\ie, commonly referred to as \emph{backscatter} or path radiance, which causes haze/visibility loss). We adapt the Sea-thru image formation model~\cite{akkaynak2018revised} that includes both these processes and is applicable to scenes 
that are either in air or underwater:
\begin{align}
    \hspace{-2mm}I = \overbrace{\underbrace{J}_{\text{clean image}} \cdot \underbrace{(e^{-\beta_d(\bv_d) \cdot z})}_{\text{attenuation}}}^{\text{{direct signal}}} + \overbrace{\underbrace{B^\infty}_{\text{color at $\infty$}} 
    \hspace{-2mm}
    \cdot \underbrace{(1 - e^{-\beta_b(\bv_b) \cdot z})}_{\text{backscatter}}}^{\text{{backscatter signal}}} \,.
    \label{eq:scattering}
\end{align}
Here, $I$ is the image captured by the camera and has attenuated colors while $J$ is the clean image ``without water''. The terms $\beta_d$ and $\beta_b$ are wideband medium parameters governing attenuation and backscatter, respectively, and they have dependencies $\bv_d$ and $\bv_b$ on object reflectance, spectrum of ambient light, spectral response of the camera, and physical attenuation coefficients of the water body, all of which are wavelength-dependent functions.  $B^\infty$ is the saturated color of water at infinity, \ie, the signal present in areas without objects, and $z$ is the scene depth, which must be known for each pixel. The \emph{direct signal} governs how the colors of the objects in the scene are affected by the distance that the light travels in a given water body, and the \emph{backscatter signal} governs the density and color of the ``fog'' occluding the scene. The backscatter signal exists due to the scattering in the water volume and is independent of the scene/objects.

It is important to note that the camera sensor is assumed to have a linear response to light, otherwise \autoref{eq:scattering} does not hold. Similarly, the image from which the parameters will be estimated must also be linear, meaning in-camera processed \texttt{.jpg} images or gamma-corrected images cannot be used, unless the non-linearities can be reversed.


%% file: figures/method/method_figure.tex
\graphicspath{ {./figures} }

\begin{figure*}[t]
    \centering
   \includegraphics[width=\textwidth]{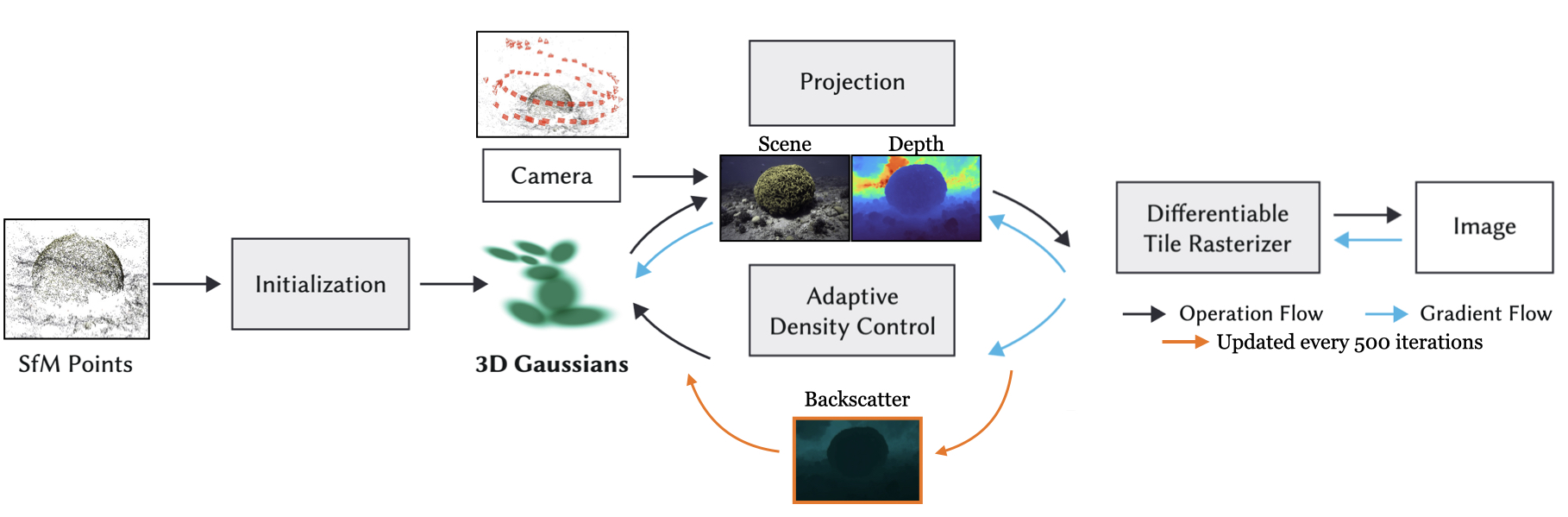}
    
   \caption{Method overview: Initially, we utilize Structure from Motion (SfM) to acquire an initial point cloud and camera poses. Subsequently, we commence the optimization process to refine the model based on our underwater rendering equation and modified tile rasterization, taking those distortions into account. We evaluate backscatter every 500 steps to ensure convergence towards the accurate medium coefficients using our approach. The base figure is adapted from the original 3D Gaussian Splatting method in~\cite{kerbl3Dgaussians}.}
    \label{fig:met_fig}
\end{figure*}

%% file: content/method/method.tex
Unsurprisingly, as we will show in~\autoref{Sec:Results}, in underwater scenes 3DGS 
is very limited. Hence, we extend the traditional 3DGS by introducing three additional learnable parameters and adapt the rendering procedure to accommodate scattering (\autoref{Sec:Method:Rendering}). Similarly, we modify depth estimation for underwater conditions (\autoref{Sec:Method:DepthEst}), thereby altering the original loss terms used in 3DGS, and introduce an additional related loss term (\autoref{Sec:Optimization}). In~\autoref{Sec:Method:Impl} we discuss our pure-CUDA implementation. 
\autoref{fig:met_fig} depicts the entire method.


\subsection{Direct Rendering for Underwater Scenes}
\label{Sec:Method:Rendering}
According to~\cite{akkaynak2018revised}, $\beta_b(\bv_d)$ can be treated as constant within an image, while $\beta_d$ mainly varies with the object distance (\ie, $z$) and to a lesser extent with object reflectance. Similarly, all dependencies of $\beta_d(\bv_d)$ can be assumed to be small~\cite{akkaynak2017}, except for that on the scene depth $z$, which we model to be linear. 
Thus, \autoref{eq:scattering_approx} is well approximated by 
\begin{align}
    I = \overbrace{\underbrace{J}_{\text{clean image}} \cdot \underbrace{(e^{-B_d \cdot z})}_{\text{attenuation}}}^{\text{{direct signal}}} + \overbrace{\underbrace{B^\infty}_{\text{color at $\infty$}} \cdot \underbrace{(1 - e^{-B_b \cdot z})}_{\text{backscatter}}}^{\text{{backscatter signal}}} \label{eq:scattering_approx}
\end{align}
where $B_b$ is a constant approximating $\beta_b(\bv_b)$
and $B_d$ is a constant such that 
$B_d \cdot z \approx \beta_d(\bv_d) \cdot z$ (\ie, the linear dependency of $\beta_d(\bv_d)$ on $z$ is absorbed into $B_d$)

Analogously to~\cite{levy2023seathru}, who extended NeRF to underwater scenes, we incorporate backscatter and attenuation effects into the 3DGS formulation. Utilizing the 
coefficients from \autoref{eq:scattering_approx}, we propose a new rendering formula for \emph{Gaussian Splashing} (namely, Gaussian Splatting underwater): 
\begin{align}
    \bC_{\text{uw}} &= \left(\sum\nolimits_{i=1}^{M} \bc_i \alpha_i  {T_i}\right)e^{-\directv z} + \binf(1-e^{-\bsv z}) 
    \label{eq:3dgs_uw}
\end{align}
where $T_i \triangleq \prod_{j=0}^{i-1}(1-\alpha_{j})$.
Here, $\bC_{\text{uw}}$ is the final color for the underwater scene, while $\directv$
is the direct attenuation parameter and the pair ($\binf$, $\bsv$) are the backscatter attenuation parameters. 
We emphasize that $\directv$, $\binf$, and $\bsv$ are \emph{learnable} parameters and are not assumed to be known beforehand.    
The other variables in \autoref{eq:3dgs_uw} are as in \autoref{eq:3dgs_orig}.  

\subsection{Estimating Dense Depth Maps}\label{Sec:Method:DepthEst}
When using 3DGS for land scenes, there is usually no need to estimate
dense depth maps. However, in the underwater case, color distortions strongly depend on scene depth (among other factors). Drawing inspiration from methodologies in~\cite{Luiten2023Dynamic3G} and~\cite{Chung2023DepthRegularizedOF}, we propose a technique for extracting the depth $z$ from (based on a current estimate of $\Theta$) while leveraging the rasterization pipeline (further details in ~\autoref{Sec:Optimization}).
Concretely, upon the rendering of an image, at each pixel we extract the depth, denoted by $z$, 
as 
\begin{align}
z = \left(\sum\nolimits_{i=1}^{M} d_i \alpha_i T_i\right)/\left(\sum\nolimits_{i=1}^{M} \alpha_i T_i\right)
\label{eqn:Depth}
\end{align}
where 
$d_i$ is the depth of 
Gaussian $i$.
As we will explain in~\autoref{Sec:Optimization}, during the optimization
we use~\autoref{eqn:Depth} to extract the $z$ values and these help in guiding the learning.

\subsection{Optimization}\label{Sec:Optimization}

The original loss function used in~\cite{kerbl3Dgaussians} is relatively simple, comprising two widely-used losses, 
\begin{align}
    &\sum\nolimits_{k=1}^K (1-\lambssim) 
    \ellOneNorm{I^\text{gt}_k-I^\text{r}_k(\Theta)} +\lambssim\loss_{\text{D-SSIM}}(I^\text{gt}_k,I^\text{r}_k(\Theta)) \,,
    \label{eqn:3DGSoriginalLoss}
    \end{align}
where $K$ is the number of images, $\lambssim\in[0,1]$ is user-defined,  $I^\text{gt}_k$ and $I^\text{r}_k$ are $i$-th ground truth (\ie, observed) image and  $i$-th rendered image,
respectively, $\ellOneNorm{\cdot}$ is the $\ell_1$ norm, and $\loss_{\text{D-SSIM}}$~\cite{Wang2003MultiscaleSS} measures the structural similarity between the two images, providing a perceptually-motivated metric.

Let $B_s=B^\infty (1-e^{-B_b\cdot z})$ denote the true-but-unknown backscatter (from \autoref{eq:scattering_approx}) at a pixel. 
One of the advantages of 3DGS over NeRF-based methods is the ability to quickly render entire images
during the optimization process. 
This is in sharp contrast to NeRFs which, due to computational reasons, 
must render only a small number (\eg, 500) of randomly-chosen pixels  
during the optimization~\cite{mildenhall2020nerf}~\cite{mueller2022instant}~\cite{li2023neuralangelo}.
The fast rendering  lets us, during the optimization,
obtain rough estimates of $\binf$ and $\bsv$ from the rendered images.
To better estimate the backscattering parameters, we introduce an additional  loss,
denoted by $\lossbest$, inspired by the backscatter-estimation method outlined in~\cite{akkaynak2019}.
Backscatter increases with depth $z$ and eventually saturates~\cite{akkaynak2018revised,akkaynak2019}. Thus, shadowed areas across the entire image provide a good estimate of the backscatter without the direct signal. Initially, we divide the depth map obtained from the current estimate of the scene (see~\autoref{Sec:Method:DepthEst}) into 10 evenly-spaced clusters covering the range from the minimal value of $z$ to the maximal. Within each  cluster, we identify, in the ground-truth image, the pixels whose RGB triplets are below the bottom $1^{\text{st}}$ percentile. 
These triplets serve as an overestimate of the backscatter, which we model as
\begin{align}
    \widehat{B}_s = \binfhat (1 - e^{-\bsvhat z})
\end{align}
where  $\widehat{B}_s$ and $z$ are the color value and depth estimate at such a pixel.
Given the pairs of $(\widehat{B}_s,z)$ values, we use nonlinear least squares fitting to estimate parameters $\binfhat$ and $\bsvhat$ for each RGB channel while ensuring they remain within the bounds [0, 1] and [0, 5], respectively. The full algorithm for estimating these parameters  appears in the Appendix.

While those estimates, $\binfhat$ and $\bsvhat$, are fairly effective, 
rather than using them directly, we opt to learn the parameters as part of the overall 
optimization, during which we let those estimates merely guide the learning of
 the $\binf$ and $\bsv$ parameters. 
Concretely, our proposed loss function is 
\begin{align}
  & \hspace{-1mm}  \loss(\Theta,\directv,\binf,\bsv) =\nonumber \\ 
  & \hspace{-1mm} 
    \sum\nolimits_{k=1}^K (1-\lambssim)
 \ellOneNorm{I^\text{gt}_k-I^\text{r}_k(\Theta,\directv,\binf,\bsv)} +\nonumber\\
 & \hspace{-1mm} 
    \lambssim\loss_{\text{D-SSIM}}(
    I^\text{gt}_k,I^\text{r}_k(\Theta,\directv,\binf,\bsv)) + \lambest \lossbest
    (\binf,\bsv) 
 \label{eqn:OurFinalLoss}
\end{align}

where $\lambest>0$ is user defined
and 
$
    \lossbest(\binf,\bsv) = \ellOneNorm{\binf - \binfhat} + \ellOneNorm{\bsv - \bsvhat}
    $.
%
As the depth maps improve during the optimization process, every 500 iterations we re-compute (using~\autoref{eqn:Depth} and the nonlinear least squares procedure mentioned earlier) the $\binfhat$ and $\bsvhat$ quantities that appear in $\lossbest$. This approach represents an alternate learning strategy, where the iterative improvement of depth maps leads to the refinement of $\binfhat$ and $\bsvhat$, which we then utilize for improving the optimization over the 3D scene (which, in turn, impacts the depth, $z$; see~\autoref{eqn:Depth}). Therefore, the additional
loss term facilitates better reconstruction and depth estimation in underwater scenes.

It is important to note that the difference between
\autoref{eqn:3DGSoriginalLoss} and \autoref{eqn:OurFinalLoss}
is not just  (the periodic inclusion of) $\lossbest$ but also 
the fact that in \autoref{eqn:OurFinalLoss}, the rendered images (which appear in the 
$\ellOneNorm{\cdot}$  and $\loss_{\text{D-SSIM}}$ loss terms in both equations) are
functions of $\Theta,\directv,\binf$, and $\bsv$ (not just $\Theta$). 

To minimize \autoref{eqn:OurFinalLoss} over $\Theta$, $\directv$, $\binf$, and $\bsv$
we employ the Adam optimizer~\cite{Kingma:arxiv:2014:Adam}, utilizing the same learning rate as in \cite{kerbl3Dgaussians}. Our additional parameters (\ie, $\directv$, $\binf$, and $\bsv$) are trained using the same learning rate as each Gaussian color parameter $\bc_i$.
The details of the partial derivatives  for all our additional parameters appear in the Appendix.

To recap, our approach is grounded in an established formulation of underwater image formation 
that we use to adapt the 3DGS rendering to aquatic environments. 
Together with our proposed additional loss term and gradual depth estimation, 
this lets us propose a fast and effective method for underwater 3D reconstruction and novel view synthesis.

\subsection{Implementation}\label{Sec:Method:Impl}
We incorporate the extraction of the depth map $z$ from the 3DGS representation using 
our own customized CUDA implementation. Empirically, the best result was achieved by mapping our depth to $ [0,1.0) $ by the (shifted and scaled) logistic function, 
$ z = 2/(1 + e^{-0.1 \cdot \widehat{z}}) - 1$
where $\widehat{z}$ is the initial output of 3DGS manipulation based on~\cite{Chung2023DepthRegularizedOF}.
This logistic function transforms the original depth values to ensure they fall within the desired range for underwater visualization.

Leveraging the insights from~\cite{kerbl3Dgaussians}, we have developed a CUDA-based package that harnesses the power of the differential rasterization pipeline. This approach ensures both speed and efficiency in integrating our extra parameters tailored to underwater scenes. 

Finally, 
in~\cite{kheradmand20243d}, one of the steps in 3DGS, the so-called densification process (addition and removal of Gaussians), was recently improved by using an MCMC approach to sample Gaussians for subsequent iterations. We adopt this improvement, as we found it beneficial for capturing high-frequency details in the scene.

%% file: content/experiments_results/comp_table.tex
\begin{table*}[t]
    \setlength\tabcolsep{6pt}
    \caption{Quantitative Comparison. 
    }
    \resizebox{1.0\textwidth}{!}{
    \begin{tabular}{@{}lcccccccccccc@{}}
    \multicolumn{1}{l|}{Dataset}                                                                                    & \multicolumn{3}{c}{Red Sea} & \multicolumn{3}{c}{Curaçao} & \multicolumn{3}{c}{Panama}       & \multicolumn{3}{c}{TableDB} \\ \midrule
    \multicolumn{1}{l|}{Method}                                                                & \multicolumn{1}{c}{PSNR$\uparrow$} & \multicolumn{1}{c}{SSIM$\uparrow$}  & \multicolumn{1}{c}{LPIPS$\downarrow$} & \multicolumn{1}{c}{PSNR$\uparrow$}  & \multicolumn{1}{c}{SSIM$\uparrow$}& \multicolumn{1}{c}{LPIPS$\downarrow$} & \multicolumn{1}{c}{PSNR$\uparrow$}     & \multicolumn{1}{c}{SSIM$\uparrow$}& \multicolumn{1}{c}{LPIPS$\downarrow$} & \multicolumn{1}{c}{PSNR$\uparrow$} & \multicolumn{1}{c}{SSIM$\uparrow$} & \multicolumn{1}{c}{LPIPS$\downarrow$}\\ \hline
    
    \multicolumn{1}{l|}{3DGS~\cite{kerbl3Dgaussians}}                                                              
    & $22.94$ &  $0.87$           &  $0.17$ &  $28.23$   &$0.88$&$0.23$     &$29.88$&$0.91$      &  $0.15$ & $31.17$ &$0.91$ &$0.12$   \\
    \multicolumn{1}{l|}{Splatfacto~\cite{liang2024analyticsplatting}}
    &$21.65$&$0.85$&$0.20$  &  $25.30$&$0.88$ & $0.19$     &     $30.61$ &$0.93$ &$\mathbf{0.07}$&      $32.51$&$0.92$&$\textbf{0.06}$   \\
    \multicolumn{1}{l|}{STNeRF~\cite{levy2023seathru}}                                                              
     &$21.83$&$0.77$&$0.25$  &  $30.48$&$0.87$ & $0.20$     &     $27.89$ &$0.83$ &$0.22$&    $29.76$&$0.86$&$0.15$   \\
    \multicolumn{1}{l|}{ STNeRfacto~\cite{subsea2023}}                                                          
    &$23.17$&$0.86$&$0.14$  &  $28.42$&$0.89$ & $\mathbf{0.12}$     &     $30.53$ &$0.92$ &$\mathbf{0.07}$&      $18.7$&$0.63$&$0.42$      \\
    \multicolumn{1}{l|}{ WS~\cite{li2024watersplatting}}                                                          
    &$24.49$&$0.90$&$0.18$  &  $\mathbf{31.43}$&$\mathbf{0.96}$ & $0.13$     &     $31.24$ &$0.93$ &$0.09$&      $21.29$&$0.73$&$0.31$      \\

        \multicolumn{1}{l|}{ SeaSplat~\cite{yang2024seasplat}}                                                          
    &$22.23$&$0.86$&$0.17$  &  $30.58$&$0.93$ & $0.15$     &     $29.82$ &$0.93$ &$0.12$&      $20.30$&$0.78$&$0.22$      \\
    \multicolumn{1}{l|}{Gaussian Splashing (\textbf{Ours})}                                                         
    &$\mathbf{24.73}$&$\mathbf{0.92}$&$\mathbf{0.11}$  &  $31.26$&$0.92$ & $0.17$     &     $\mathbf{31.35}$ &$\mathbf{0.94}$ &$0.11$&    
       $\mathbf{32.61}$&$\mathbf{0.96}$&$0.07$ 
\end{tabular}
    \label{tab:comparison}    
    }
    \end{table*}

%% file: content/dataset/dataset.tex
Noting that existing public underwater datasets 
are limited in terms of the variation in camera-to-scene distances, 
we captured a new underwater dataset (TableDB). It consists of 172 images with a resolution of $1384\times918$ pixels
(see our Appendix for example images), and, unlike existing public datasets,
it is unbounded in terms camera-to-scene distances.

%% file: content/experiments_results/experiments.tex
\textbf{Datasets and Camera Pose Extraction.} 
In addition to our TableDB we
also experimented on several publicly available underwater datasets from~\cite{levy2023seathru}. 
We used COLMAP~\cite{schoenberger2016sfm, schoenberger2016mvs} to extract camera poses and utilized its sparse 3D point initialization to determine the initial positions of the Gaussians.


\textbf{Competing methods and evaluation metrics.} 
The most relevant prior work addressing 3D reconstruction and rendering of underwater scenes in natural environments is Sea-Thru NeRF (STNeRF)~\cite{levy2023seathru}, which serves as our primary quality benchmark. 
Two implementations of STNeRF exist: one built upon the NeRF backbone from~\cite{multinerf2022}, and another implemented within the Nerfstudio framework~\cite{Tancik2023NerfstudioAM}, where the more advanced Nerfacto variant is used as the backbone. 
We refer to the latter as \textit{STNeRFacto} and compare our method against both versions. 
Our approach extends the original 3D Gaussian Splatting (3DGS)~\cite{kerbl3Dgaussians} by adapting it to aquatic environments through the integration of ocean optics models. 
Accordingly, we also include comparisons with the original 3DGS and with Splatfacto~\cite{Ye2023MathematicalSF}, which currently achieves state-of-the-art results on in-the-wild scenes. 
Additionally, we report qualitative and runtime comparisons with WaterSplatting (WS)~\cite{li2024watersplatting} and SeaSplat~\cite{yang2024seasplat}. We measure performance using three widely-used metrics: PSNR, LPIPS, and SSIM. Peak Signal-to-Noise Ratio (PSNR) quantifies image quality by measuring the ratio between the maximal possible power of a signal and the power of the corrupting noise. Learned Perceptual Image Patch Similarity (LPIPS) assesses perceptual similarity by comparing deep features extracted from neural networks, as in~\cite{Krizhevsky2012ImageNetCW}, providing a human-aligned evaluation of visual similarity. Structural Similarity Index (SSIM) evaluates image similarity based on luminance, contrast, and structure, thereby capturing perceptual differences that align closely with human visual perception.

\textbf{Hyper-parameters, memory and computational resources.} 
 In our experiments, we set $\lambest= 0.1$ and $\lambssim = 0.3$. 
All of the other hyper-parameters values are as in~\cite{kerbl3Dgaussians}; see our appendix for details.
Training was conducted on a single NVIDIA GeForce RTX 4090 GPU. The average memory consumption per scene over 30,000 training steps was approximately 500 MB, which is slightly less than STNeRF's average of around 600 MB.

%% file: figures/experiments_results/fig_results.tex
\graphicspath{ {./figures/experiments_results/figs} }
\begin{figure*}[t]
\newcommand{\myW}{0.24\linewidth}
\newcommand{\myH}{2cm} 
\centering
\subcaptionbox{Ground Truth}[\myW]
{\includegraphics[width=\linewidth, height=\myH]{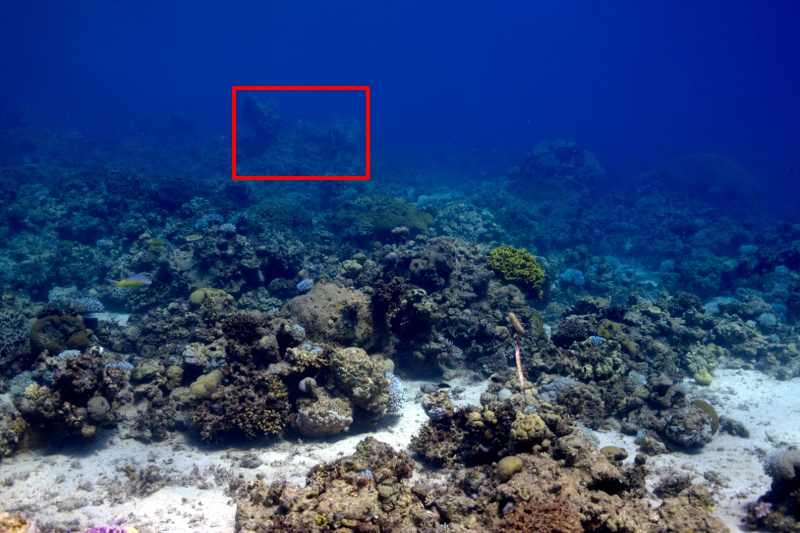}\\
 \includegraphics[width=\linewidth, height=\myH]{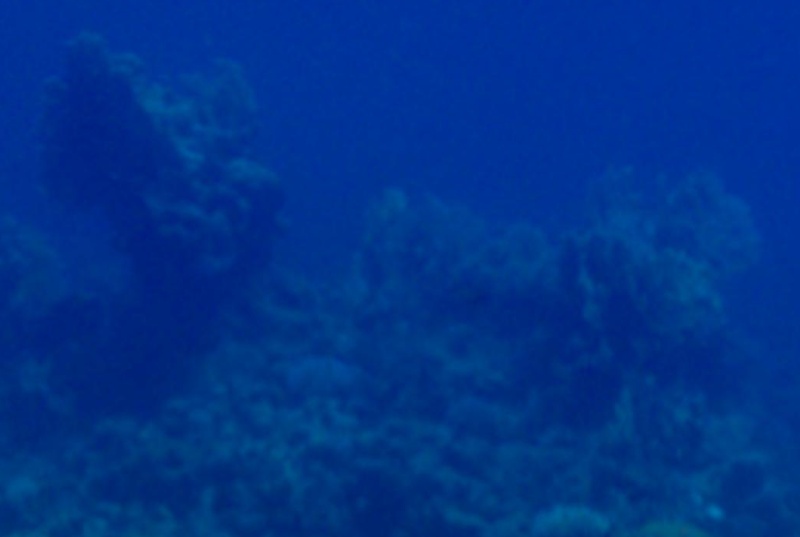}\\
 \includegraphics[width=\linewidth, height=\myH]{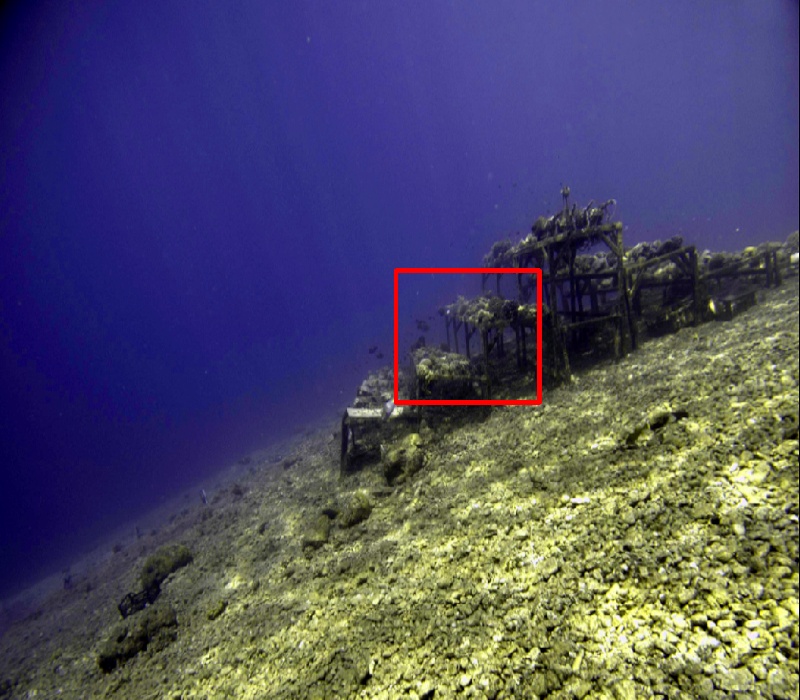}\\
 \includegraphics[width=\linewidth, height=\myH]{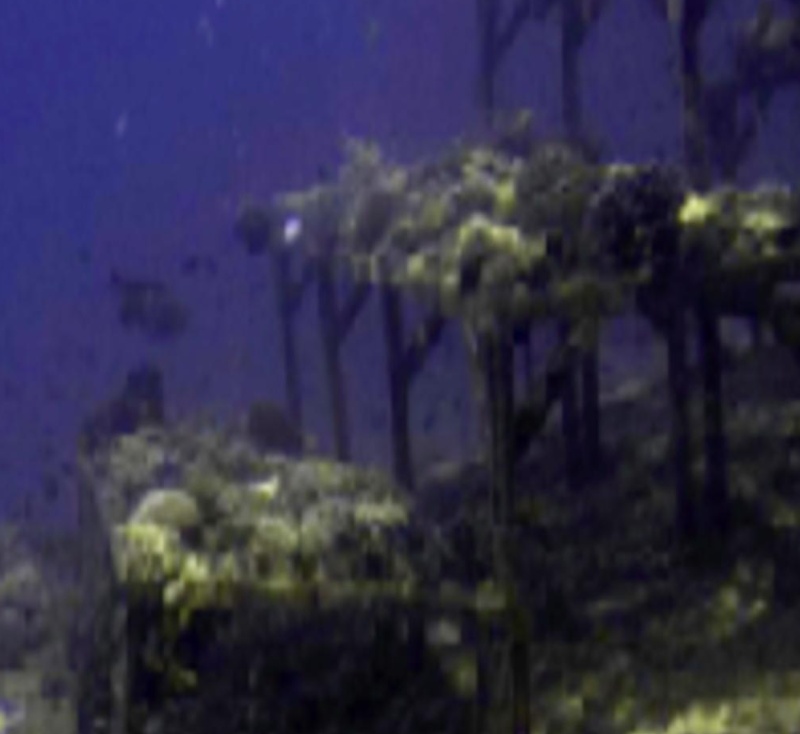}\\
 \includegraphics[width=\linewidth, height=\myH]{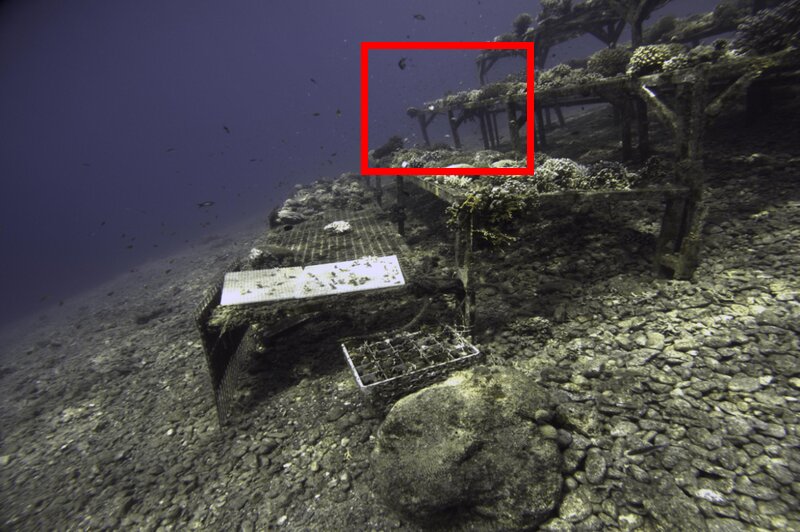}\\
 \includegraphics[width=\linewidth, height=\myH]{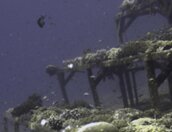}\\
 \includegraphics[width=\linewidth, height=\myH]{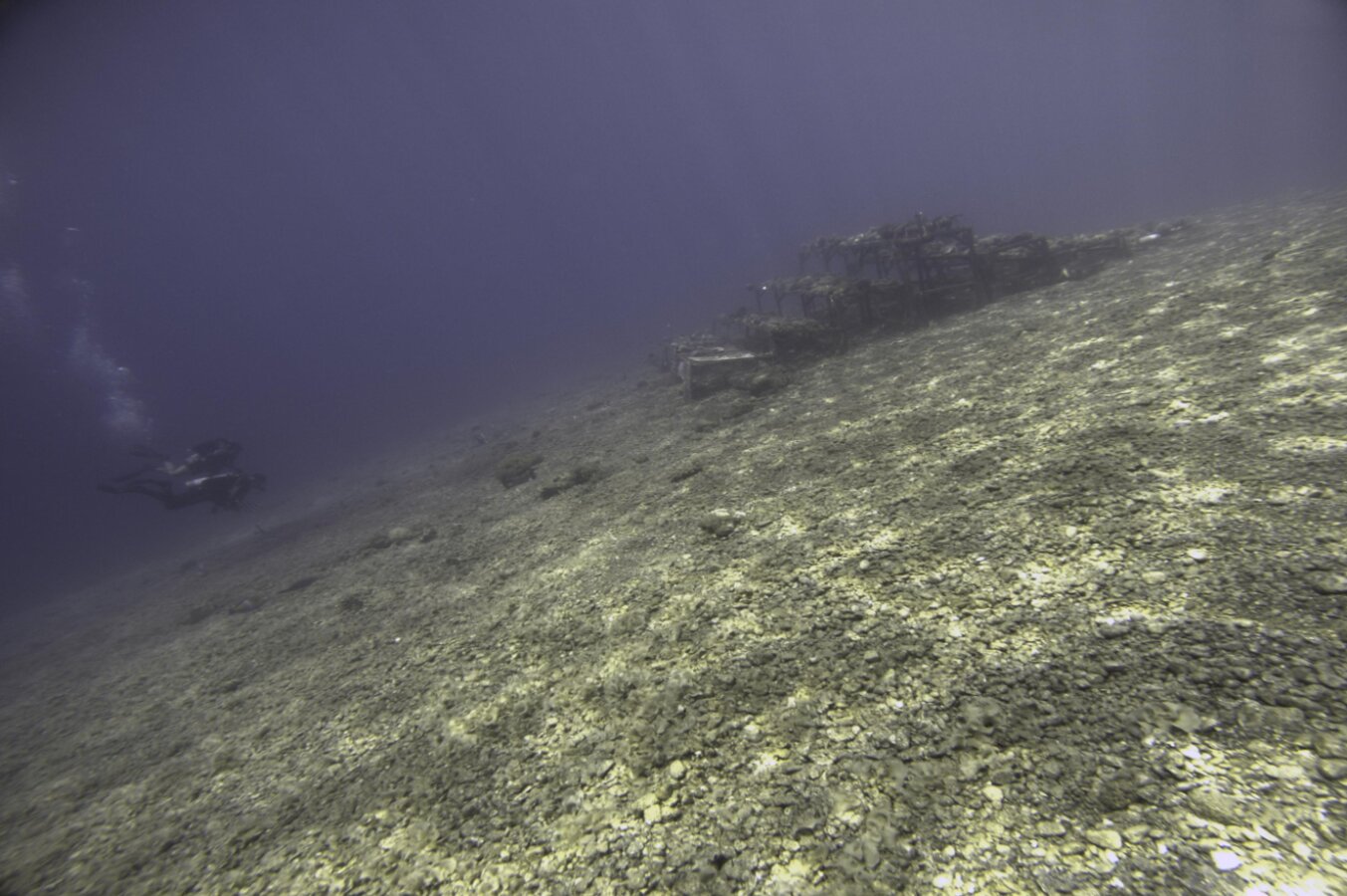}}
\subcaptionbox{3DGS~\cite{kerbl3Dgaussians}}[\myW]
{\includegraphics[width=\linewidth, height=\myH]{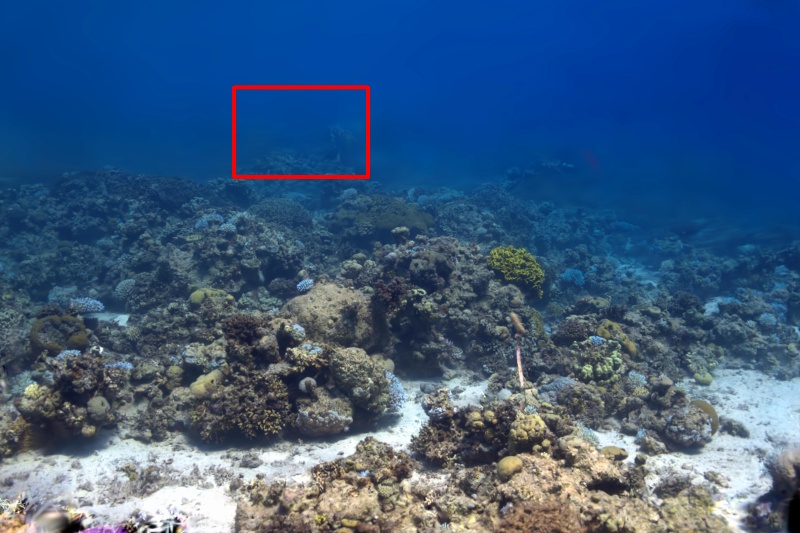}\\
 \includegraphics[width=\linewidth, height=\myH]{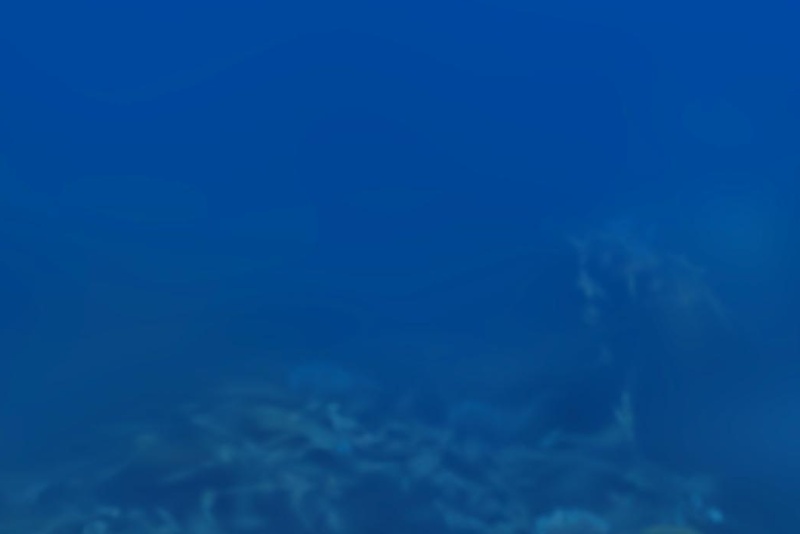}\\
 \includegraphics[width=\linewidth, height=\myH]{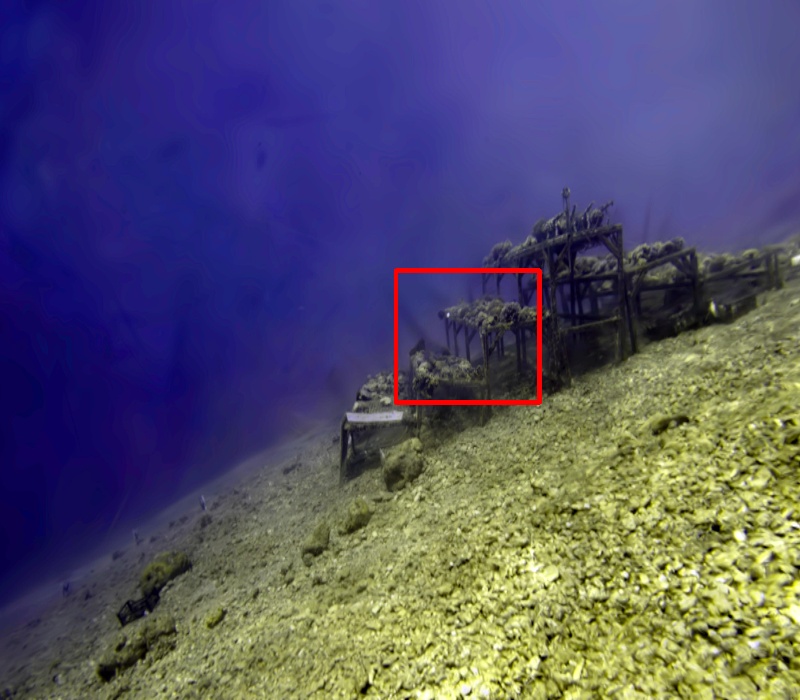}\\
 \includegraphics[width=\linewidth, height=\myH]{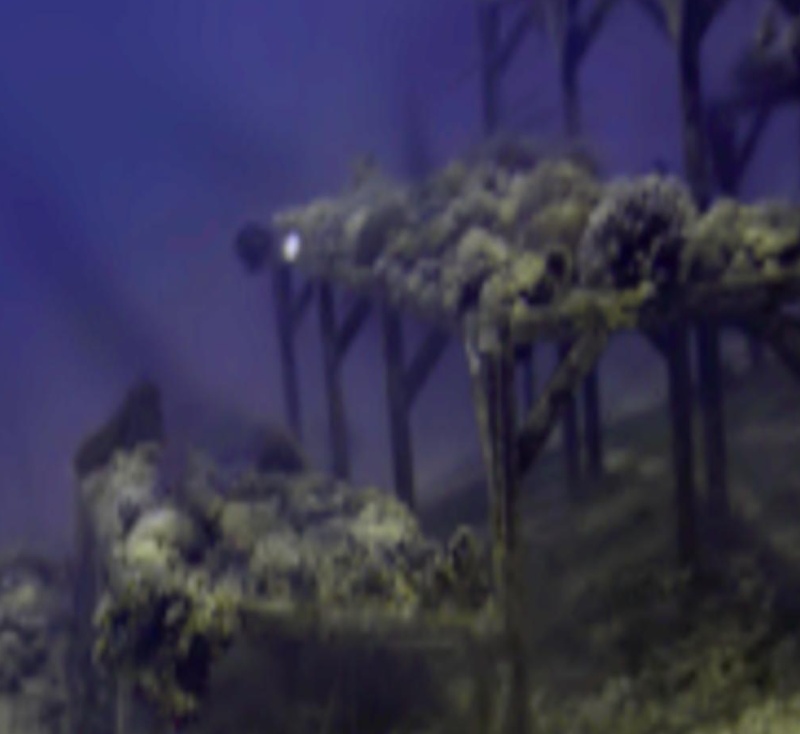}\\
 \includegraphics[width=\linewidth, height=\myH]{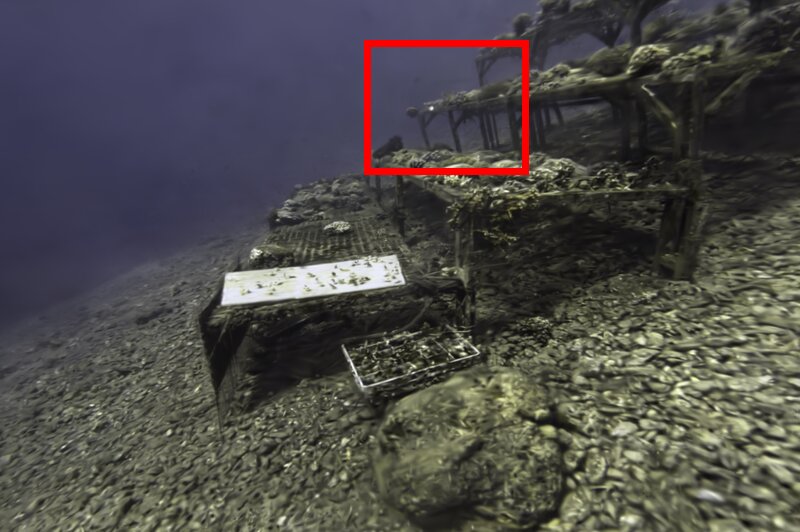}\\
 \includegraphics[width=\linewidth, height=\myH]{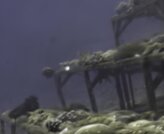}\\
 \includegraphics[width=\linewidth, height=\myH]{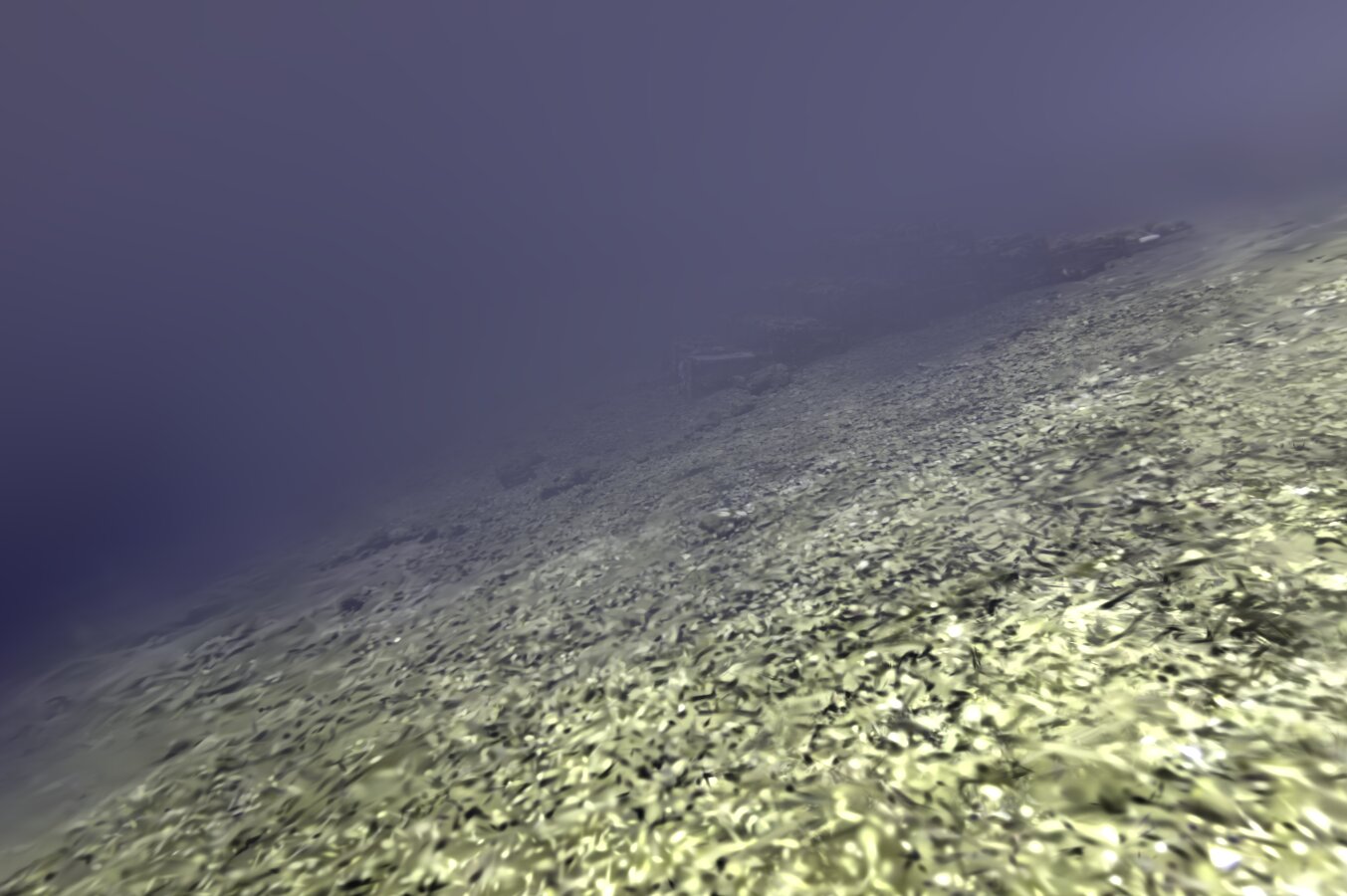}}
\subcaptionbox{STNeRFacto~\cite{subsea2023}}[\myW]
{\includegraphics[width=\linewidth, height=\myH]{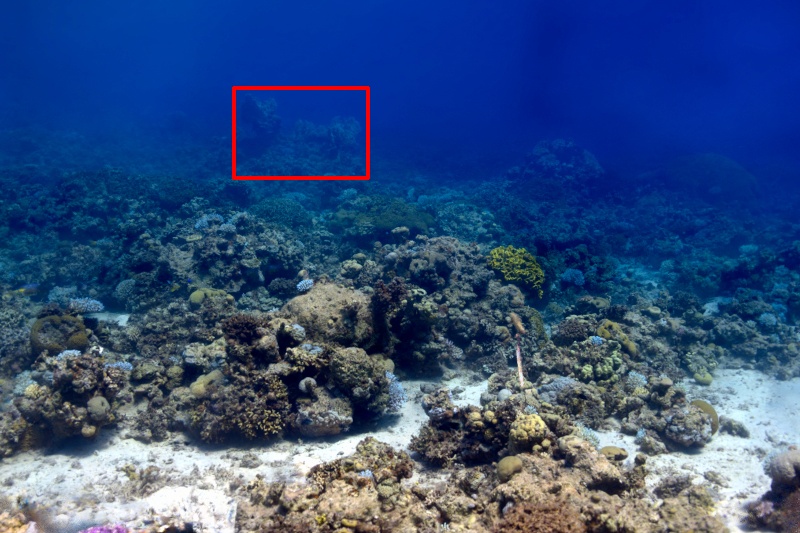}\\
 \includegraphics[width=\linewidth, height=\myH]{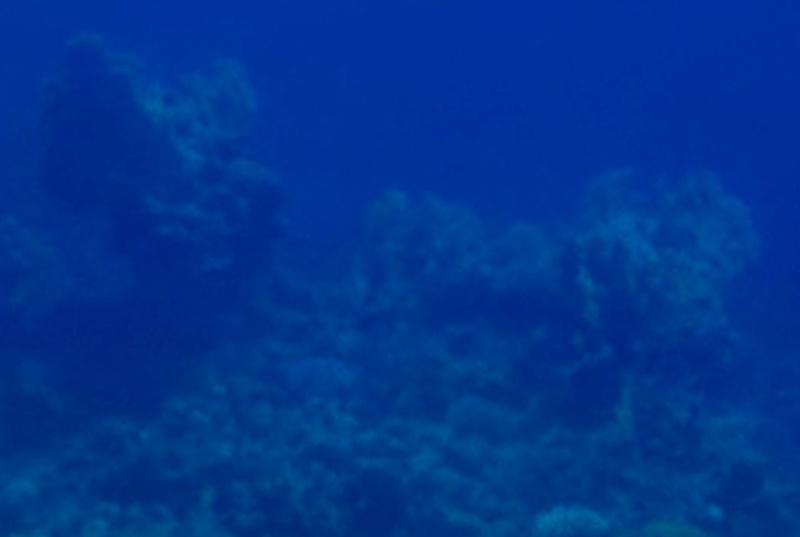}\\
 \includegraphics[width=\linewidth, height=\myH]{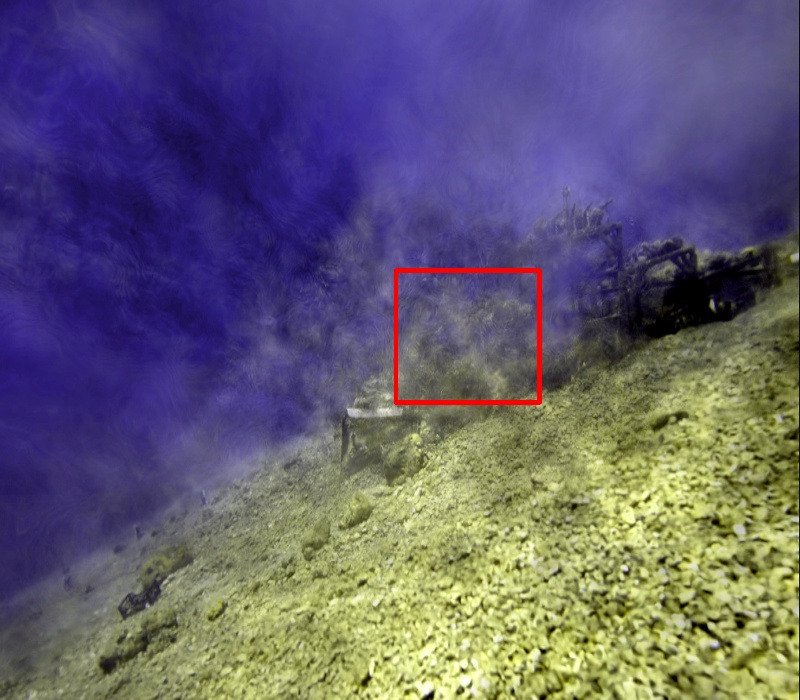}\\
 \includegraphics[width=\linewidth, height=\myH]{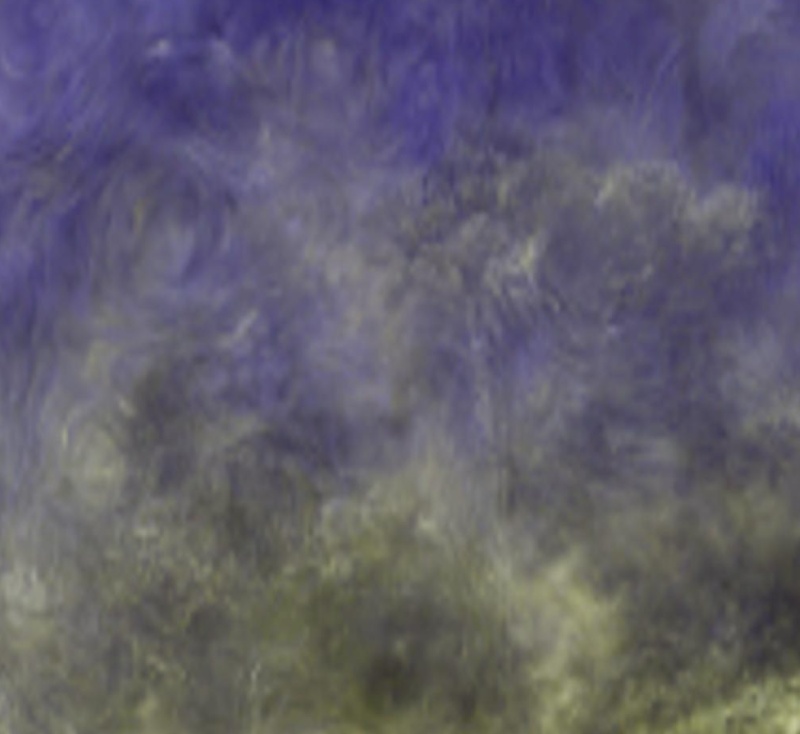}\\
 \includegraphics[width=\linewidth, height=\myH]{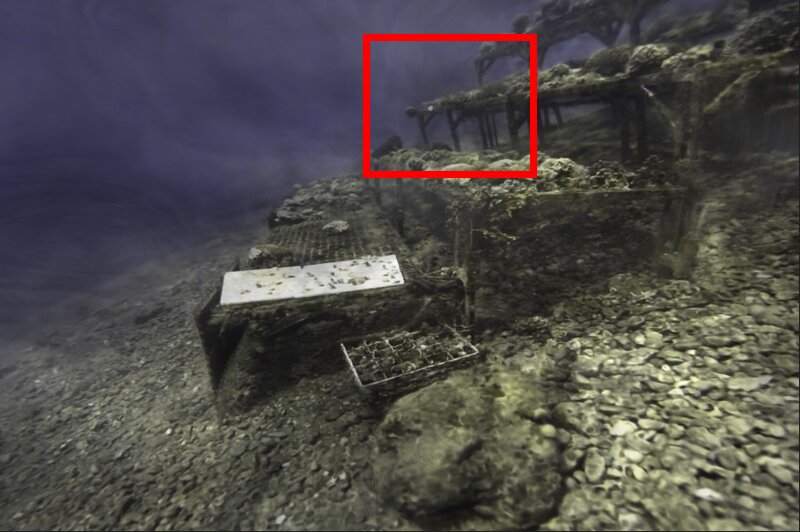}\\
 \includegraphics[width=\linewidth, height=\myH]{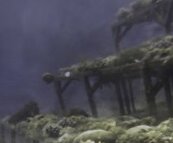}\\
 \includegraphics[width=\linewidth, height=\myH]{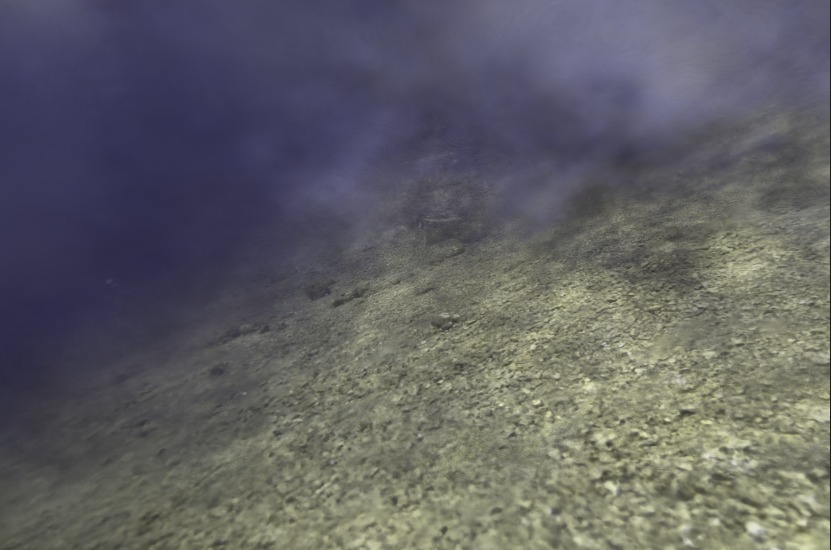}}
 \subcaptionbox{Ours}[\myW]
{\includegraphics[width=\linewidth, height=\myH]{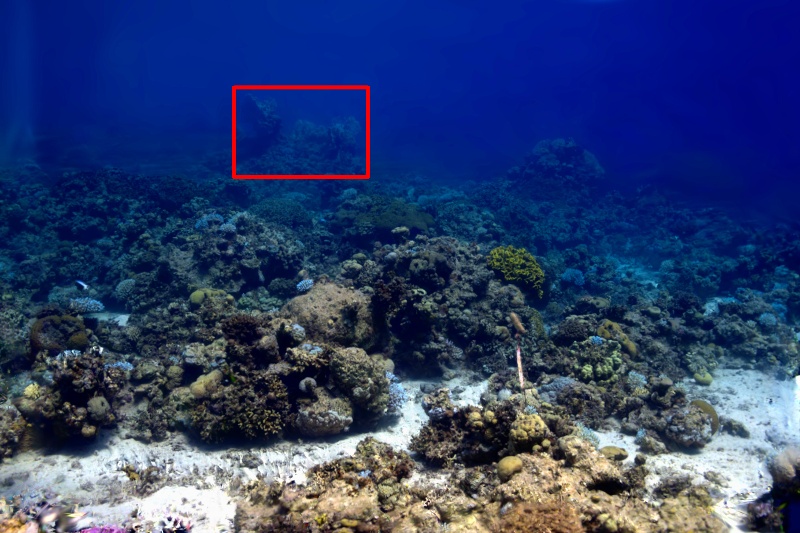}\\
 \includegraphics[width=\linewidth, height=\myH]{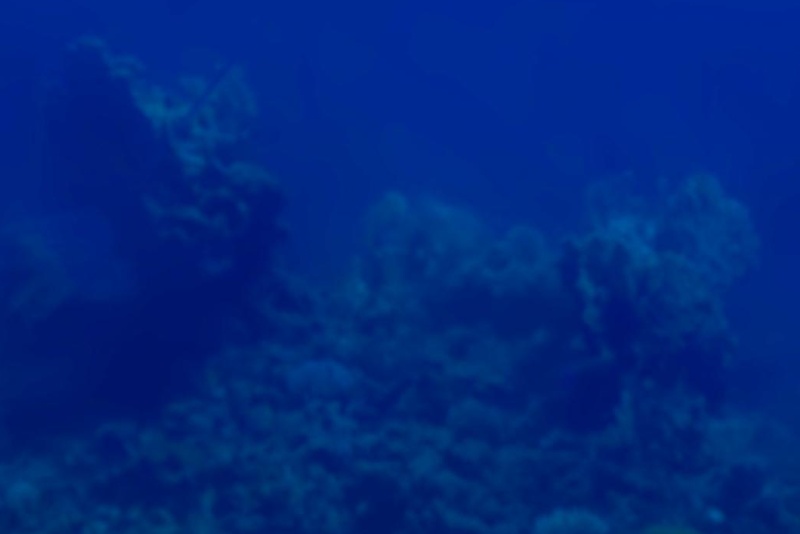}\\
 \includegraphics[width=\linewidth, height=\myH]{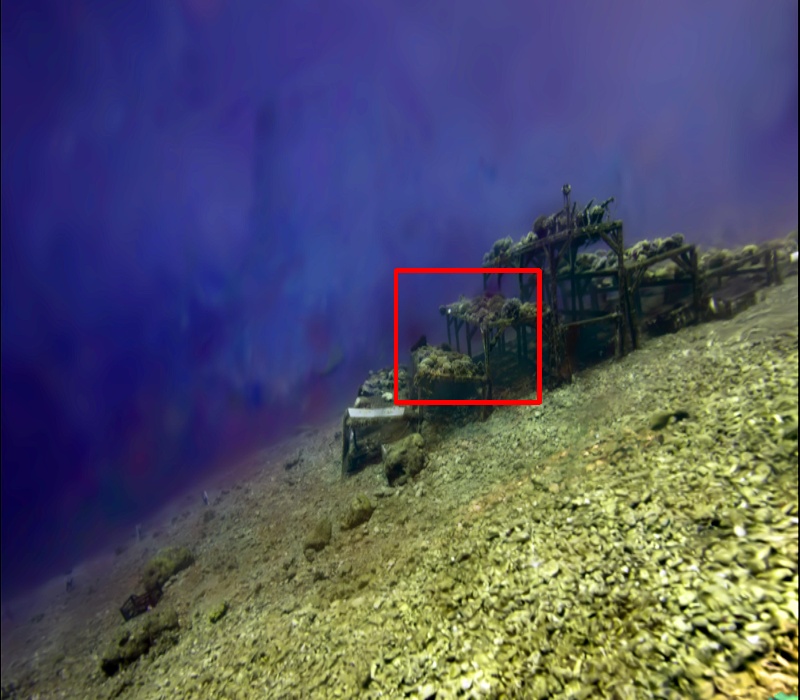}\\
 \includegraphics[width=\linewidth, height=\myH]{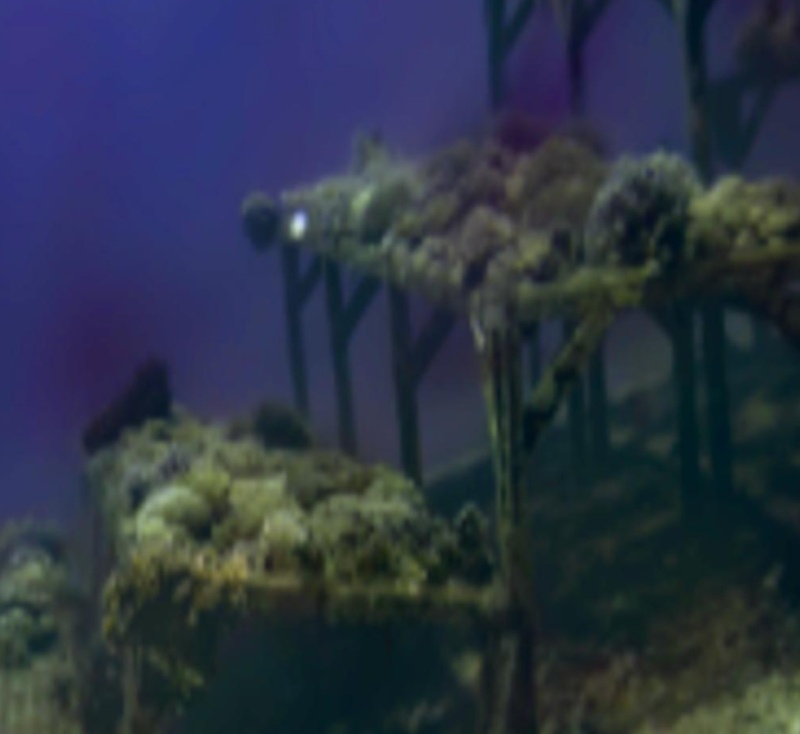}\\
 \includegraphics[width=\linewidth, height=\myH]{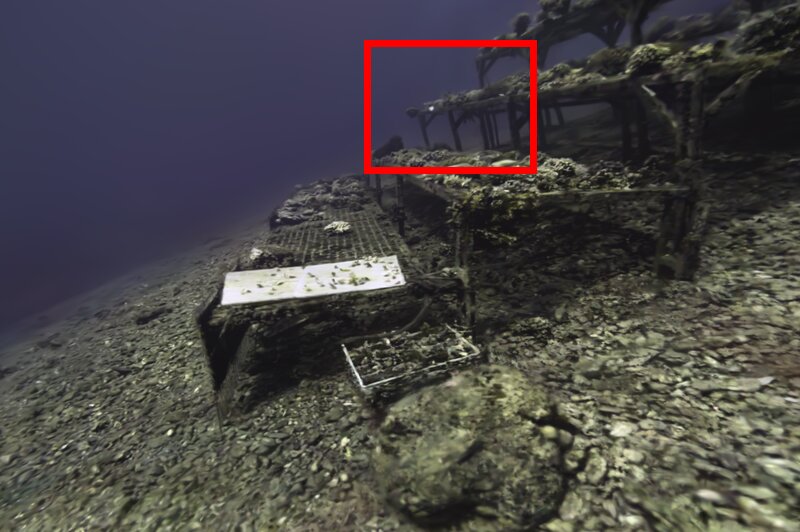}\\
 \includegraphics[width=\linewidth, height=\myH]{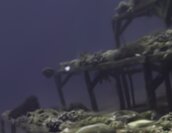}\\
 \includegraphics[width=\linewidth, height=\myH]{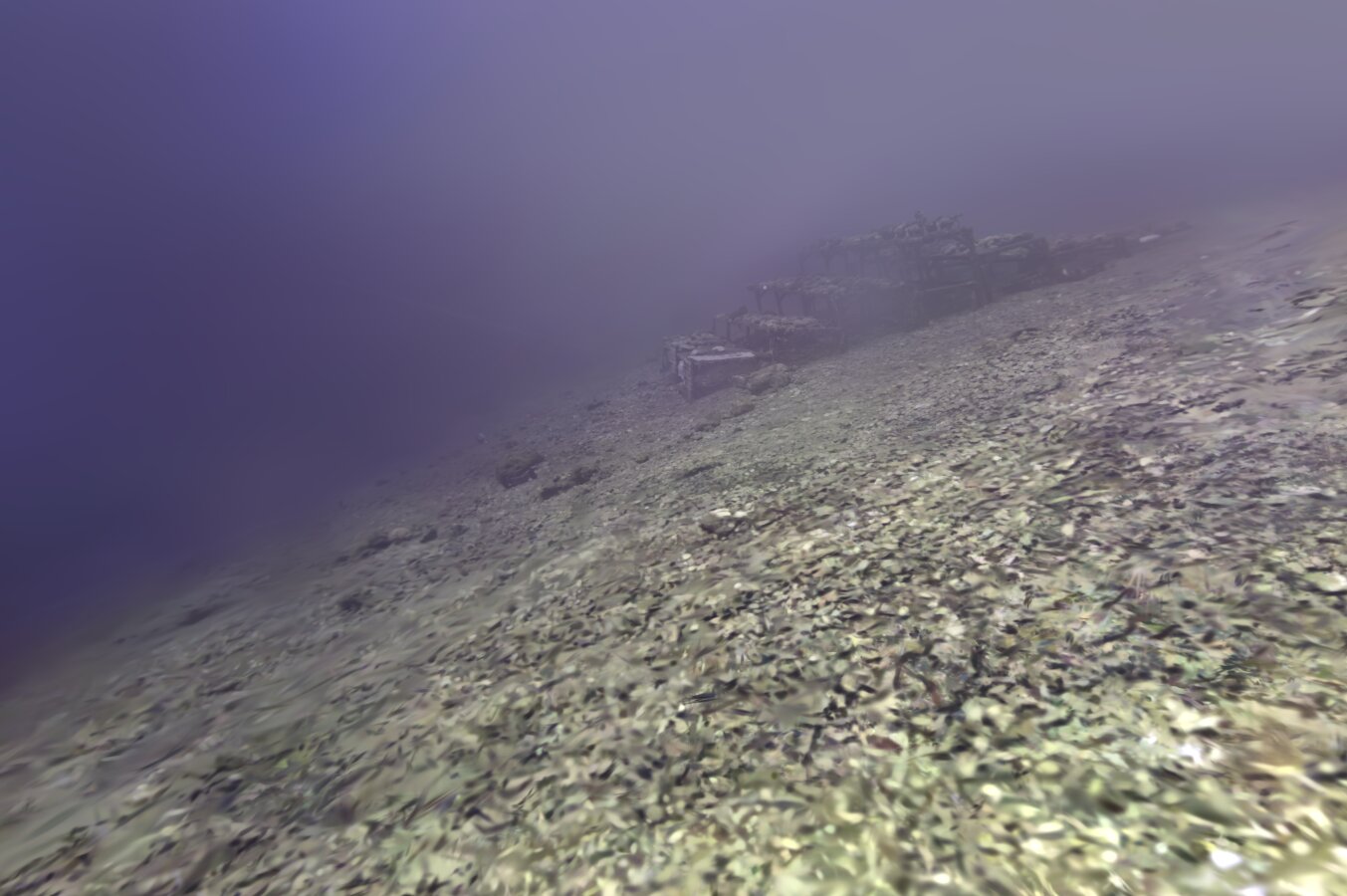}}

\caption[Visual Comparison of Novel Views]{Visual Comparison of novel views.  Row 1: an example from the Red-Sea scene. Row 3/5/7: examples from the TableDB dataset. 
Rows 2/4/6:  zoom in on the red rectangles 
in Rows 1/3/5. Note that Row 7 highlights STNeRFacto's failure on unbounded scenes.}
\label{fig:visual_comparison}
\end{figure*}

%% file: figures/experiments_results/fig_results2.tex
\graphicspath{ {./figures/experiments_results/figs2} }
\begin{figure*}[!ht]
\newcommand{\myW}{0.3\linewidth}
\newcommand{\myH}{3.2cm} 
\centering
\subcaptionbox{WS ~\cite{li2024watersplatting}}[\myW]
{\includegraphics[width=\linewidth, height=\myH]{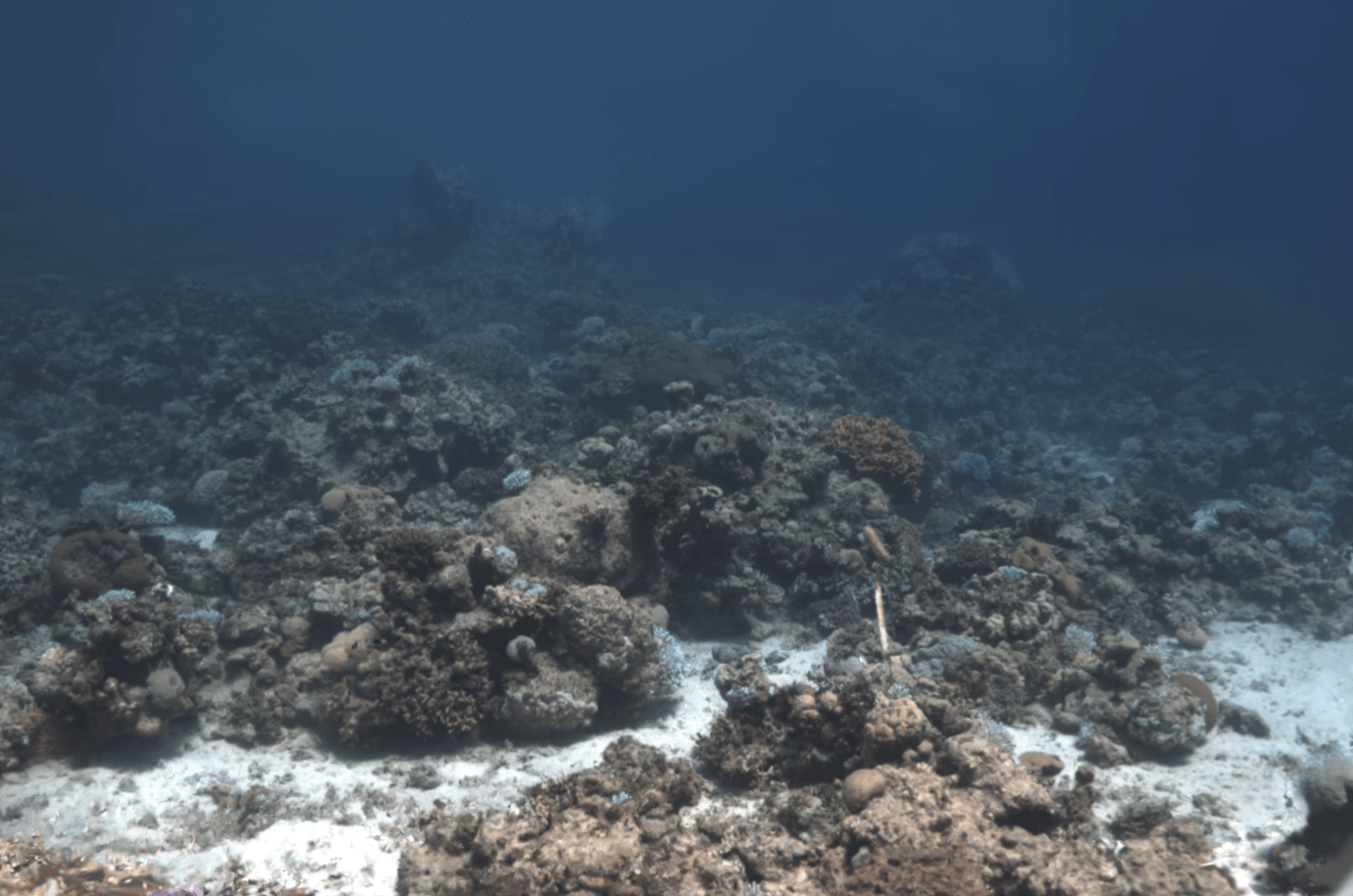}\\
 \includegraphics[width=\linewidth, height=\myH]{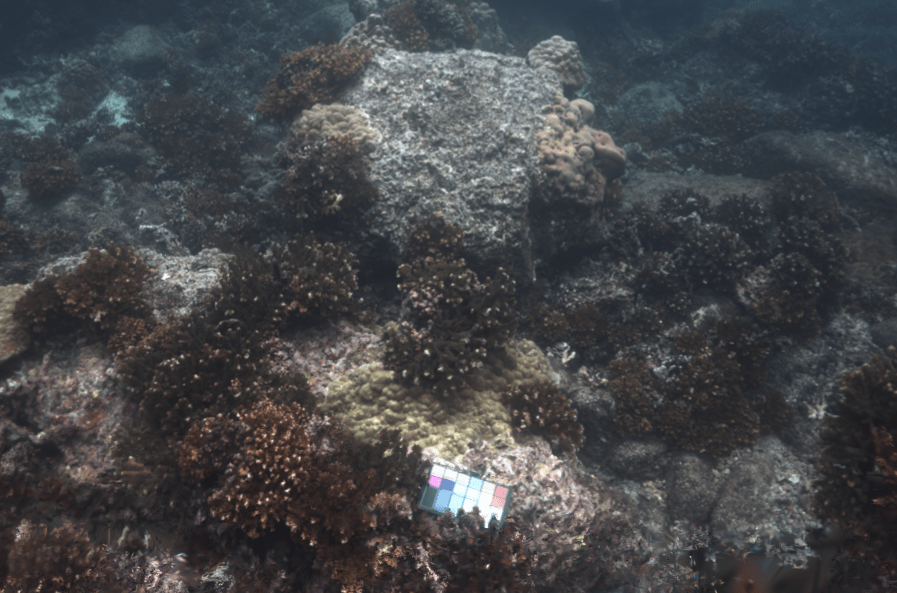}\\
 \includegraphics[width=\linewidth, height=\myH]{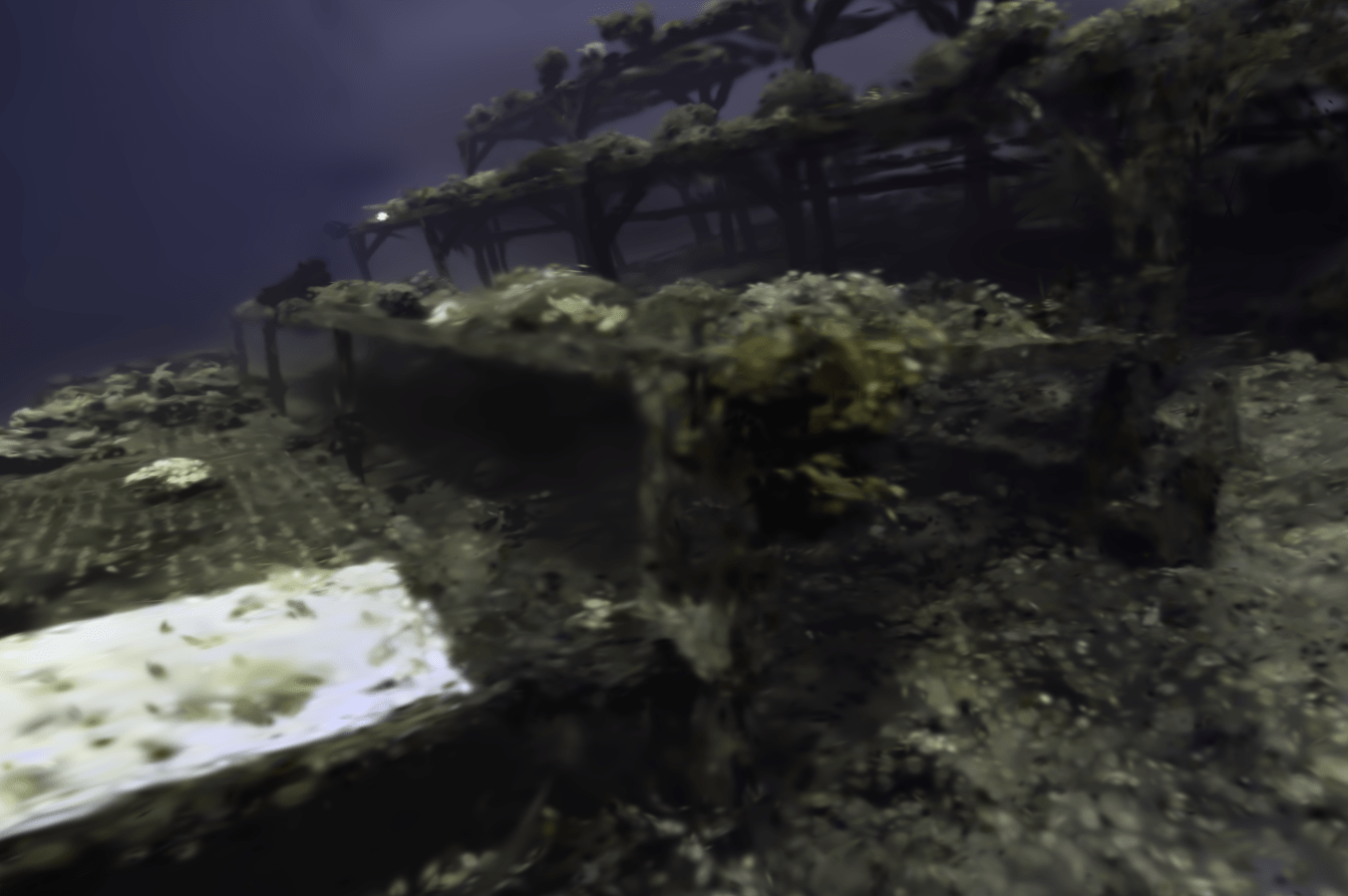}\\
 \includegraphics[width=\linewidth, height=\myH]{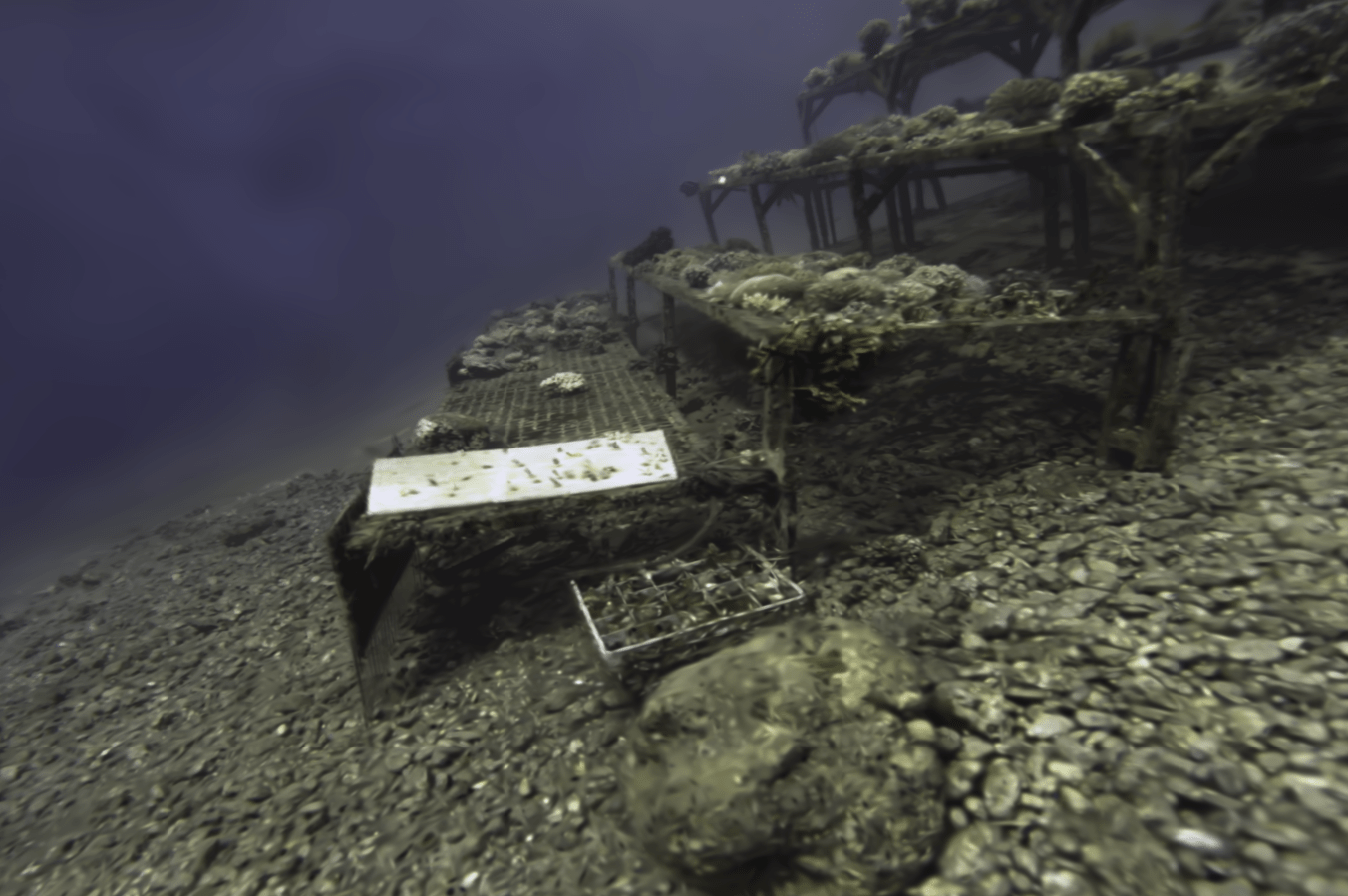}\\
 \includegraphics[width=\linewidth, height=\myH]{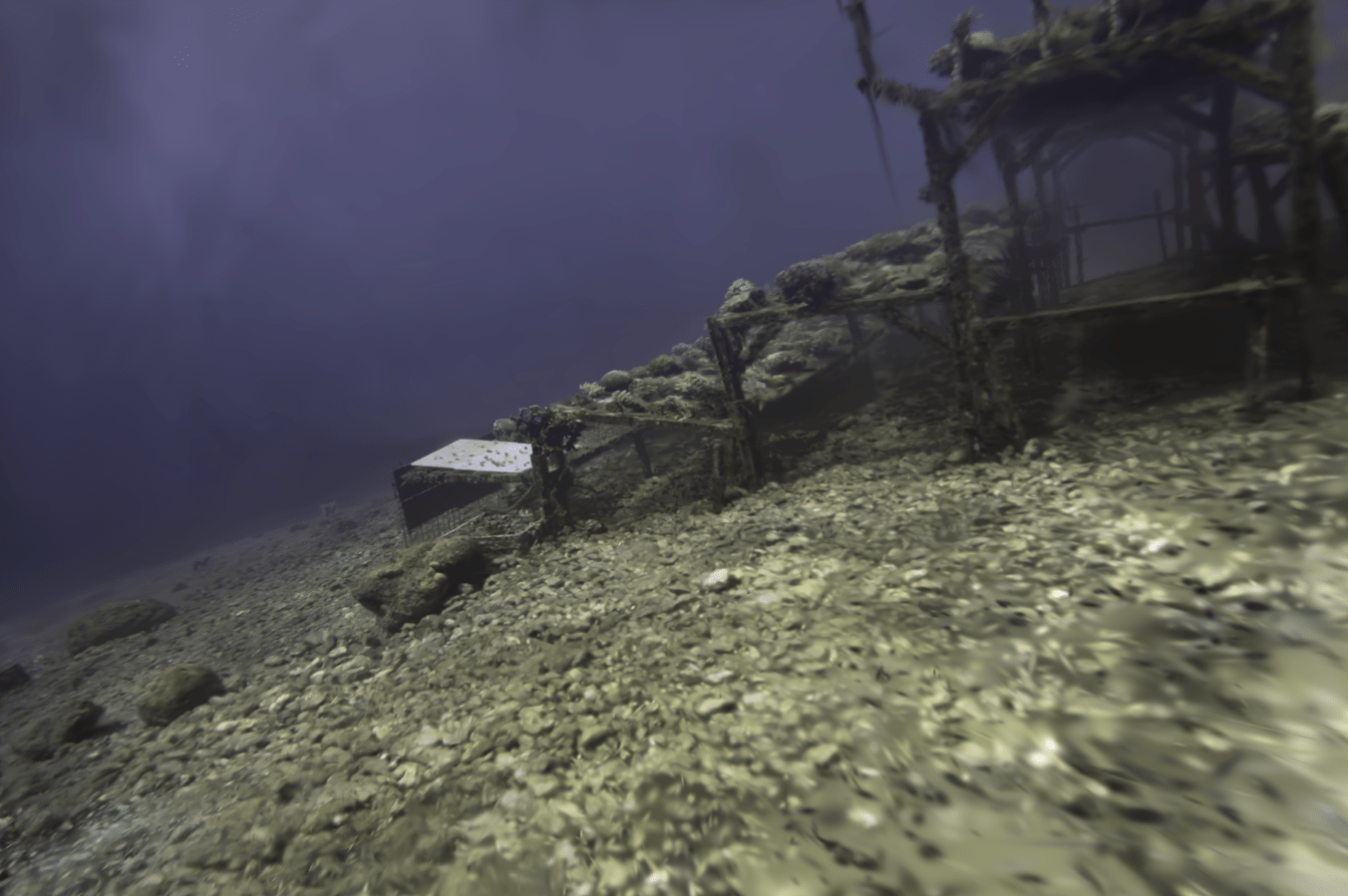}\\
 \includegraphics[width=\linewidth, height=\myH]{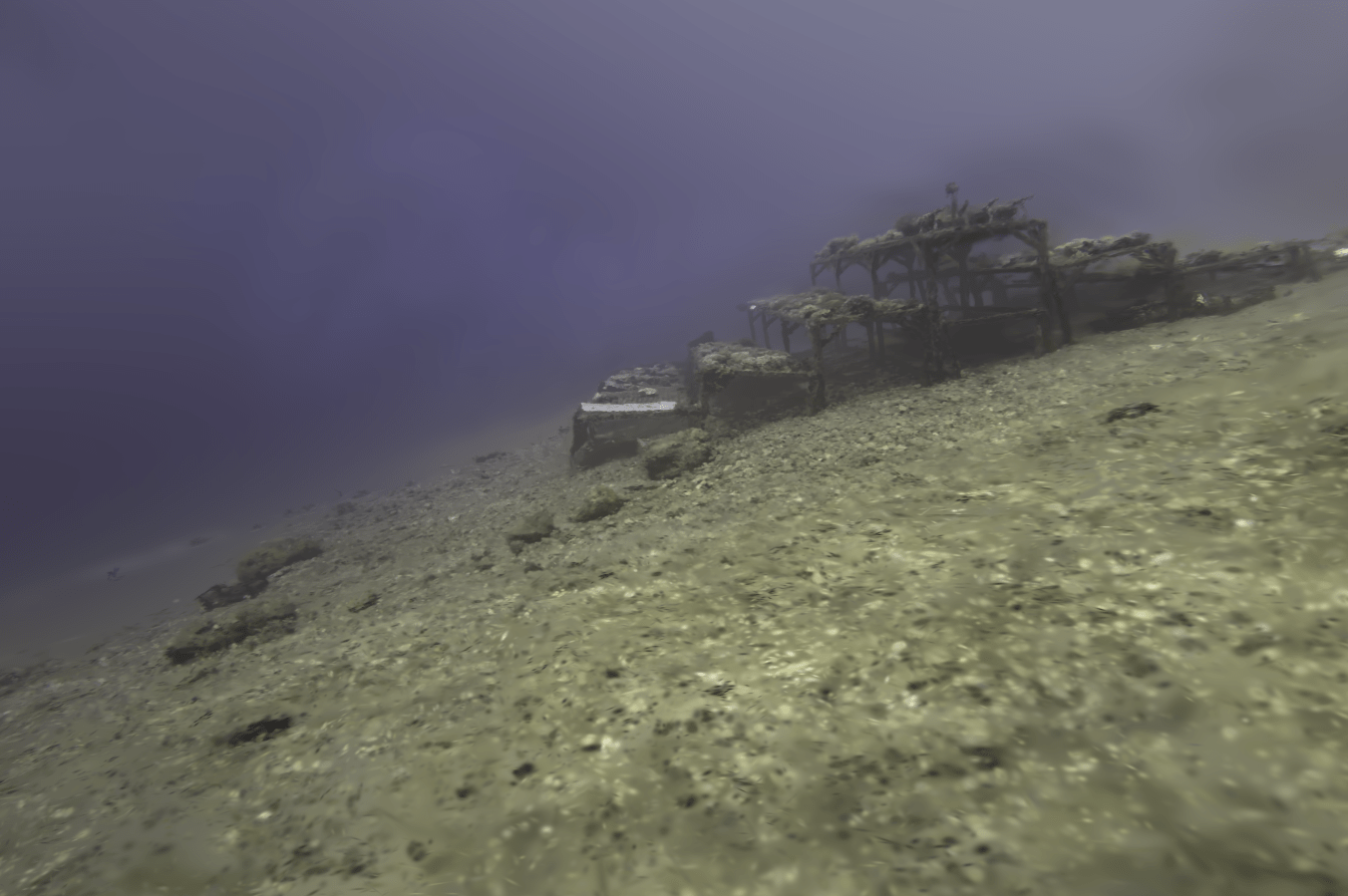}}
\subcaptionbox{SeaSplat ~\cite{yang2024seasplat}}[\myW]
{\includegraphics[width=\linewidth, height=\myH]{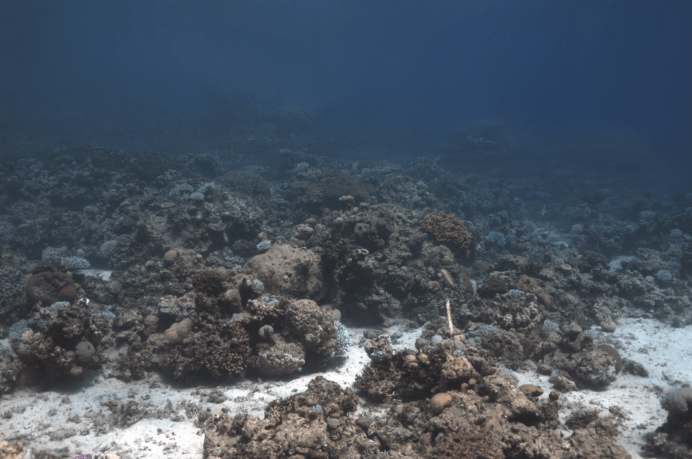}\\
 \includegraphics[width=\linewidth, height=\myH]{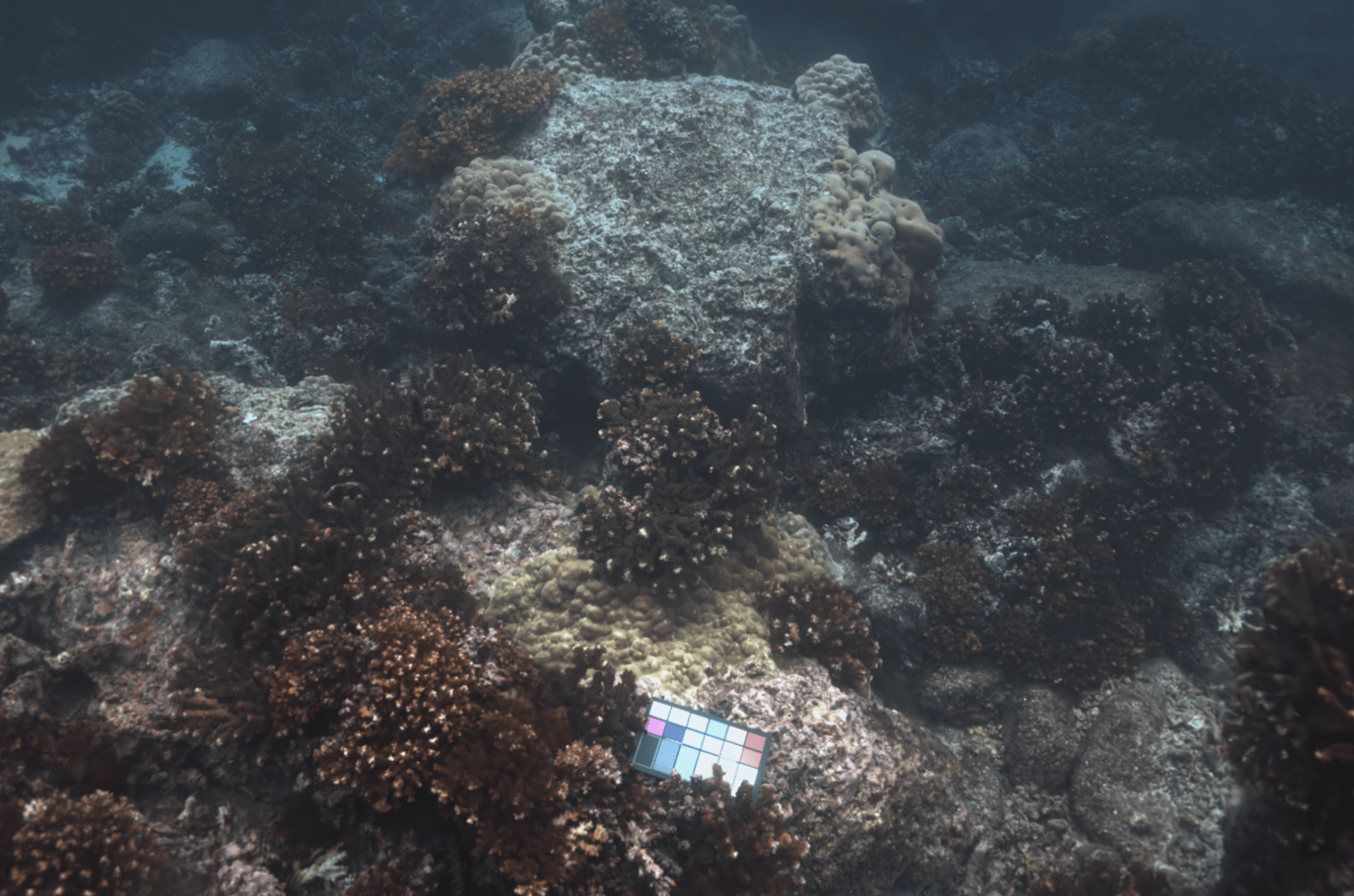}\\
 \includegraphics[width=\linewidth, height=\myH]{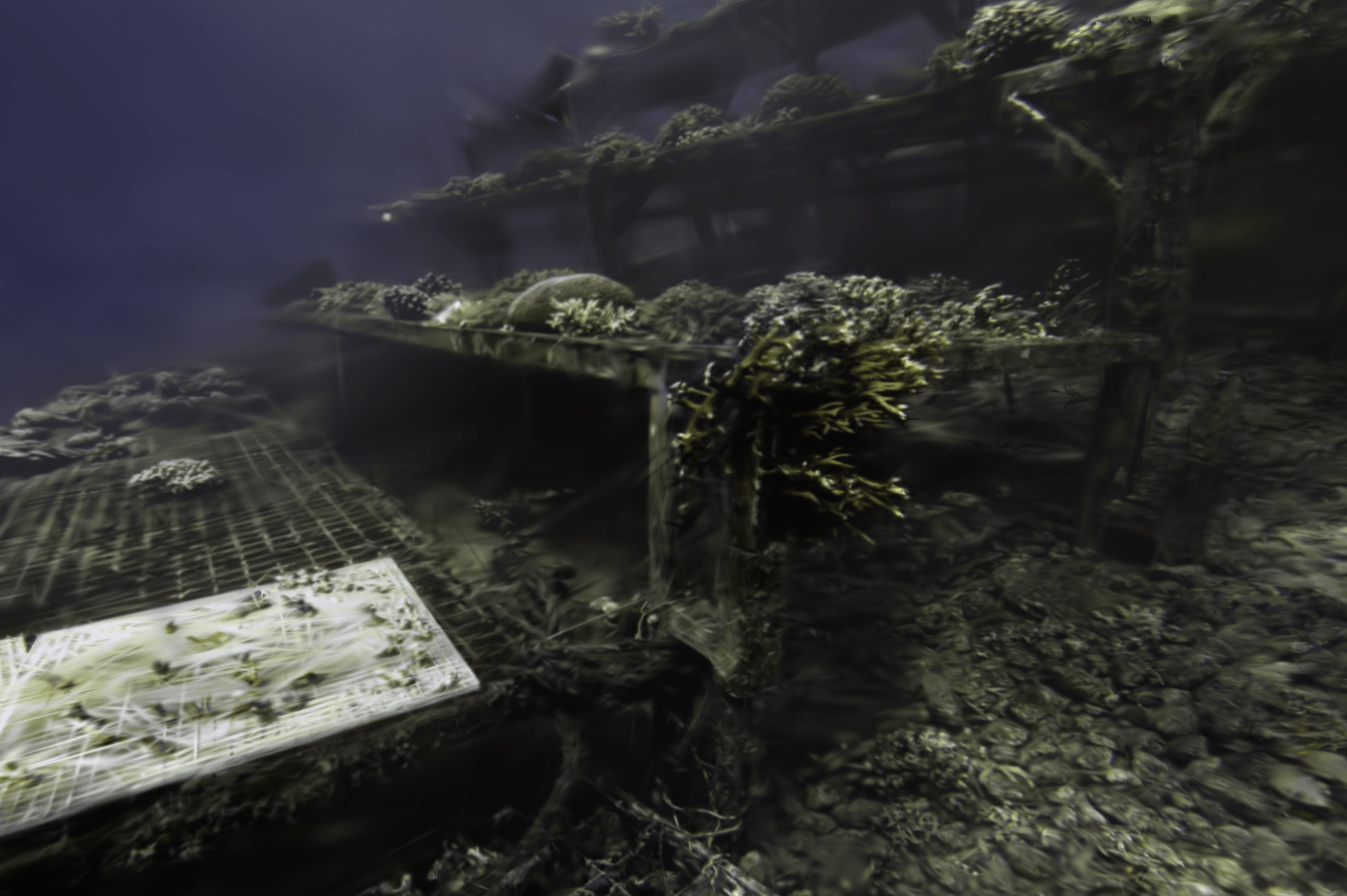}\\
 \includegraphics[width=\linewidth, height=\myH]{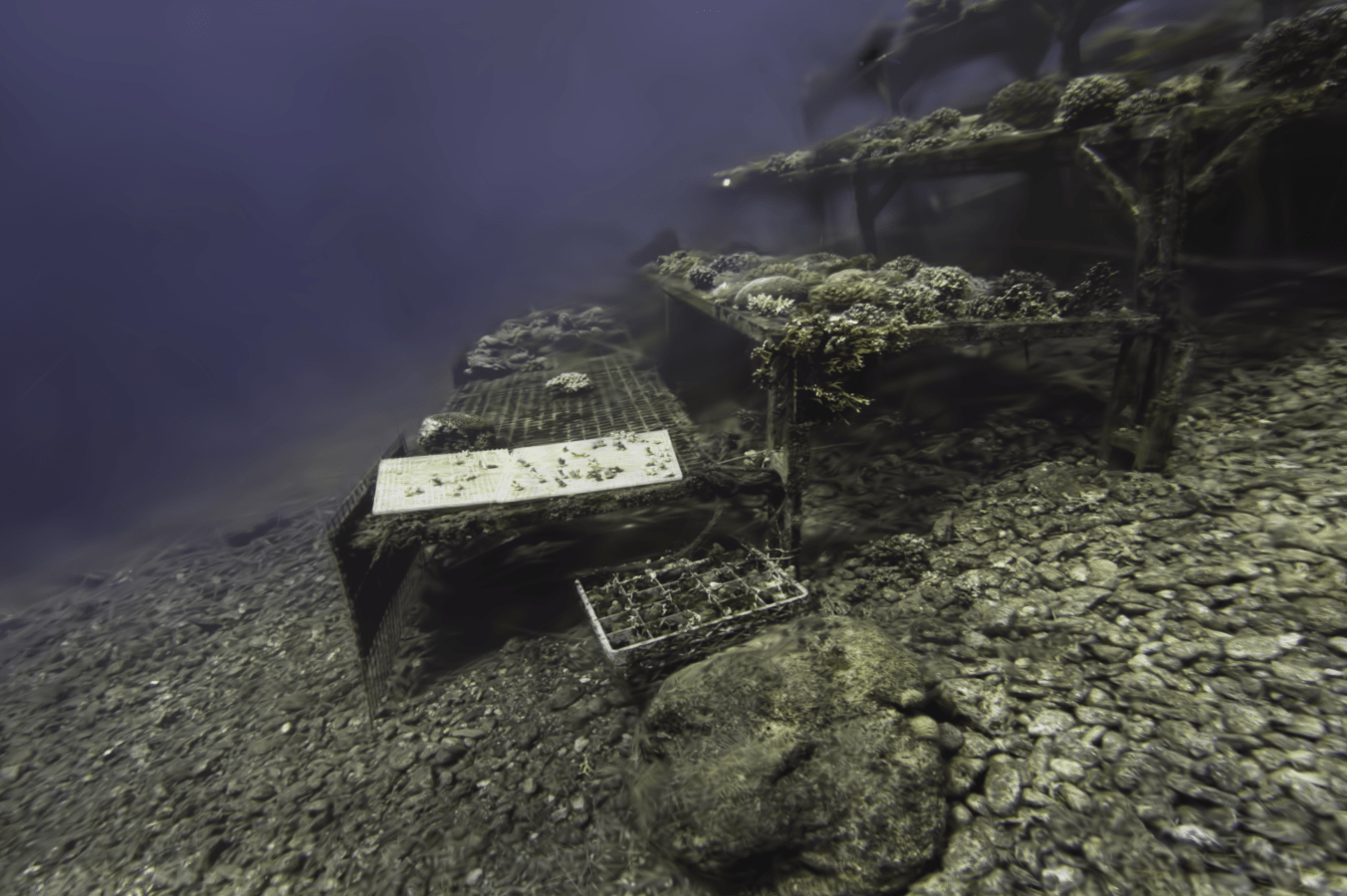}\\
 \includegraphics[width=\linewidth, height=\myH]{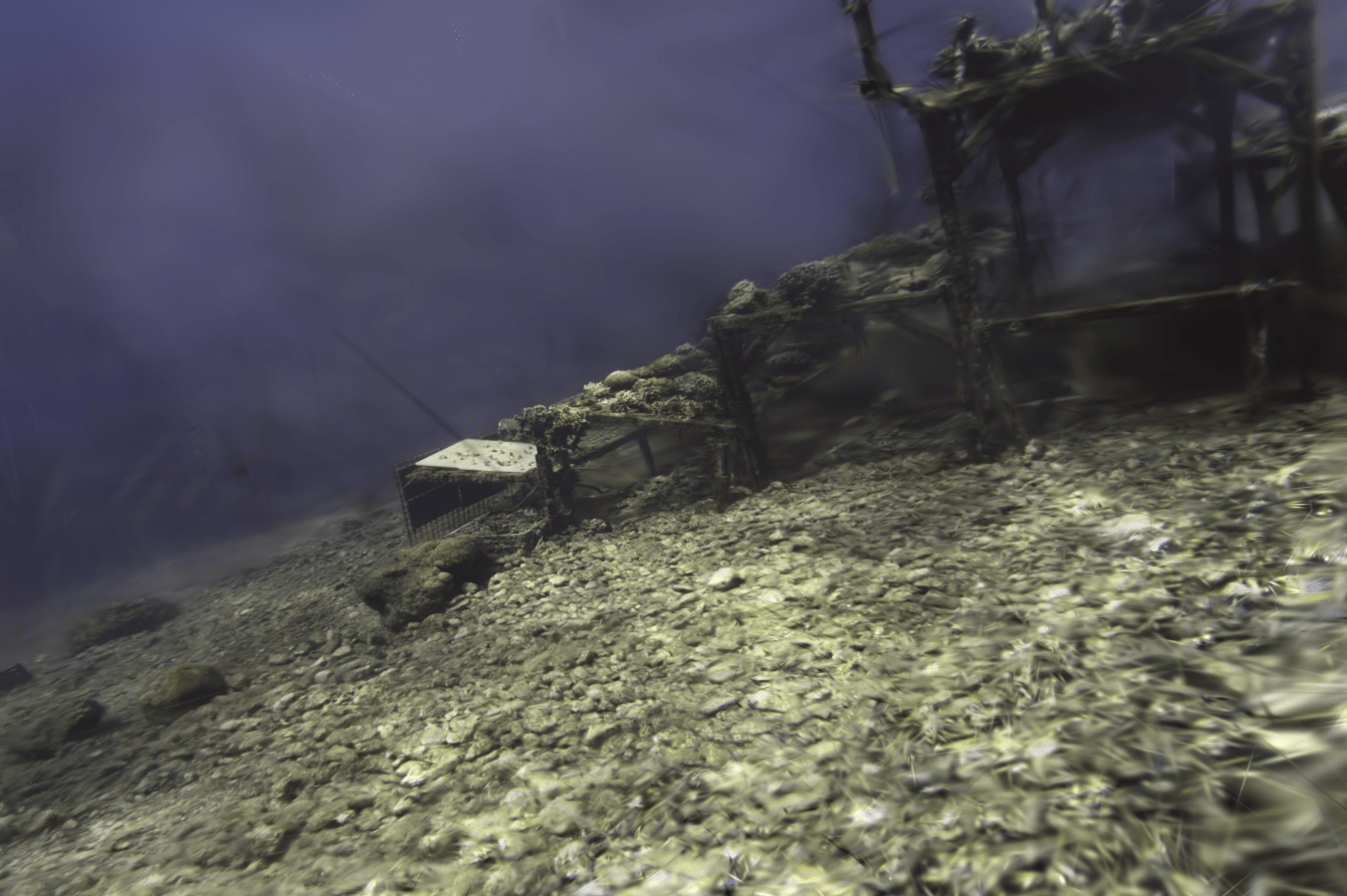}\\
 \includegraphics[width=\linewidth, height=\myH]{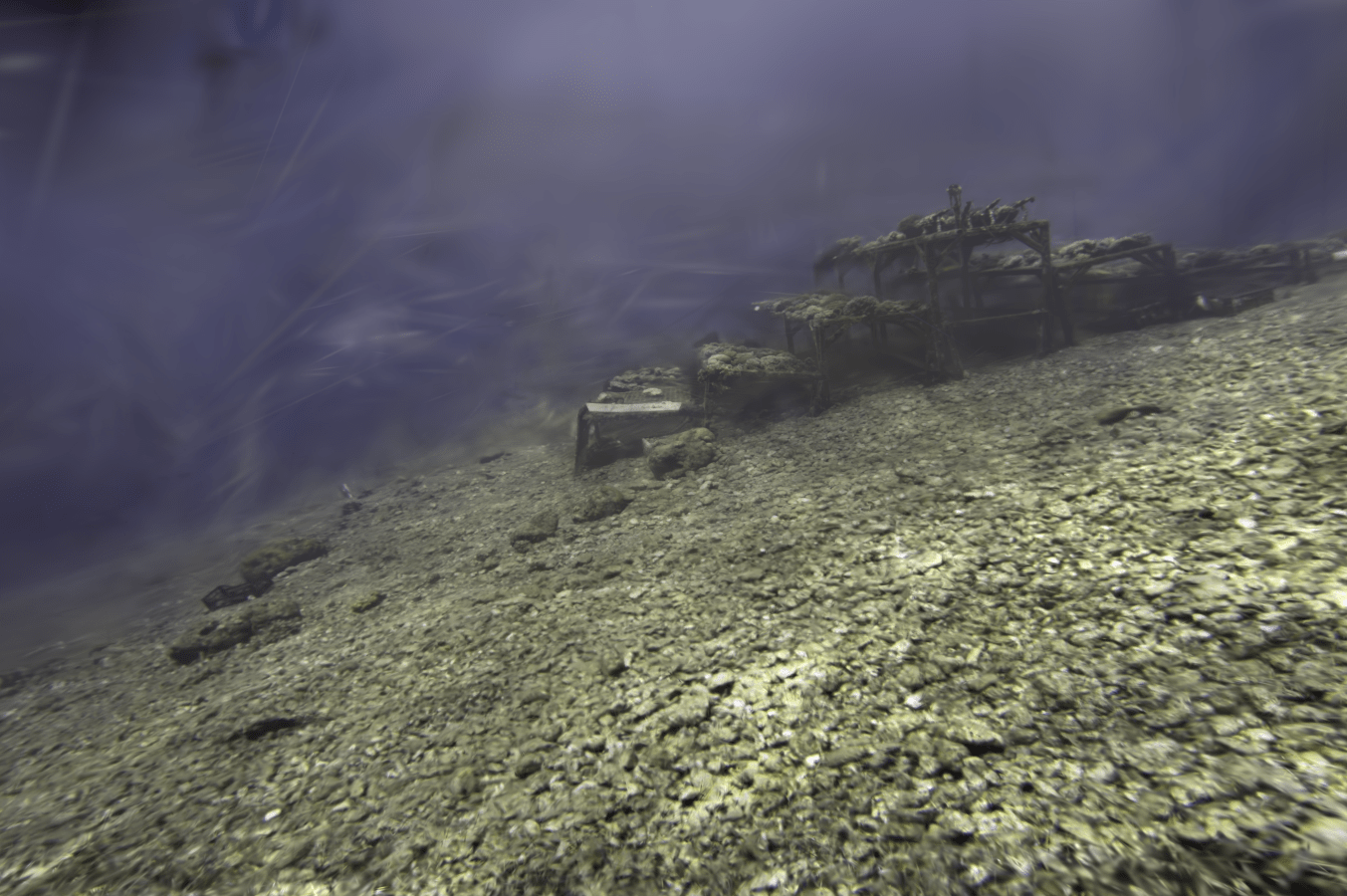}}
 \subcaptionbox{Ours}[\myW]
{\includegraphics[width=\linewidth, height=\myH]{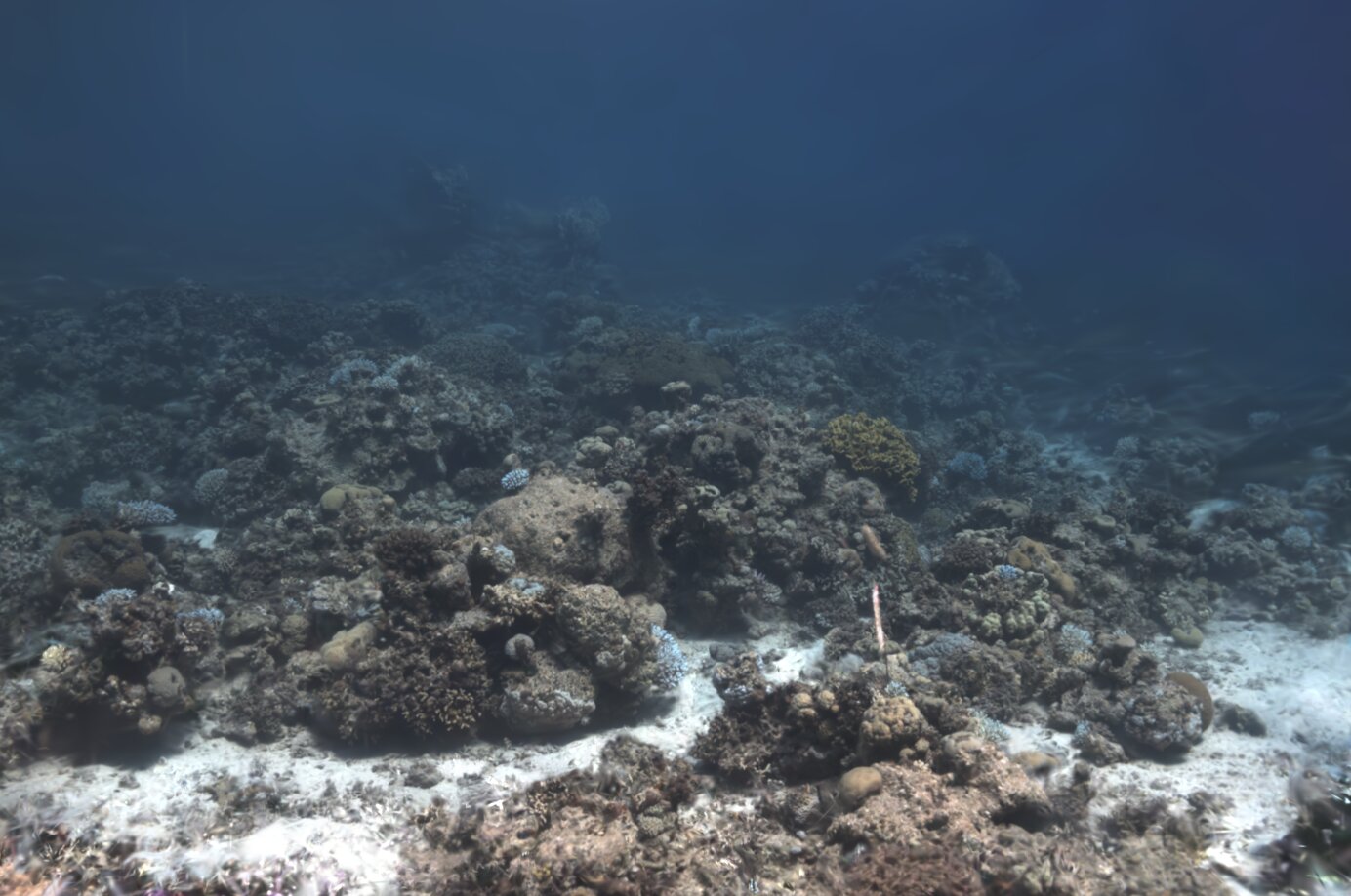}\\
 \includegraphics[width=\linewidth, height=\myH]{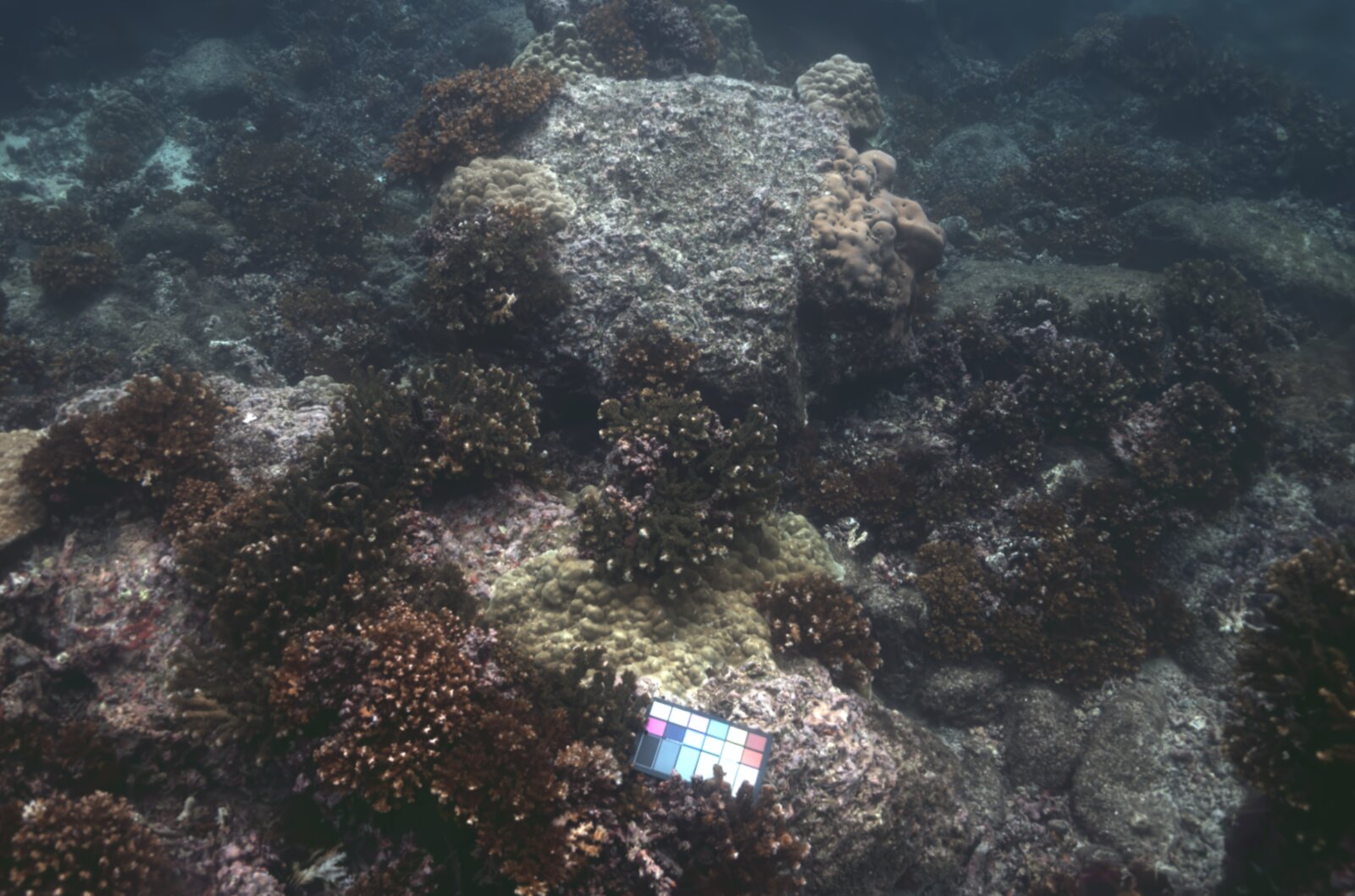}\\
 \includegraphics[width=\linewidth, height=\myH]{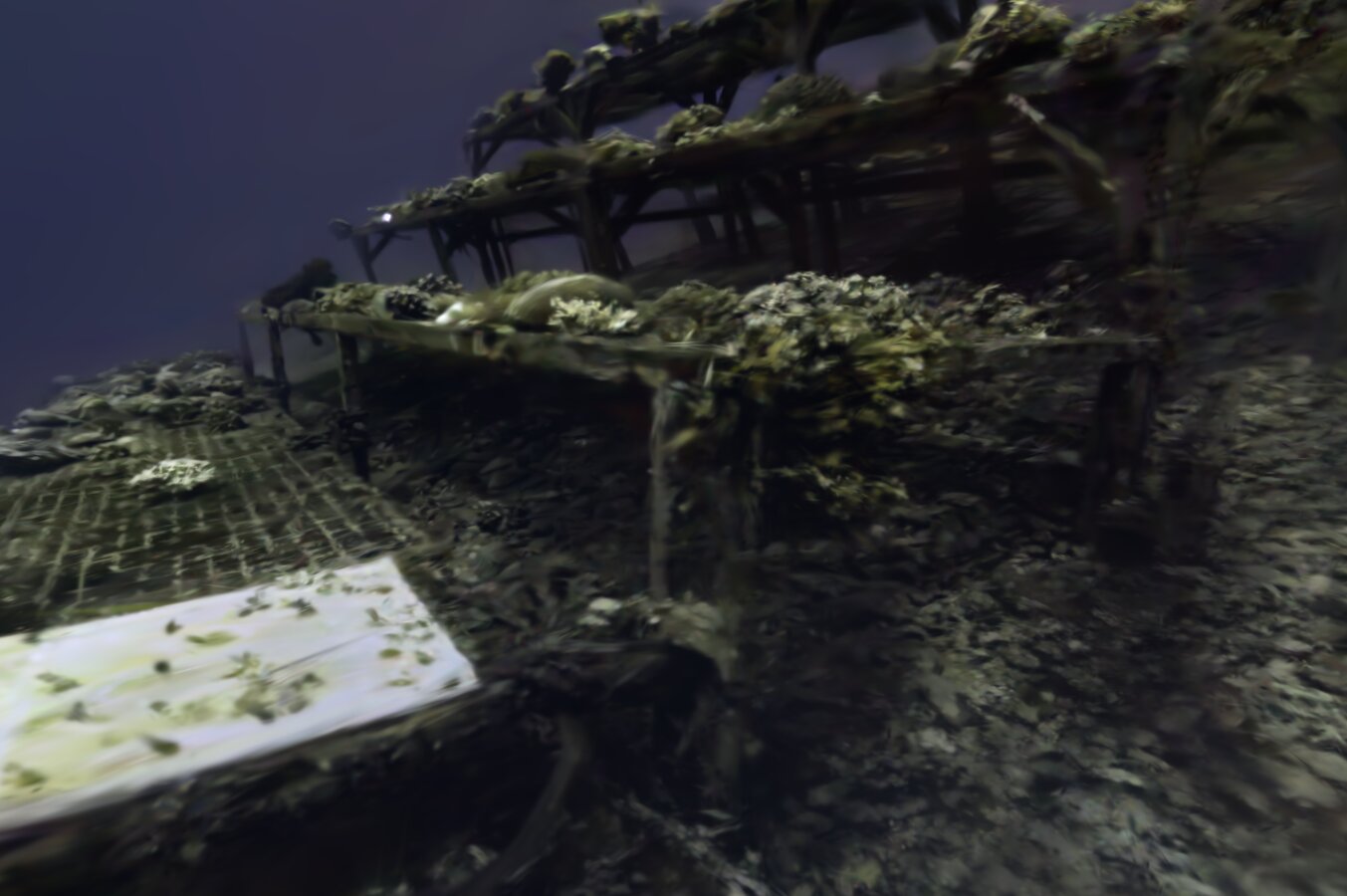}\\
 \includegraphics[width=\linewidth, height=\myH]{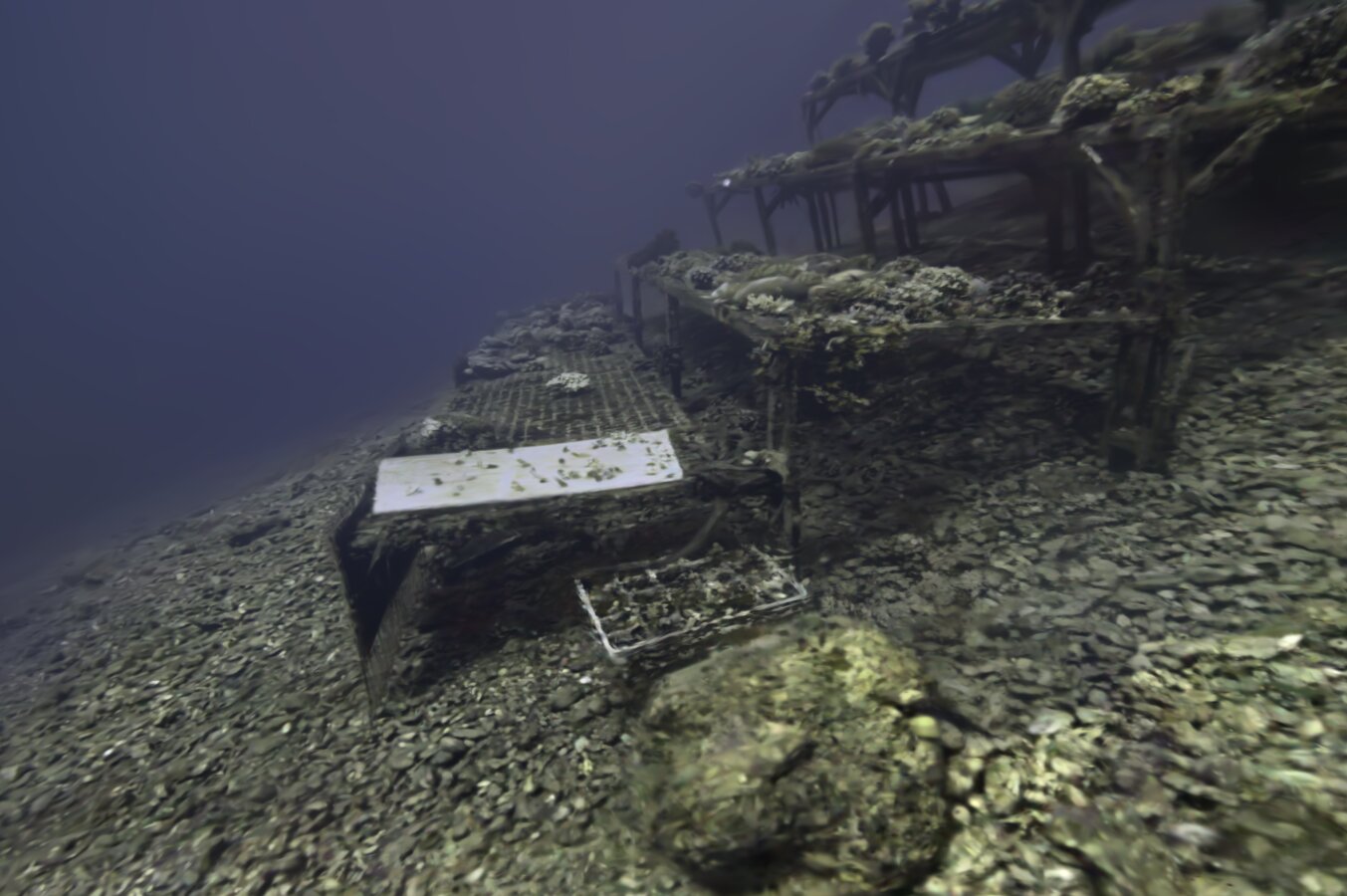}\\
 \includegraphics[width=\linewidth, height=\myH]{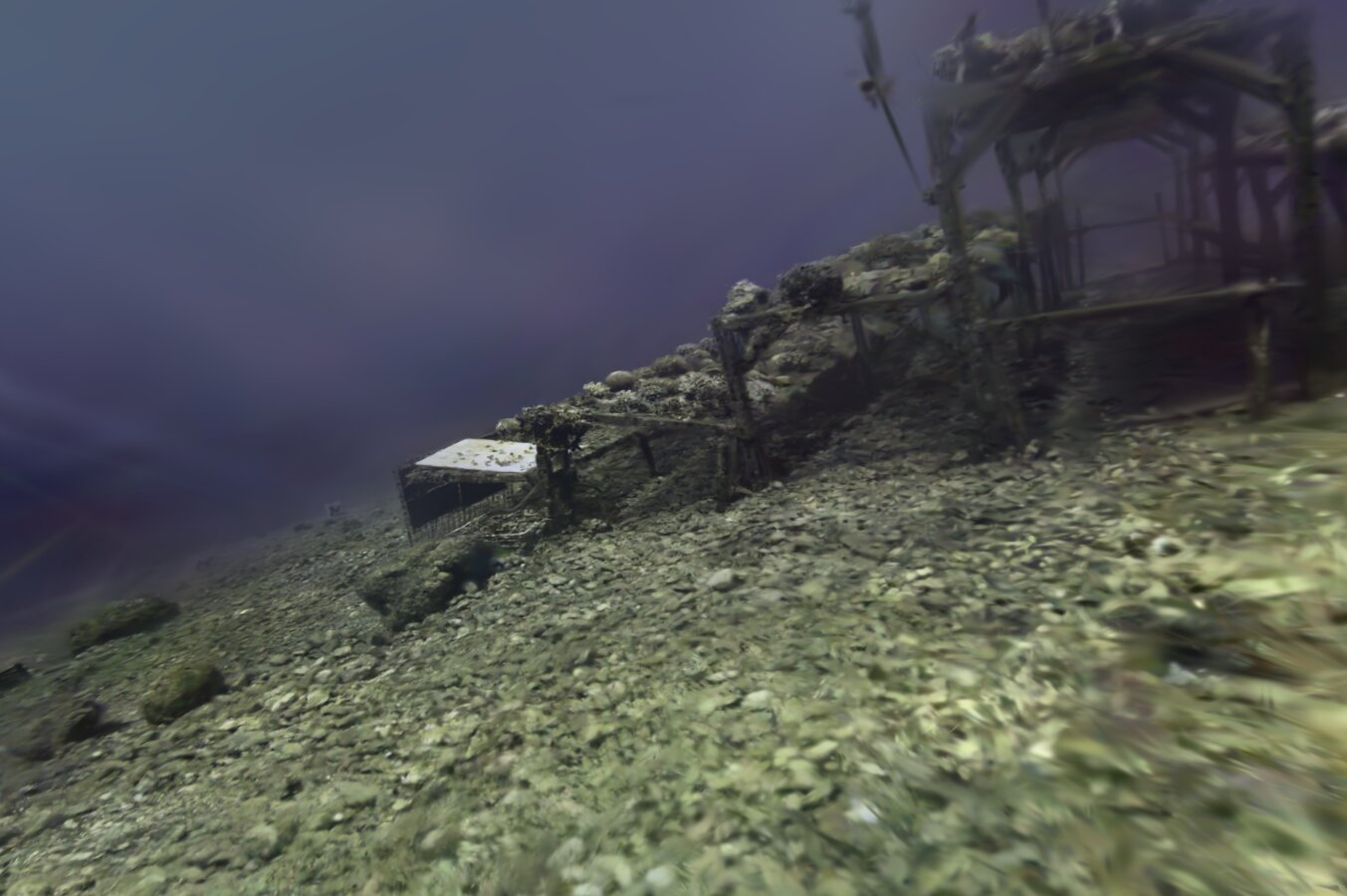}\\
 \includegraphics[width=\linewidth, height=\myH]{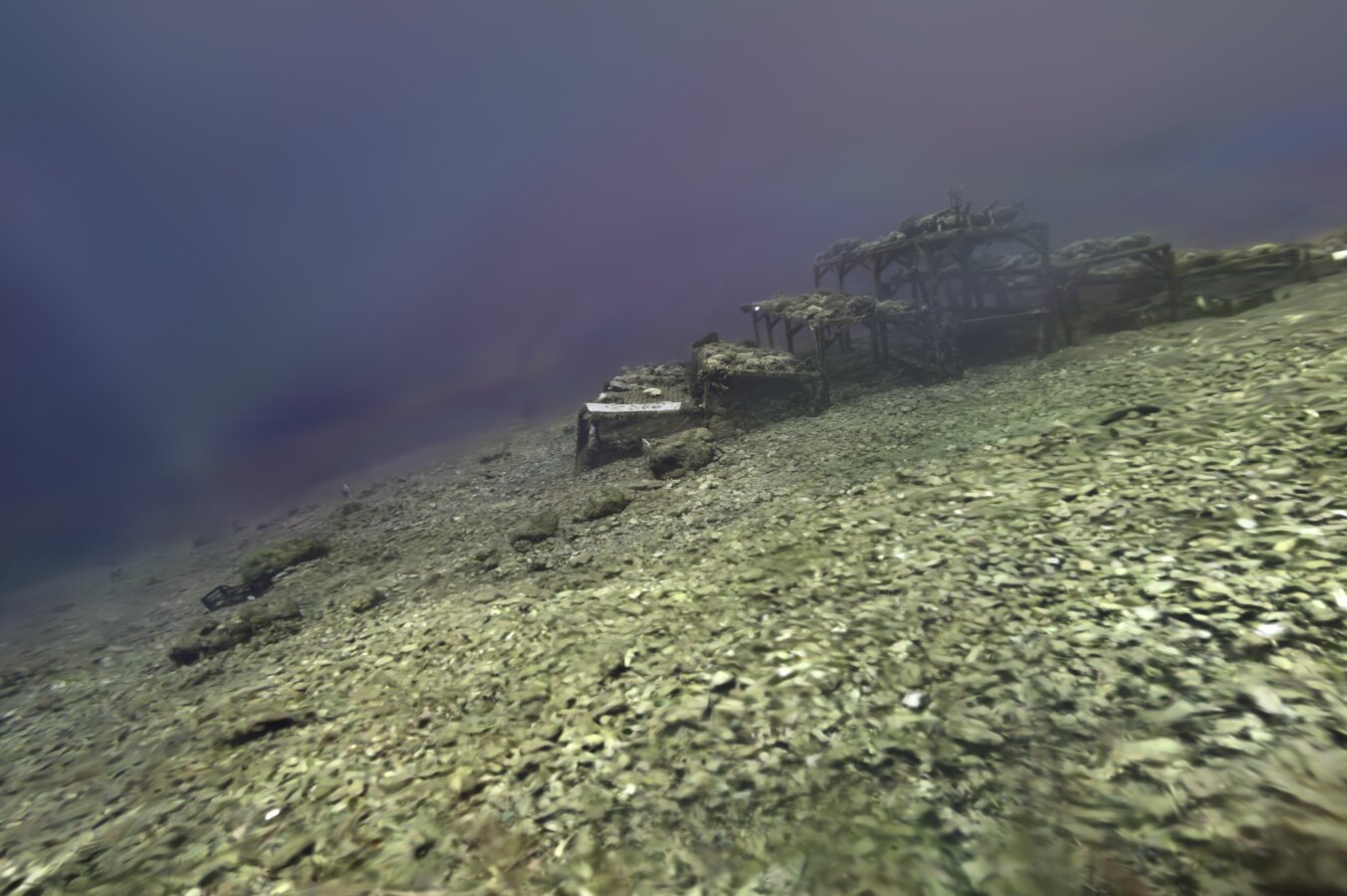}}
\caption[Visual comparison of novel views across underwater splatting variants]{Visual comparison of novel views generated by different underwater splatting methods. Evidently, our results are better than SeaSplat's, and are comparable 
to those of WS. That said, our method is much faster. Also, in regions far from the cameras,
our method typically yields sharper results than WS.  

}

\label{fig:results2}
\end{figure*}

%% file: figures/appendix/appendix_fig_examples3.tex
\graphicspath{ {./figures/appendix/figs} }
\begin{figure*}[!ht]
\newcommand{\myW}{0.24\linewidth}
\newcommand{\myH}{2.2cm} 
\centering
\subcaptionbox{Rendered (train)}[\myW]
{\includegraphics[width=\linewidth, height=\myH]{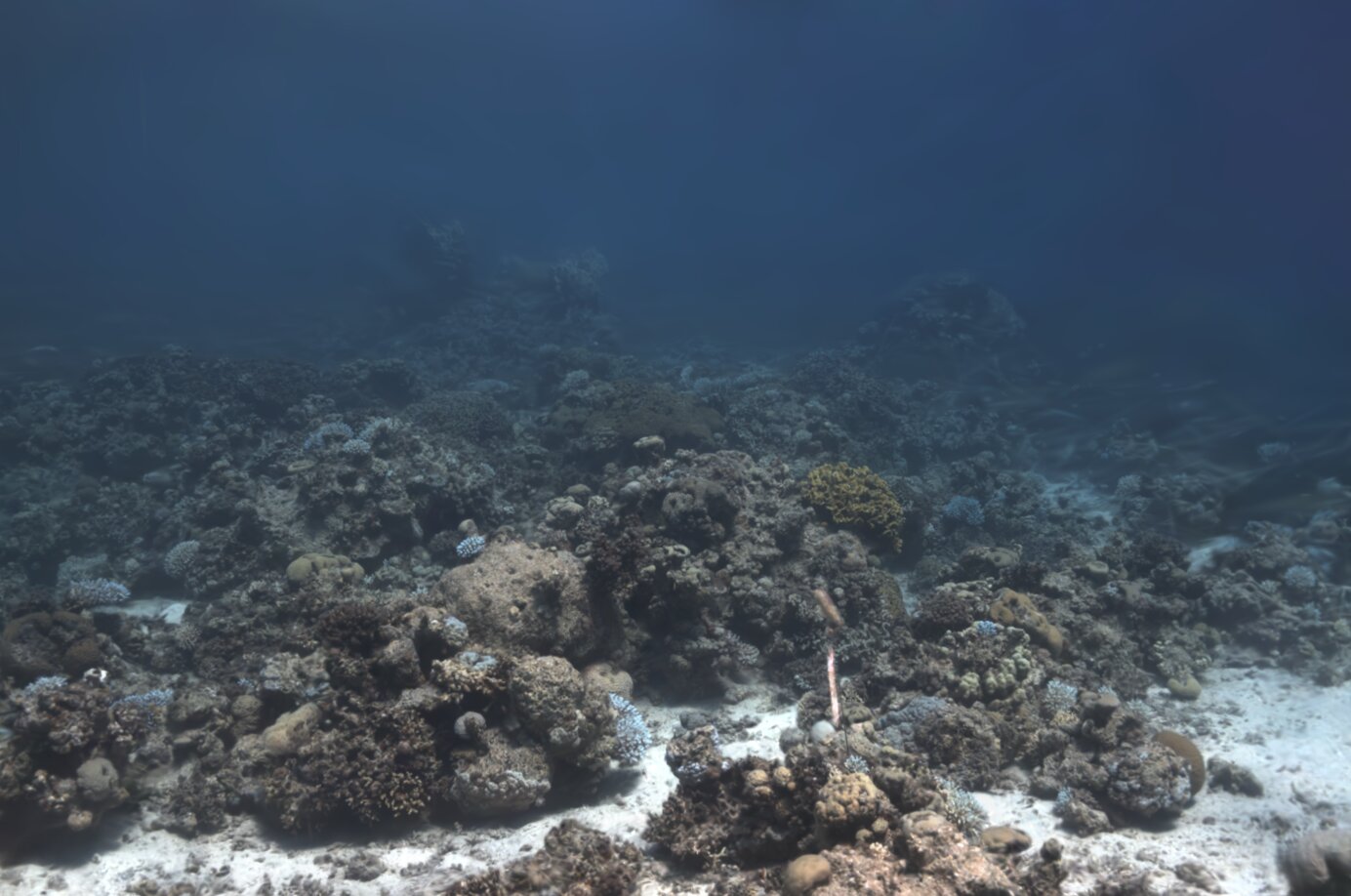}\\
 \includegraphics[width=\linewidth, height=\myH]{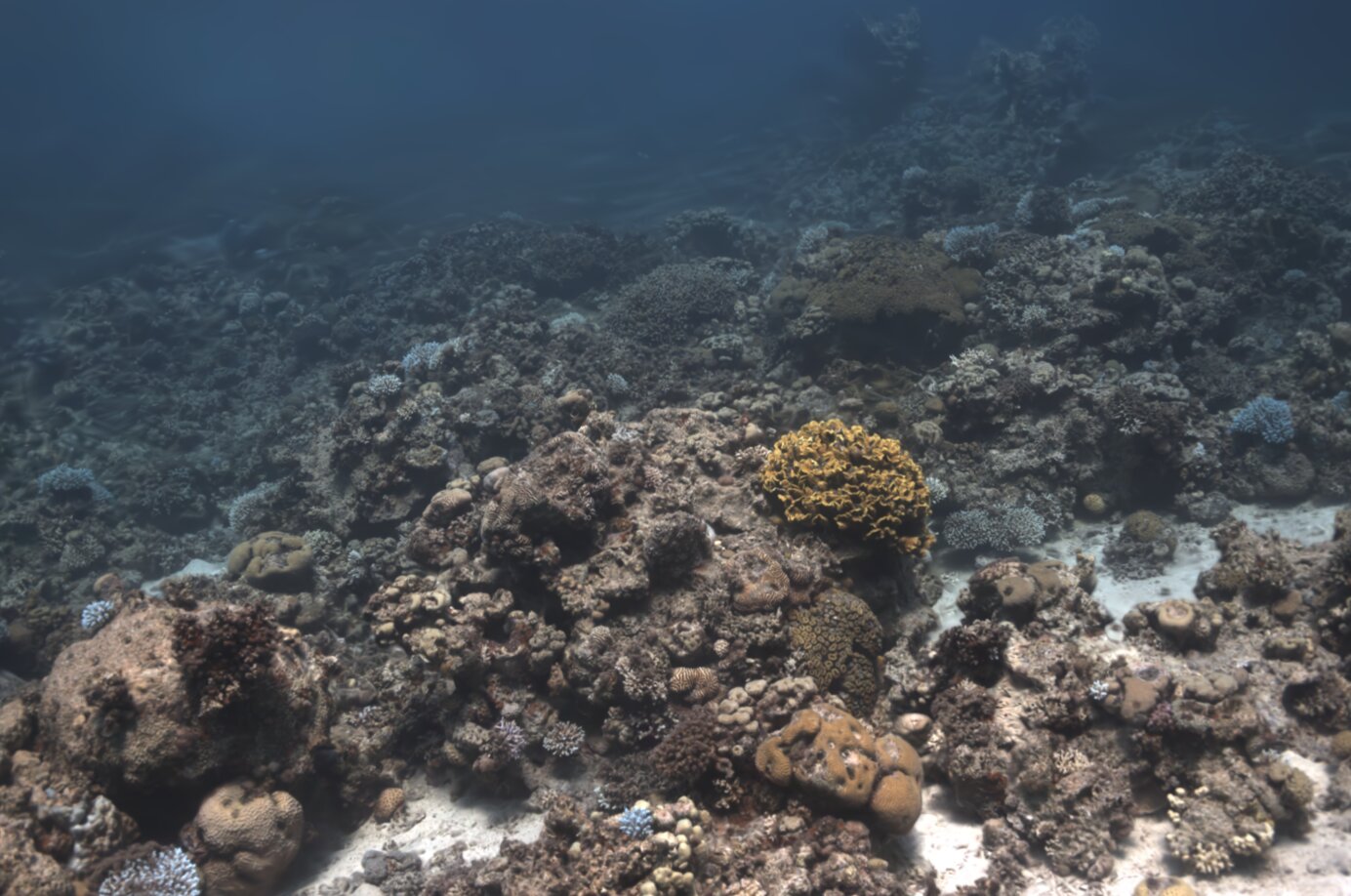}\\
 \includegraphics[width=\linewidth, height=\myH]{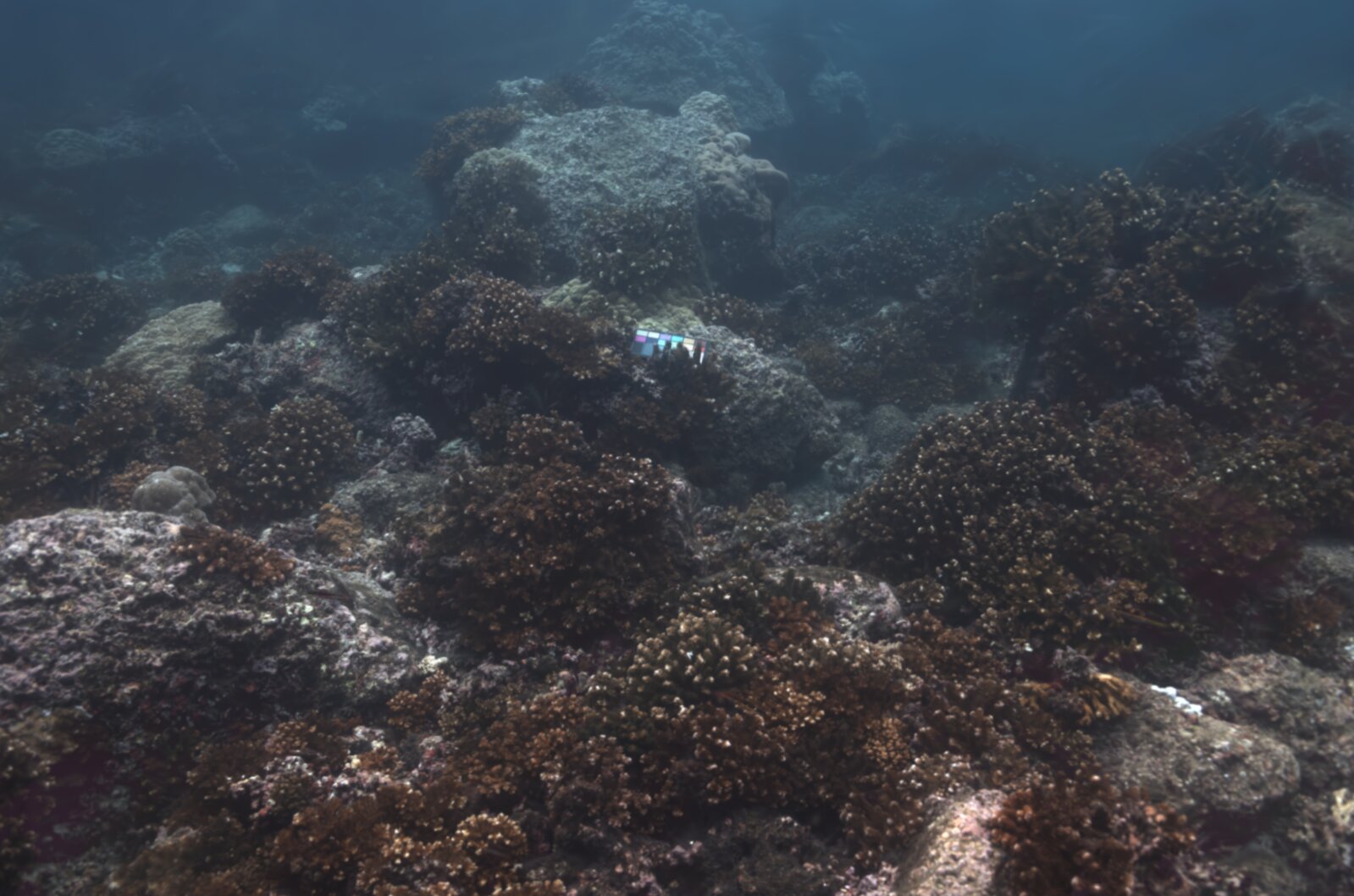}\\
 \includegraphics[width=\linewidth, height=\myH]{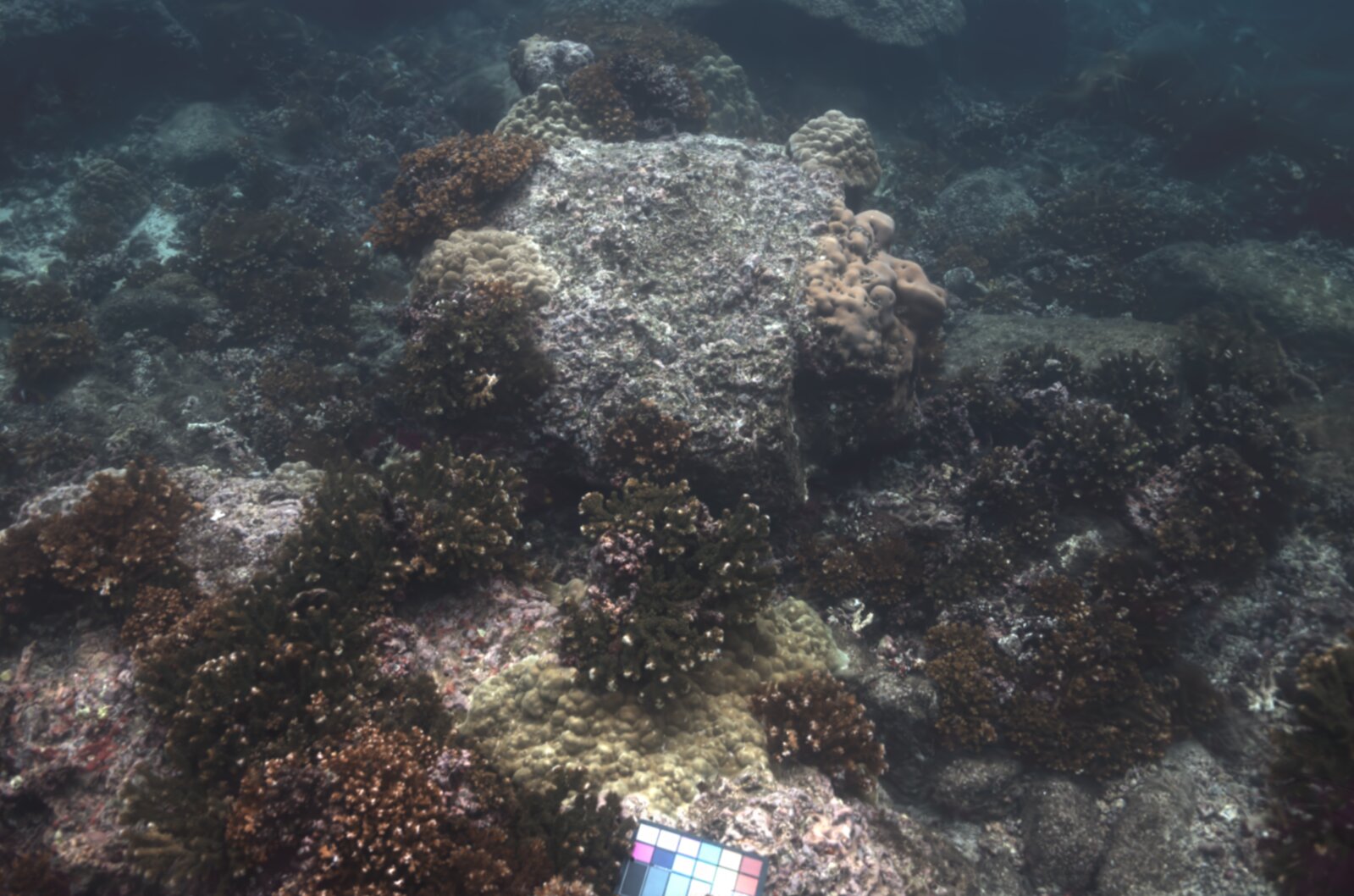}\\
 \includegraphics[width=\linewidth, height=\myH]{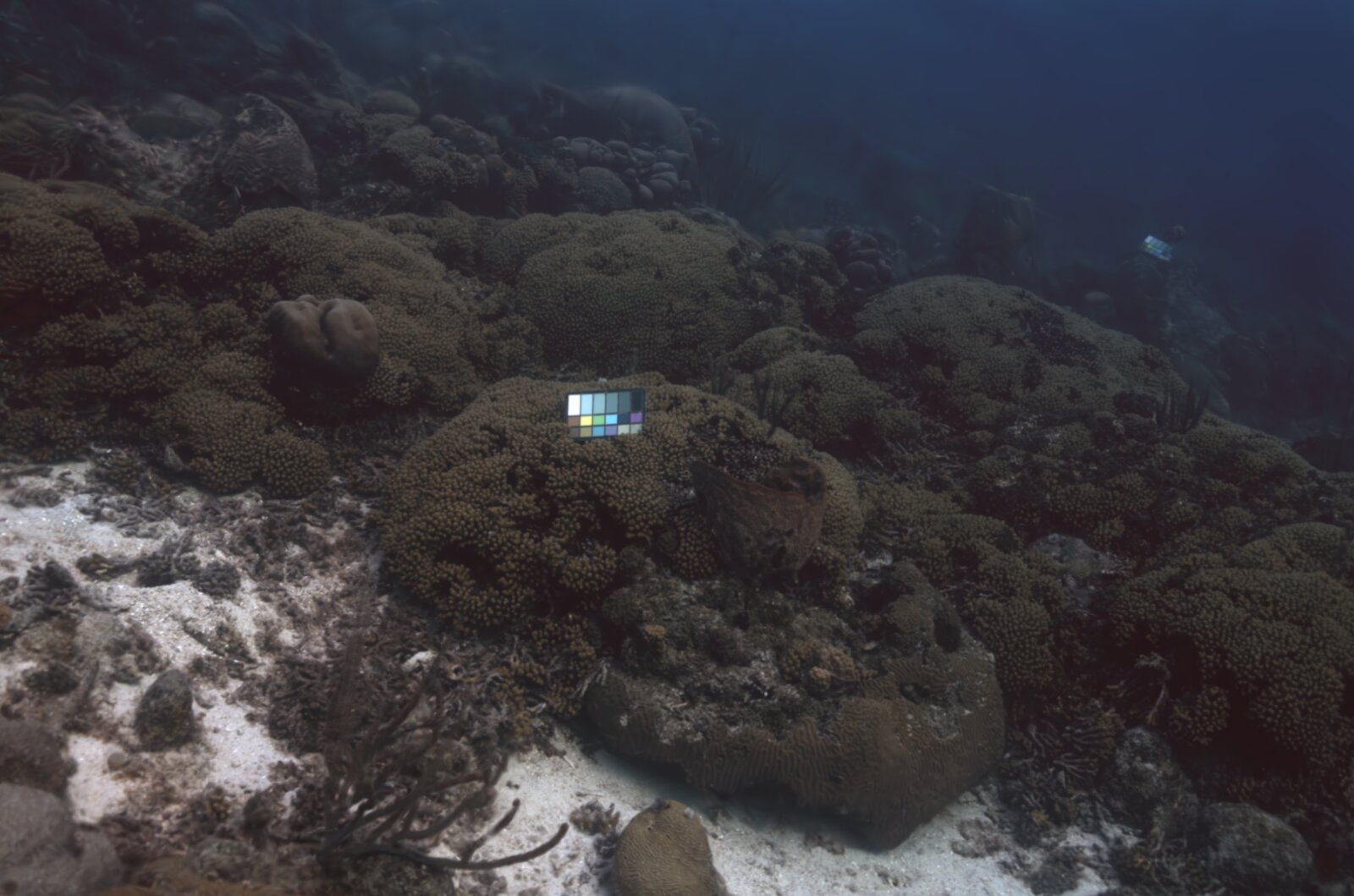}\\
 \includegraphics[width=\linewidth, height=\myH]{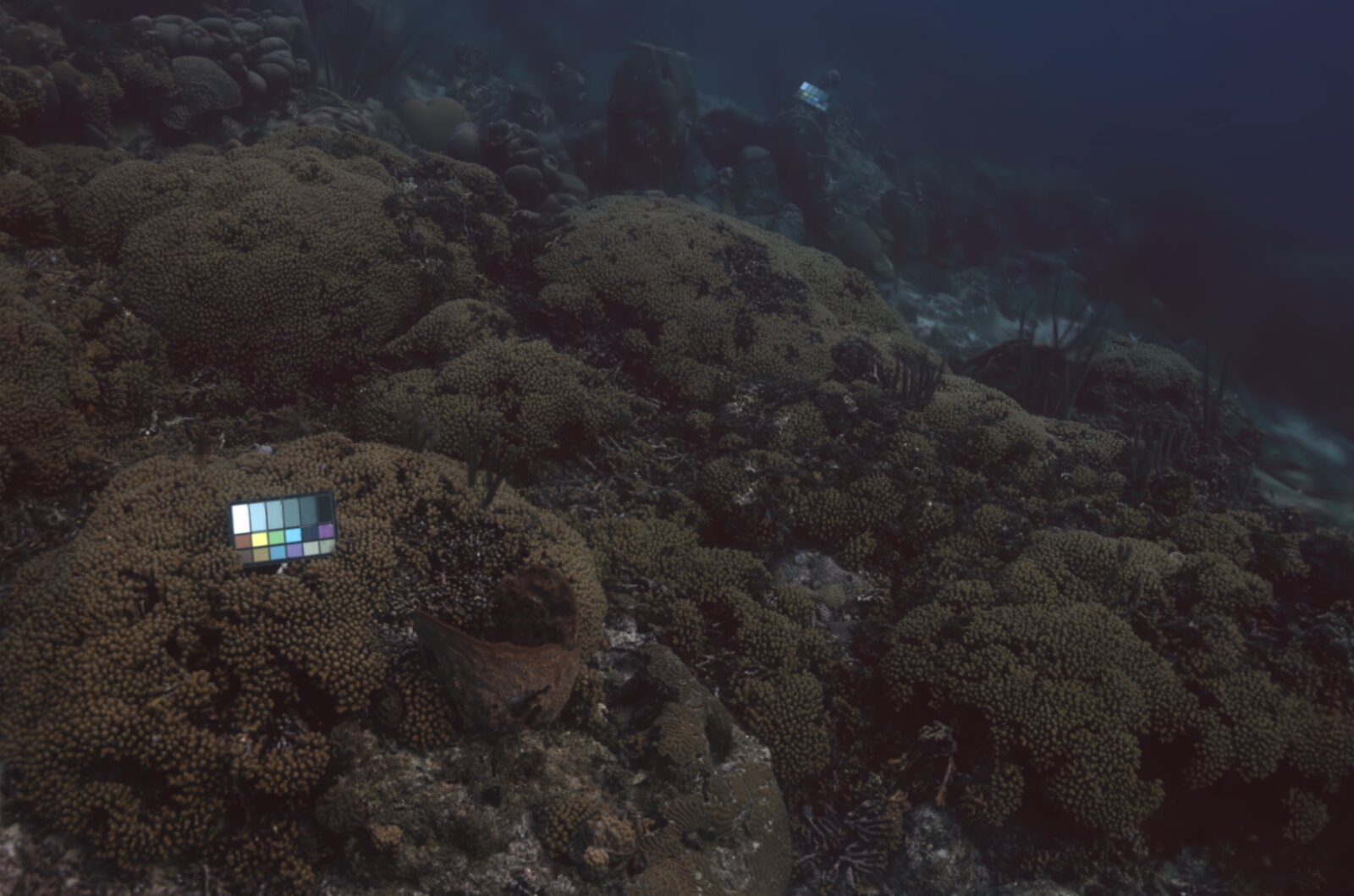}}
\subcaptionbox{Depth (train)}[\myW]
{\includegraphics[width=\linewidth, height=\myH]{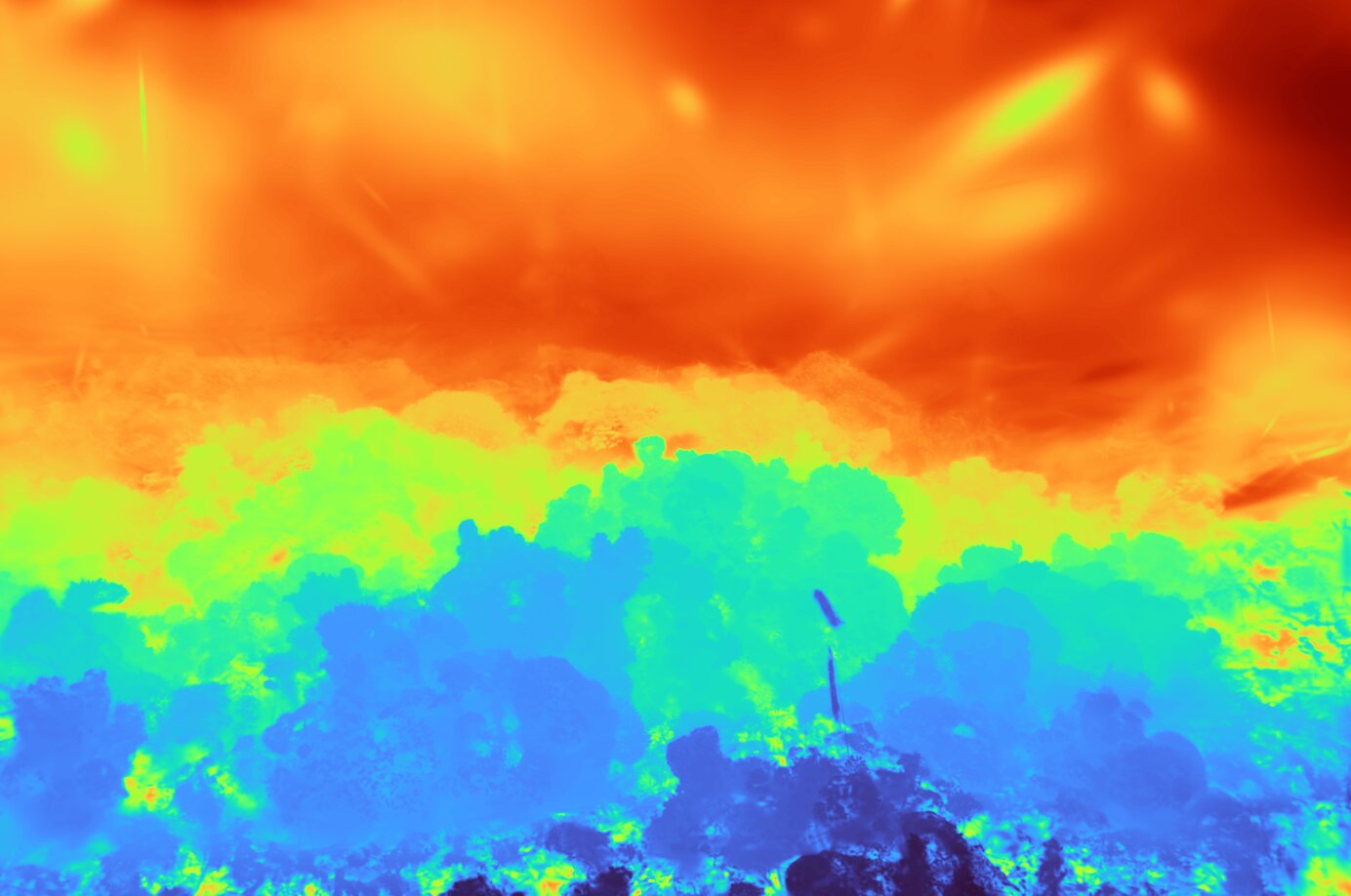}\\
 \includegraphics[width=\linewidth, height=\myH]{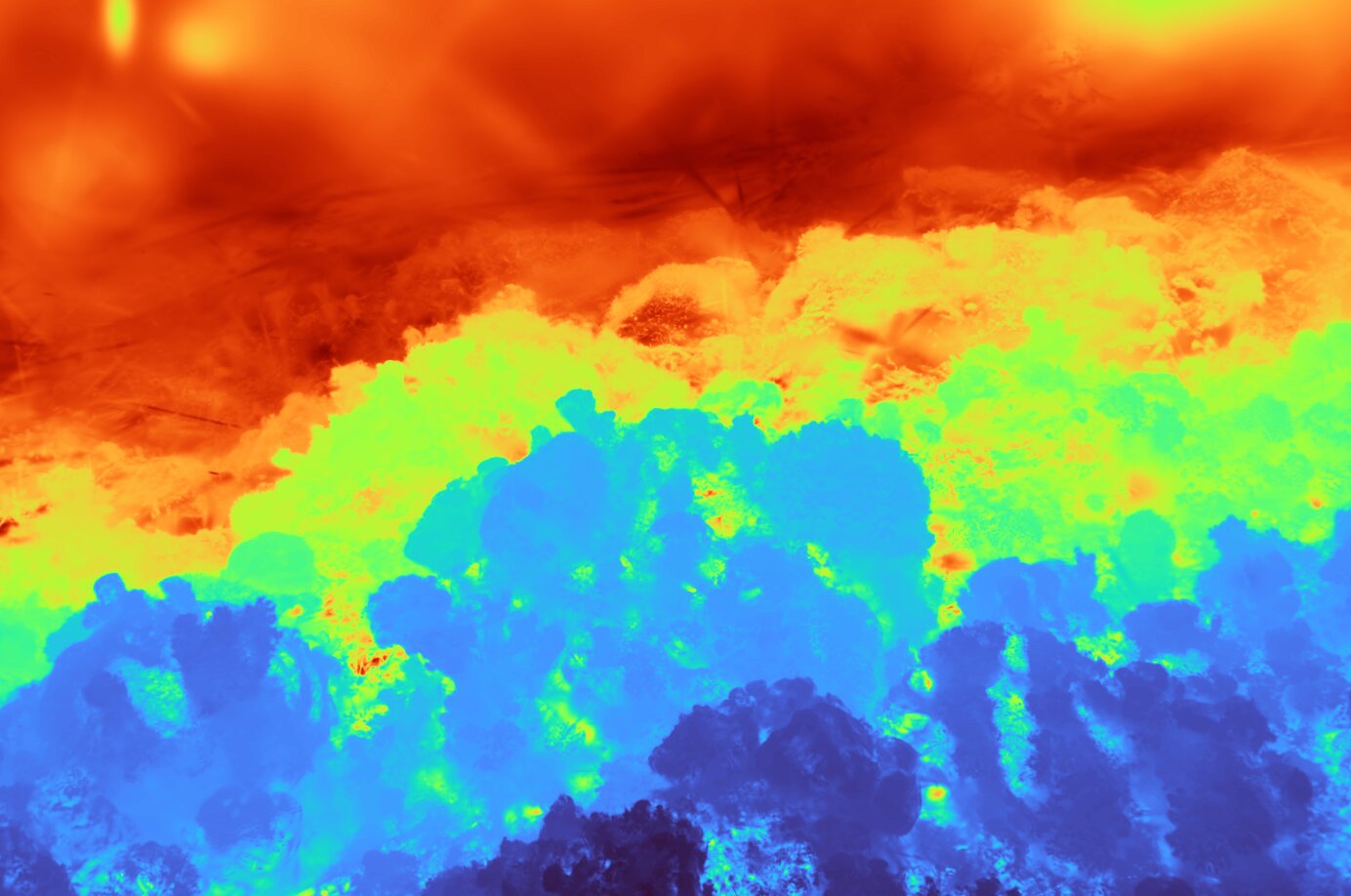}\\
 \includegraphics[width=\linewidth, height=\myH]{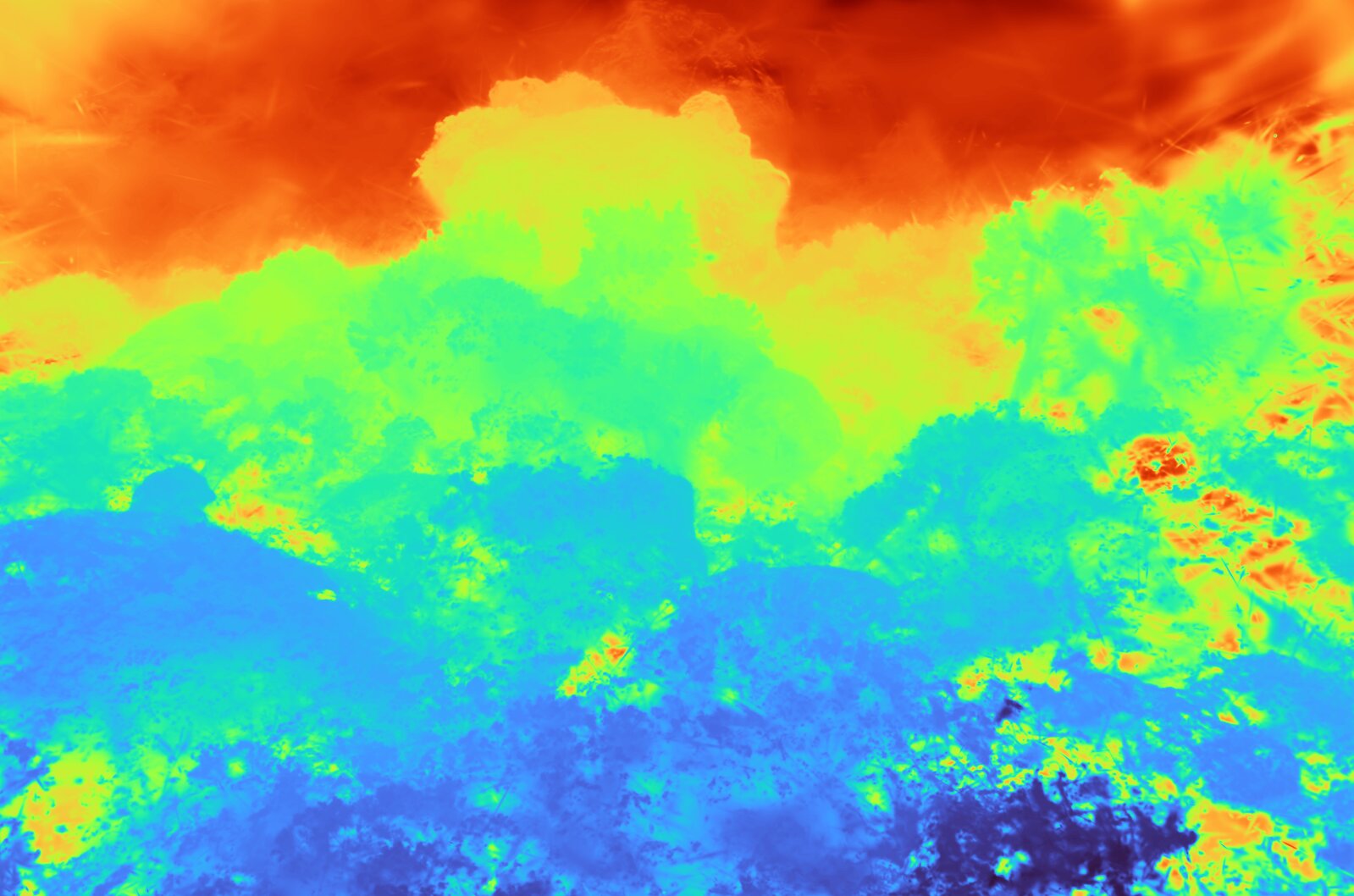}\\
 \includegraphics[width=\linewidth, height=\myH]{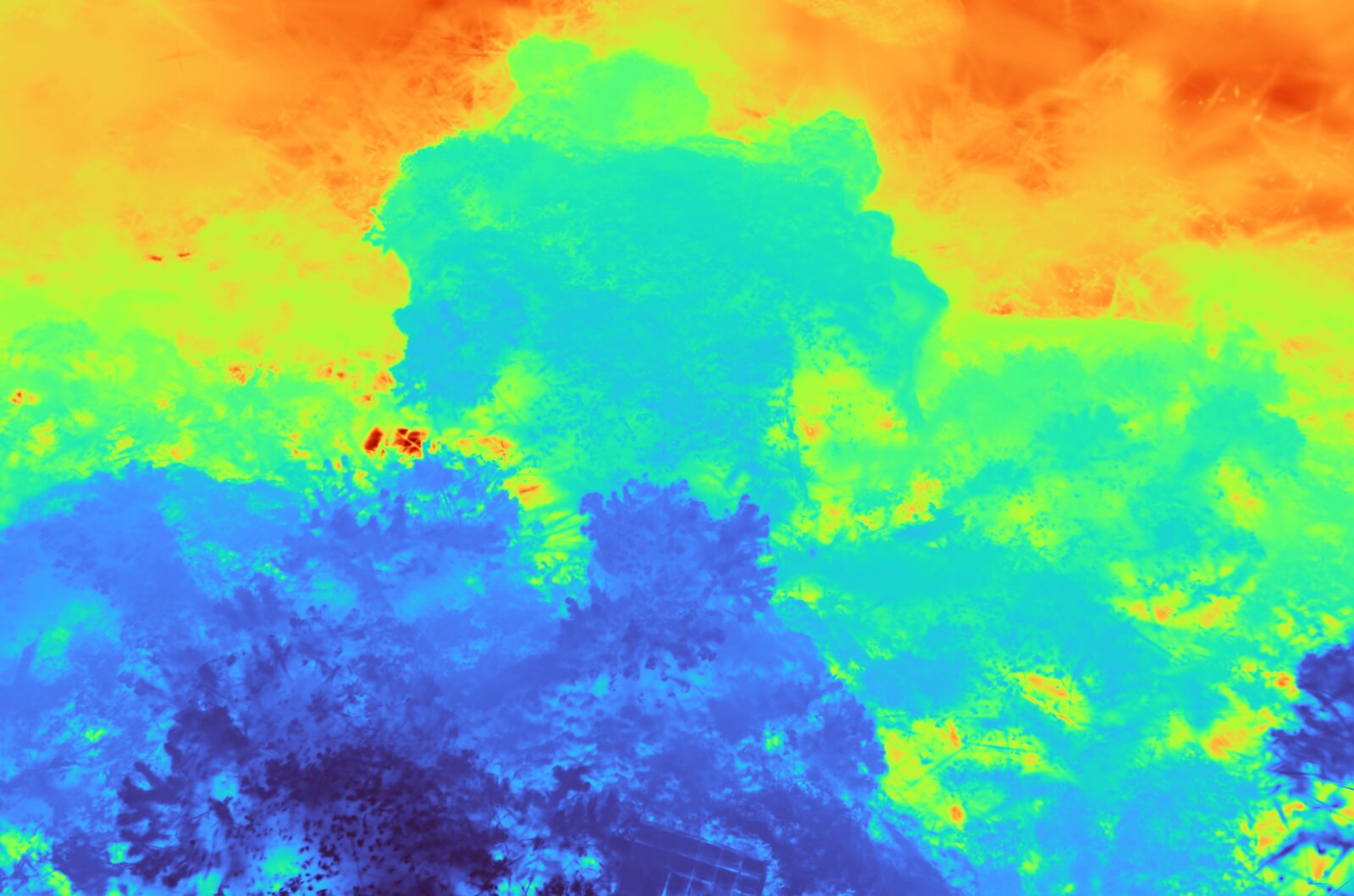}\\
 \includegraphics[width=\linewidth, height=\myH]{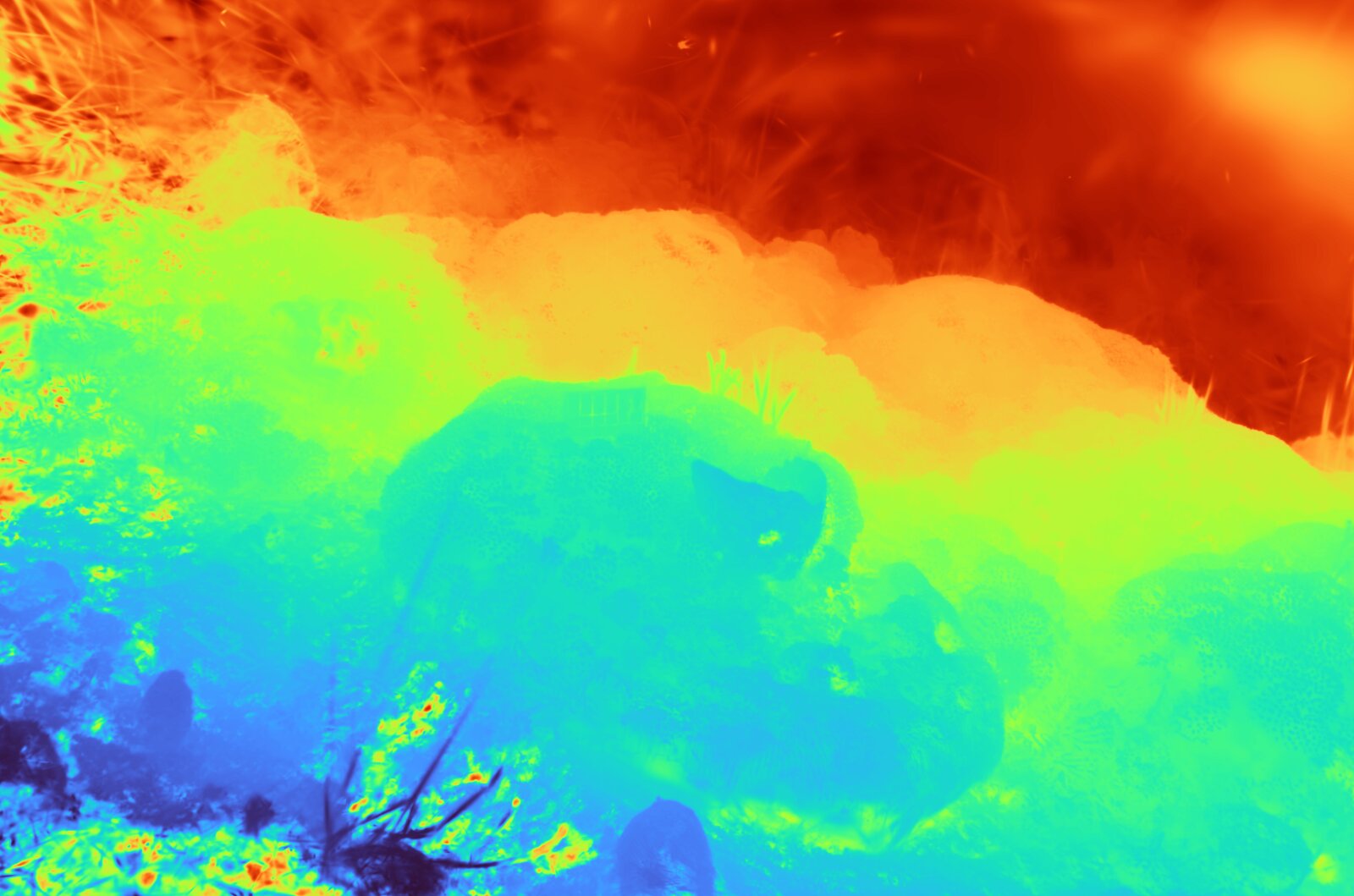}\\
 \includegraphics[width=\linewidth, height=\myH]{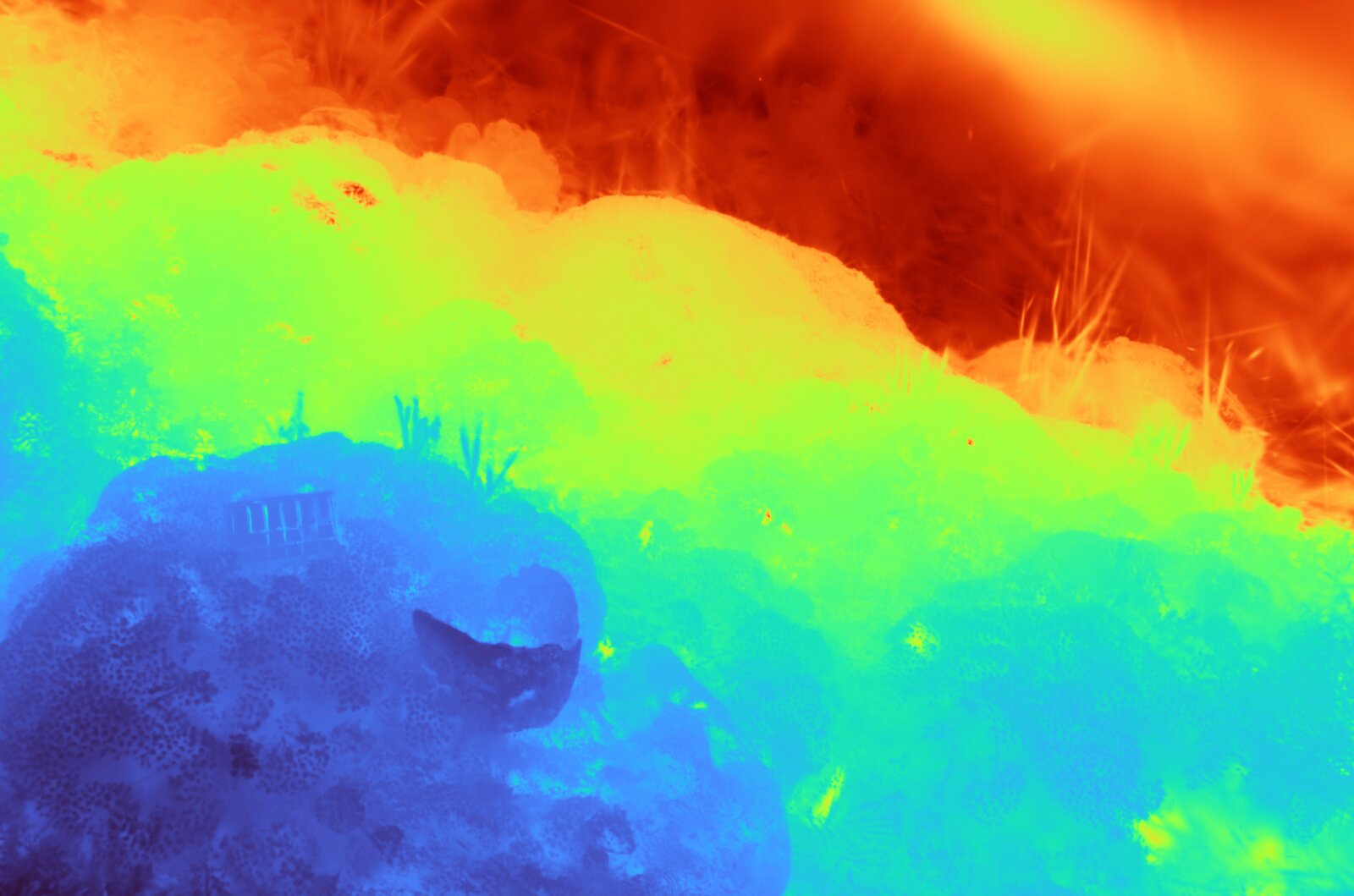}}
\subcaptionbox{Rendered (test)}[\myW]
{\includegraphics[width=\linewidth, height=\myH]{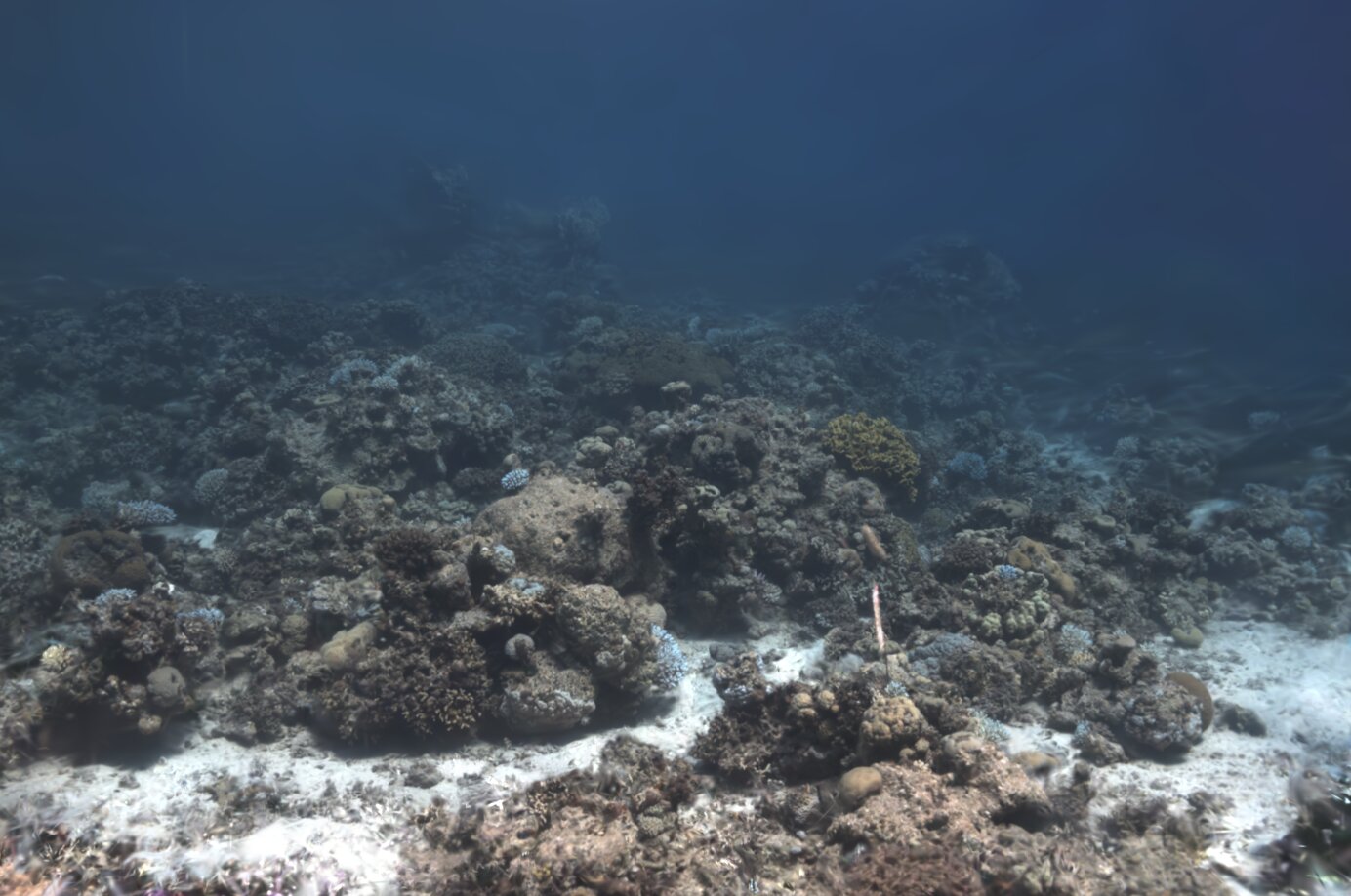}\\
 \includegraphics[width=\linewidth, height=\myH]{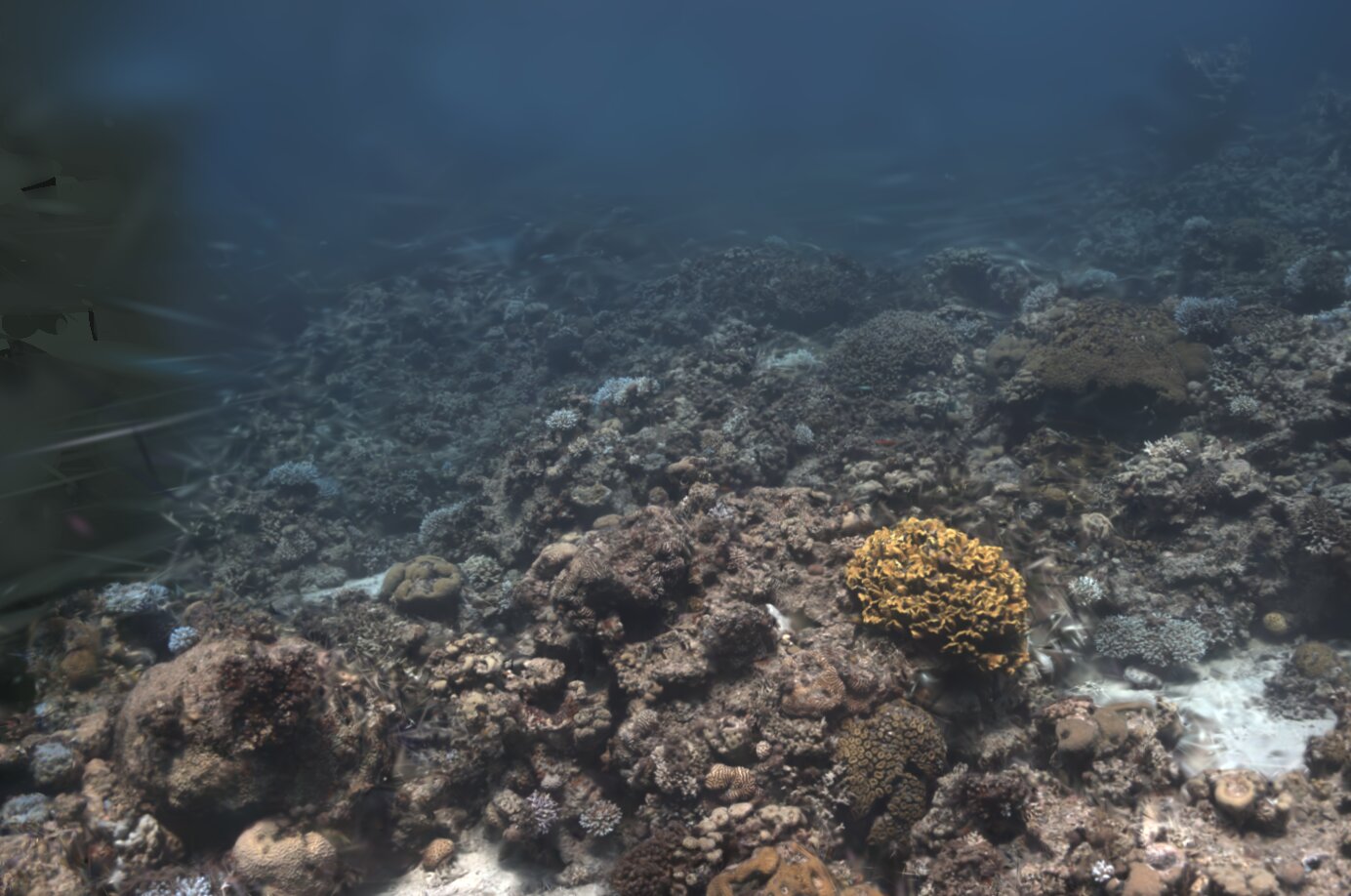}\\
 \includegraphics[width=\linewidth, height=\myH]{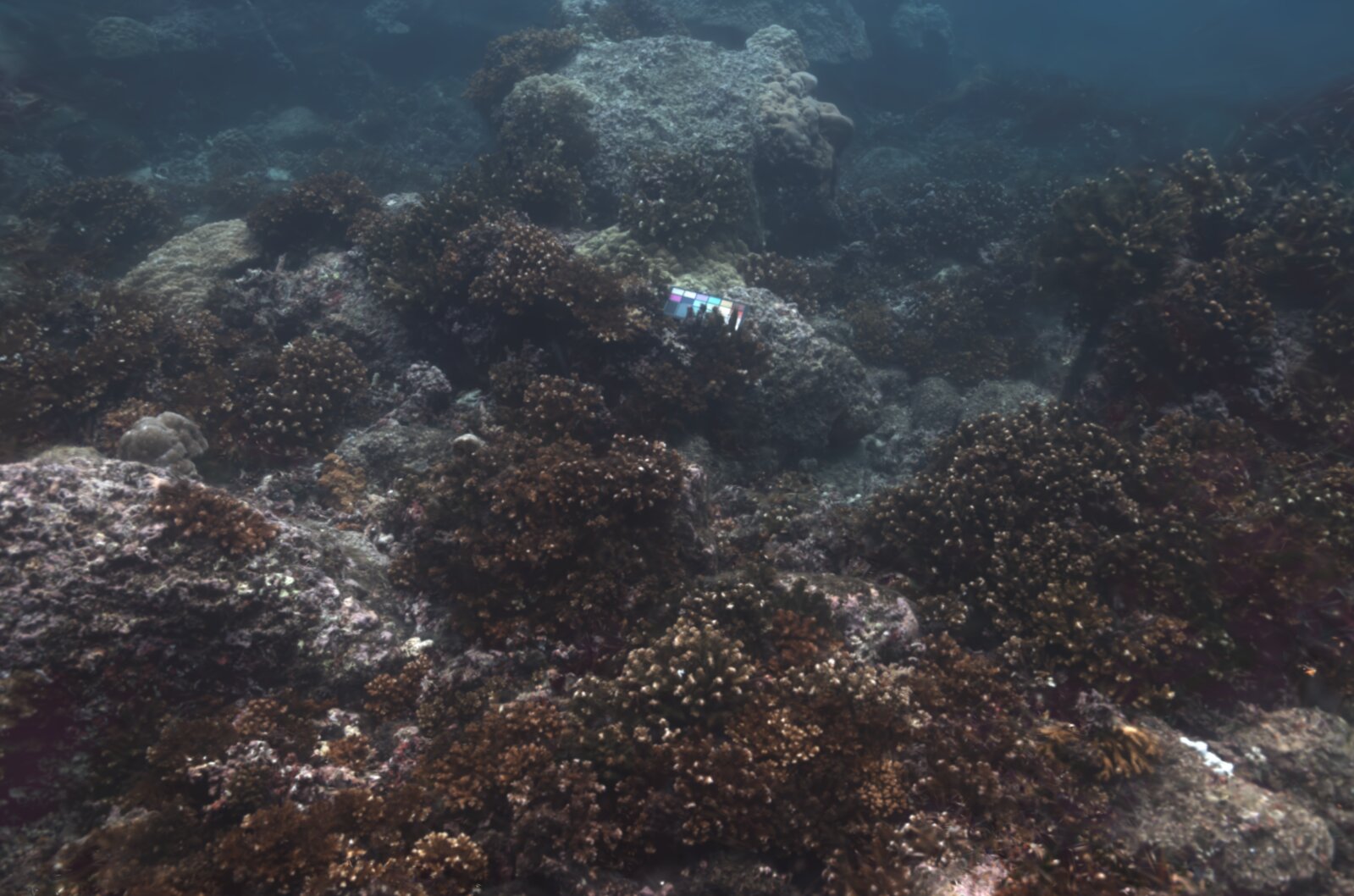}\\
 \includegraphics[width=\linewidth, height=\myH]{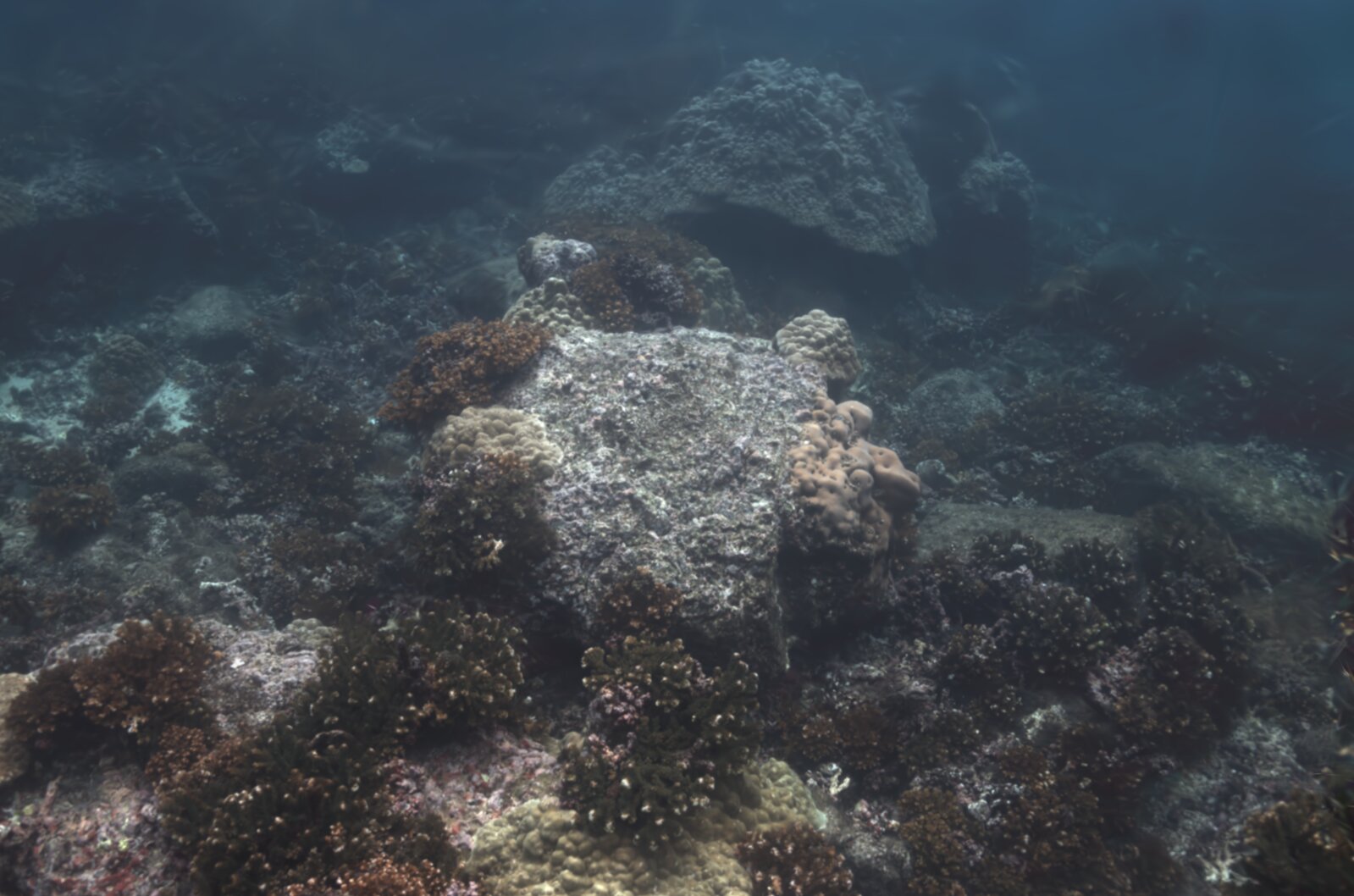}\\
 \includegraphics[width=\linewidth, height=\myH]{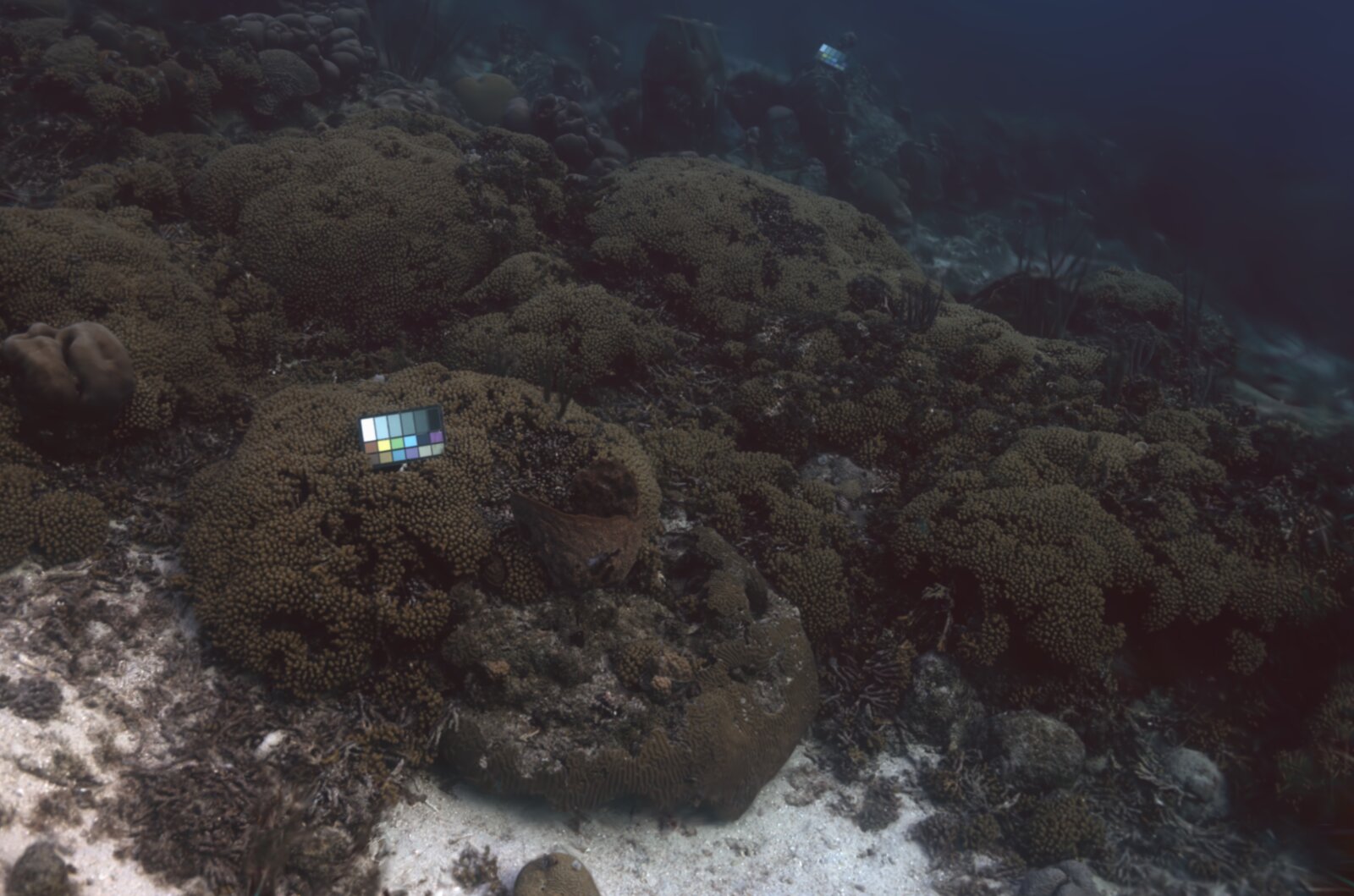}\\
 \includegraphics[width=\linewidth, height=\myH]{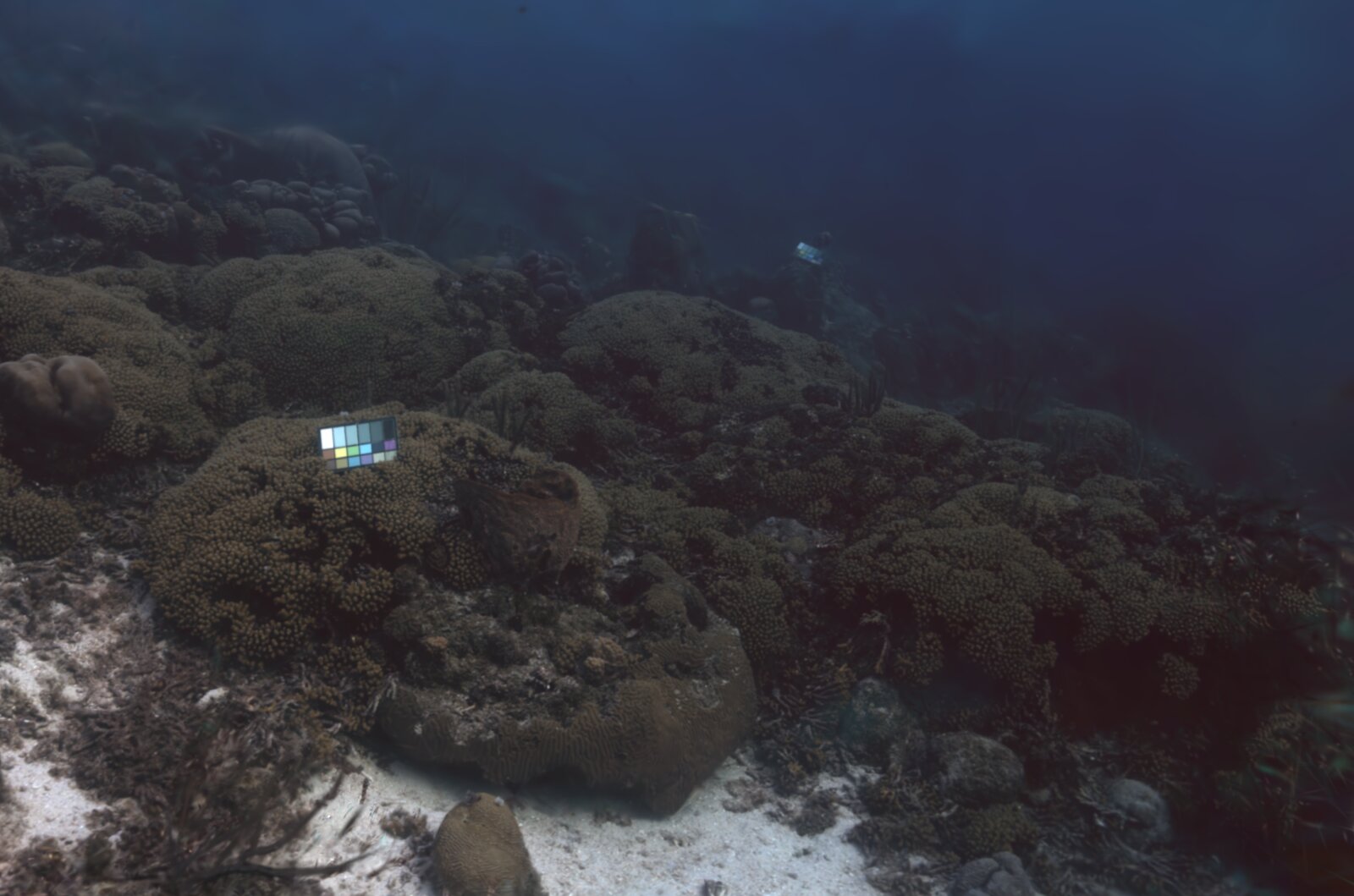}}
 \subcaptionbox{Depth (test)}[\myW]
{\includegraphics[width=\linewidth, height=\myH]{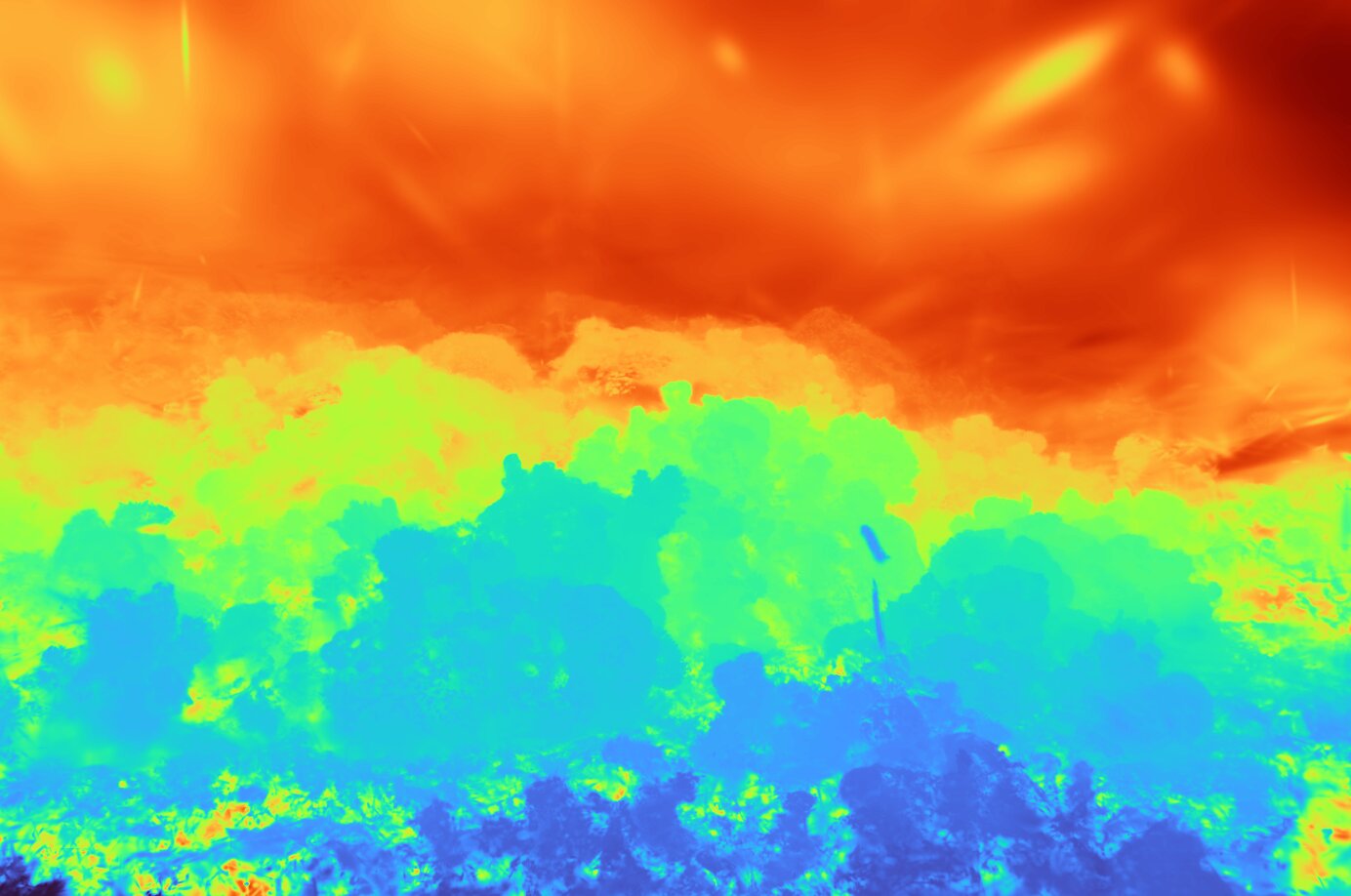}\\
 \includegraphics[width=\linewidth, height=\myH]{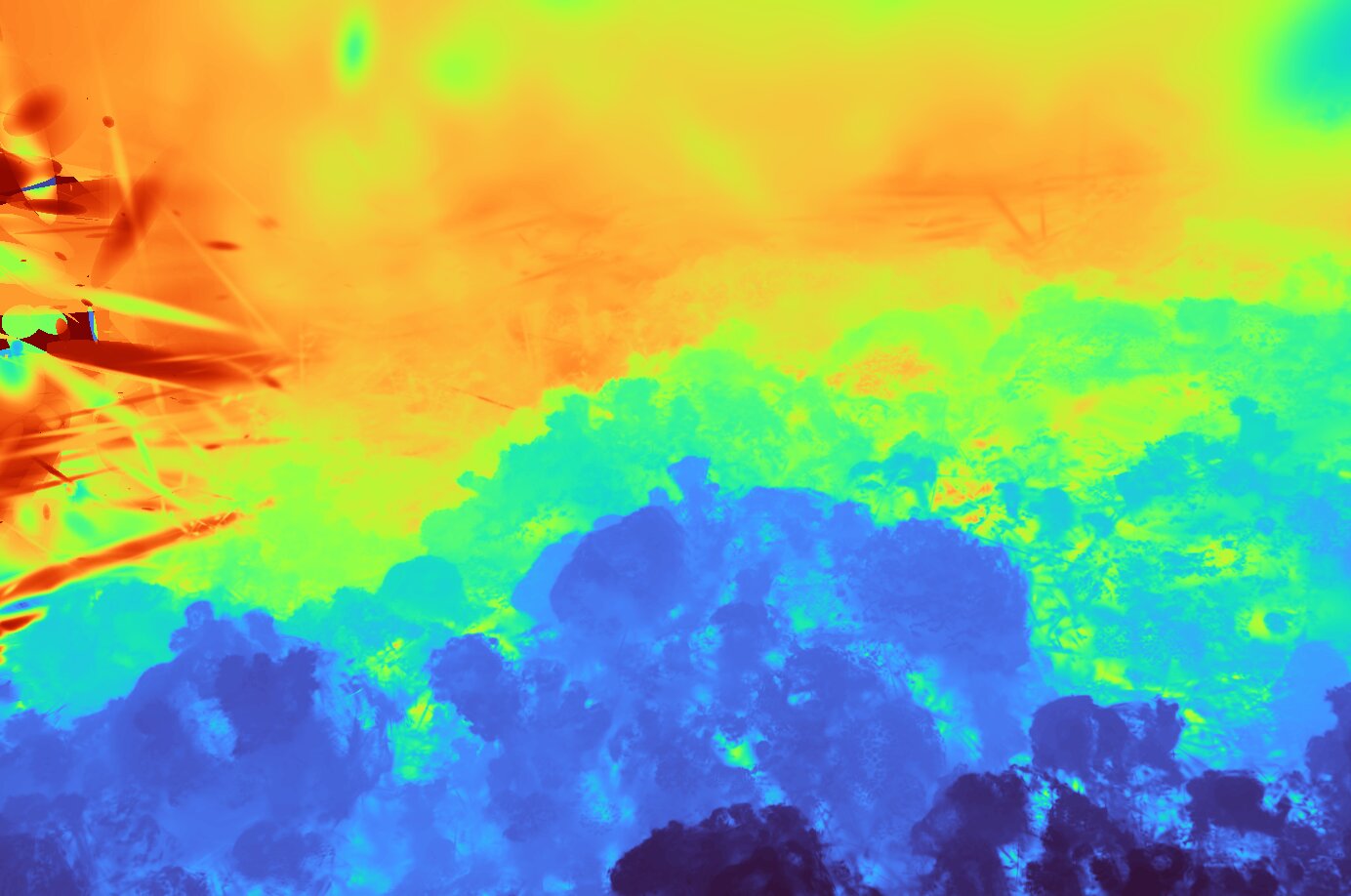}\\
 \includegraphics[width=\linewidth, height=\myH]{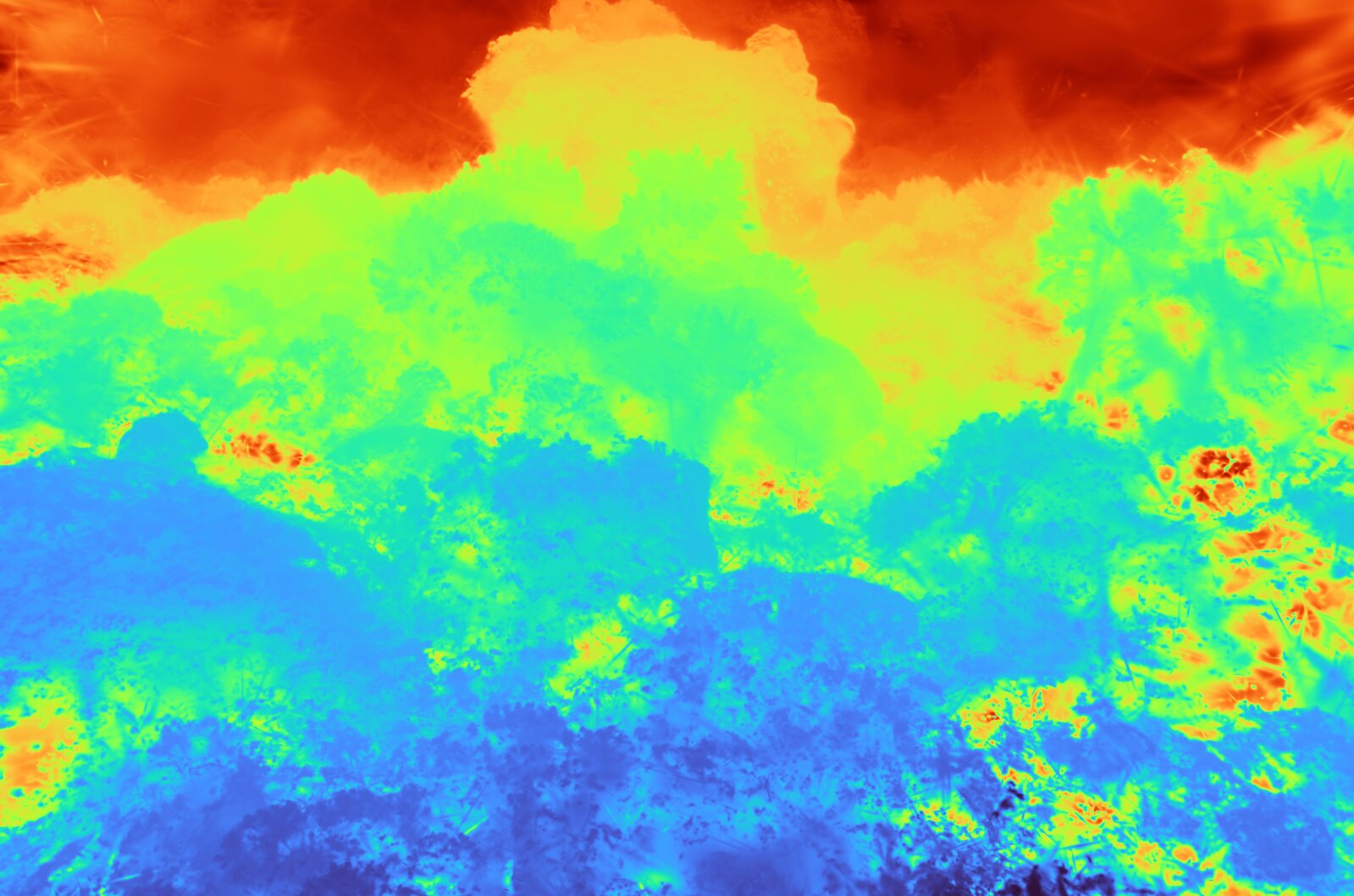}\\
 \includegraphics[width=\linewidth, height=\myH]{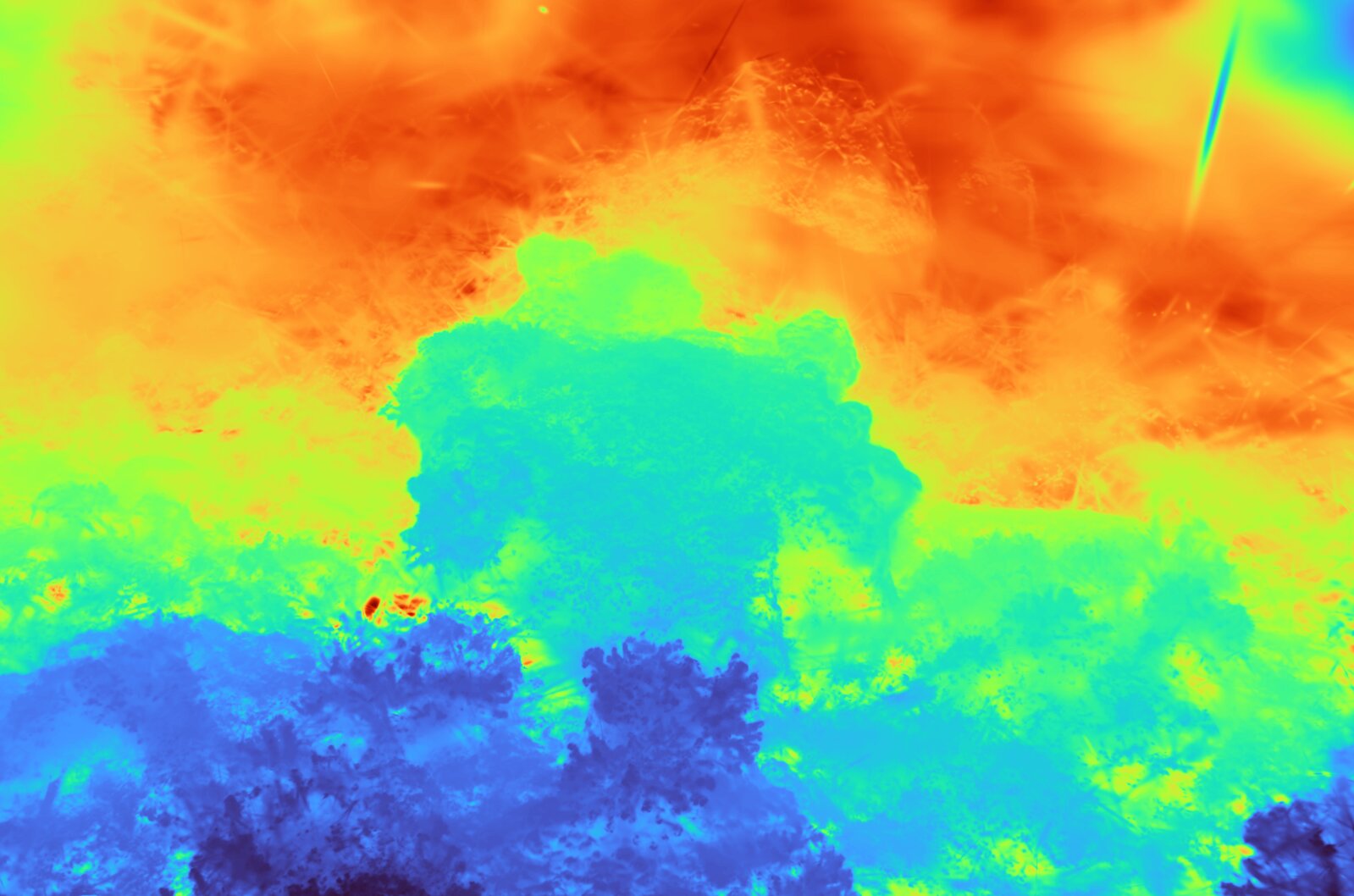}\\
 \includegraphics[width=\linewidth, height=\myH]{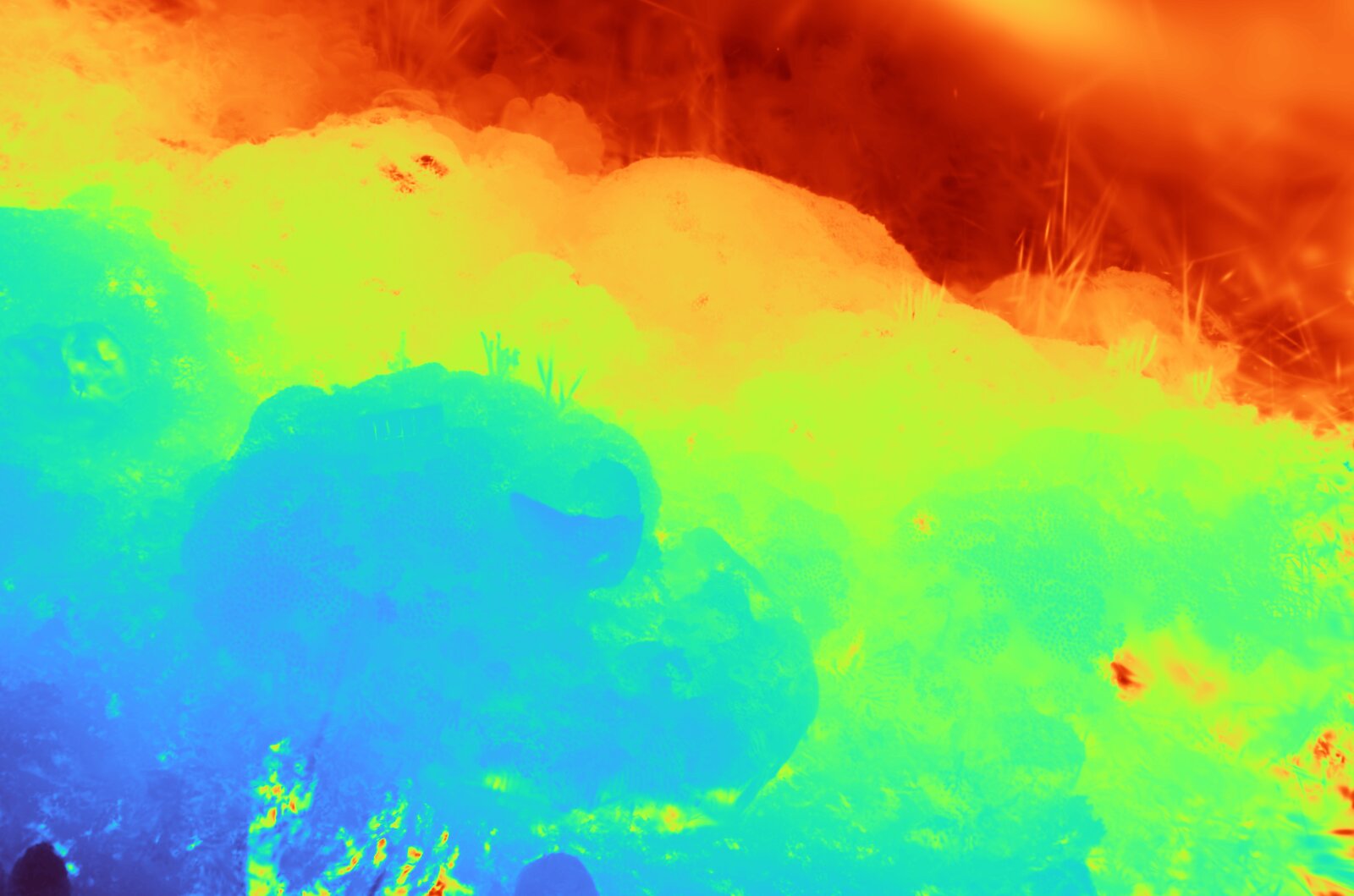}\\
 \includegraphics[width=\linewidth, height=\myH]{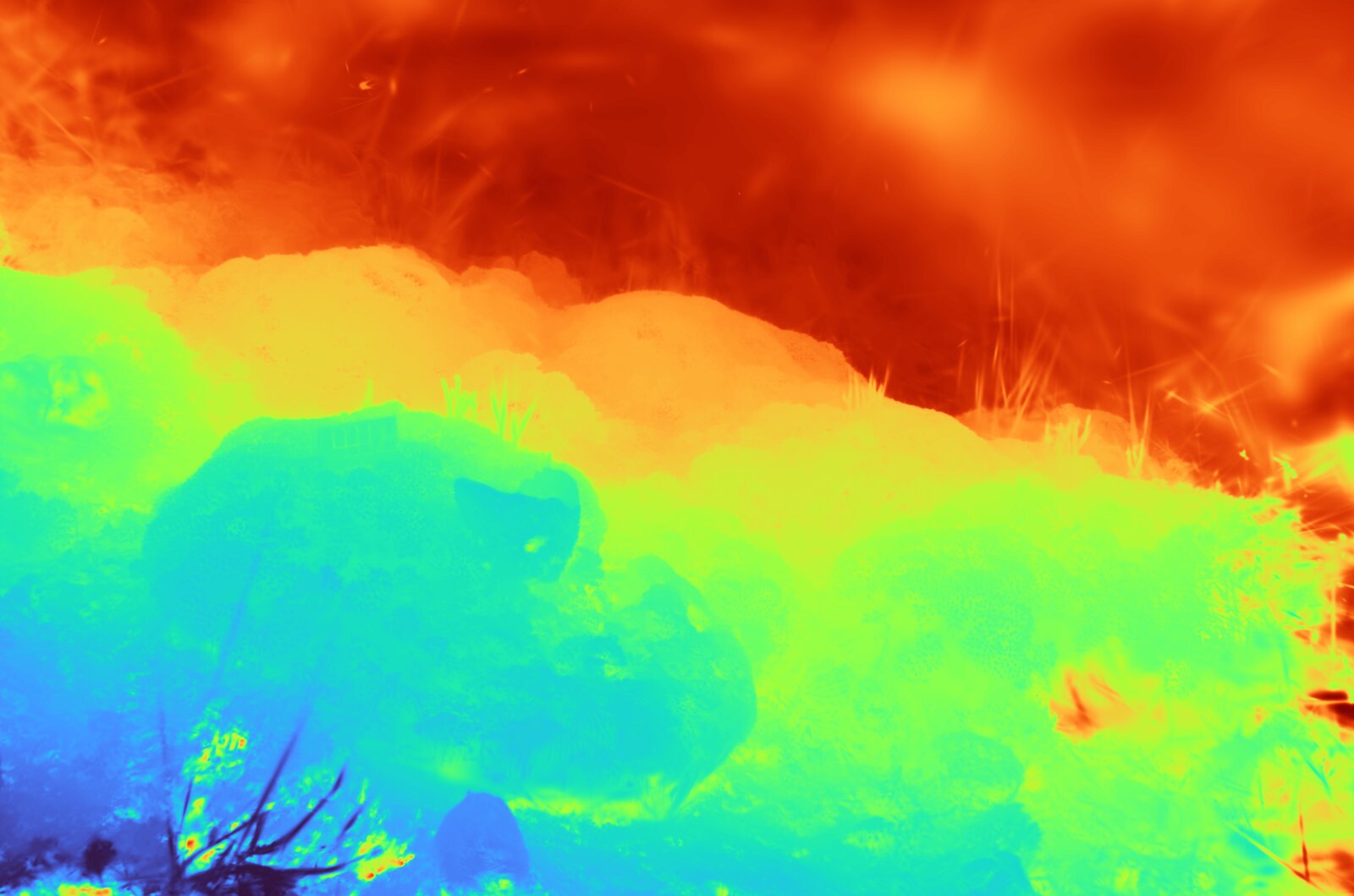}}
\caption[Supplementary Rendering Depth Results]{Rendered views and novel views, along with the depth maps generated by our method, shown for scenes from the Red Sea, Panama, and Curaçao. }
\label{fig:appendix3}
\end{figure*}

%% file: figures/appendix/fig_appendix_color_reconstruction.tex
\begin{figure*}[!h]
    \centering
    \includegraphics[height=7cm, keepaspectratio]{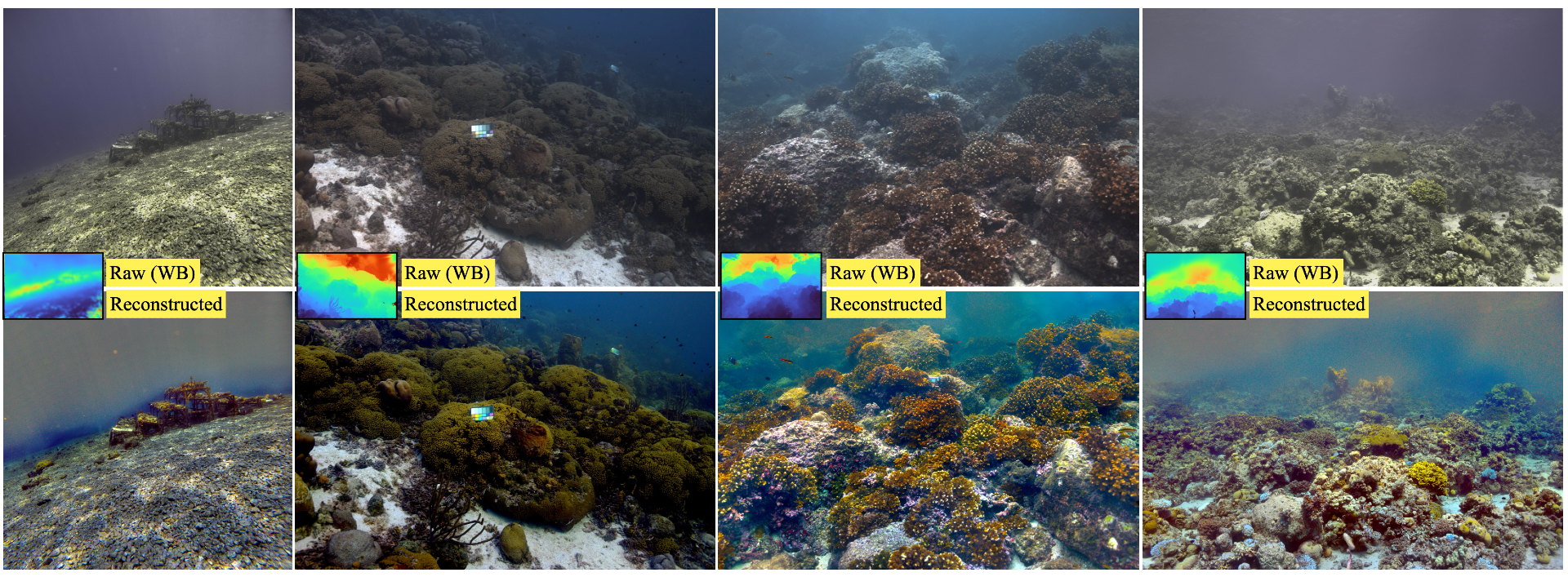}
    \caption{Color reconstruction application utilizing depth maps generated by our method, with raw (white-balanced, WB) images as inputs to the original Sea-thru algorithm~\cite{akkaynak2019}. Results are presented across all datasets.}
    \label{fig:Intro_appendix}
\end{figure*}

%% file: content/experiments_results/results.tex

\autoref{fig:visual_comparison} presents a visual comparison between the different methods, while \autoref{fig:results2} shows results from other underwater splatting approaches, which are slower than ours, as demonstrated in \autoref{fig:performance_comparison}. While all methods achieve high-quality performance at close ranges, a significant difference in reconstruction quality becomes apparent at larger distances.

A quantitative analysis, summarized  in \autoref{tab:comparison},  shows 
the effectiveness of the proposed method. 
Being based on 3DGS, our method is very fast, 
much faster than the NeRF-based methods. 
For example,
its training takes only several minutes and, during inference, it renders images at 140 frames per second (FPS) on average (across multiple datasets).
Please see the appendix for a comprehensive comparison 
between the different methods in terms of running times during both training
and inference.  
%
\autoref{tab:comparison} also shows that vanilla 3DGS (while being fast) struggles to effectively handle underwater scenes (see also~\autoref{fig:visual_comparison}). 
This is in sharp contrast with our method which offers both speed and quality. 

Although traditional metrics like PSNR might be misleading in underwater scenarios due to the predominant bluish color palette, our method's superiority becomes evident upon closer inspection, particularly when examining distant objects within the scene. This advantage is particularly pronounced with datasets like TableDB, which unlike publicly-available datasets, is unbounded
in viewpoints and scene depth. 
Of note, our method stands out in preserving the original resolution of rendered scenes and the geometry (see figure ~\autoref{fig:appendix3}) due to its efficient rendering speed, as shown in \autoref{fig:performance_comparison}. In contrast, STNeRF, STNeRFacto and WS which must downsize the images during training.
Futhermore, our work can use for color reconstruction applications using our depth maps, see ~\autoref{fig:Intro_appendix}.
For additional visual results, see the appendix. 
\textbf{Also, for best impression, we strongly encourage the reader to see our videos
on} \href{https://bgu-cs-vil.github.io/gaussiansplashingUW.github.io/}{our webpage}.

%% file: content/experiments_results/time_comparison.tex
Measuring the rendering time of our method reveals that it is significantly faster than competitors in both training and inference. The high frame rate achieved during rendering enables its use in real-time applications while maintaining accuracy at such speeds. Detailed results are summarized in ~\autoref{tab:comparisonTime_appendix}.

%% file: figures/experiments_results/table_compTime.tex
\begin{table*}[h]
    \setlength\tabcolsep{6pt}
    \resizebox{\textwidth}{!}{
    \begin{tabular}{@{}lcccccc@{}}
    \toprule
    \multicolumn{1}{l|}{Dataset} & \multicolumn{2}{c}{Red Sea} & \multicolumn{2}{c}{Curaçao} & \multicolumn{2}{c}{TableDB} \\ 
    \cmidrule(lr){2-3} \cmidrule(lr){4-5} \cmidrule(lr){6-7}
    \multicolumn{1}{l|}{Method} & FPS$\uparrow$ & Training Time (min)$\downarrow$ & FPS$\uparrow$ & Training Time (min)$\downarrow$ & FPS$\uparrow$ & Training Time (min)$\downarrow$ \\ 
    \midrule
    \multicolumn{1}{l|}{3DGS~\cite{kerbl3Dgaussians}} 
    & 174 & 10 & 154 & 13 & 162 & 16 \\ 
    \multicolumn{1}{l|}{Splatfacto~\cite{liang2024analyticsplatting}} 
    & 129 & 28 & 111 & 25 & 119 & 37 \\ 
    \multicolumn{1}{l|}{STNeRF~\cite{levy2023seathru}} 
    & 0.05 & 582 & 0.05 & 547 & 0.06 & 612 \\ 
    \multicolumn{1}{l|}{STNeRfacto~\cite{subsea2023}} 
    & 0.72   & 168 & 0.76 & 151 & 0.53 & 213\\ 
     \multicolumn{1}{l|}
    {WS~\cite{li2024watersplatting}} 
    & 53   & 23 & 82 & 32 & 61 & 37\\
    \multicolumn{1}{l|}{SeaSplat~\cite{yang2024seasplat}} 
    & 41   & 85 & 110 & 98 & 112 & 124\\ 
    \multicolumn{1}{l|}{Gaussian Splashing (\textbf{Ours})} 
    & 187 & 11 & 135 & 12 & 161 & 15 \\ 
    \bottomrule
    \end{tabular}
       }
       \caption{Comprehensive comparison table across scenes, highlighting the frames per second (FPS) performance during rendering and the training time (minutes) required to create the model.} 
    \label{tab:comparisonTime_appendix}   
    
\end{table*}

%% file: content/ablation/ablation_table.tex
\begin{table*}[t]
    \setlength\tabcolsep{6pt}
    \caption{Ablation Study}
    \resizebox{1.0\textwidth}{!}{
    \begin{tabular}{@{}lcccccccccccc@{}}
    \multicolumn{1}{l|}{Dataset} & \multicolumn{3}{c}{Red Sea} & \multicolumn{3}{c}{Curaçao} & \multicolumn{3}{c}{Panama} & \multicolumn{3}{c}{Avg.} \\ \midrule
    \multicolumn{1}{l|}{Method} & \multicolumn{1}{c}{PSNR$\uparrow$} & \multicolumn{1}{c}{SSIM$\uparrow$} & \multicolumn{1}{c}{LPIPS$\downarrow$} & \multicolumn{1}{c}{PSNR$\uparrow$} & \multicolumn{1}{c}{SSIM$\uparrow$} & \multicolumn{1}{c}{LPIPS$\downarrow$} & \multicolumn{1}{c}{PSNR$\uparrow$} & \multicolumn{1}{c}{SSIM$\uparrow$} & \multicolumn{1}{c}{LPIPS$\downarrow$} & \multicolumn{1}{c}{PSNR$\uparrow$} & \multicolumn{1}{c}{SSIM$\uparrow$} & \multicolumn{1}{c}{LPIPS$\downarrow$} \\ \hline
    \multicolumn{1}{l|}{3DGS~\cite{kerbl3Dgaussians} (the baseline)}                                                              
    & $22.94$ &  $0.87$           &  $0.17$ &  $28.23$   &$0.88$&$0.23$     &$29.88$&$0.91$      &  $0.15$ & $27.02$ &$0.89$ &$0.18$   \\
    
    \multicolumn{1}{l|}{Ours w/o $\lossbest$} 
    & $22.93$ & $0.88$ & $0.14$      & $20.65$ & $0.79$ & $0.27$ &       $27.67$ & $0.87$ & $0.21$ & $23.75$ & $0.85$ & $0.20$ \\

    \multicolumn{1}{l|}{Ours w/o $\directv$} & $22.12$ & $0.88$ & $0.15$ & $30.12$ & $0.89$ & $0.20$ & $29.47$ & $0.91$ & $0.13$ & $27.24$ & $0.89$ & $0.16$ \\

    \multicolumn{1}{l|}{Ours w/o MCMC densification} & $23.04$ & $0.88$ & $0.15$ & $28.40$ & $0.89$ & $0.23$ & $29.99$ & $0.91$ & $0.14$ & $27.14$ & $0.89$ & $0.31$ \\
        
    \multicolumn{1}{l|}{Ours w/o depth} & $21.16$ & $0.84$ & $0.23$ & $23.86$ & $0.80$ & $0.24$ &
    $27.69$ & $0.88$ & $0.21$ & $24.23$ & $0.84$ & $0.23$ \\

    \multicolumn{1}{l|}{Our complete method} & $\mathbf{24.73}$ & $\mathbf{0.92}$ & $\mathbf{0.11}$ & $\mathbf{31.26}$ & $\mathbf{0.92}$ & $\mathbf{0.17}$ & $\mathbf{31.35}$ & $\mathbf{0.94}$ & $\mathbf{0.11}$ & $\mathbf{29.11}$ & $\mathbf{0.92}$ & $\mathbf{0.13}$ \\
    
    \bottomrule
    \end{tabular}
    }
    \label{tab:ablation}
\end{table*}

%% file: content/ablation/ablation.tex
Our ablation study, summarized in~\autoref{tab:ablation}, considered several conditions:
1) Removing our additional 
loss term $\lossbest$; 
2) Forcing  $\directv$ to be zero, to test if $B_s$ alone can capture enough of the underwater scene distortions. 
3) Using original densification process from~\cite{kerbl3Dgaussians} instead of MCMC densification process approach ~\cite{kheradmand20243d}.
4) Replacing our depth estimation 
with a standard  monocular depth model (\cite{Ranftl2020,Ranftl2021}).
For completeness, \autoref{tab:ablation} also includes, as a baseline, the original 3DGS~\cite{kerbl3Dgaussians},
as well as our complete model. 
As \autoref{tab:ablation} shows,  all of the components turn out to be important.

%% file: content/conclusion/conclusion.tex
We presented \emph{Gaussian Splashing}, a variant of 3D-Gaussian Splatting specialized for underwater imagery. Our method accurately estimates geometry for underwater scenes, from as few as five images, which is a necessary step to maximize the amount of useful information that can be extracted from the original images containing attenuated colors. Perhaps more importantly, our method performs geometry estimation from a handful of scenes within a few minutes (compared to several hours for NeRF-based methods).  
We are unaware of any potential negative societal impact of this work.
Having accurate scene geometry is a necessary component for consistent color reconstruction, following which underwater imagery can be processed using powerful computer-vision and machine-learning methods developed for in-air images. Thus, getting close to near real-time training opens the possibility for real-time color reconstruction of underwater scenes, which would improve capabilities of autonomous or remotely operated underwater vehicles for better navigation, SLAM, and obstacle avoidance. Lastly, leveraging well-estimated scene geometry, we are able to render novel-views of underwater scenes in real time. This capability can have a broad and positive societal impact as it immediately enables applications for subsea training (\eg, divers or submarine pilots, search and rescue teams, commercial divers, \etc), computer games, marine science education, and cultural heritage documentation. 


%% file: content/conclusion/limitations.tex
Our method shares three limitations with~\cite{levy2023seathru,subsea2023}:
%
%
1) While, like~\cite{levy2023seathru,subsea2023}, we rely on a SOTA image formation model, that model does not address phenomena such as multiple scattering or artificial illumination.
2) The method requires 
the extraction of camera poses, which is hard if the visibility is too poor (as in, \eg, turbid water). This  limitation is shared with not only~\cite{levy2023seathru,subsea2023} but also other NeRF-based or 3DGS-based methods.
3) 
While our formulation's strength lies in its ability to learn
the medium's characteristics, the success of that learning
depends on having enough variation in scene range. 
However, while~\cite{levy2023seathru,subsea2023} struggle if the inter-image range
variability is \emph{too large} (see~\autoref{fig:visual_comparison}), our method handles such cases gracefully. 


%% file: content/conclusion/discussion.tex
Having accurate scene geometry is a necessary component for consistent color reconstruction, following which underwater imagery can be processed using powerful computer-vision and machine-learning methods developed for in-air images. Thus, getting close to near real-time training opens the possibility for real-time color reconstruction of underwater scenes, which would improve capabilities of autonomous or remotely operated underwater vehicles for better navigation, SLAM, and obstacle avoidance. Lastly, leveraging well-estimated scene geometry, we are able to render novel-views of underwater scenes in real time. This capability can have a broad and positive societal impact as it immediately enables applications for subsea training (\eg, divers or submarine pilots, search and rescue teams, commercial divers, \etc), computer games, marine science education, and cultural heritage documentation.

%% file: content/acknowledgments/acknowledgments.tex
This project was partially funded by the U.S. Office of Naval Research (US ONR), whose support is gratefully acknowledged.

%% file: refs.bbl

%% file: content/biography/biography.tex
\graphicspath{ {./content/biography} }

\begin{IEEEbiography}[{\includegraphics[width=1in,height=1.25in,clip,trim=0 0 0 0, keepaspectratio]{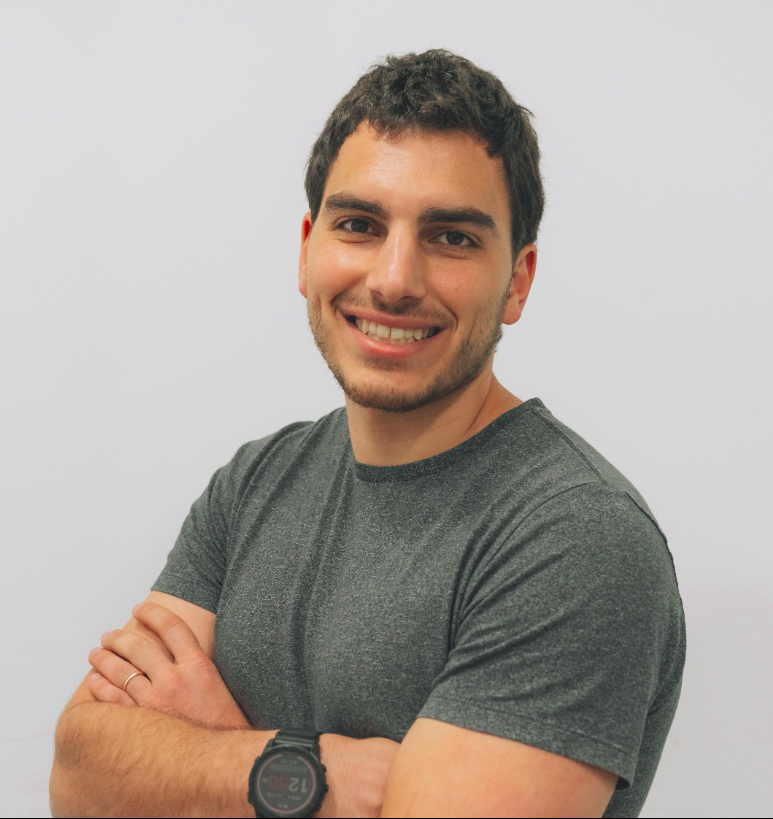}}]{Nir Mualem}
is a computer vision researcher in the tech industry. He received his M.Sc. in Computer Science (AI track) from Ben-Gurion University and his B.Sc. in Computer Science from the Technion – Israel Institute of Technology, graduating with excellence in both degrees. His master's research focused on 3D reconstruction for underwater scenes and CUDA optimization. He works on problems in 3D reconstruction, computer vision, and machine learning.
\end{IEEEbiography}

 \vspace{11pt}

 \begin{IEEEbiographynophoto}{Roy Amoyal}
is a Ph.D. candidate in Computer Science at Ben-Gurion University of the Negev, under the supervision of Prof. Oren Freifeld. He received his B.Sc. and M.Sc. degrees in Computer Science from Ben-Gurion University. His research focuses on 3D Gaussian Splatting and 3D registration. He also has industry experience in 3D computer vision for autonomous drone navigation and led the Autonomous Vision Division of the BGRacing Formula Student team, under whose leadership the team built Israel’s first autonomous race car.
\end{IEEEbiographynophoto}

\vspace{11pt}

\begin{IEEEbiographynophoto}{Oren Freifeld}
received his BSc and MSc degrees in Biomedical Engineering from Tel-Aviv University, in 2005 and 2007, respectively. He received the ScM and PhD degrees in Applied Mathematics from Brown University, in 2009 and 2013, respectively. Later, he was a postdoc with the MIT Computer Science \& Artificial Intelligence Laboratory. Next, he joined Ben-Gurion University where he is currently an Associate Professor at the Faculty of Computer and Information Science. His fields of research are computer vision and machine learning.
\end{IEEEbiographynophoto}

\vspace{11pt}

\begin{IEEEbiography}[{\includegraphics[width=1in,height=1.25in,clip,trim=0 0 0 0, keepaspectratio]{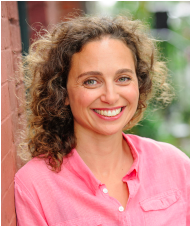}}]{Derya Akkaynak}
is an assistant professor in the Department of Marine Technologies at the University of Haifa, where she directs the COLOR
Lab located at the Interuniversity Institute of Marine Sciences in Eilat. She received her PhD in
Mechanical \& Oceanographic Engineering from
MIT and Woods Hole Oceanographic Institution,
MSc in Aeronautics and Astronautics from MIT,
and BSc in Aerospace Engineering from METU
(Turkey). She studies light and vision in the
ocean, at the intersection of which lies color.
\end{IEEEbiography}

%% file: content/appendix/appendix.tex
\appendix

\section{Supplemental Material}





\subsection{Additional Visual Results}
Qualitative videos showcasing comparisons, depth estimation, and backscattering estimation, including videos for frames from our new TableDB dataset, are available on \href{https://bgu-cs-vil.github.io/gaussiansplashingUW.github.io/}{\textbf{our webpage}}, and we strongly encourage viewing them to get a better impression and deeper understanding. Additional rendering results for training images and novel views are presented in ~\autoref{fig:appendix}, where ~\autoref{fig:appendix2} includes several images from our new TableDB dataset
(more examples from TableDB
 are available in our webpage). 


\input{figures/appendix/appendix_fig_examples}
\clearpage
\input{figures/appendix/appendix_fig_examples2}

\clearpage



\clearpage
\onecolumn 
\subsection{Partial Derivatives of the New Parameters}

\begin{align}
    \Cuw &= \left[\sum_{i=1}^{N}\bc_i\alpha_i \cdot {T_i}\right]e^{-\directv z} + \binf(1-e^{-\bsv z}) \nonumber\\
    &= \left[\sum_{i=1}^{N}\bc_i\alpha_i \cdot {\prod_{j=0}^{i-1}(1-\alpha_j)}\right]e^{-\directv z} + \binf(1-e^{-\bsv z})
\end{align}

\begin{align}
    \frac{\partial \loss}{\partial \bc_i} &=  \frac{\partial \Cuw}{\partial \bc_i}\cdot e^{-\directv z} \cdot \frac{\partial \loss}{\partial \Cuw} \nonumber\\
    &= \alpha_i{T_i}\cdot e^{-\directv z} \cdot \frac{\partial \loss}{\partial \Cuw} \nonumber\\
    &= \alpha_i{\prod_{j=0}^{i-1}(1-\alpha_j)}\cdot e^{-\directv z} \cdot \frac{\partial \loss}{\partial \Cuw}
\end{align}

\begin{align}
    \frac{\partial \loss}{\partial \alpha_i} &=\frac{\partial \Cuw}{\partial \alpha_i}\cdot e^{-\directv z} \cdot \frac{\partial \loss}{\partial \Cuw} \nonumber\\
    &= \left[\bc_i \cdot T_i -\sum_{j=i+1}^{N-1}\bc_j\alpha_j \frac{T_j}{(1-\alpha_i)}\right] \cdot e^{-\directv z} \cdot \frac{\partial \loss}{\partial \Cuw} \nonumber\\
    &= \left[\bc_i \prod_{j=0}^{i-1}(1-\alpha_j) -\sum_{j=i+1}^{N-1}\bc_j\alpha_j \left({\prod_{k=0}^{i-1}(1-\alpha_k)} {\prod_{k=i+1}^{j-1}(1-\alpha_k)} \right)\right]  \cdot e^{-\directv z} \cdot \frac{\partial \loss}{\partial \Cuw}
\end{align}

\begin{align}
    \frac{\partial \loss}{\partial \directv} &=
    -\sum_{i=1}^{N}\left( \bc_i\alpha_iT_i \right)\cdot e^{-\directv z} \cdot \frac{\partial \loss}{\partial \Cuw} \nonumber\\
    &=-\sum_{i=1}^{N}\left( \bc_i\alpha_i\prod_{j=0}^{i-1}(1-\alpha_j) \right)\cdot e^{-\directv z} \cdot \frac{\partial \loss}{\partial \Cuw}
\end{align}

\begin{align}
    \frac{\partial \loss}{\partial \bsv } &= \left( \frac{\partial \loss}{\partial \Cuw}  \right) \binf e^{-\bsv z} 
    + \lambest 
\end{align}

\begin{align}
    \frac{\partial \loss}{\partial \binf} &= \left( \frac{\partial \loss}{\partial \Cuw}  \right) (1-e^{-\bsv z})
    + \lambest 
\end{align}

\subsection{Backscatter Estimation}
\input{figures/appendix/fig_attenuation}

The backscatter estimation algorithm is outlined in ~\autoref{Alg:1}, which calls ~\autoref{Alg:2} as a subroutine. A visual explanation of underwater attenuation is provided in ~\autoref{fig:atten_figure}.
{
    \SetKwComment{Comment}{}{}
    \begin{algorithm}
\KwIn{$I_{sm}, z_{sm}, p_{dark}=0.01, intervals\_num=25, resized\_height=300$}
\begin{enumerate}
    \item Resize $I_{sm}$ to $resized\_height$
    \item Set negative values in $I_{sm}$ to 0
    \item Resize $z_{sm}$ to $resized\_height$
    \item Set negative values in $z_{sm}$ to 0
    \item $darkZ, B_c \leftarrow \text{getBackscatterByCurveFittingMultipleImages}(I_{sm}, z_{sm}, p_{dark})$
    \Comment{(\ie, \autoref{Alg:2})}
    \item $intervals \leftarrow \text{linspace}(\min(darkZ), \max(darkZ), intervals\_num)$
    \item Initialize $min\_values\_depth$ and $minvals$ as empty lists for each color channel
    \item \textbf{for} each color channel $k$ \textbf{do}
    \begin{enumerate}
        \item \textbf{for} each interval $i$ in $intervals$ \textbf{do}
        \begin{enumerate}
            \item Find indices $ind$ within the current interval
            \item \textbf{if} valid data exists \textbf{then}
            \begin{enumerate}
                \item Append $\min(B_c[ind, k])$ to $minvals[k]$
                \item Append corresponding $darkZ$ value to $min\_values\_depth[k]$
            \end{enumerate}
        \end{enumerate}
    \end{enumerate}
    \item Initialize $out$ as a zero matrix
    \item \textbf{for} each color channel $k$ \textbf{do}
    \begin{enumerate}
        \item Fit model to $min\_values\_depth[k]$ and $minvals[k]$
        \item Extract parameters $b_{\infty}$ and $b_{cb}$ into $out$
    \end{enumerate}
    \item $\binf \leftarrow out[0, :]$
    \item $\bsv \leftarrow out[1, :]$
    \item \textbf{return} $\{\binf, \bsv\}$
\end{enumerate}
\caption{Estimate Backscatter}
\label{Alg:1}
\end{algorithm}

\begin{algorithm}
\KwIn{$I, z_{sm}, pdark=0.01, rhoflag=0, edges\_num=10$}
\begin{enumerate}   
    \item Initialize $edges$ with $edges\_num$ evenly spaced values between $\min(z_{sm})$ and $\max(z_{sm})$
    \item Compute $zcluster = \text{clusterRange}(z_{sm}, edges)$
    \item Initialize $maskRho$ as a zero matrix with the same shape as $z_{sm}$
    \item \textbf{for} $i$ in range($\text{len}(edges) - 1$) \textbf{do}
    \begin{enumerate}
        \item Set $thisMask = (zcluster == i)$
        \item \textbf{if} $\sum(thisMask) > 0$ \textbf{then}
        \begin{enumerate}
            \item Extract $thisRho$ using $thisMask$
            \item Find darkest pixels in $thisRho$ with threshold $pdark$
            \item Update $maskRho$ by adding $thisMaskRho$
        \end{enumerate}
    \end{enumerate}
    \item Convert $maskRho$ to boolean
    \item Extract darkest pixels and their corresponding depths using $maskRho$ and $dm$, resulting in $darkZ$ and $Bc$
    \item \textbf{return} $darkZ, Bc$
\end{enumerate}
\caption{Get Backscatter By Curve Fitting Multiple Images}
\label{Alg:2}
\end{algorithm}
}
\clearpage
\twocolumn 
\subsection{Experimental Setup}

We adopted hyperparameters consistent with those used in \cite{kerbl3Dgaussians}, as they have shown to yield optimal results empirically. For underwater-specific parameters, we maintained the same learning rates as those used for color. Our data splits involved testing every 8 images, and we utilized the Adam optimizer.

\begin{itemize}
    \item \textbf{Iterations:} 30,000
    \item \textbf{Position Learning Rate Initial:} 0.00016
    \item \textbf{Position Learning Rate Final:} 0.0000016
    \item \textbf{Position Learning Rate Delay Multiplier:} 0.01
    \item \textbf{Position Learning Rate Maximum Steps:} 30,000
    \item \textbf{Feature Learning Rate:} 0.0025
    \item \textbf{Direct Volume Absorption Learning Rate:} 0.0025
    \item \textbf{Backscatter Learning Rate:} 0.0025
    \item \textbf{Opacity Learning Rate:} 0.05
    \item \textbf{Scaling Learning Rate:} 0.005
    \item \textbf{Rotation Learning Rate:} 0.001
    \item \textbf{Lambda SSIM:} 0.3
    \item \textbf{Lambda Backscatter:} 0.1
    \item \textbf{Densification Interval:} 100
    \item \textbf{Opacity Reset Interval:} 3000
    \item \textbf{Densify From Iteration:} 1500
    \item \textbf{Densify Until Iteration:} 15,000
    \item \textbf{Densify Gradient Threshold:} 0.0002
    \item \textbf{Minimum Opacity Threshold:} 0.1
\end{itemize}

\subsection{License Restrictions}

Utilizing the \cite{kerbl3Dgaussians} code package entails adhering to the following license restrictions:

\begin{itemize}
    \item \textbf{Redistribution:} You may reproduce or distribute the Work only if (a) you do so under this License, (b) include a complete copy of this License with your distribution, and (c) retain all copyright, patent, trademark, or attribution notices present in the Work.
    \item \textbf{Derivative Works:} You may specify additional or different terms for your derivative works of the Work ("Your Terms") only if (a) Your Terms include the use limitation in Section 2, and (b) you identify the specific derivative works subject to Your Terms. The original License still applies to the Work itself.
    \item \textbf{Other Uses:} Any other use without prior consent of the Licensors is prohibited. Research users must ensure they have all necessary information to use the Software safely.
    \item \textbf{Publications:} If using the Software for a publication, users are encouraged to cite the relevant publications as detailed in the Software documentation.
\end{itemize}

Additionally, the dataset distribution restrictions from \cite{levy2023seathru} are also enforced.

\clearpage

%% file: figures/appendix/appendix_fig_examples.tex
\graphicspath{ {./figures/appendix/figs} }
\begin{figure*}[!ht]
\newcommand{\myW}{0.24\linewidth}
\newcommand{\myH}{2cm} 
\centering
\subcaptionbox{Ground Truth (train)}[\myW]
{\includegraphics[width=\linewidth, height=\myH]{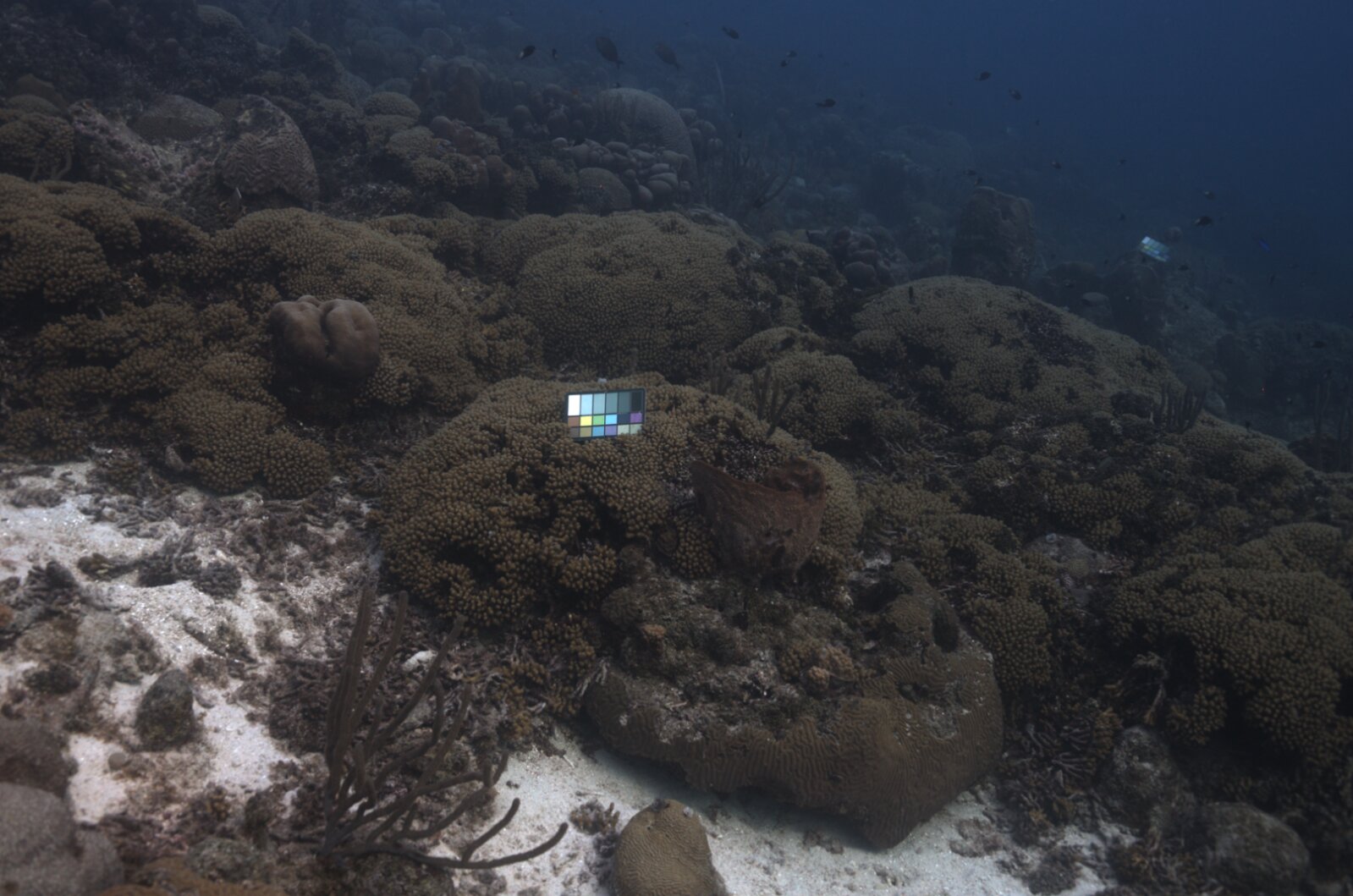}\\
 \includegraphics[width=\linewidth, height=\myH]{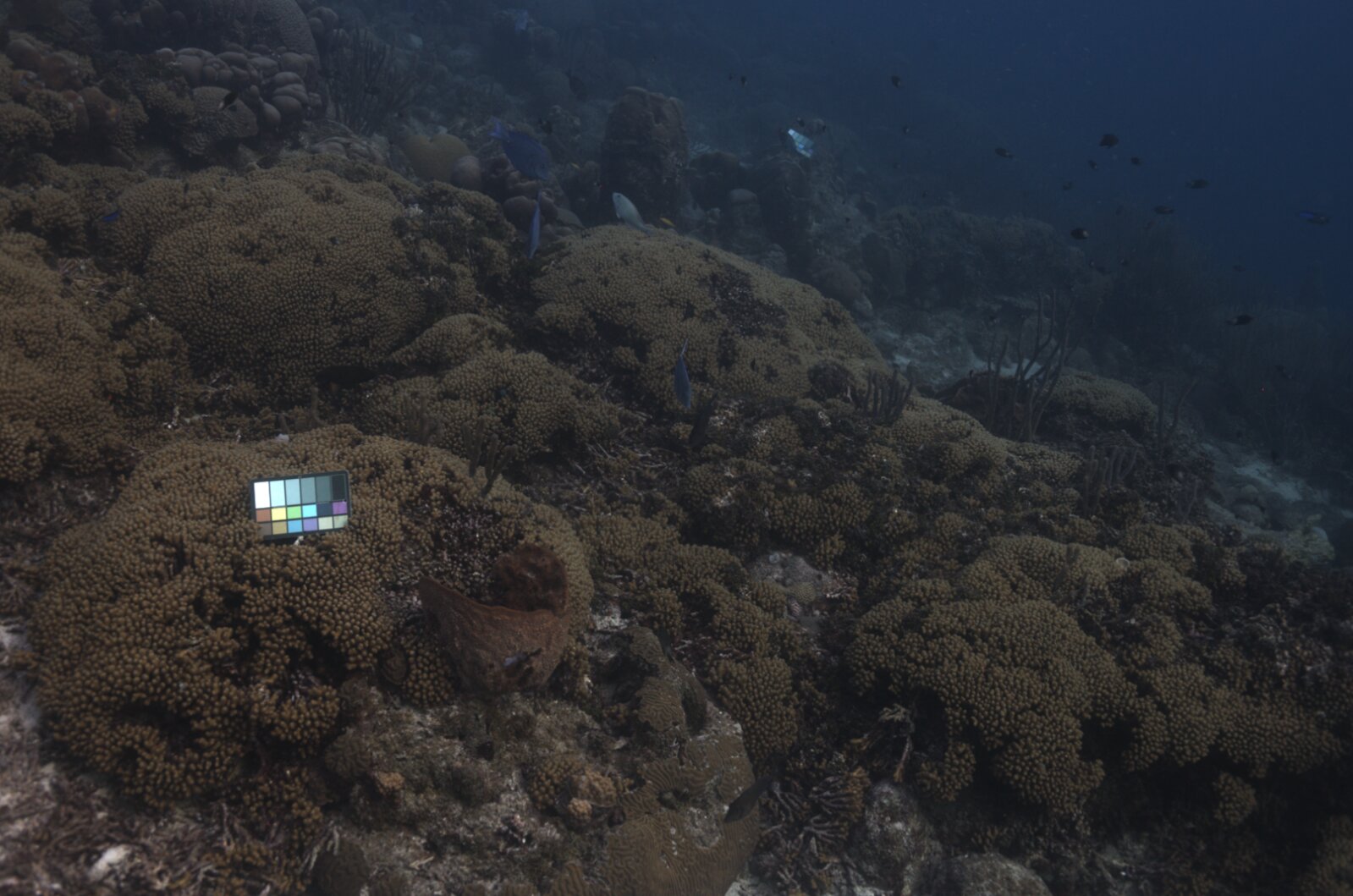}\\
 \includegraphics[width=\linewidth, height=\myH]{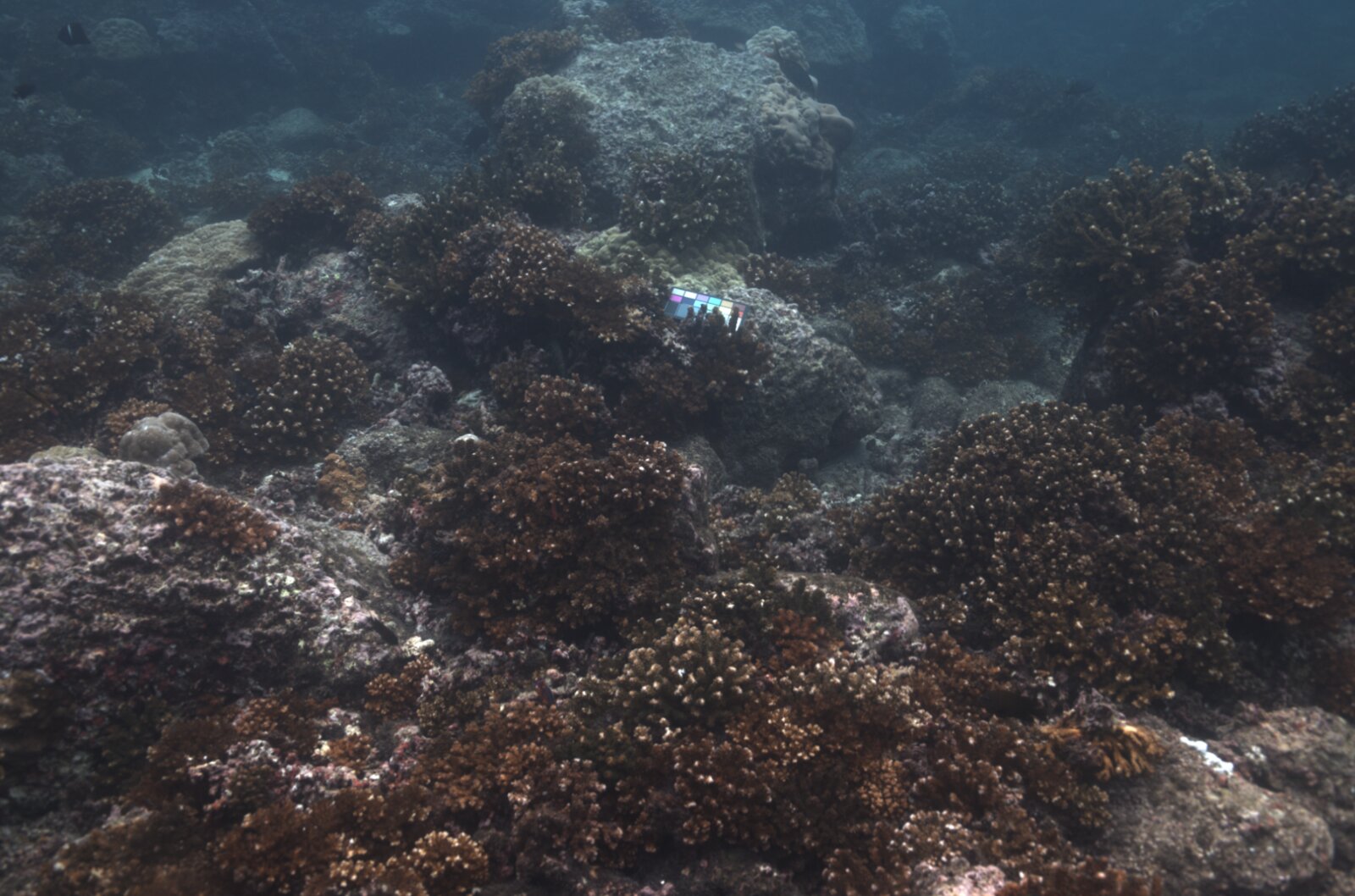}\\
 \includegraphics[width=\linewidth, height=\myH]{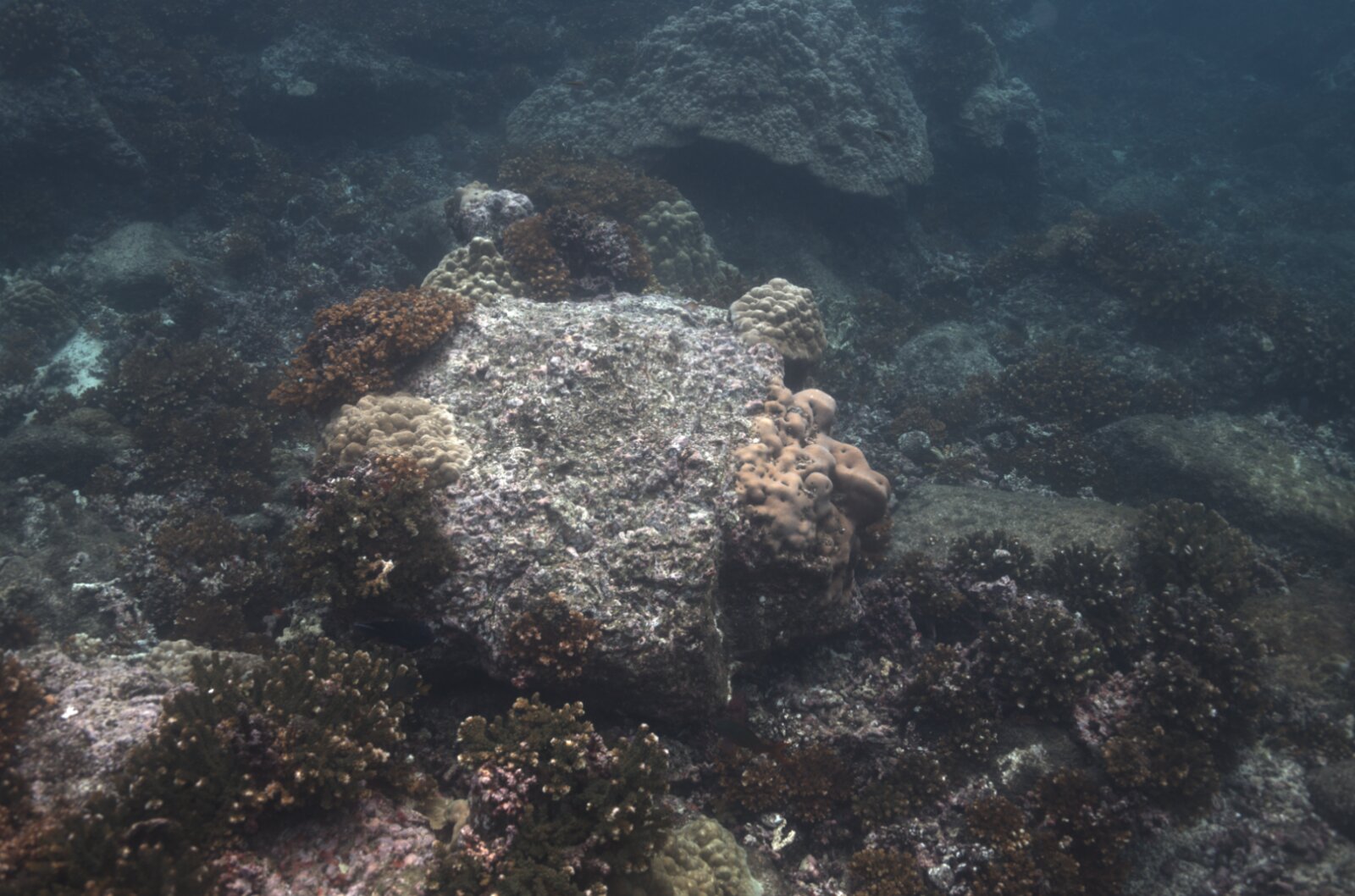}\\
 \includegraphics[width=\linewidth, height=\myH]{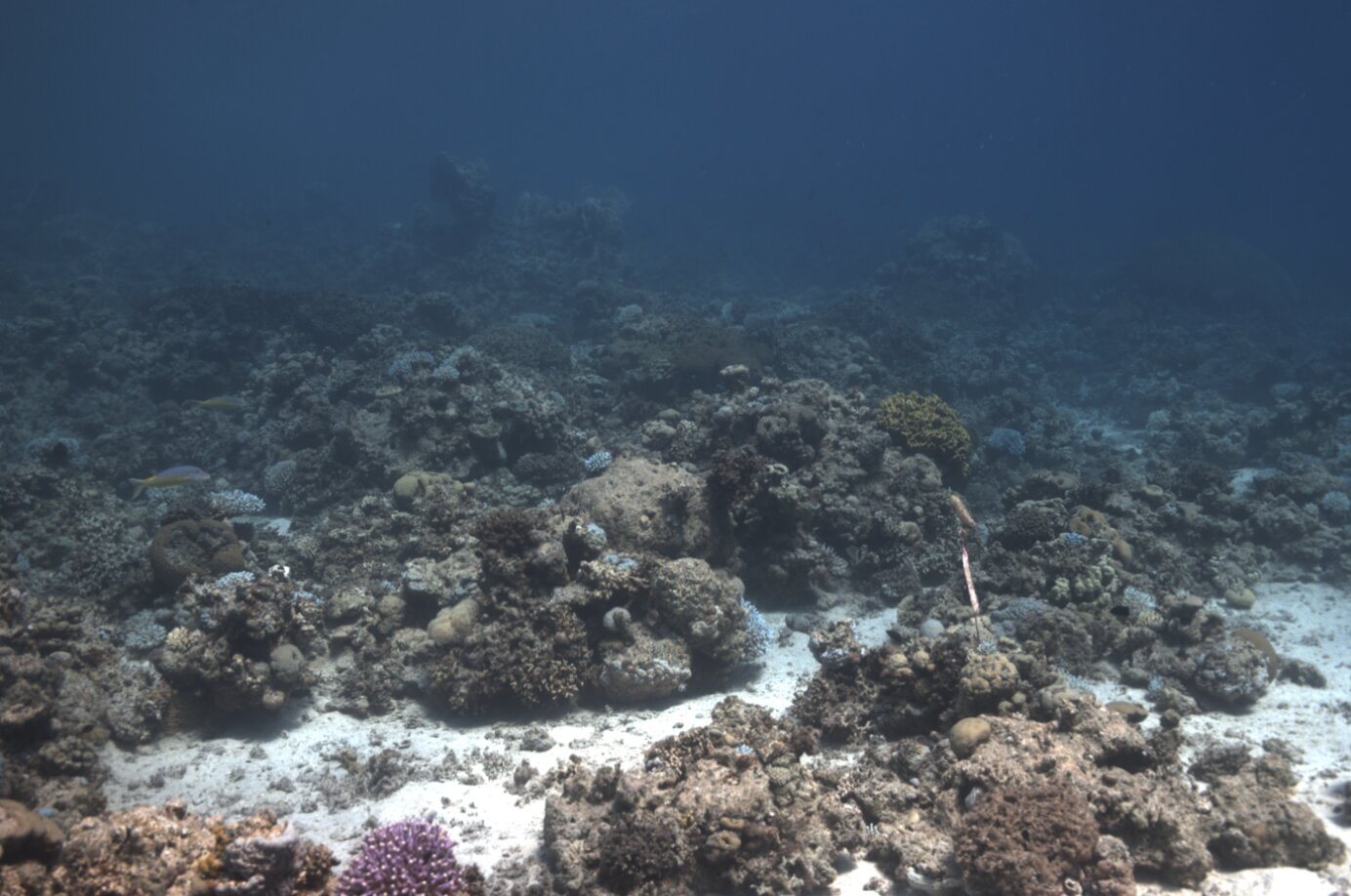}\\
 \includegraphics[width=\linewidth, height=\myH]{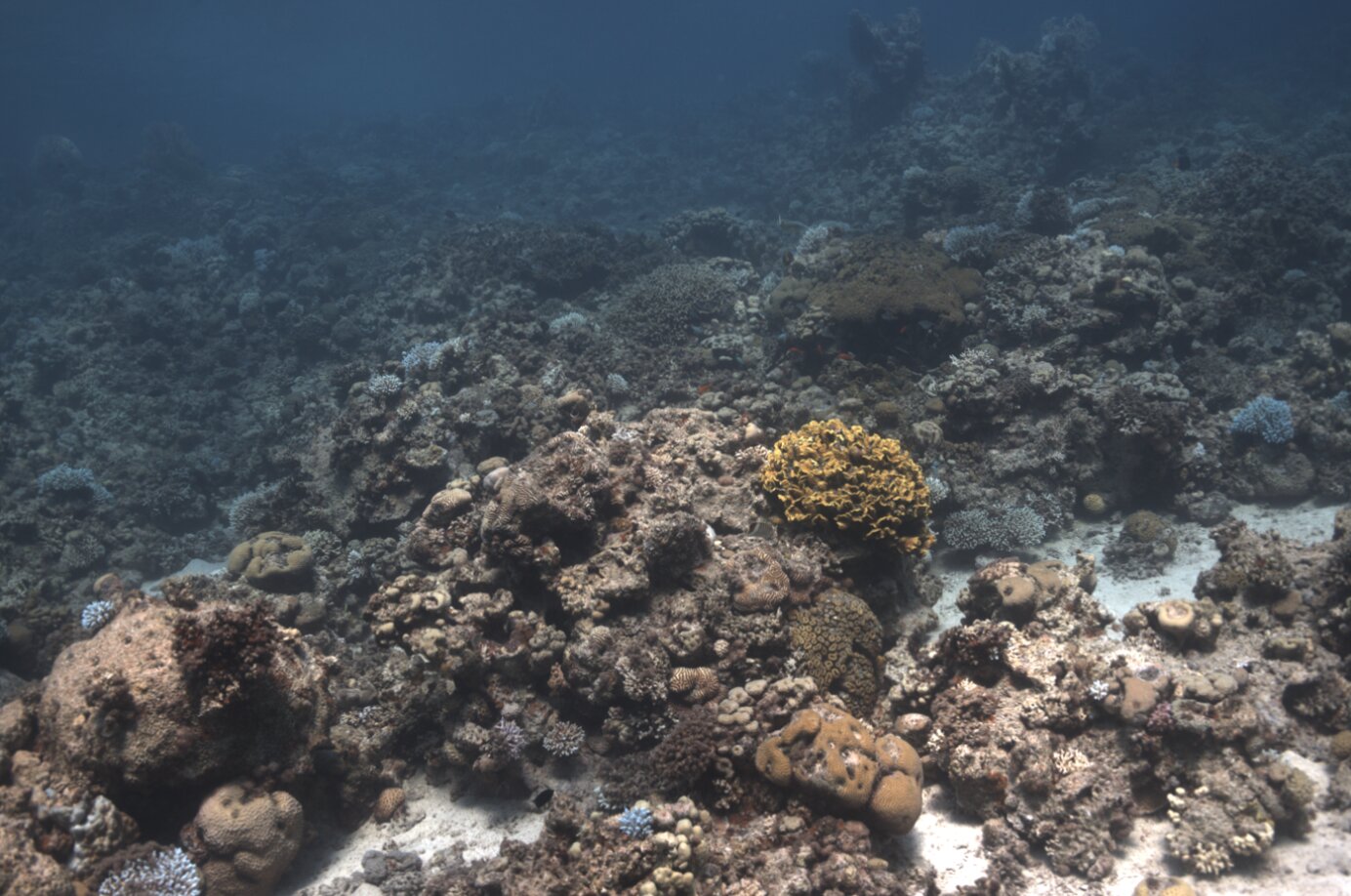}}
\subcaptionbox{Rendered (train)}[\myW]
{\includegraphics[width=\linewidth, height=\myH]{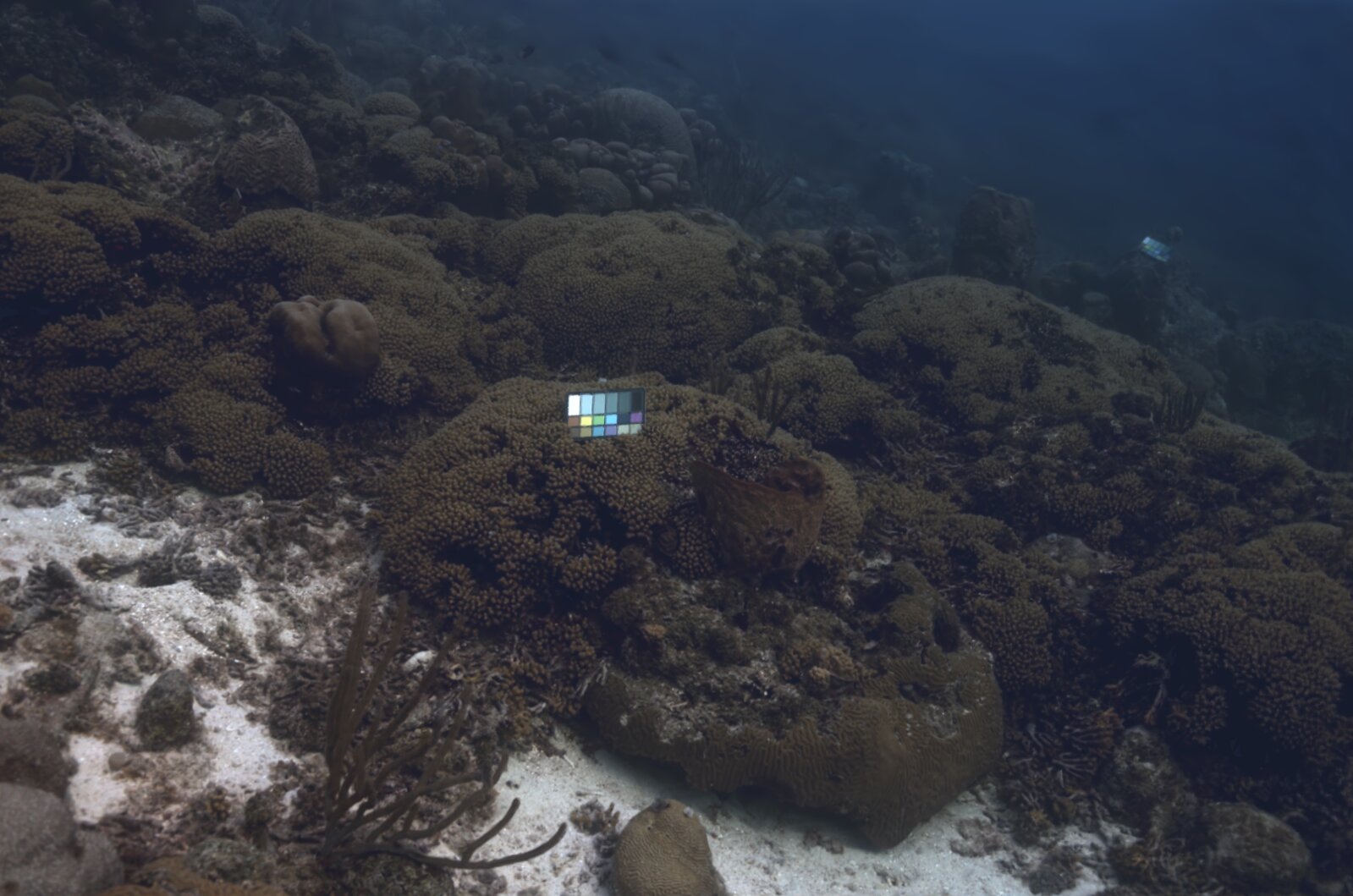}\\
 \includegraphics[width=\linewidth, height=\myH]{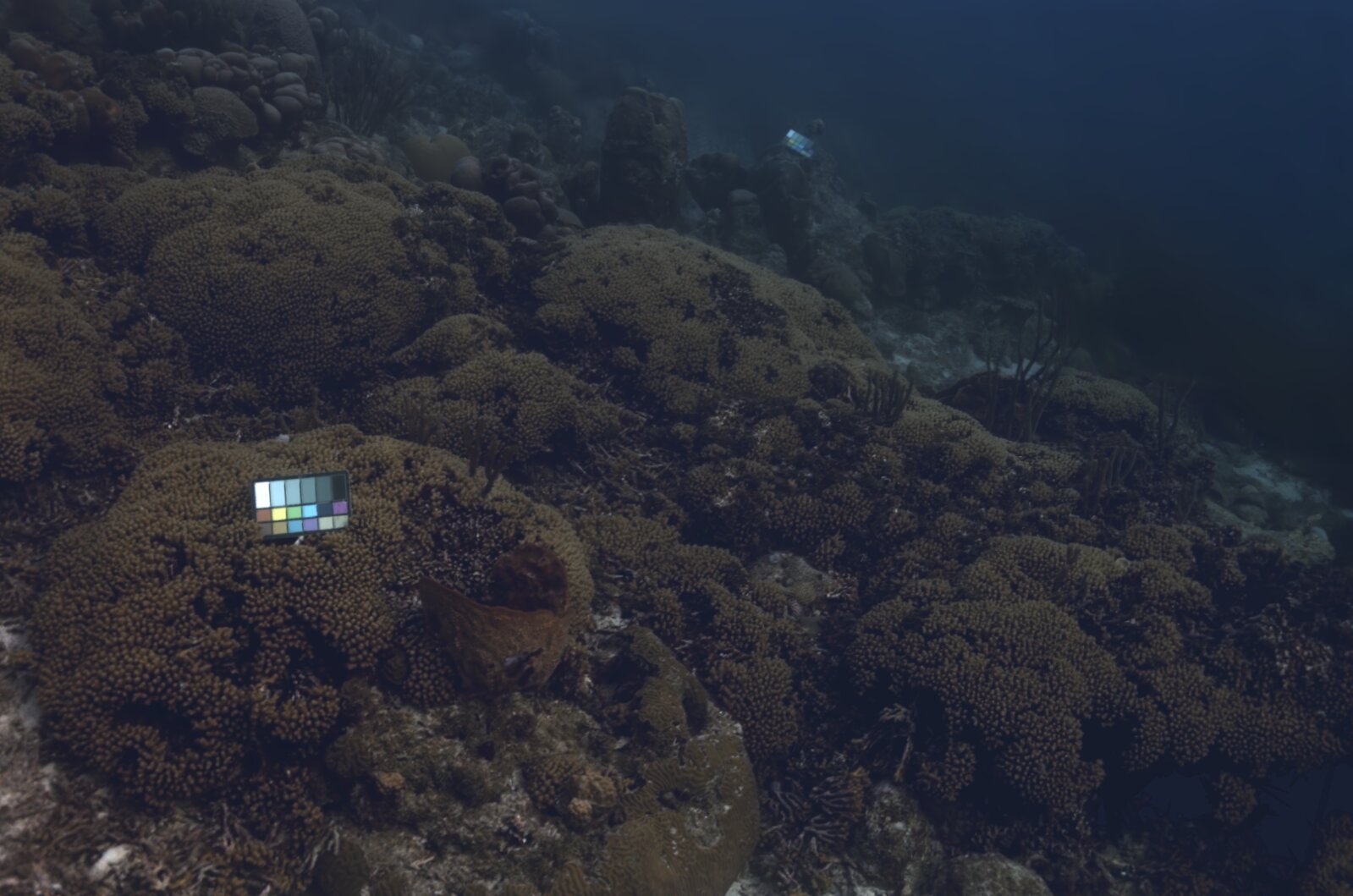}\\
 \includegraphics[width=\linewidth, height=\myH]{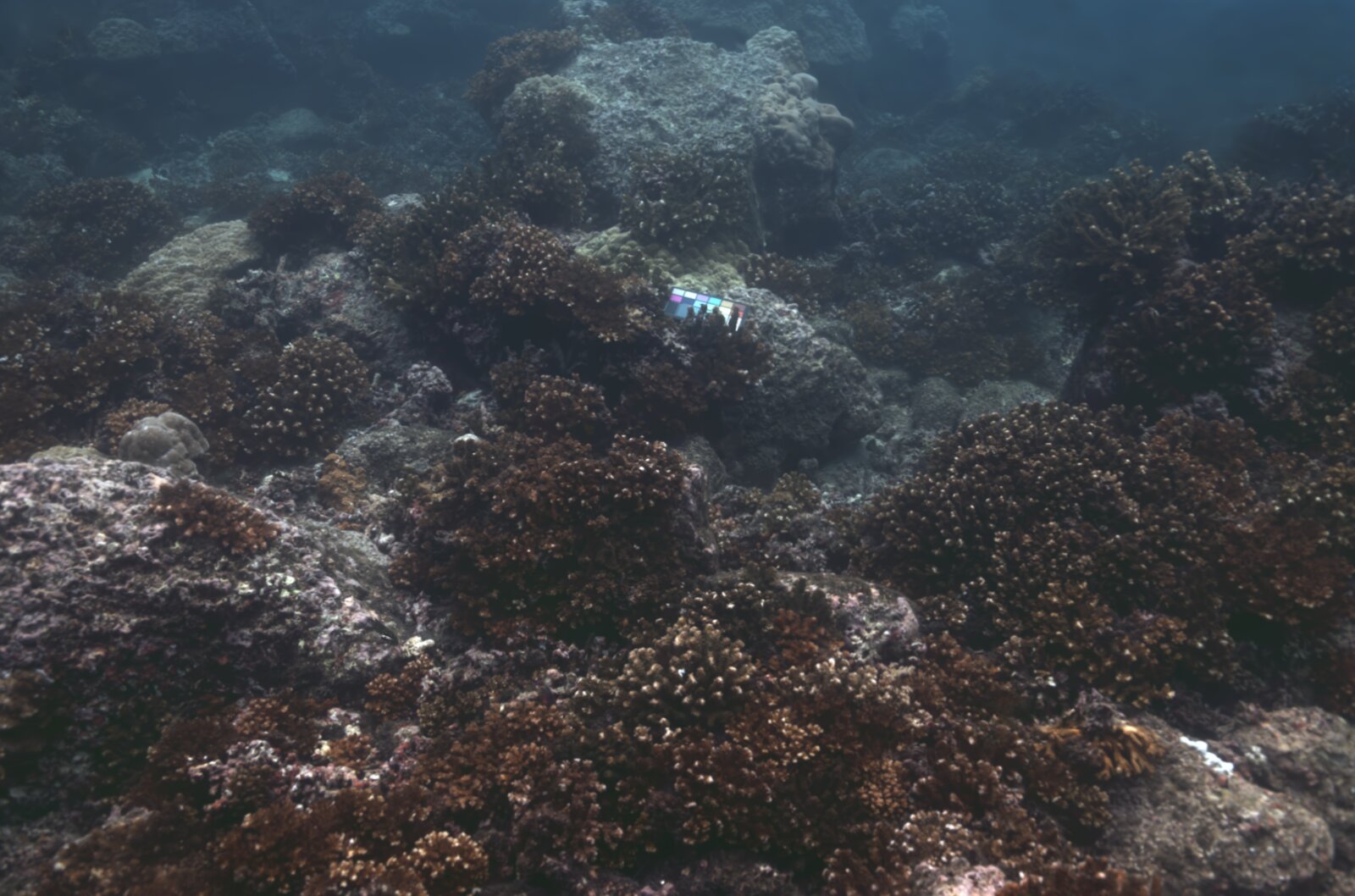}\\
 \includegraphics[width=\linewidth, height=\myH]{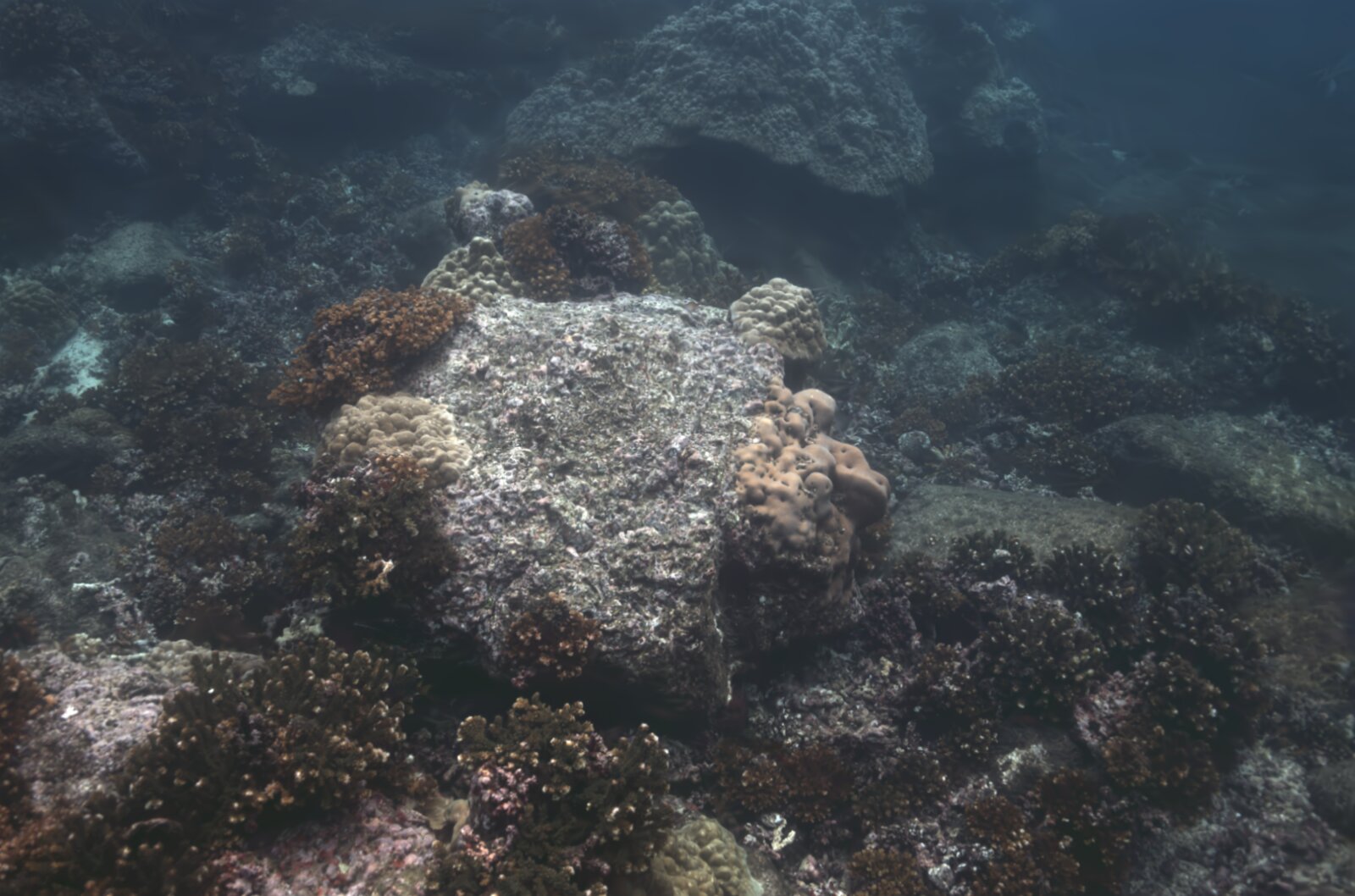}\\
 \includegraphics[width=\linewidth, height=\myH]{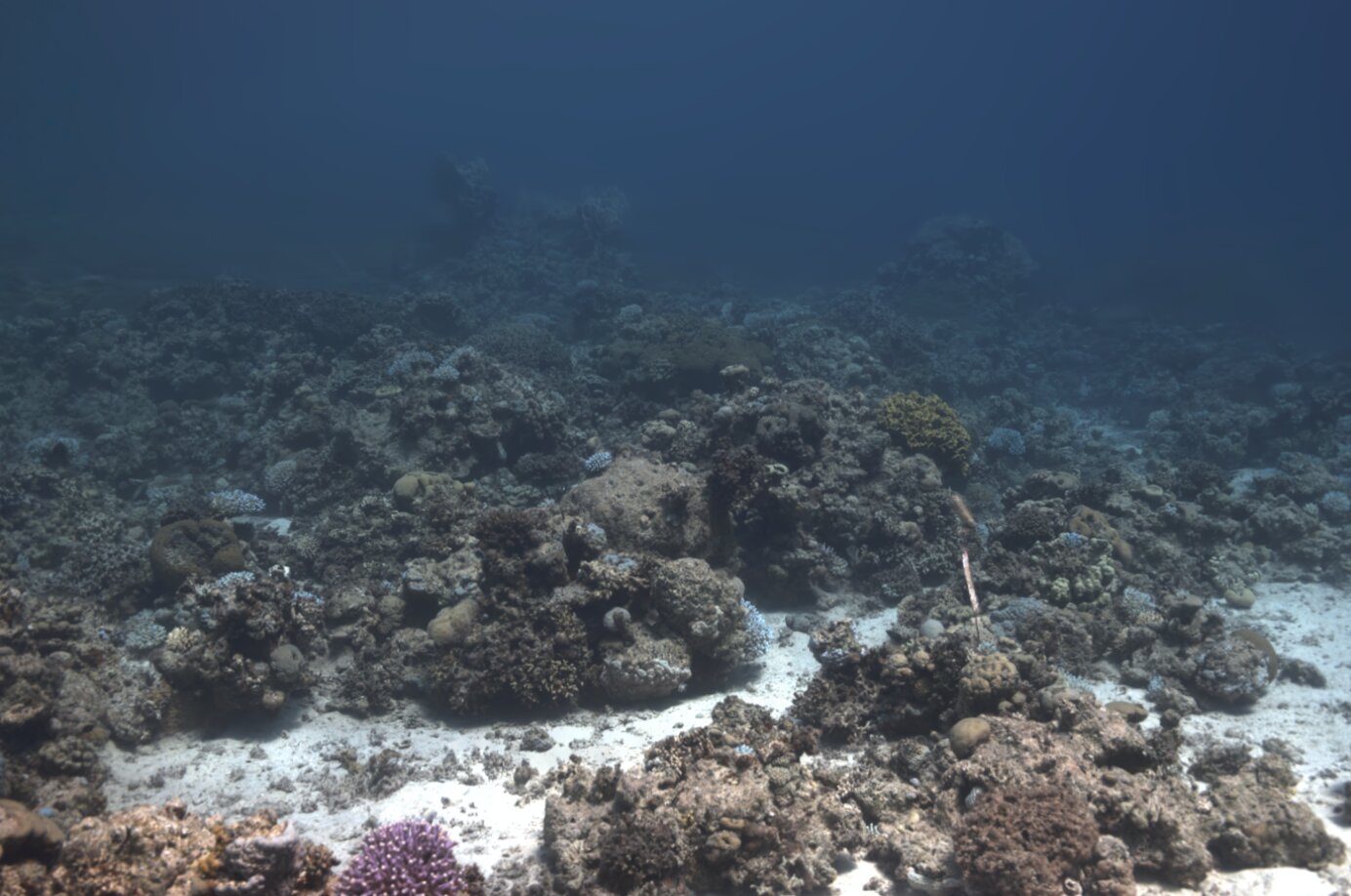}\\
 \includegraphics[width=\linewidth, height=\myH]{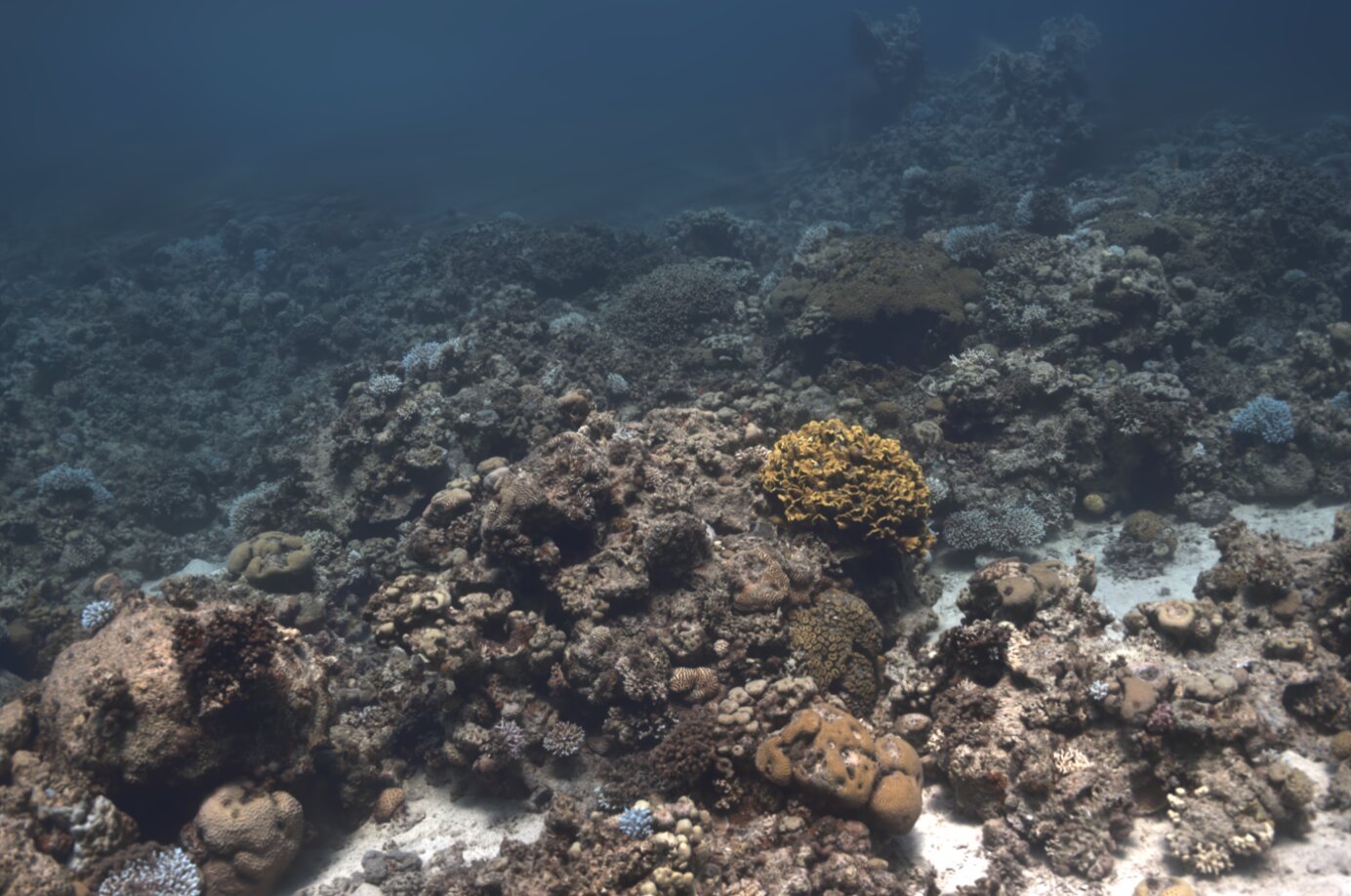}}
\subcaptionbox{Ground Truth (test)}[\myW]
{\includegraphics[width=\linewidth, height=\myH]{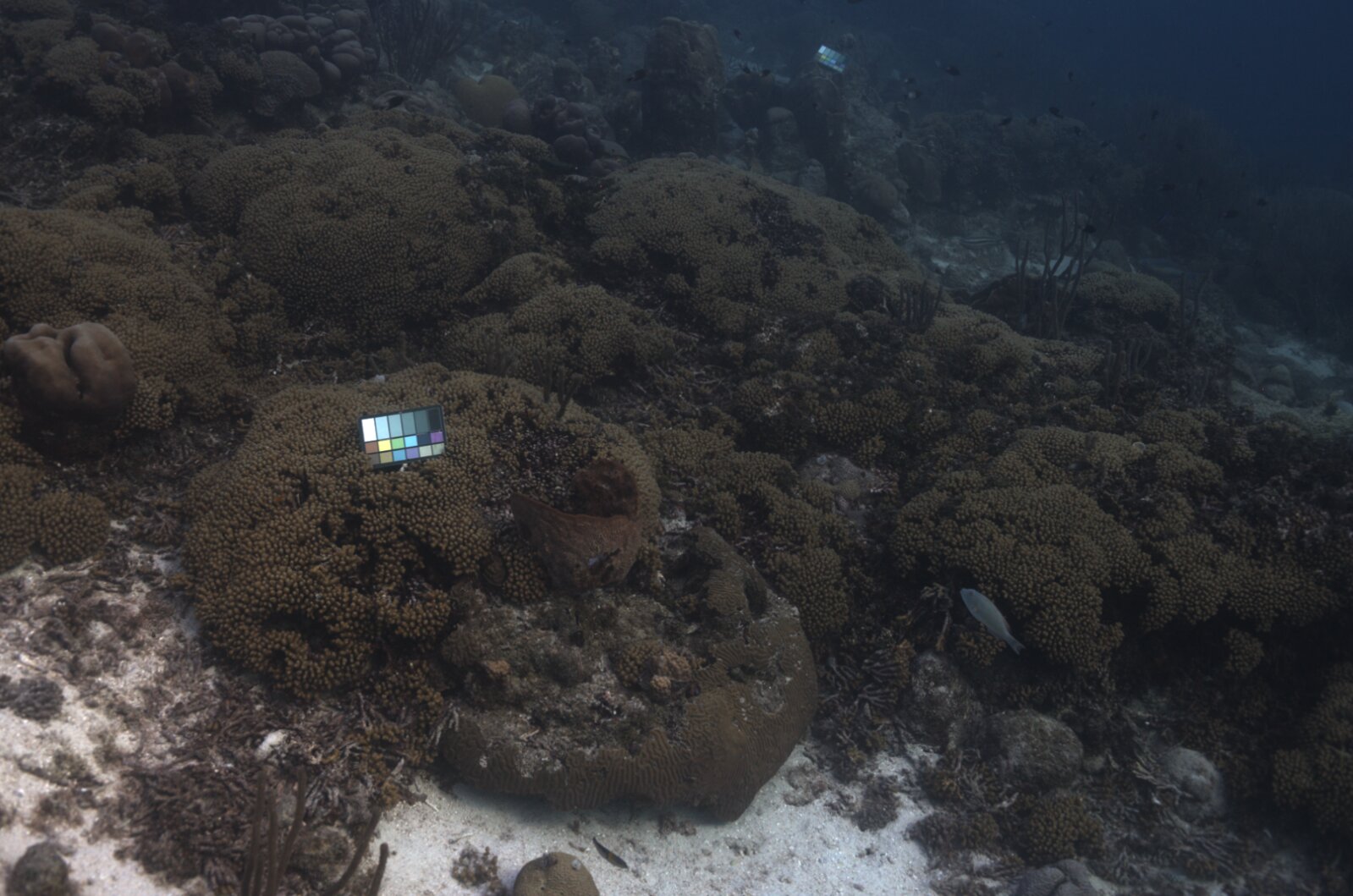}\\
 \includegraphics[width=\linewidth, height=\myH]{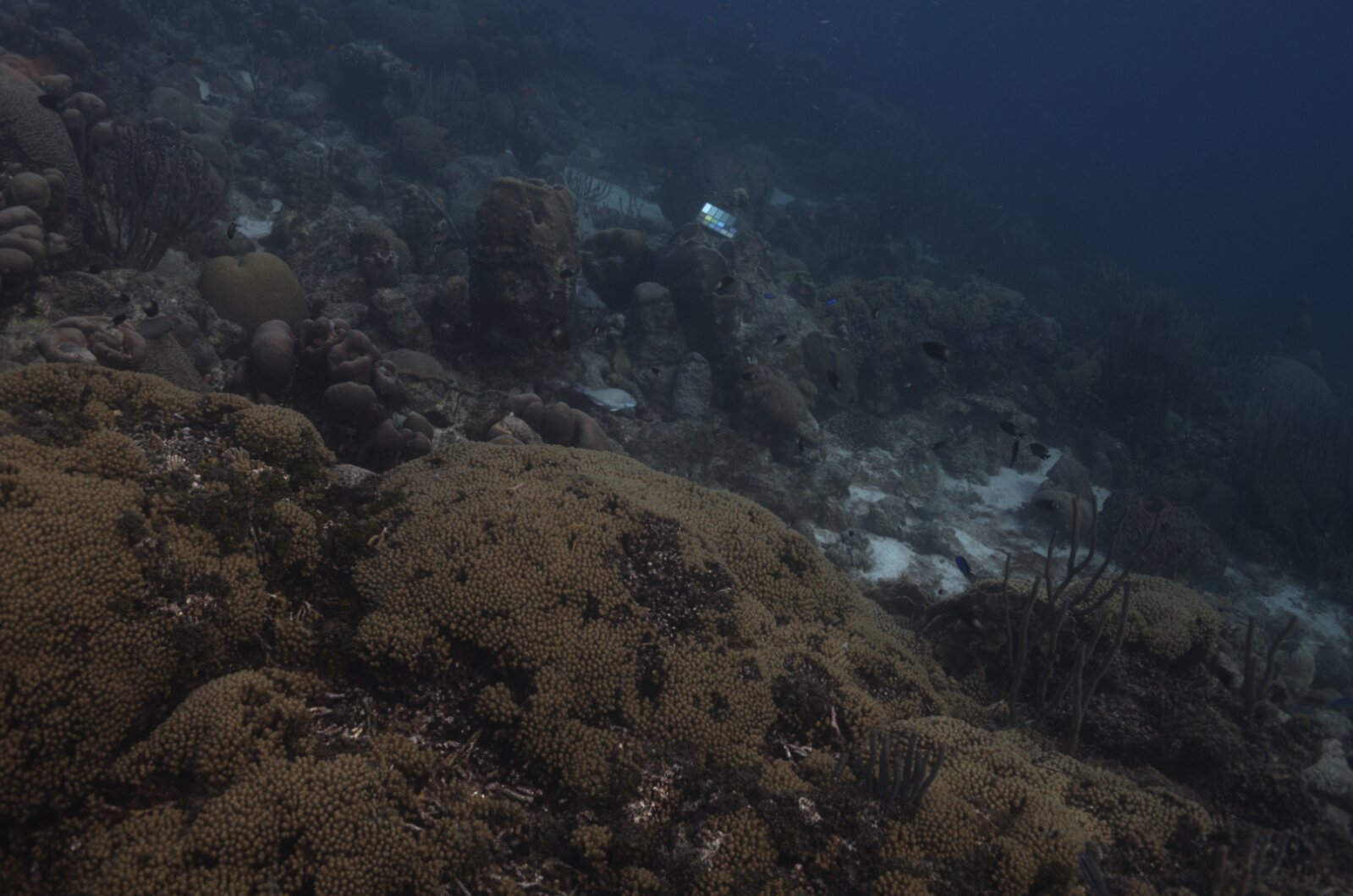}\\
 \includegraphics[width=\linewidth, height=\myH]{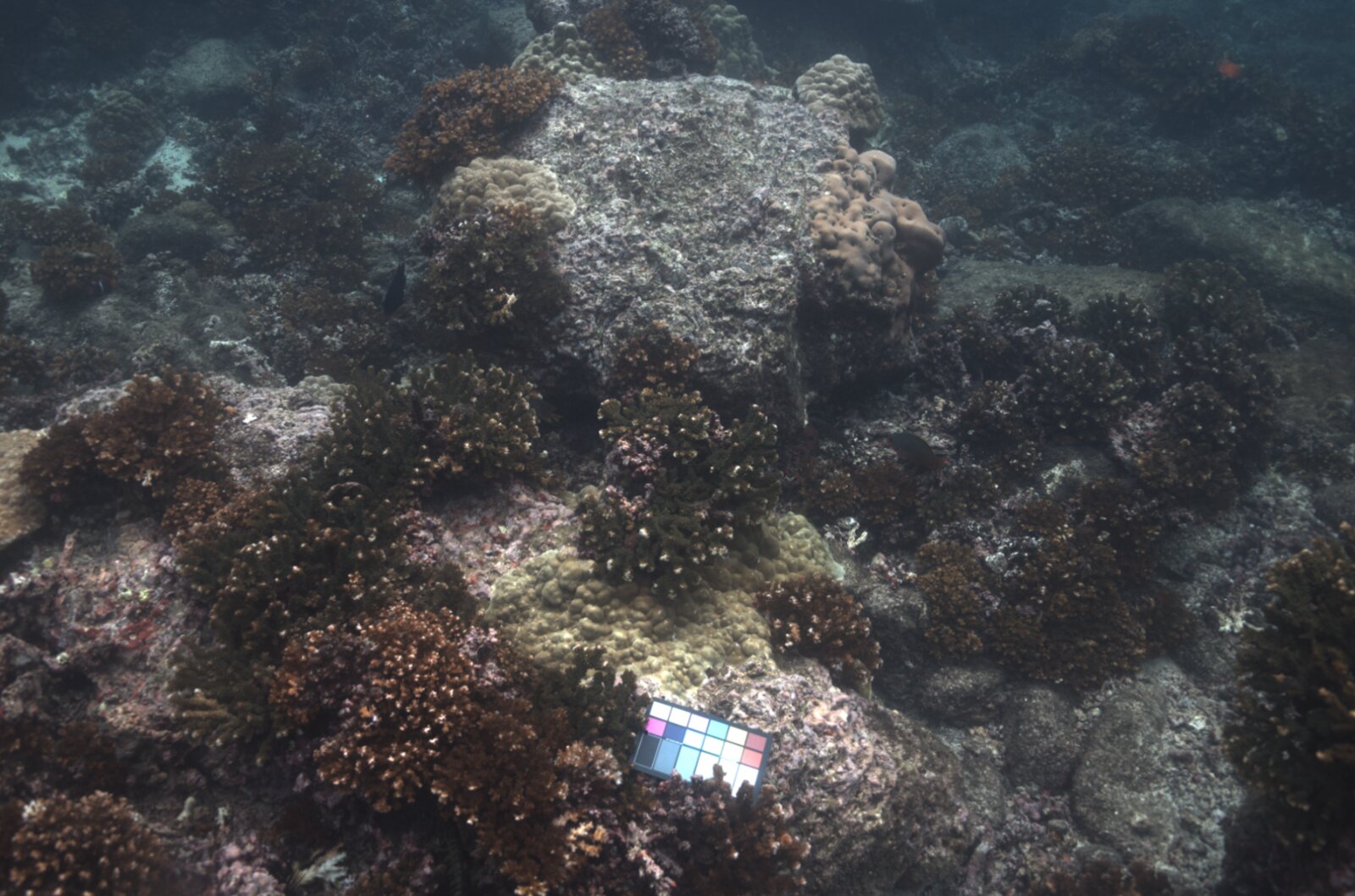}\\
 \includegraphics[width=\linewidth, height=\myH]{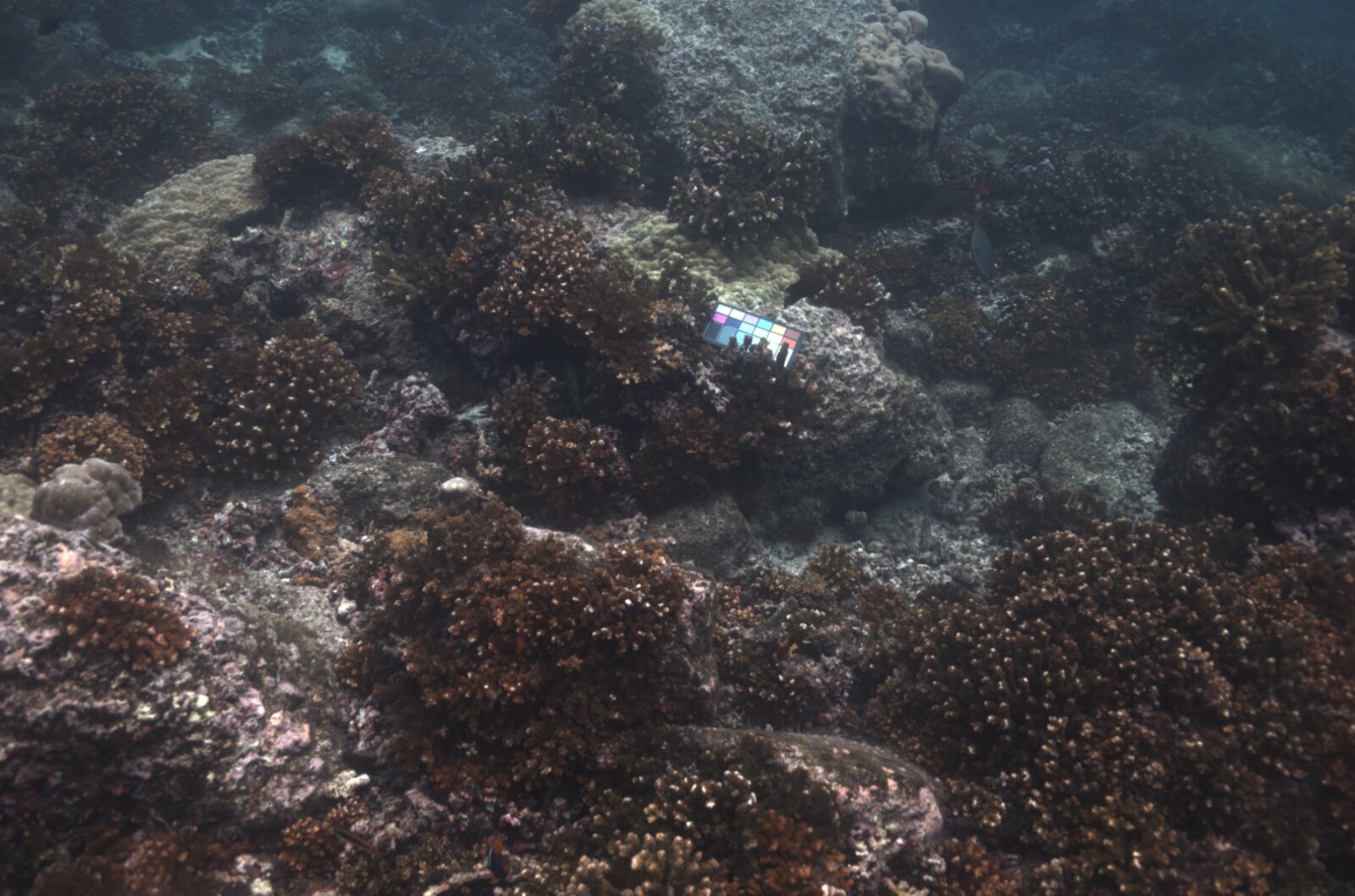}\\
 \includegraphics[width=\linewidth, height=\myH]{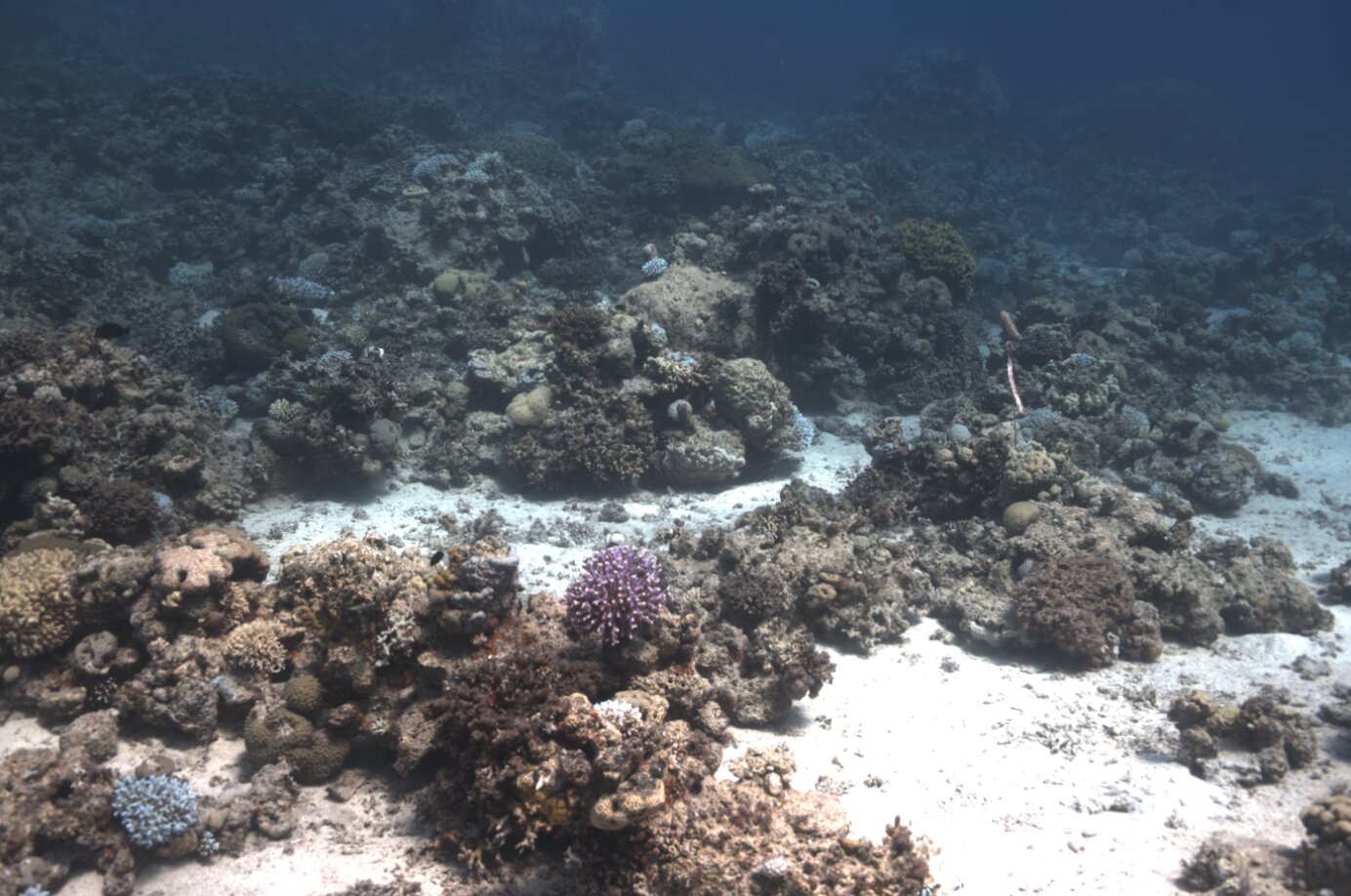}\\
 \includegraphics[width=\linewidth, height=\myH]{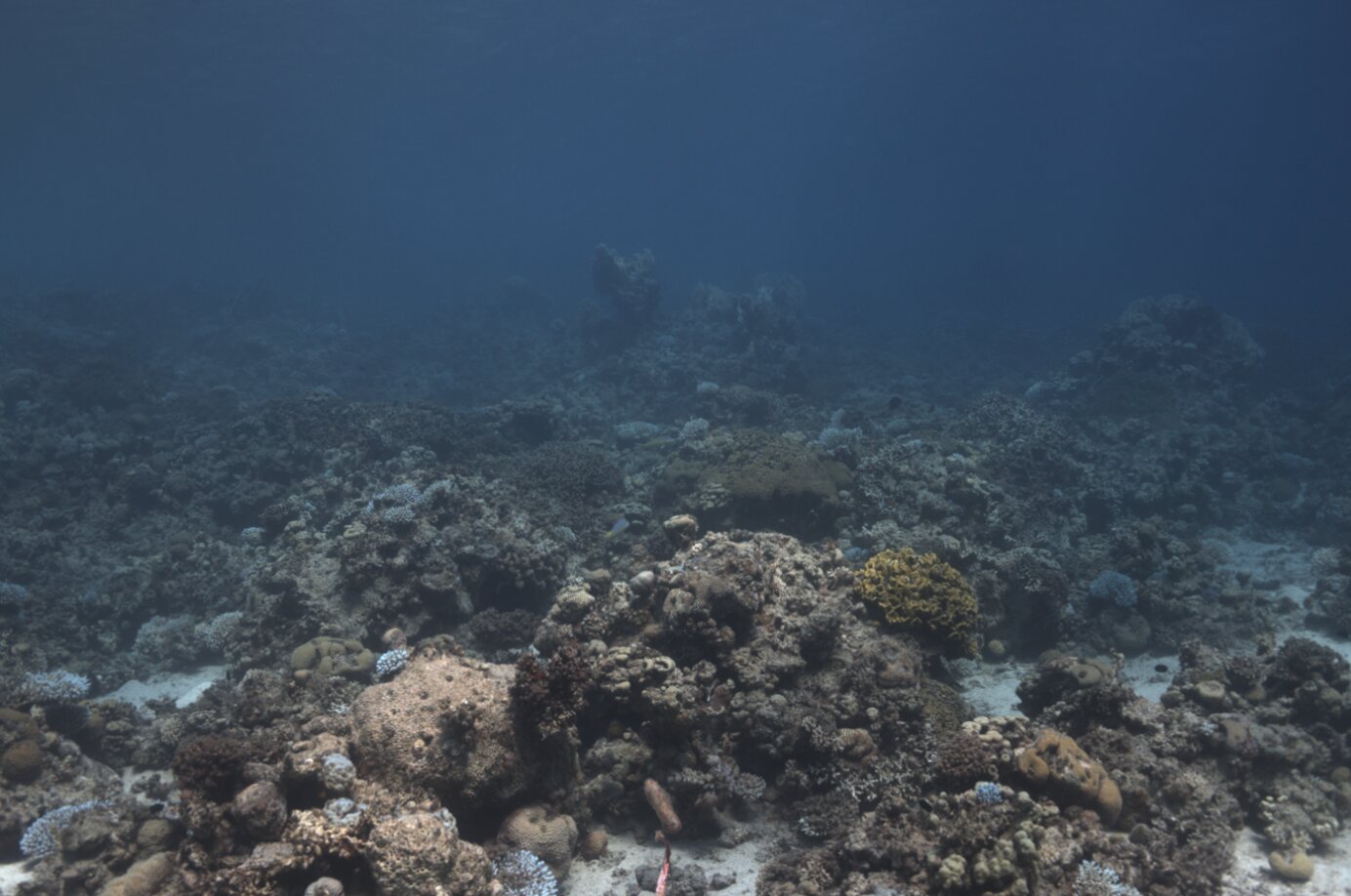}}
 \subcaptionbox{Rendered (test)}[\myW]
{\includegraphics[width=\linewidth, height=\myH]{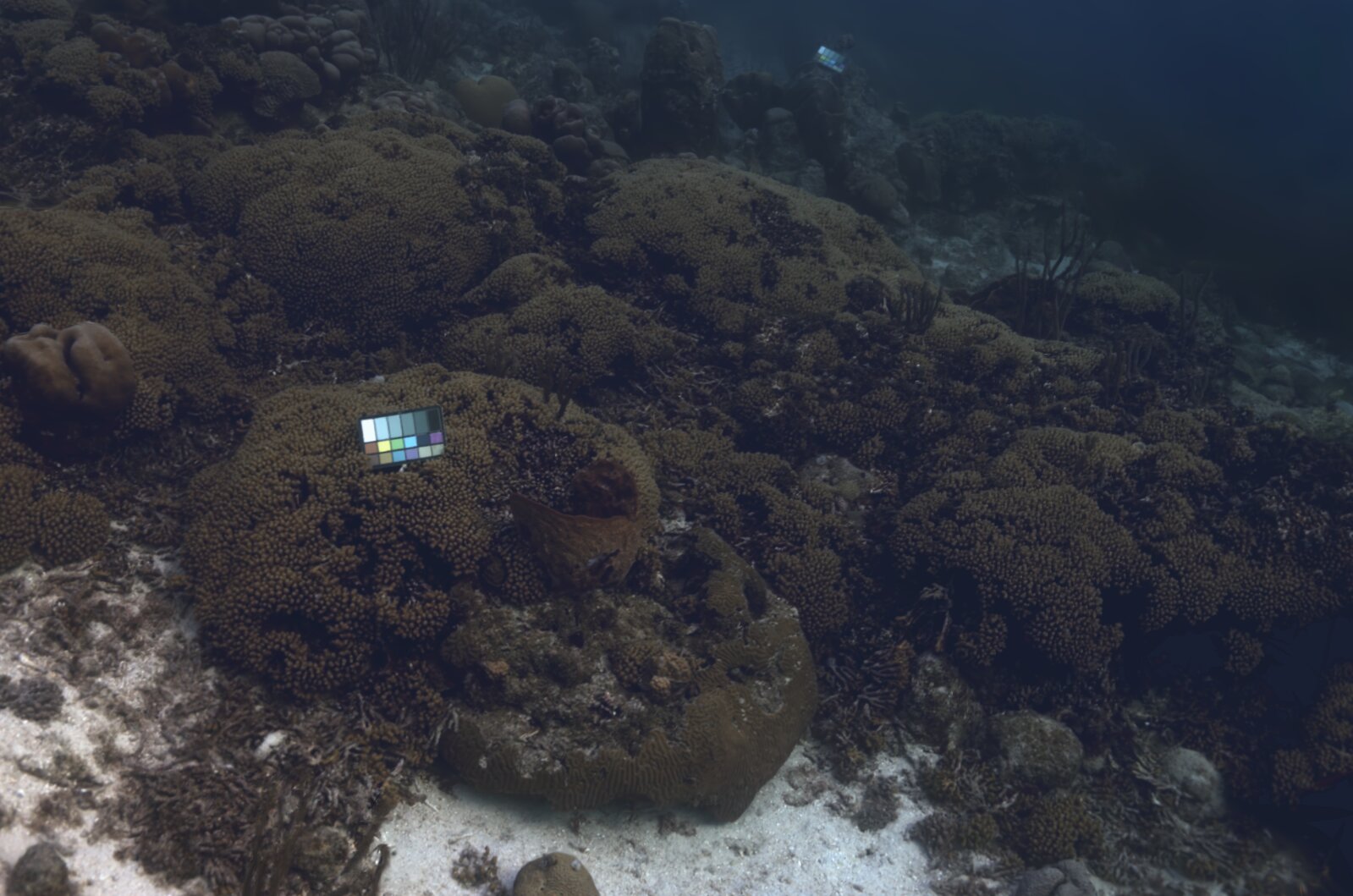}\\
 \includegraphics[width=\linewidth, height=\myH]{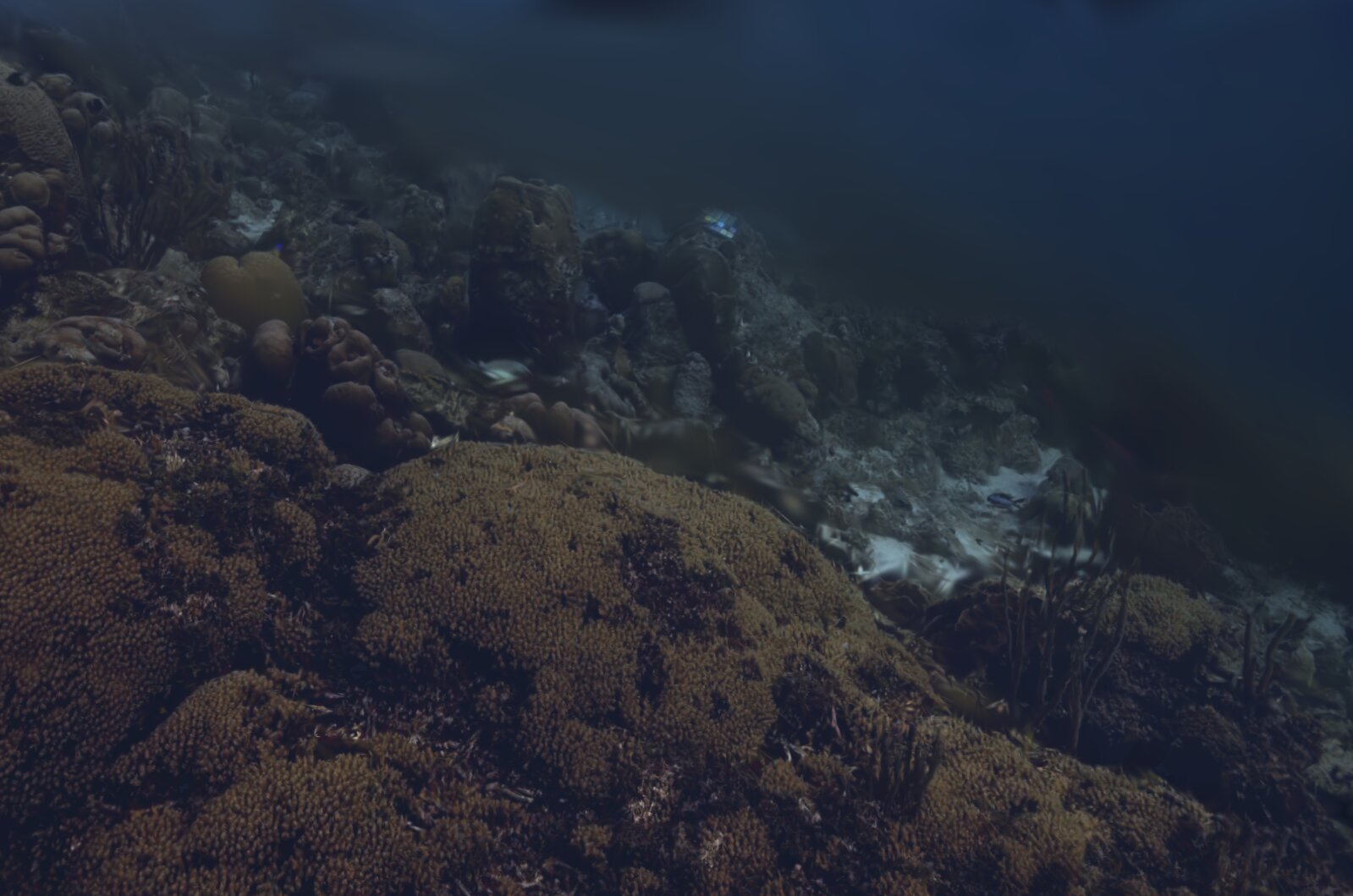}\\
 \includegraphics[width=\linewidth, height=\myH]{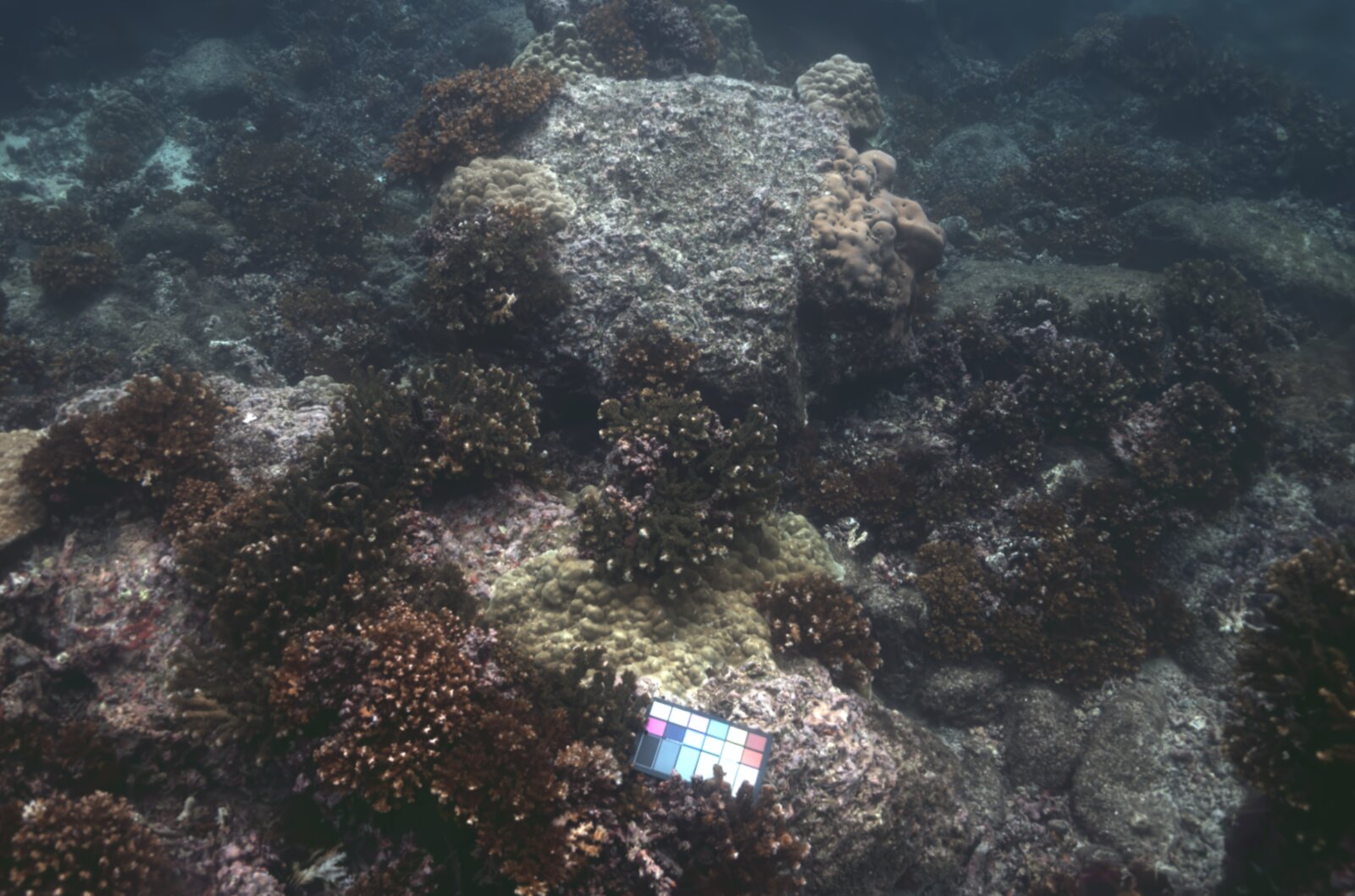}\\
 \includegraphics[width=\linewidth, height=\myH]{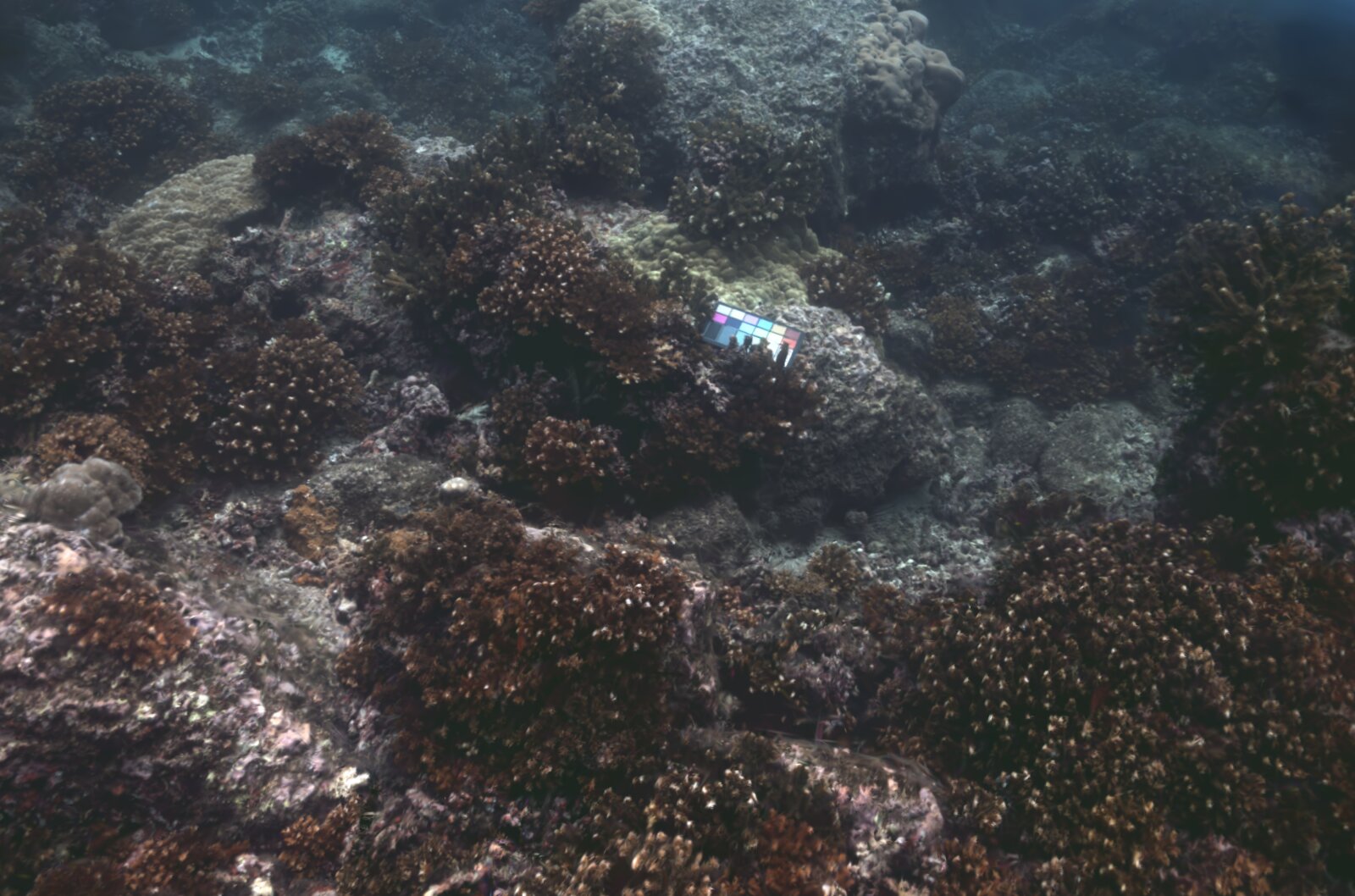}\\
 \includegraphics[width=\linewidth, height=\myH]{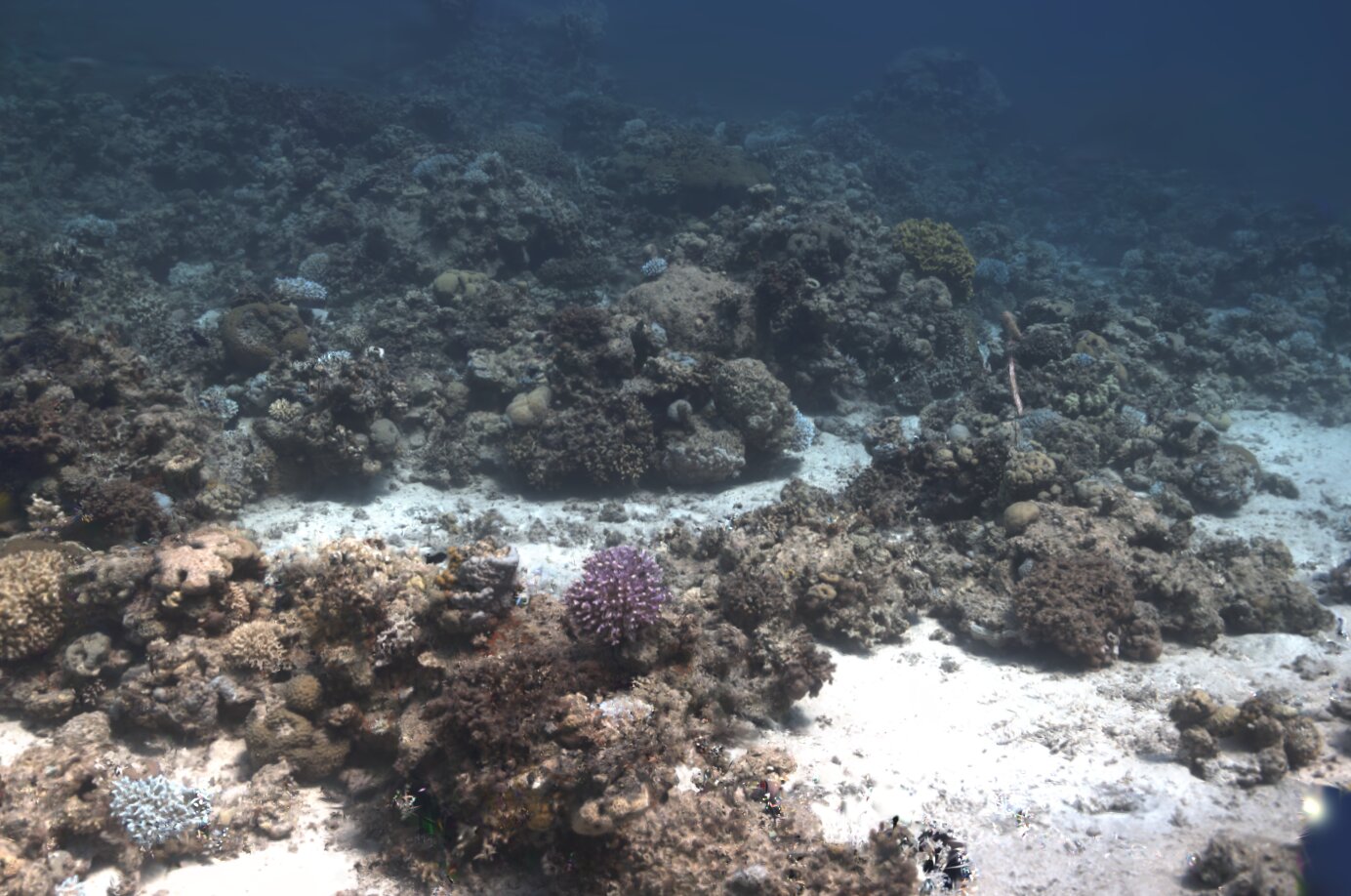}\\
 \includegraphics[width=\linewidth, height=\myH]{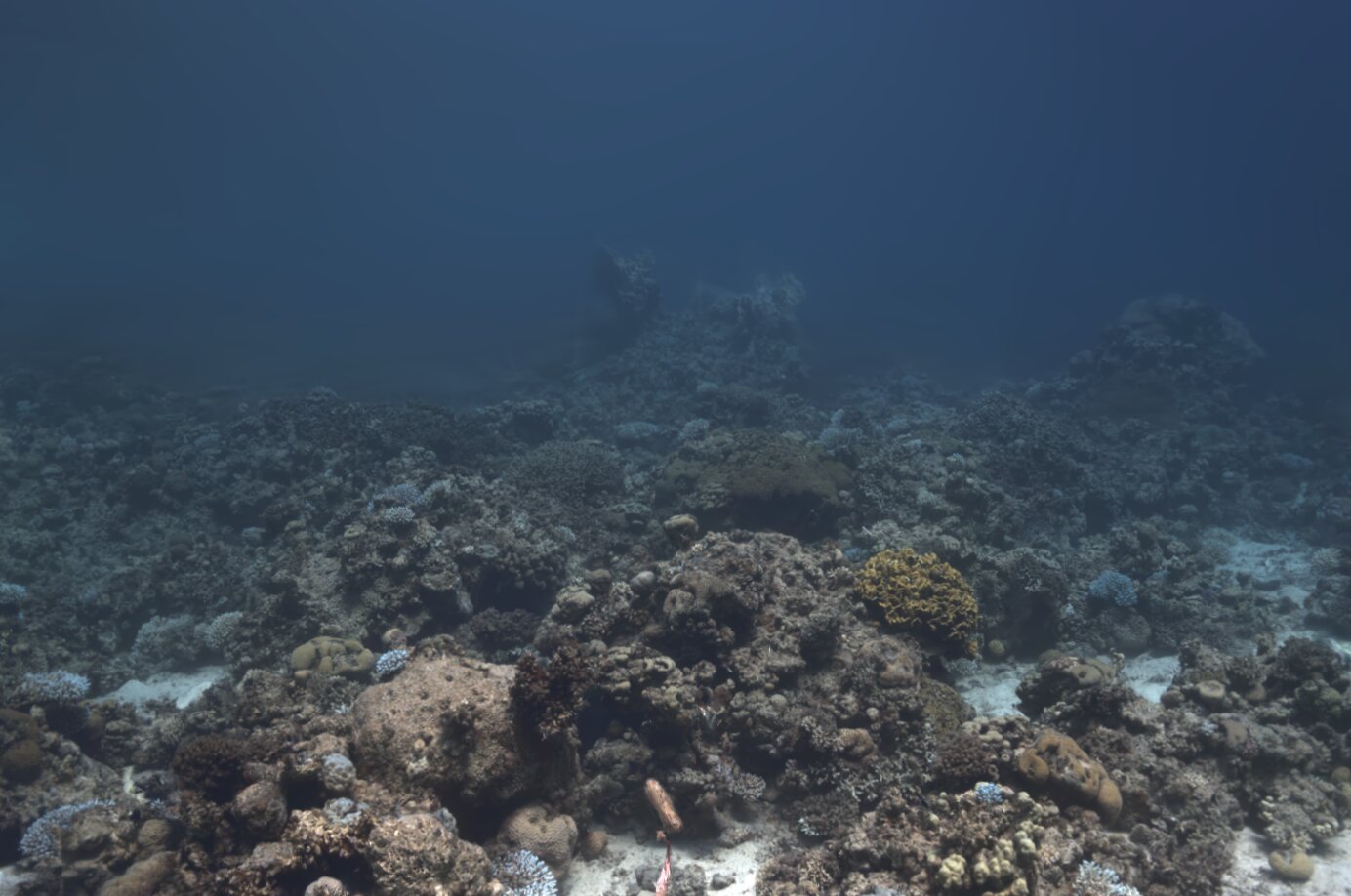}}
\caption[Supplementary Rendering Results]{Supplementary rendering results exemplifying the effectiveness and visual quality of our method across three different scenes and conditions.
Train refers to the data used during the reconstruction. Test refers to novel views. }
\label{fig:appendix}
\end{figure*}

%% file: figures/appendix/appendix_fig_examples2.tex
\begin{figure*}[!ht]
\newcommand{\myW}{0.24\linewidth}
\newcommand{\myH}{2cm} 
\centering
\subcaptionbox{Ground Truth (train)}[\myW]
{\includegraphics[width=\linewidth, height=\myH]{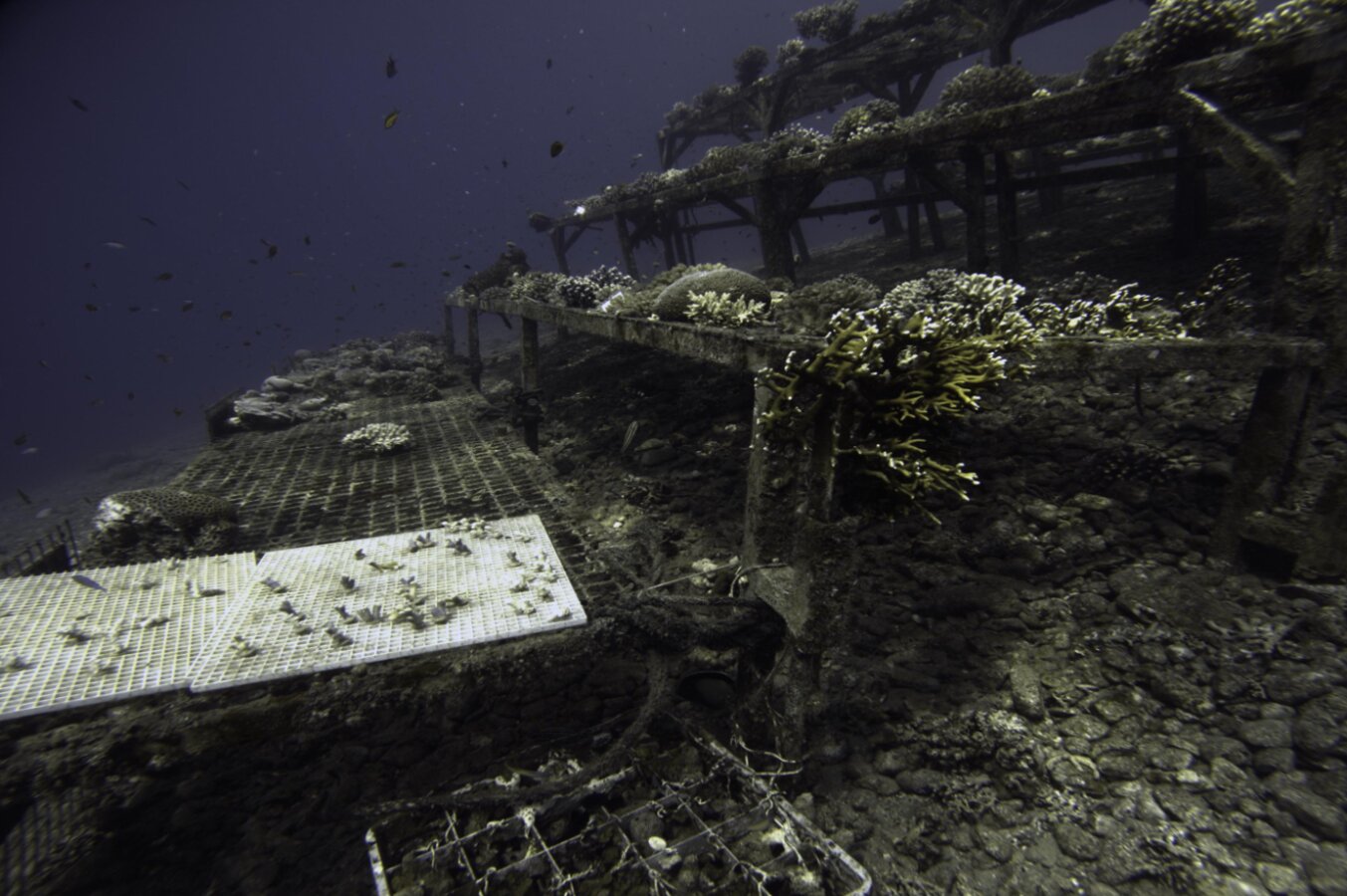}\\
 \includegraphics[width=\linewidth, height=\myH]{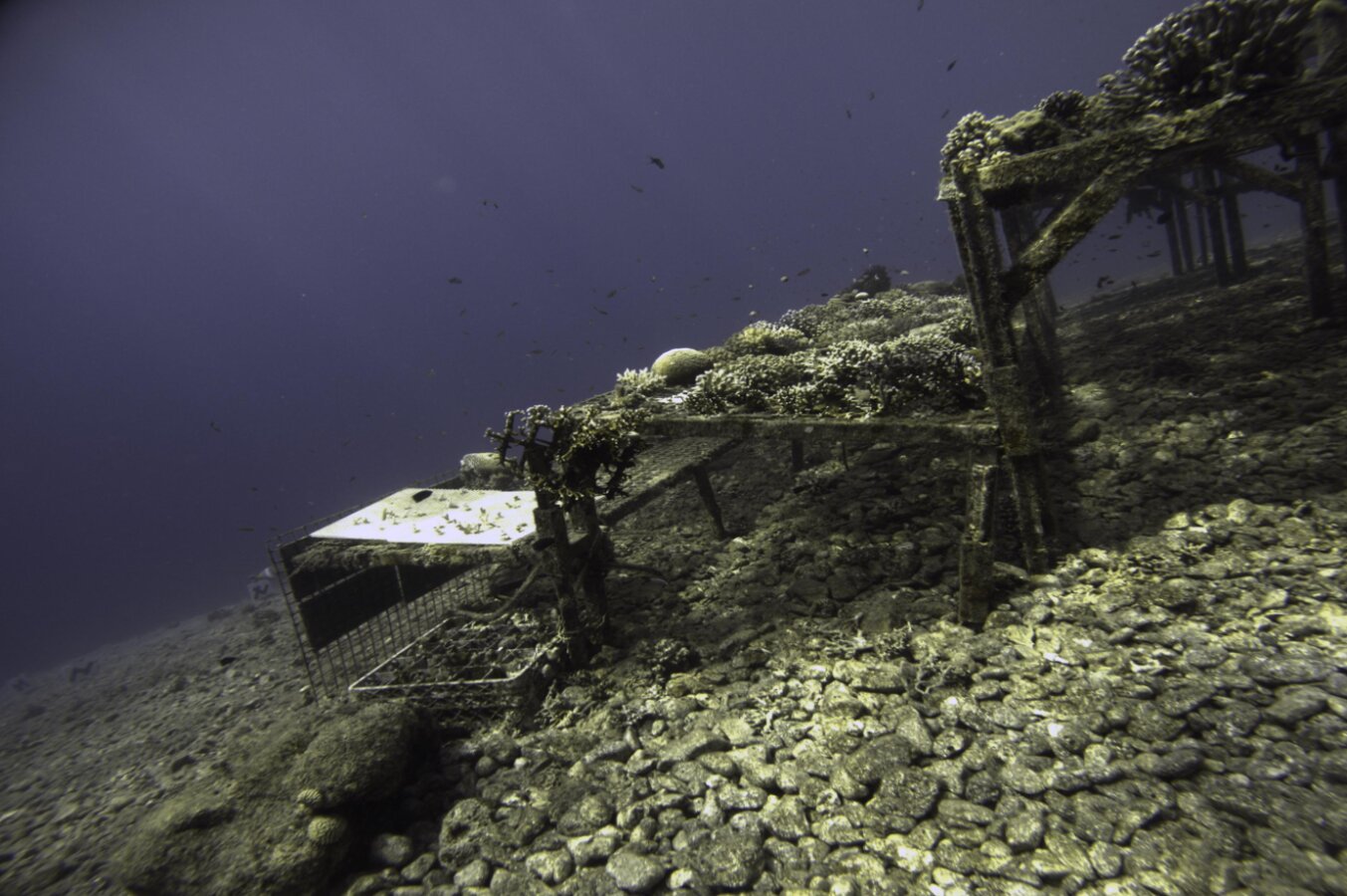}\\
 \includegraphics[width=\linewidth, height=\myH]{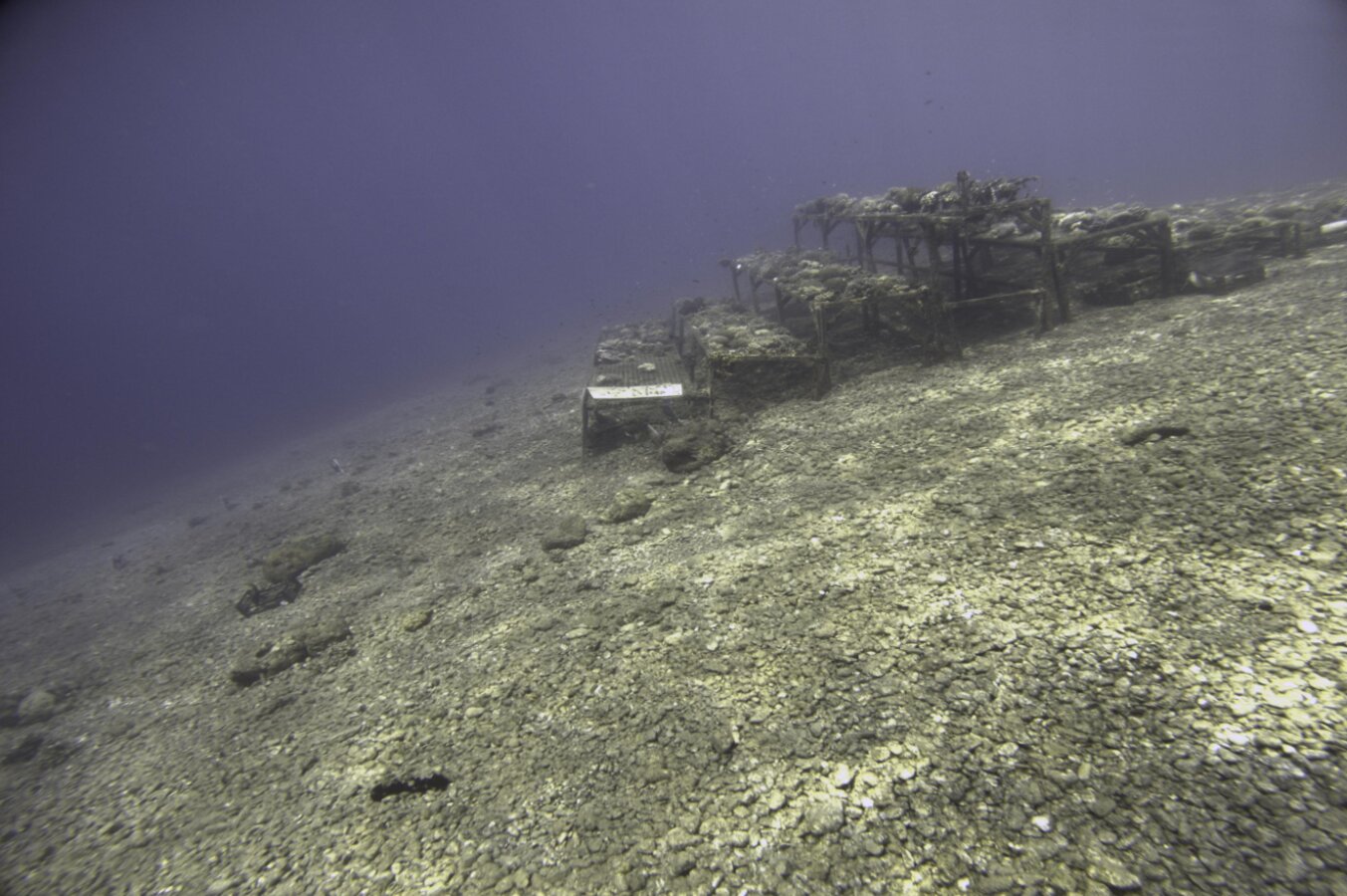}\\
 \includegraphics[width=\linewidth, height=\myH]{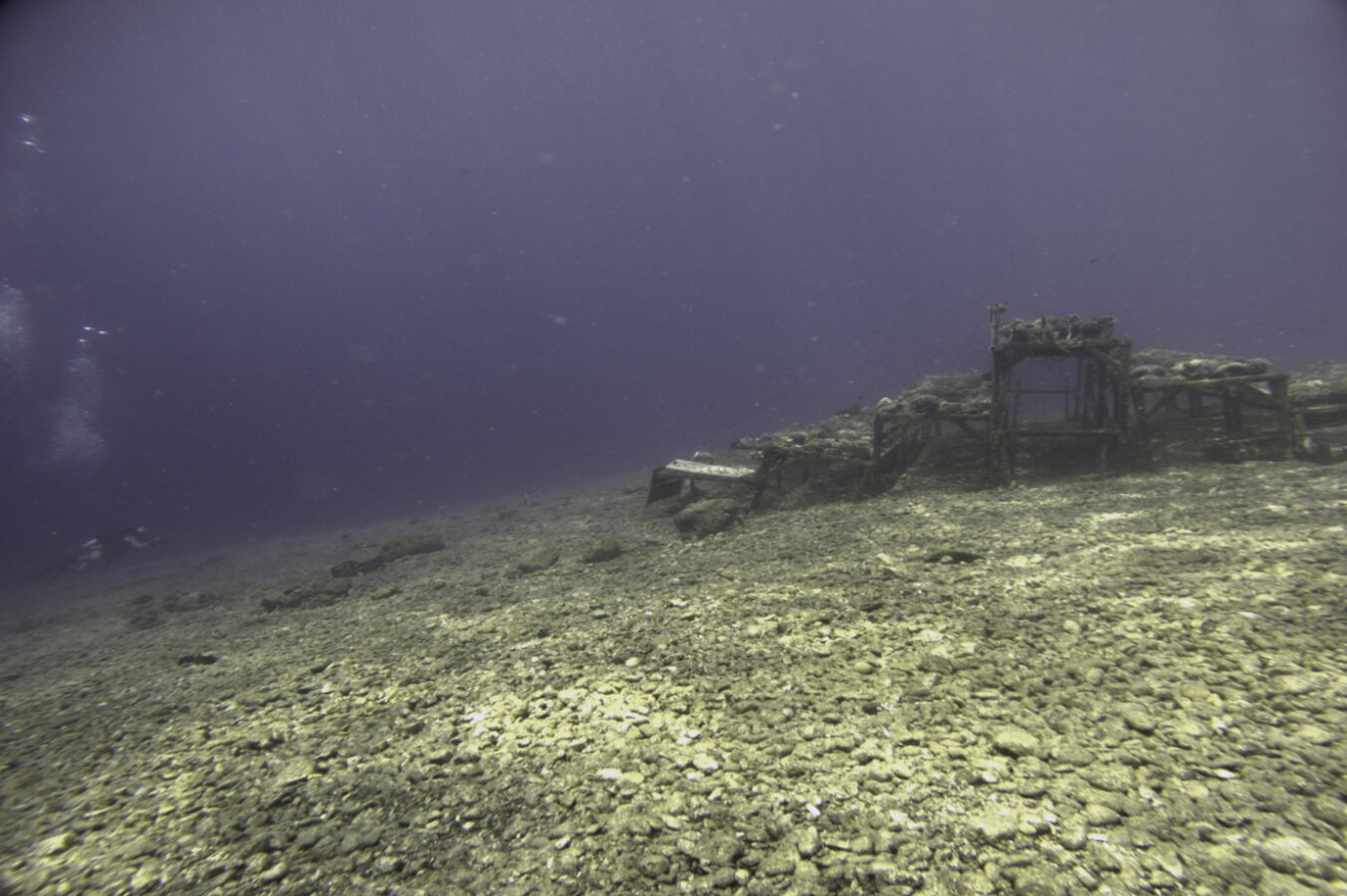}\\
 \includegraphics[width=\linewidth, height=\myH]{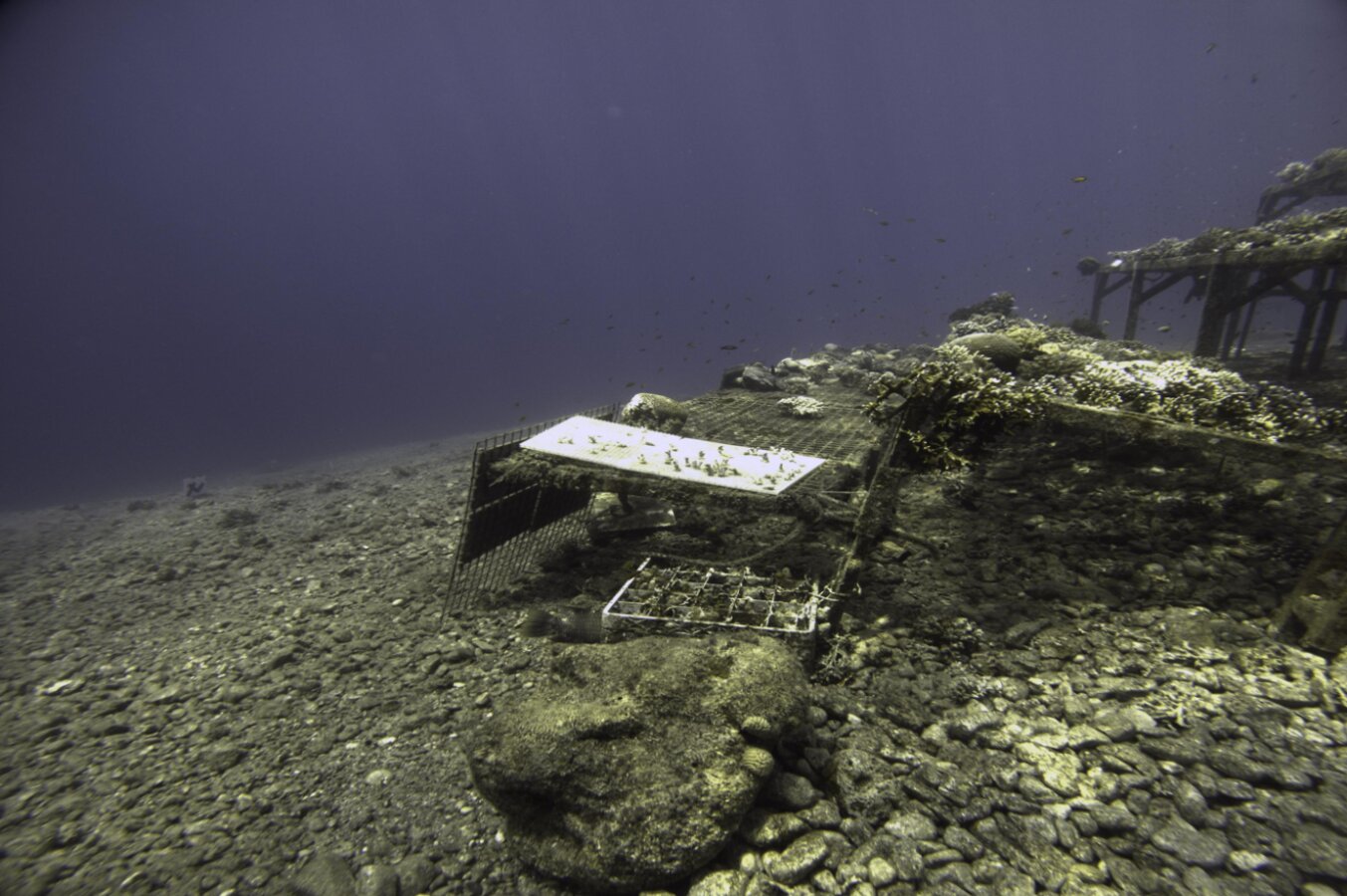}\\
 \includegraphics[width=\linewidth, height=\myH]{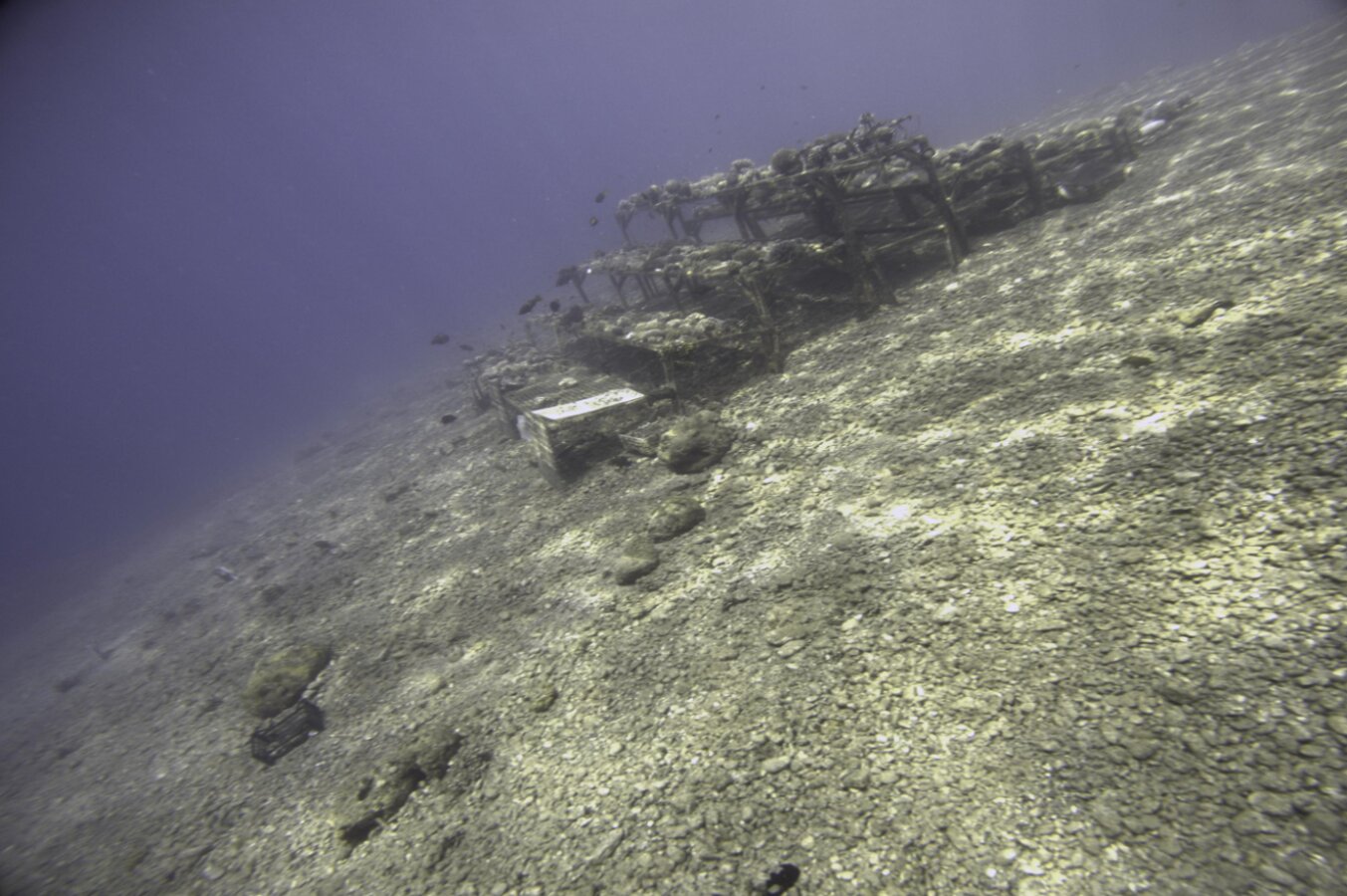}}
\subcaptionbox{Rendered (train)}[\myW]
{\includegraphics[width=\linewidth, height=\myH]{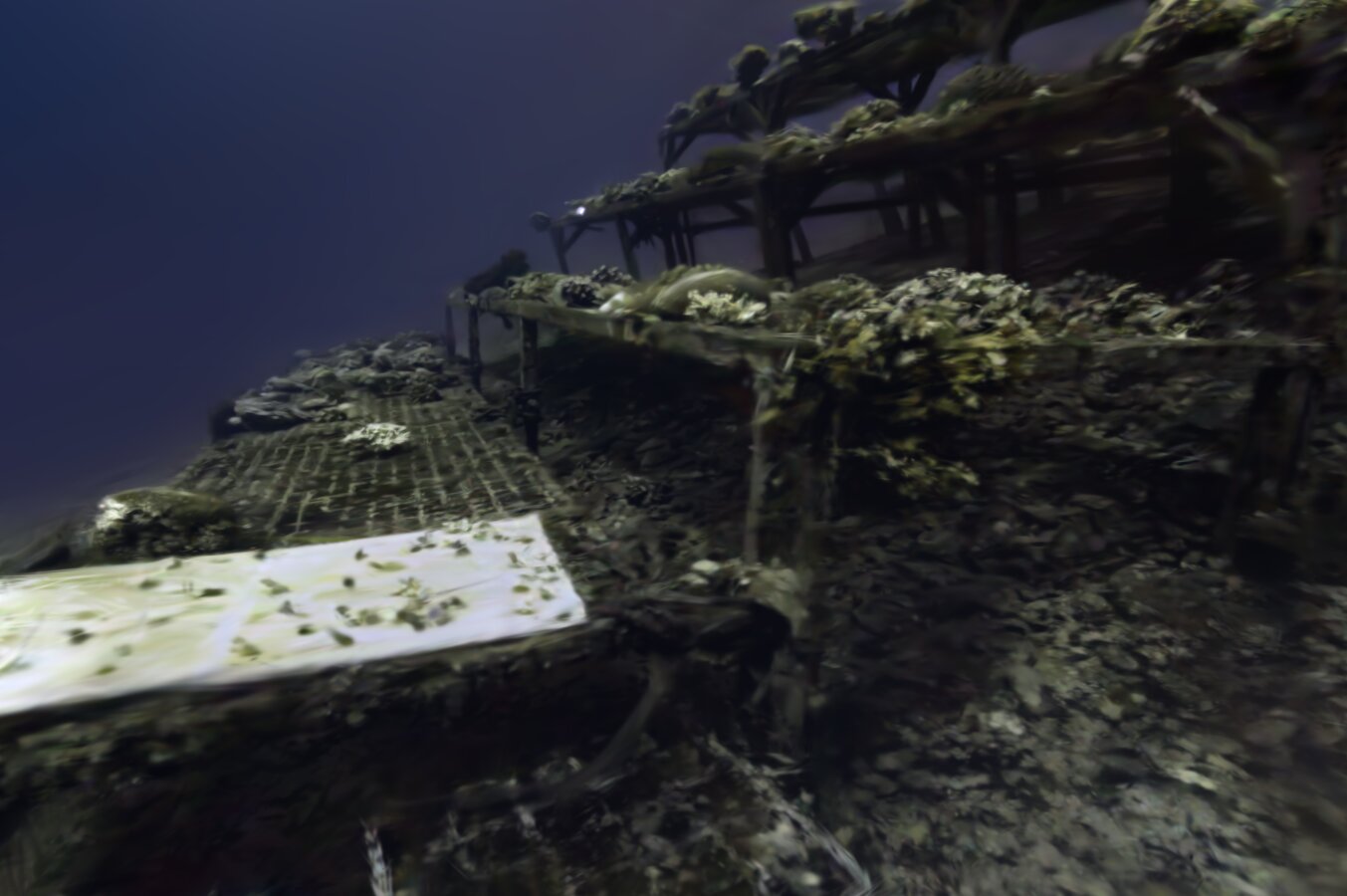}\\
 \includegraphics[width=\linewidth, height=\myH]{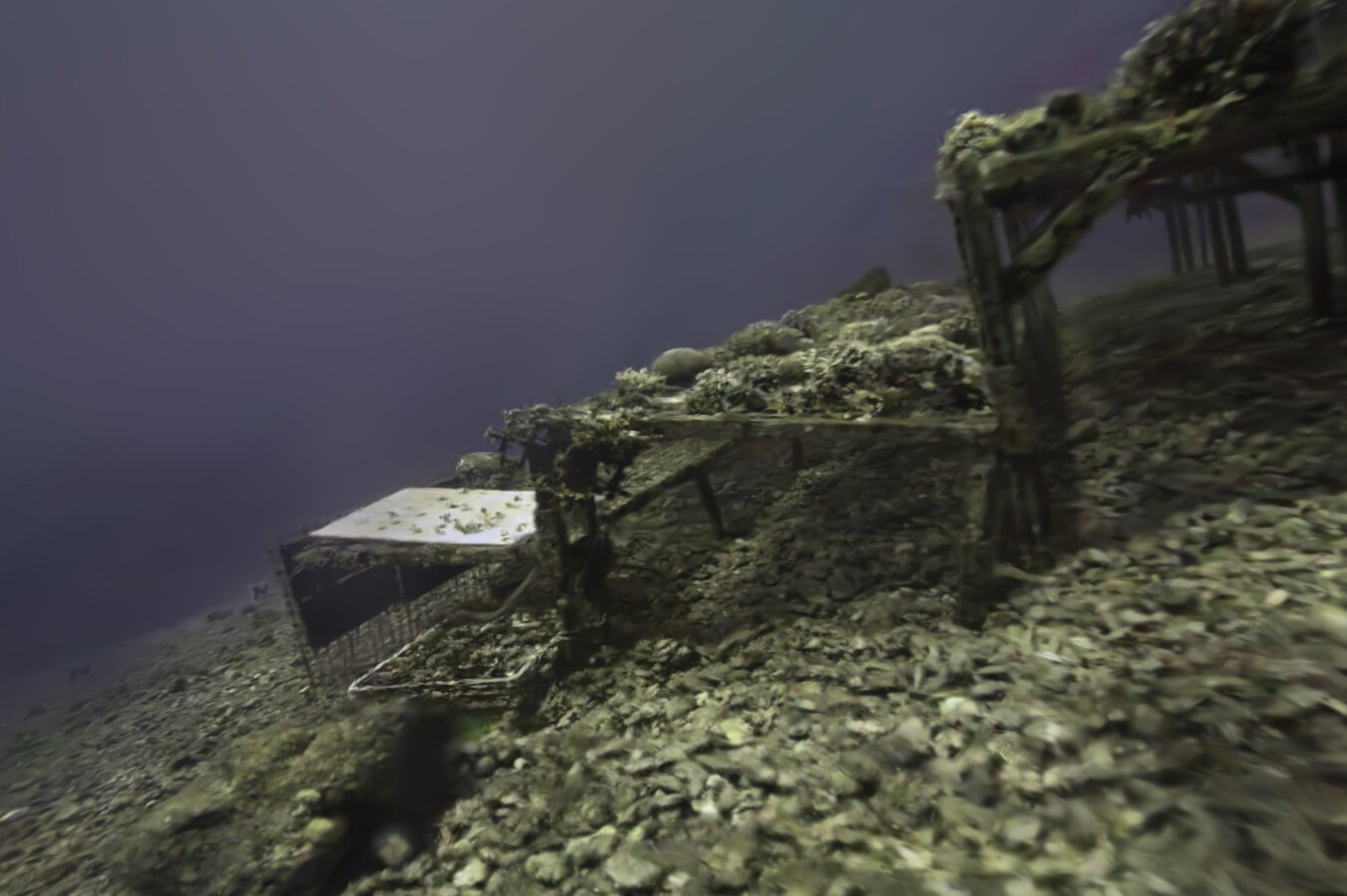}\\
 \includegraphics[width=\linewidth, height=\myH]{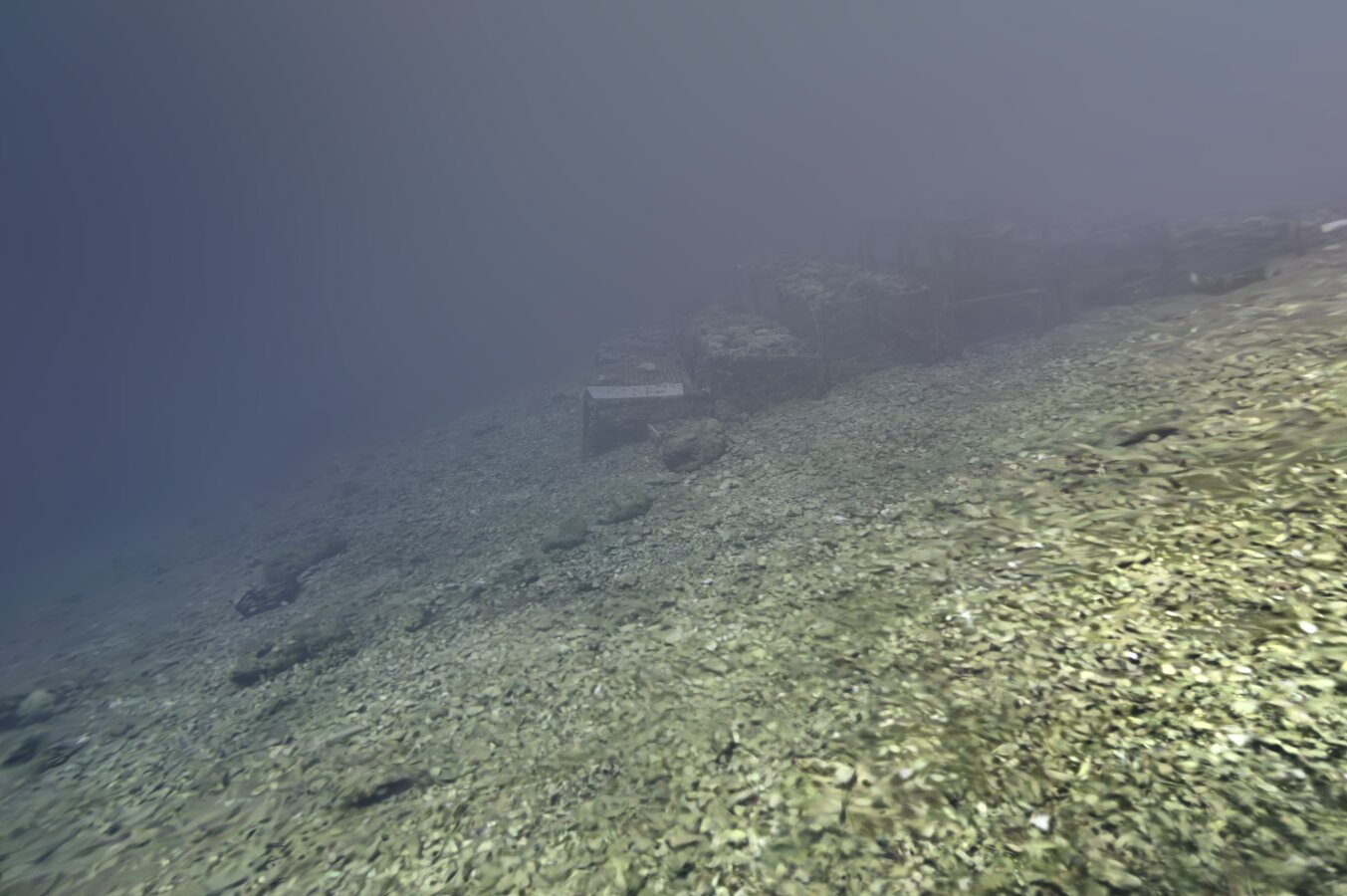}\\
 \includegraphics[width=\linewidth, height=\myH]{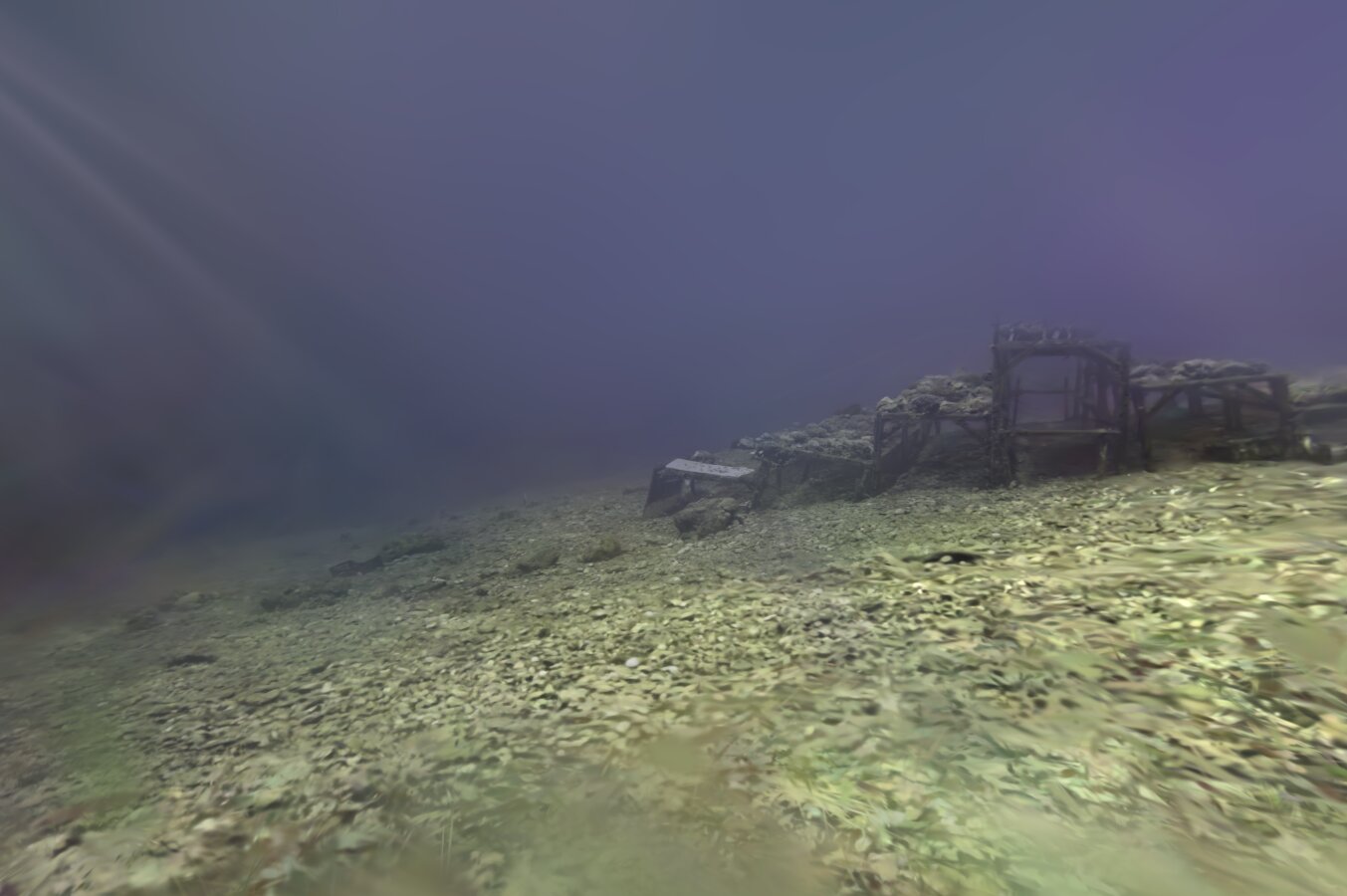}\\
 \includegraphics[width=\linewidth, height=\myH]{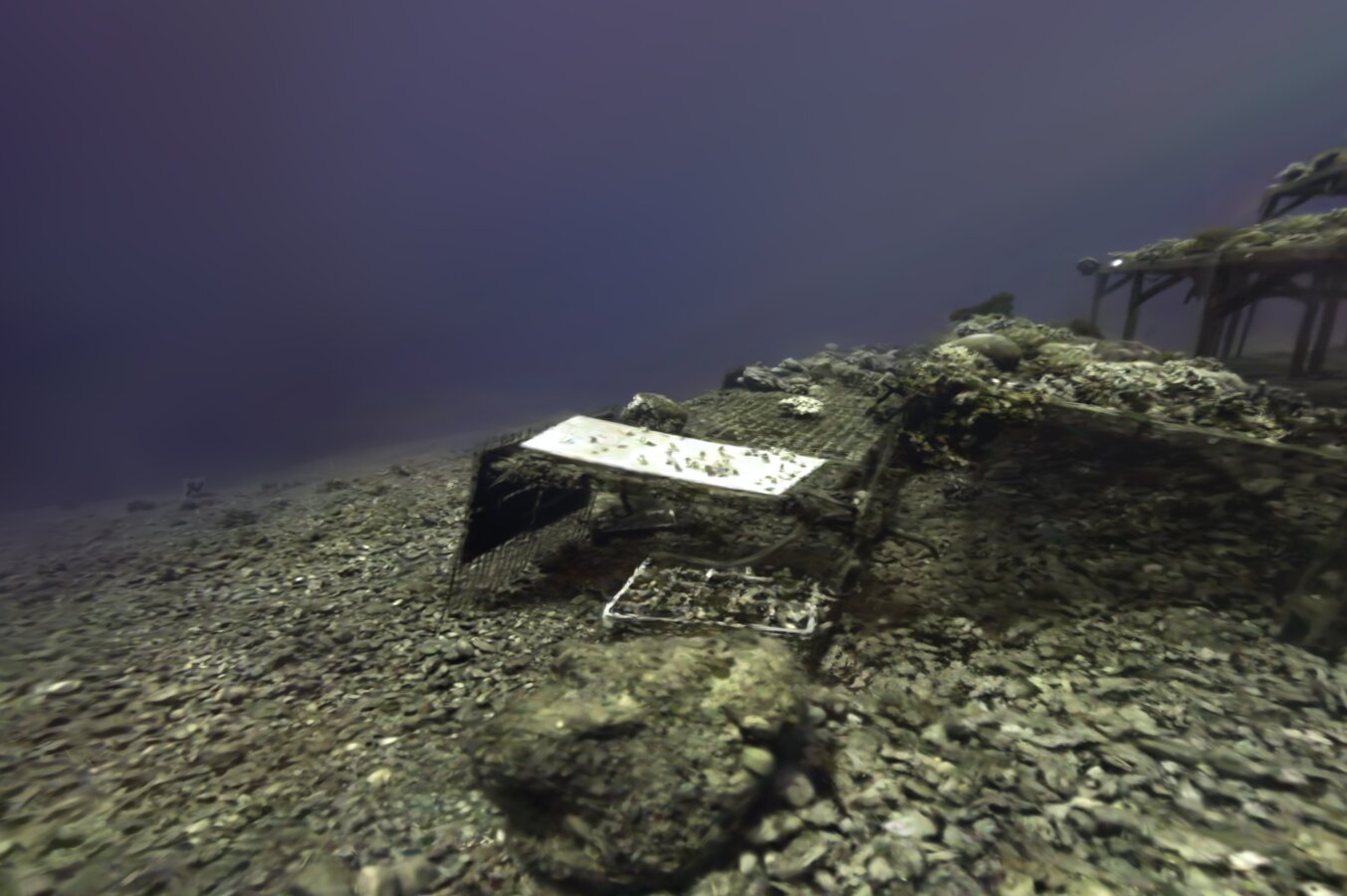}\\
 \includegraphics[width=\linewidth, height=\myH]{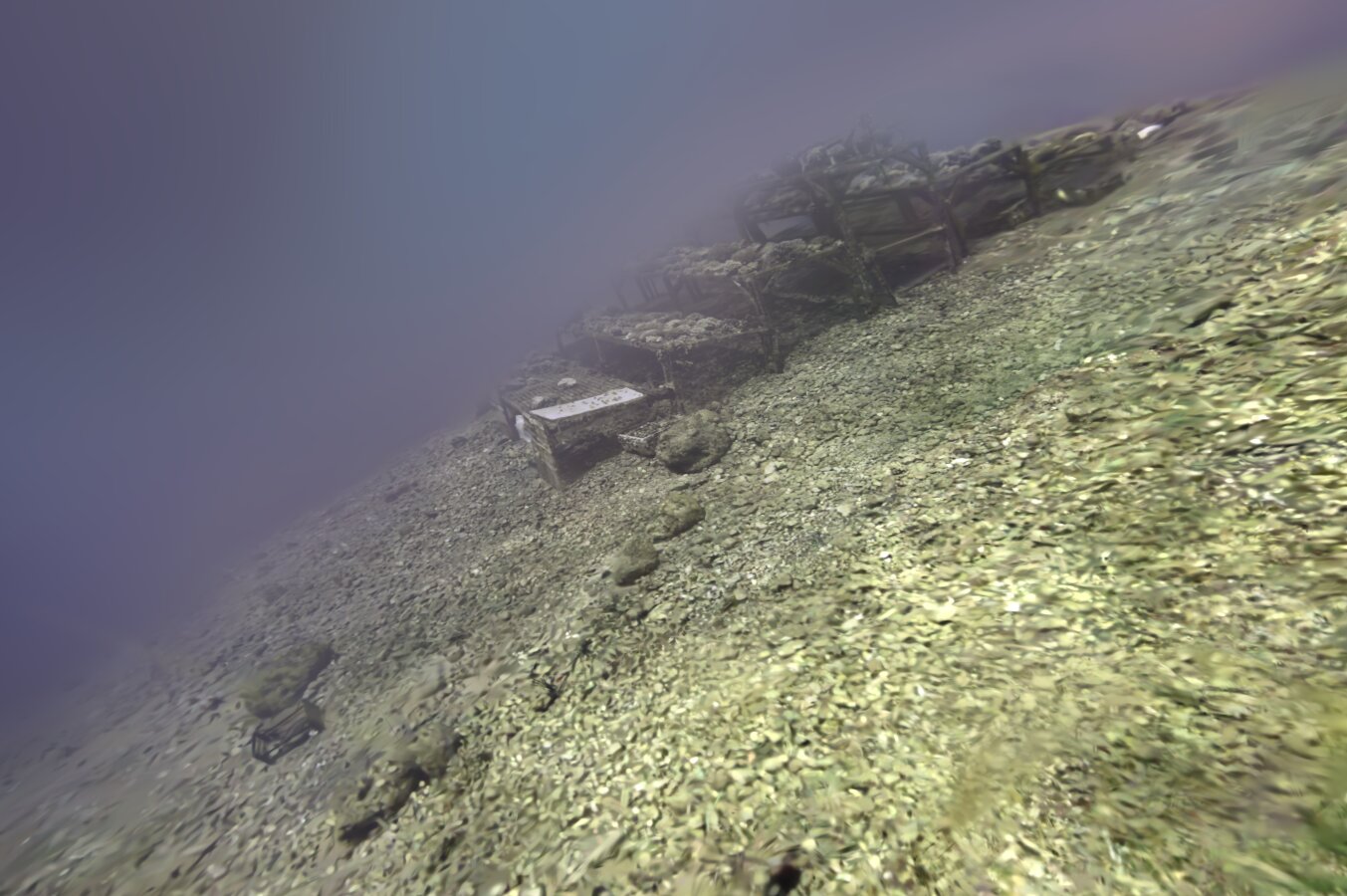}}
\subcaptionbox{Ground Truth (test)}[\myW]
{\includegraphics[width=\linewidth, height=\myH]{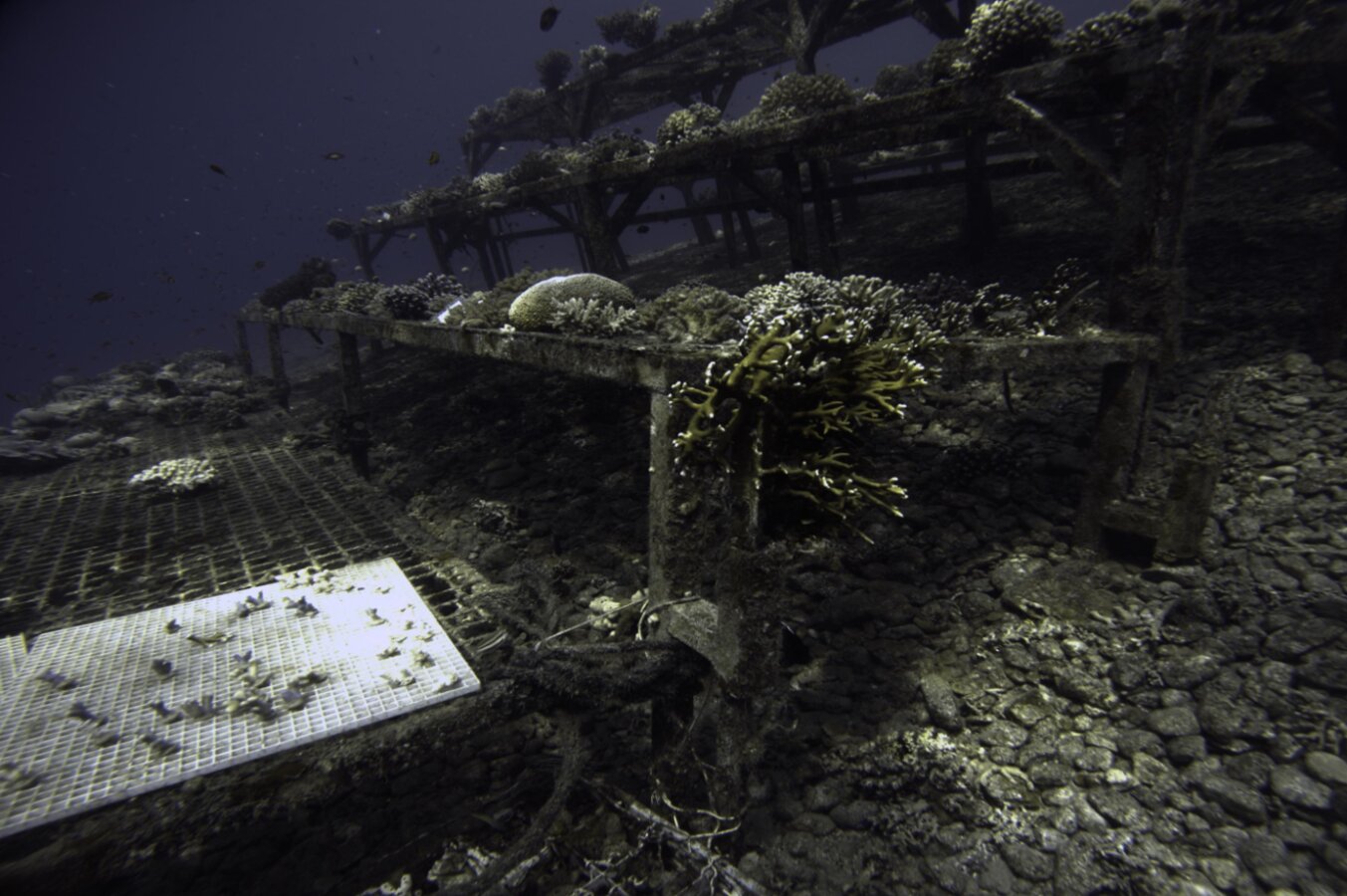}\\
 \includegraphics[width=\linewidth, height=\myH]{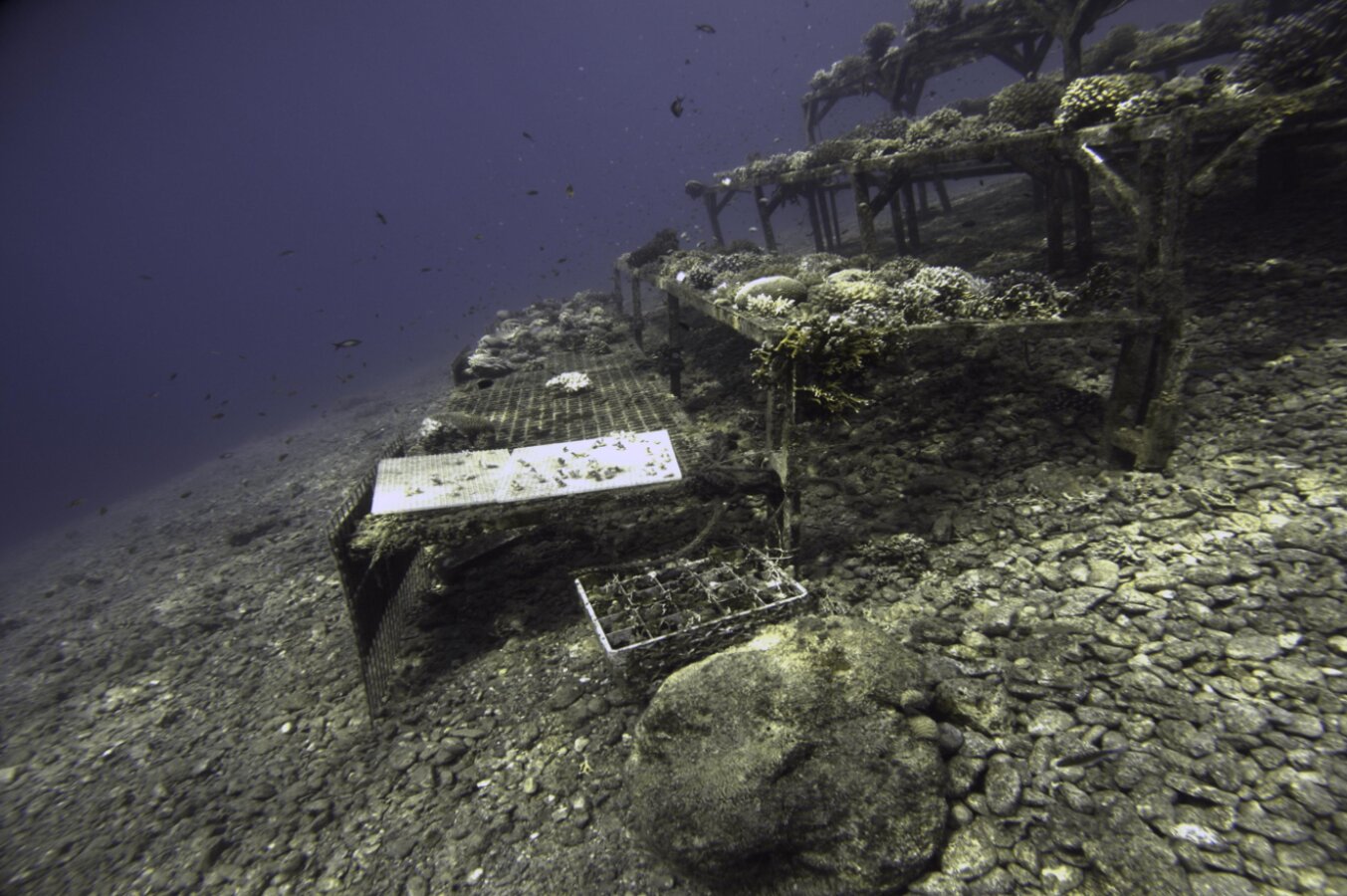}\\
 \includegraphics[width=\linewidth, height=\myH]{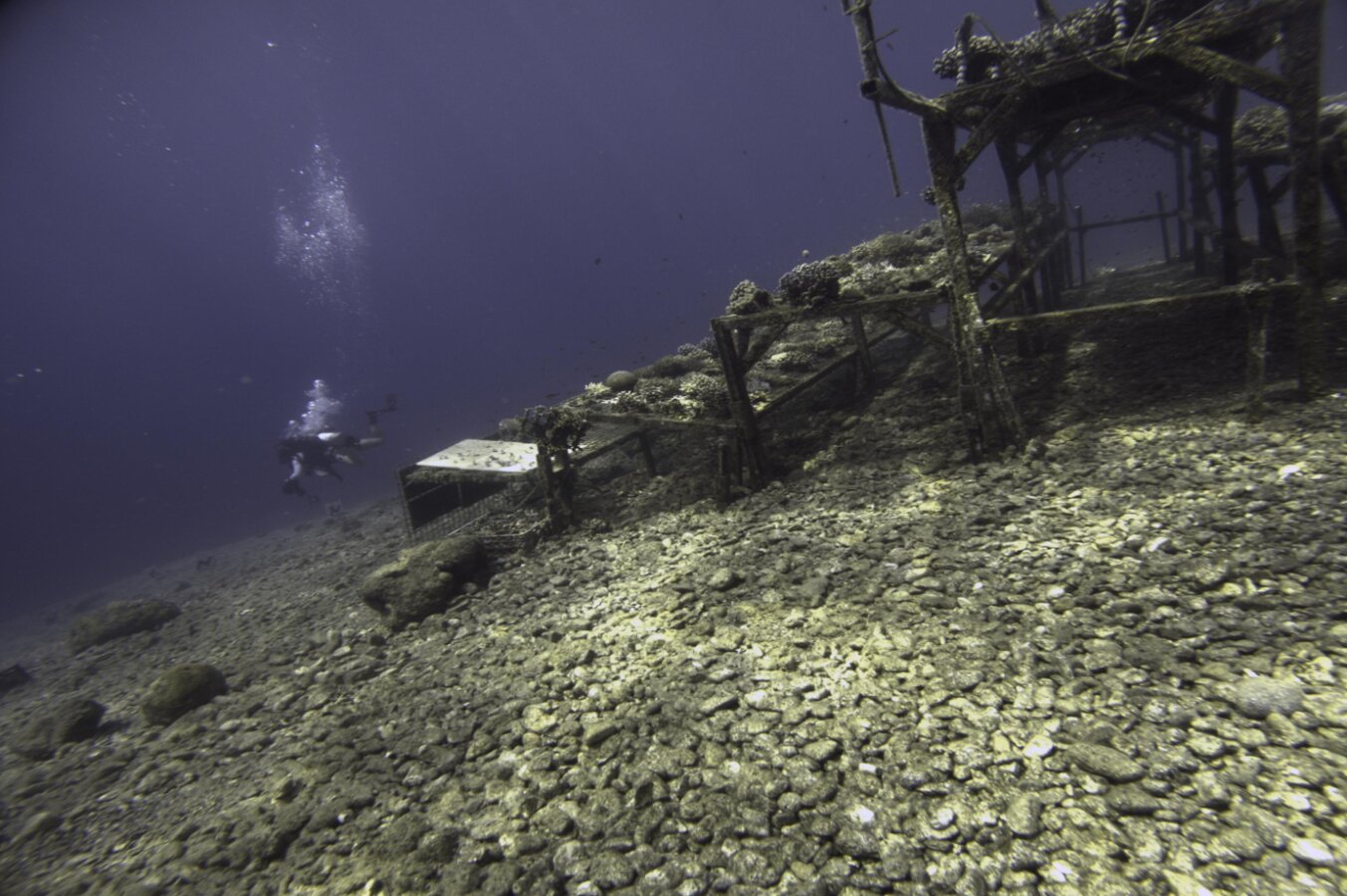}\\
 \includegraphics[width=\linewidth, height=\myH]{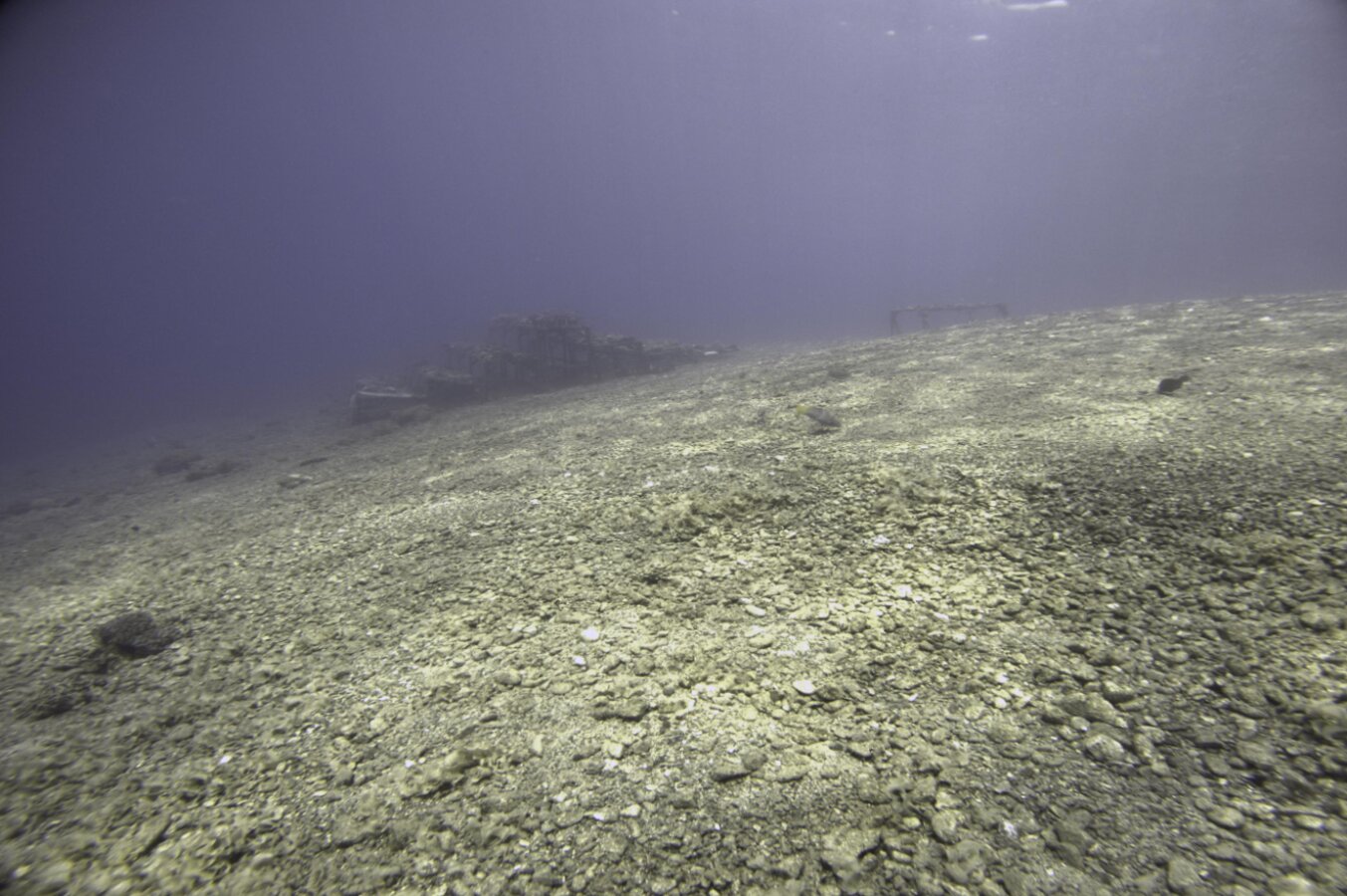}\\
 \includegraphics[width=\linewidth, height=\myH]{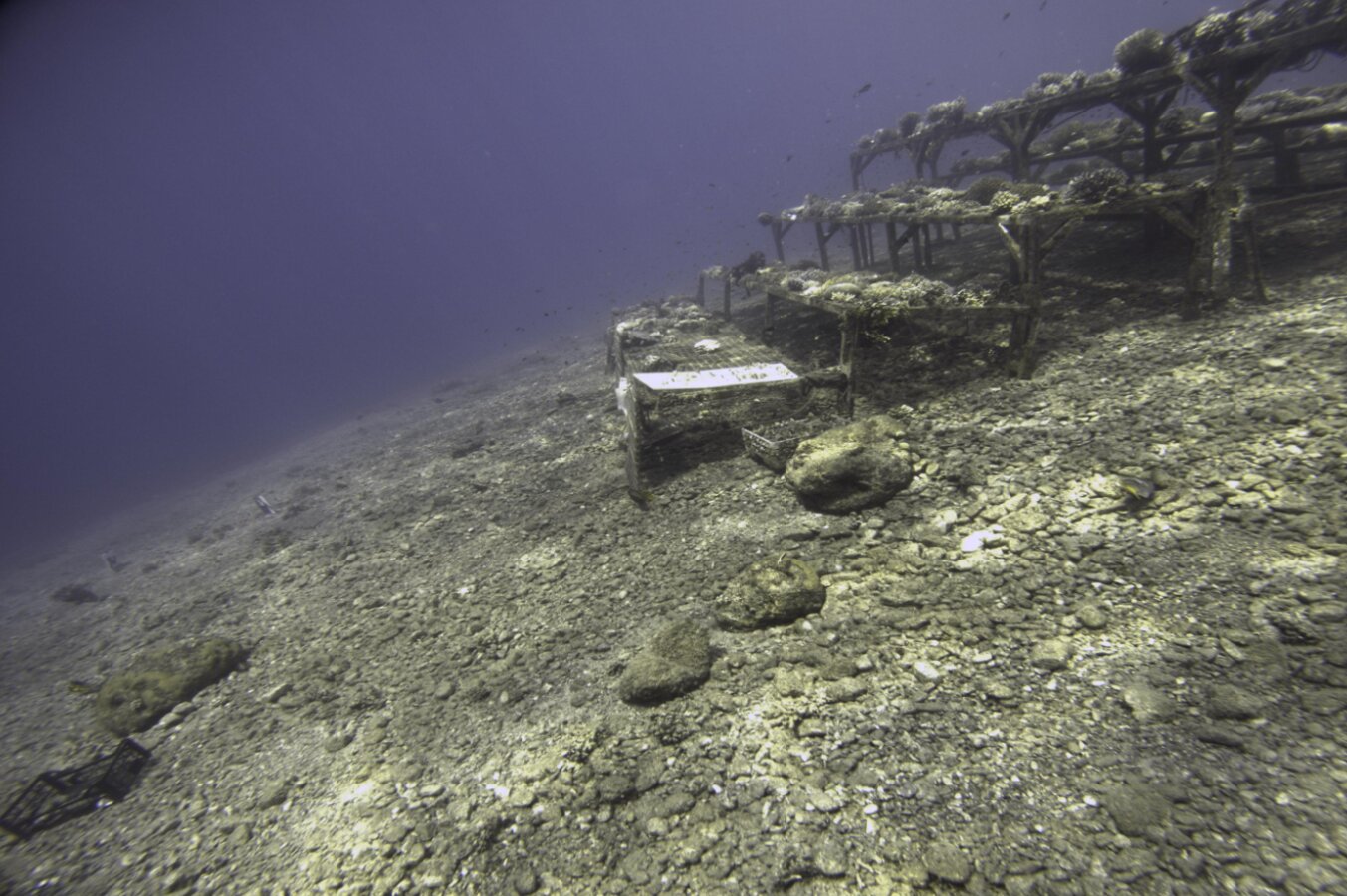}\\
 \includegraphics[width=\linewidth, height=\myH]{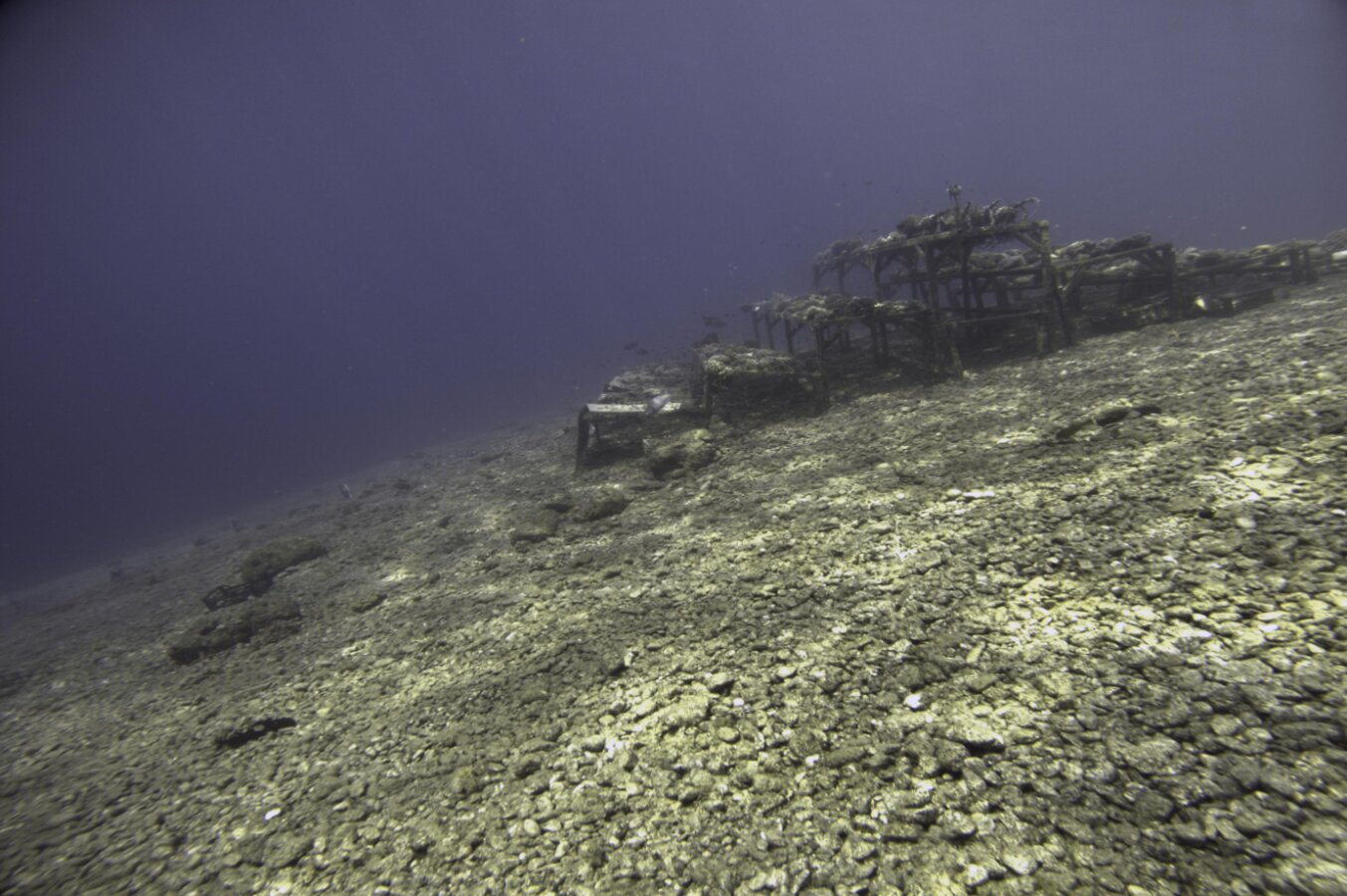}}
 \subcaptionbox{Rendered (test)}[\myW]
{\includegraphics[width=\linewidth, height=\myH]{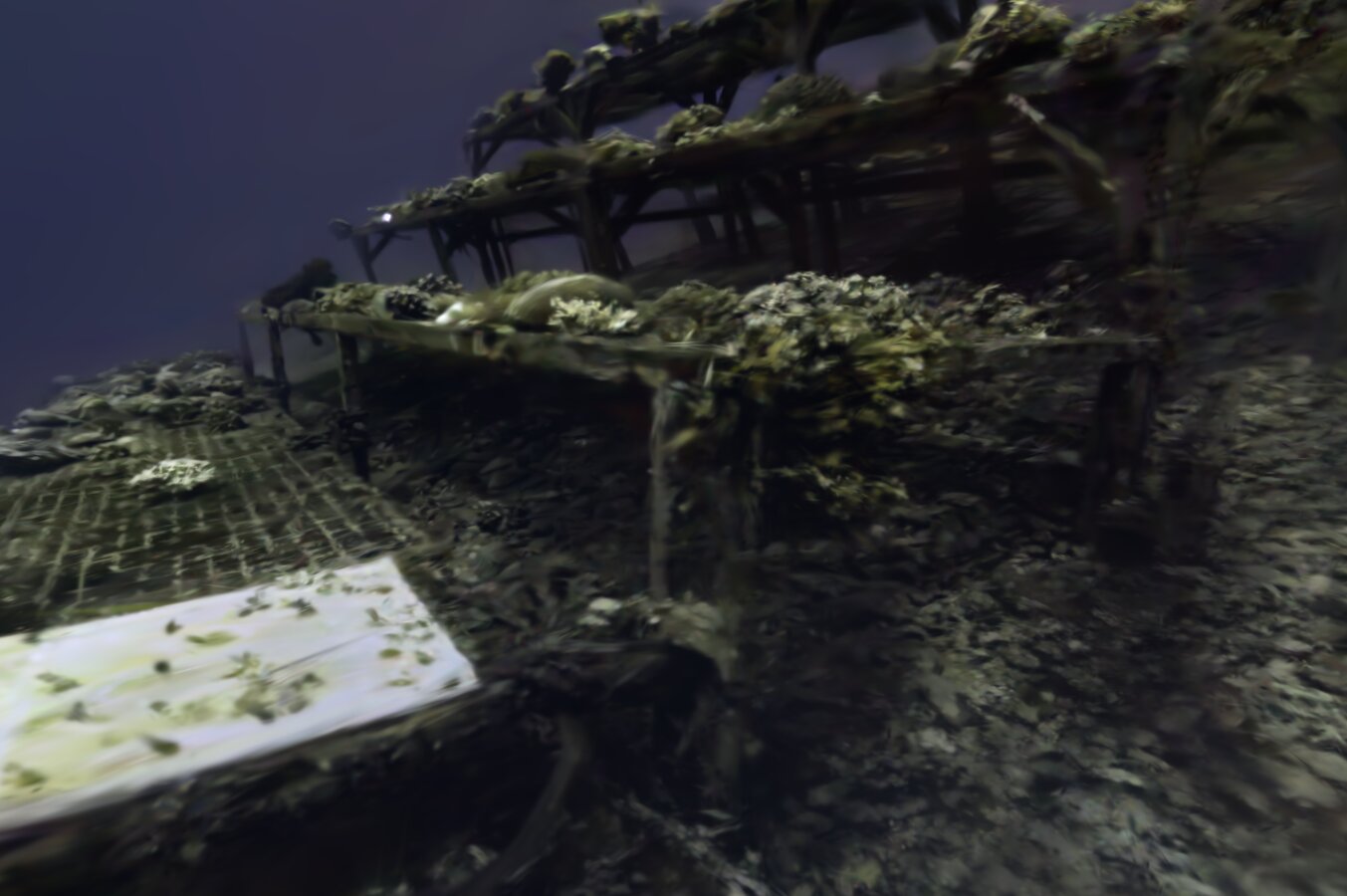}\\
 \includegraphics[width=\linewidth, height=\myH]{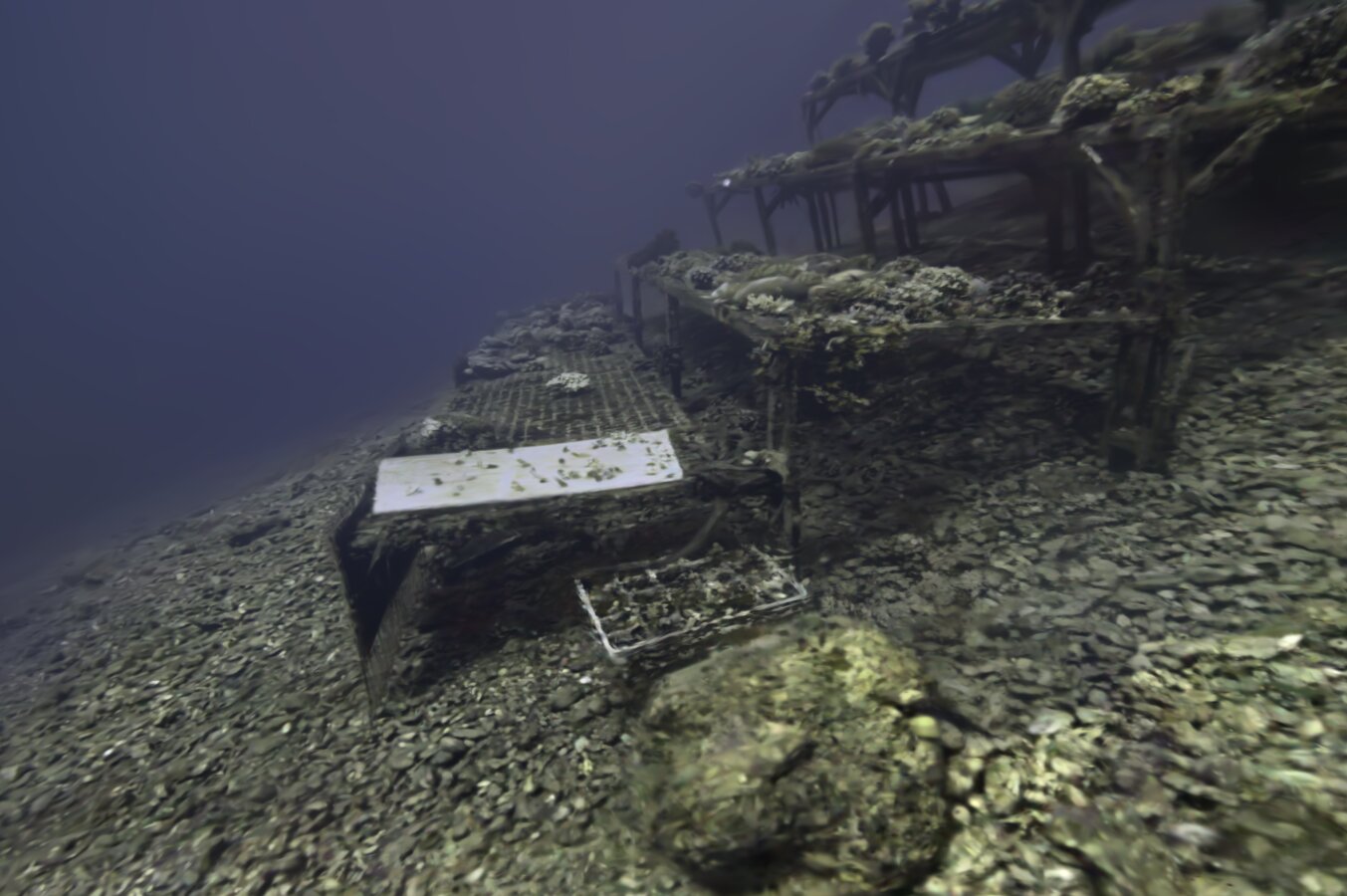}\\
 \includegraphics[width=\linewidth, height=\myH]{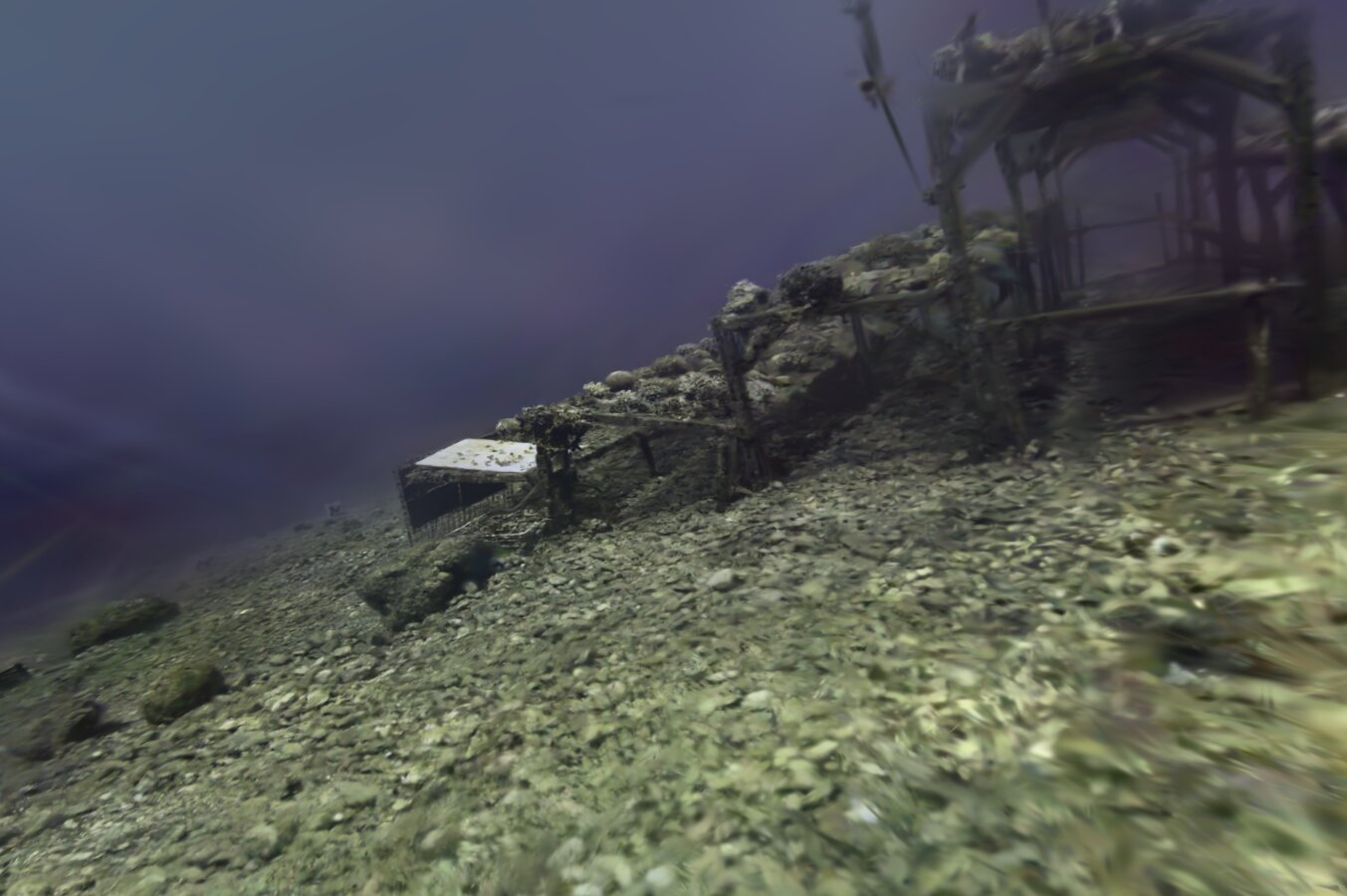}\\
 \includegraphics[width=\linewidth, height=\myH]{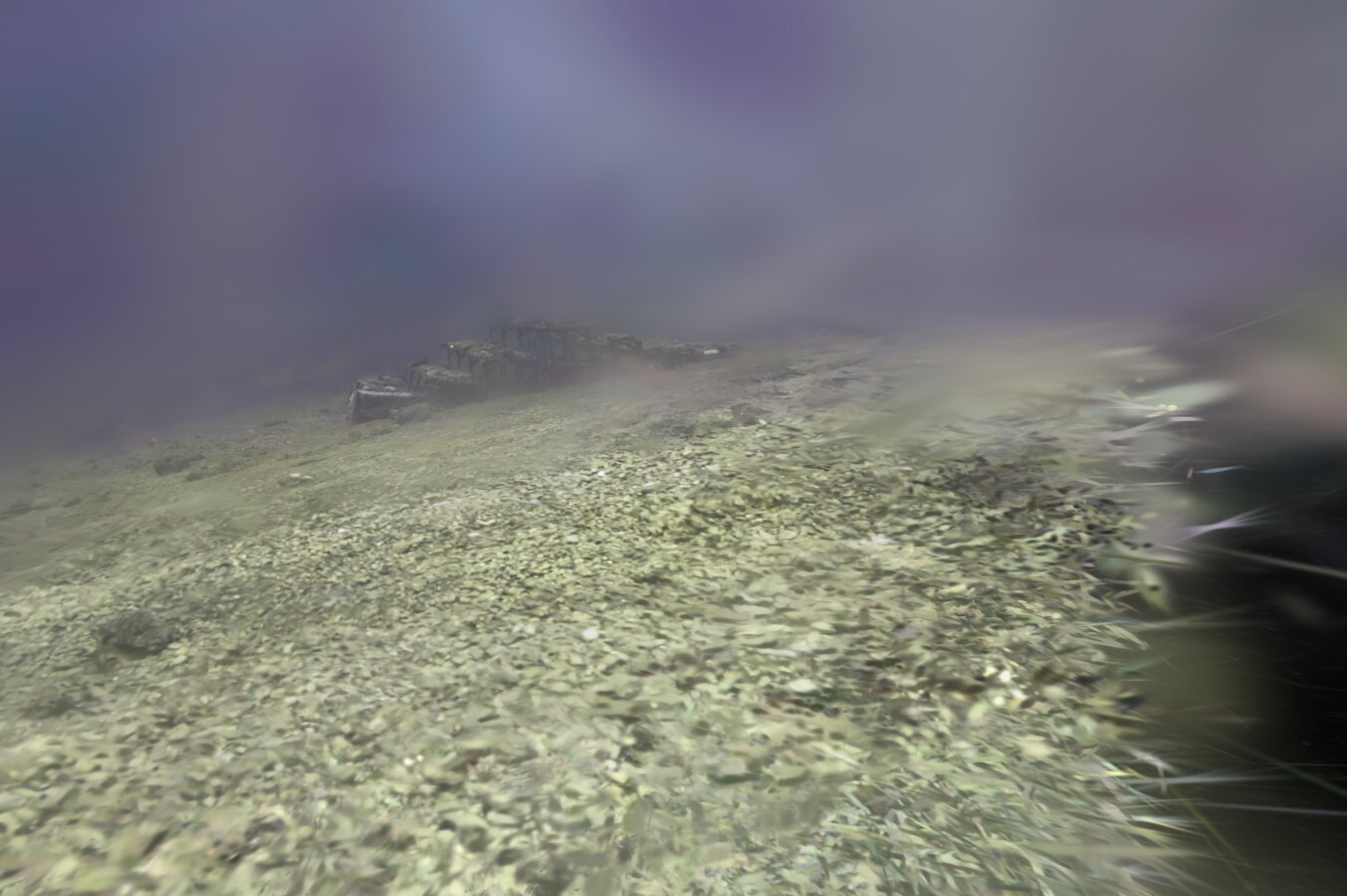}\\
 \includegraphics[width=\linewidth, height=\myH]{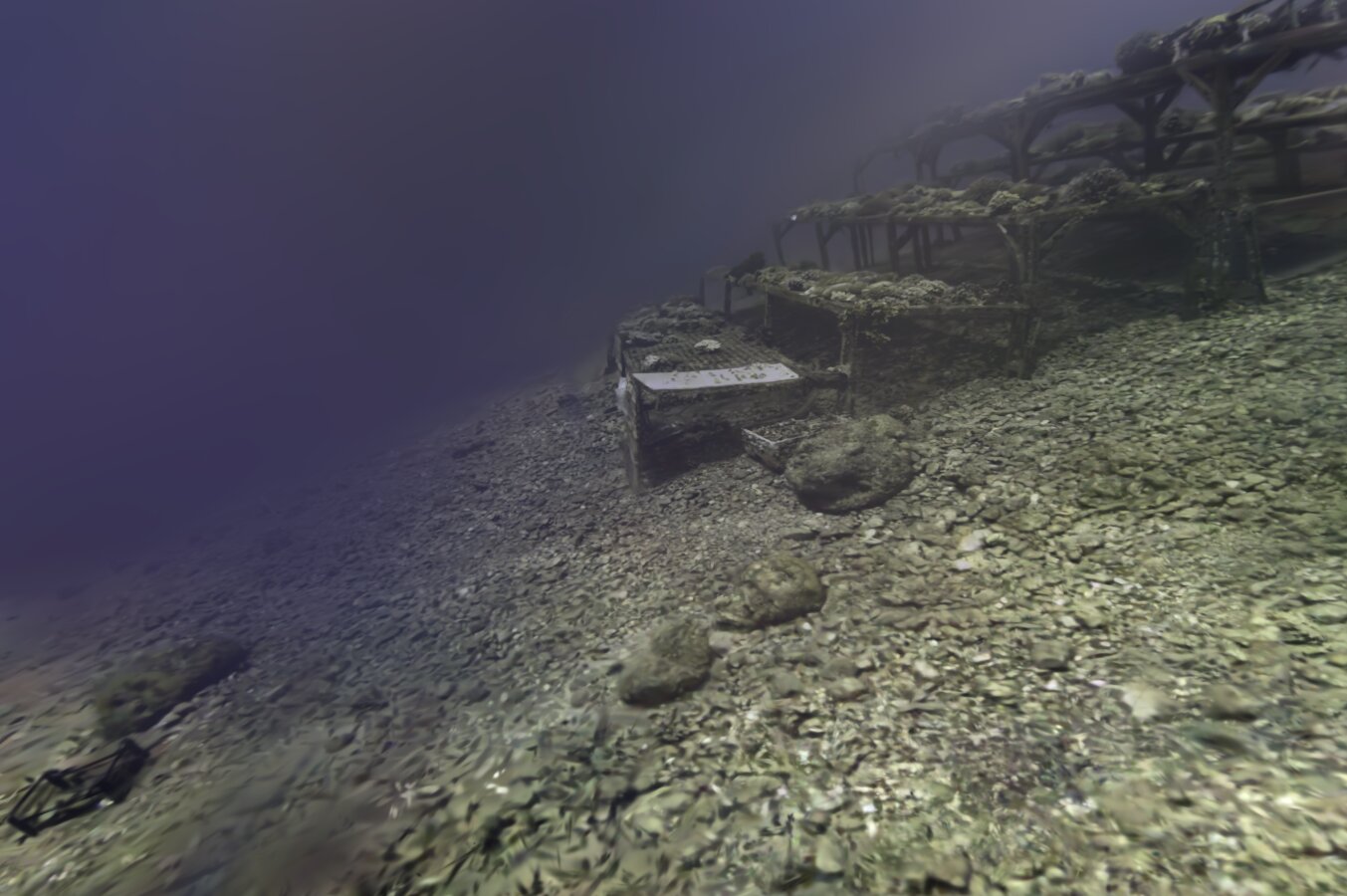}\\
 \includegraphics[width=\linewidth, height=\myH]{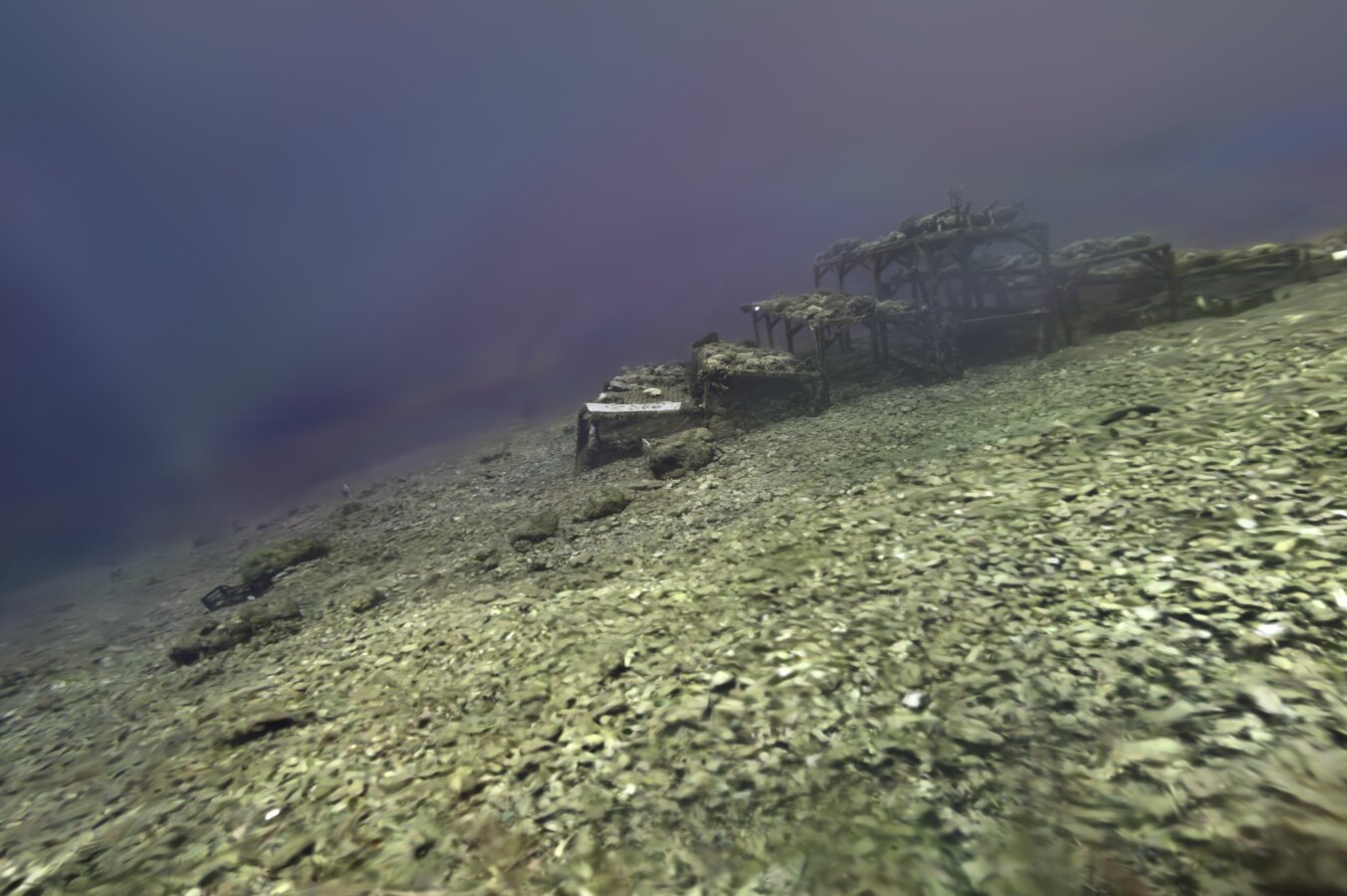}}
\caption[Supplementary Rendering Results - TableDB]{Supplementary rendering results demonstrating the effectiveness and visual quality of our method on the TableDB dataset. TableDB is a distinctive underwater dataset featuring a wide range of depths. ``Train'' refers to data used for reconstruction, while ``Test'' represents novel views. }
\label{fig:appendix2}
\end{figure*}

%% file: figures/appendix/fig_attenuation.tex
\graphicspath{{./figures/appendix/}}
\begin{figure*}
    \centering
    \begin{overpic}[width=\linewidth, height=6cm]{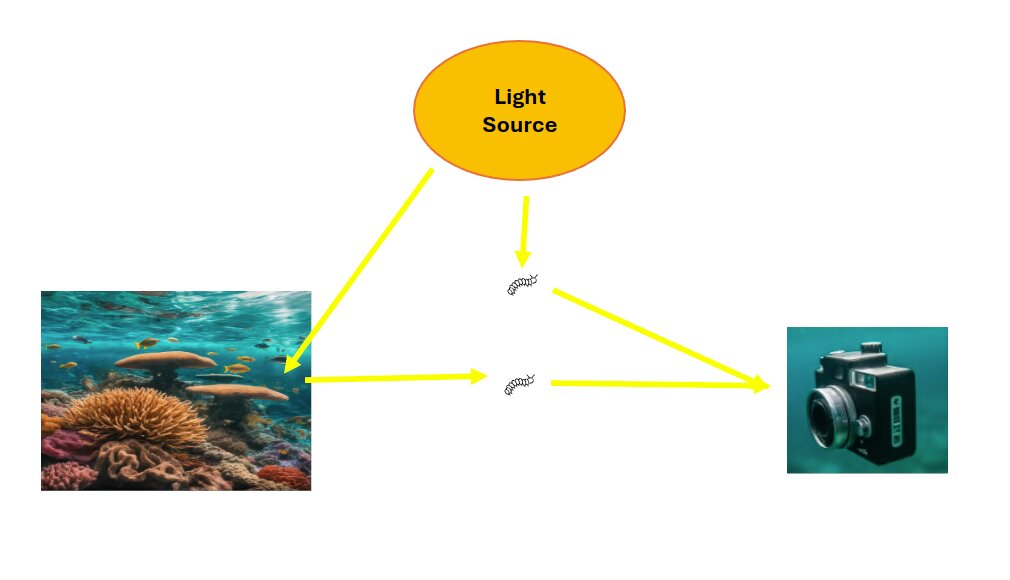} 
        \put(37,14){\colorbox{white}{\textcolor{black}{\small\textbf{\emph{attenuation}}}}}
        \put(60,17){\colorbox{white}{\textcolor{black}{\small\textbf{\emph{backscatter}}}}}
    \end{overpic}
    \caption{Illustration of the scattering and absorption of light by waterborne particles, which then reflect back to the camera. This phenomenon leads to image degradation by reducing contrast, altering colors, and obscuring fine details, posing significant challenges for underwater imaging and rendering.}
    \label{fig:atten_figure}
\end{figure*}